%% file: main.tex
\documentclass[10pt,twocolumn,letterpaper]{article}

\usepackage[pagenumbers]{cvpr_arxiv} %

\usepackage{booktabs} %

\def\arxiv{1}

\usepackage[ruled]{algorithm2e} %
\usepackage{soul} 

\usepackage{commath} 
\usepackage{float}
\usepackage{pifont} %

\usepackage{multirow}

\SetAlFnt{\small}
\SetAlCapFnt{\small}
\SetAlCapNameFnt{\small}
\SetAlCapHSkip{0pt}
\usepackage{amsmath}

\usepackage[pagebackref,breaklinks,colorlinks]{hyperref}

\usepackage[capitalize, sort]{cleveref}
\crefname{section}{Sec.}{Secs.}
\Crefname{section}{Section}{Sections}
\Crefname{table}{Table}{Tables}
\crefname{table}{Tab.}{Tabs.}

\newcommand{\shortcite}[1]{\cite{#1}}

\input{macros}

\begin{document}

\title{
Self-Conditioned Generative Adversarial Networks for Image Editing}

\author{Yunzhe Liu$^{1}$ \and Rinon Gal$^{2}$ \and Amit H. Bermano$^{2}$ \and Baoquan Chen$^{1}$ \and Daniel Cohen-Or$^{2}$ \and \\ CFCS, Peking University$^{1}$ \and \\ Tel Aviv University$^{2}$}

\maketitle

\input{abstract_arxiv}

\input{intro}

\input{related}

\input{method}

\input{experiments}

\input{future}

\bibliographystyle{ieee_fullname}
\bibliography{main}

\end{document}

%% file: macros.tex
\usepackage{comment}
\usepackage{multirow,bigdelim}
\usepackage{lipsum}
\usepackage{array}
\usepackage{wrapfig}

\usepackage{adjustbox}
\usepackage[percent]{overpic}
\usepackage{makecell}
\usepackage[normalem]{ulem}

\usepackage{blindtext}

\usepackage{xcolor}
\usepackage{soul}

\newcolumntype{C}[1]{>{\centering\let\newline\\\arraybackslash\hspace{0pt}}m{#1}}

\newcommand{\myarrow}[1][3cm]{\mathrel{%
    \hbox{\rule[\dimexpr\fontdimen22\textfont2-.2pt\relax]{#1}{.4pt}}%
    \mkern-4mu\hbox{\usefont{U}{lasy}{m}{n}\symbol{41}}}}

\newcommand{\myleftarrow}[1][3cm]{\mathrel{%
    \hbox{\usefont{U}{lasy}{m}{n}\symbol{40}}\mkern-4mu%
    \hbox{\rule[\dimexpr\fontdimen22\textfont2-.2pt\relax]{#1}{.4pt}}%
    }}

\newif\ifdraft
\drafttrue

\ifdraft
\newcommand{\dcc}[1]{{\color{red}[\textbf{DC:} #1]}}
\newcommand{\rgc}[1]{{\color{purple}[\textbf{RG:} #1]}}
\newcommand{\ylc}[1]{{\color{blue}[\textbf{YL:} #1]}}

\else
\newcommand{\dcc}[1]{}
\newcommand{\rgc}[1]{}
\newcommand{\ylc}[1]{}

\fi

\newcommand{\w}{$\mathcal{W}$\xspace}
\newcommand{\wplus}{$\mathcal{W}+$\xspace}

\makeatletter
\DeclareRobustCommand\onedot{\futurelet\@let@token\@onedot}
\def\@onedot{\ifx\@let@token.\else.\null\fi\xspace}

\def\eg{\emph{e.g}\onedot}

\def\ie{\emph{i.e}\onedot}

\def\etal{\emph{et al}\onedot}
\makeatother

\usepackage[bottom]{footmisc}
\makeatletter
\def\blfootnote{\xdef\@thefnmark{}\@footnotetext}
\makeatother

%% file: abstract_arxiv.tex
\begin{abstract}
    Generative Adversarial Networks (GANs) are susceptible to bias, learned from either the unbalanced data, or through mode collapse. The networks focus on the core of the data distribution, leaving the tails --- or the edges of the distribution --- behind. We argue that this bias is responsible not only for fairness concerns, but that it plays a key role in the collapse of latent-traversal editing methods when deviating away from the distribution's core. Building on this observation, we outline a method for mitigating generative bias through a self-conditioning process, where distances in the latent-space of a pre-trained generator are used to provide initial labels for the data. By fine-tuning the generator on a re-sampled distribution drawn from these self-labeled data, we force the generator to better contend with rare semantic attributes and enable more realistic generation of these properties. We compare our models to a wide range of latent editing methods, and show that by alleviating the bias they achieve finer semantic control and better identity preservation through a wider range of transformations. Our code and models will be available at \url{https://github.com/yzliu567/sc-gan}.
    
\end{abstract}

%% file: intro.tex
\section{Introduction}
\label{sec:intro}

\input{resources/figures/teaser}

Generative Adversarial Networks~\cite{goodfellow2014generative} (GANs) have shown remarkable performance on a wide range of synthesis-related tasks. Recently, the structure of their latent space has been thoroughly explored, giving birth to a wide range of methods designed to manipulate the generated images in semantically meaningful ways. 
However, GANs tend to suffer from two kinds of biases. The first, and most intuitive type of bias, is the biased \textit{learned} from the data. If two attributes are highly correlated in the data (\eg, glasses and age) the network learns to tie them together, leading to entanglement in the latent space. The second kind of bias is an inherent \textit{Generative Bias}, so called because it is derived from the training process rather than merely the data~\cite{yu2020inclusive}. This bias can be found near the edges of the distribution, where data exists - but in insufficient quantities. As these data points appear rarely, the GAN naturally prefers to pay the cost of their miss-classification, freeing it to devote more resources to denser regions.

We argue that this Generative Bias plays a key role in the poor performance of GANs when tasked with synthesizing such rare instances. We focus on latent-based image manipulations, and postulate that the failures of classical GAN-based editing techniques can be traced in part to this same root. We propose to tackle this bias by converting a pre-trained generator into a self-conditioned model~\cite{liu2020diverse}, where continuous conditioning labels are derived from the latent structure of the generator itself. In doing so, we build on existing linear editing methods and achieve superior quality, control and identity preservation.

Our method consists of four steps: First, we find a linear editing direction responsible for a faulty attribute which we wish to enhance. Such directions can be found with weak supervision \cite{shen2020interpreting}, in a zero-shot manner \cite{patashnik2021styleclip} or even in an unsupervised fashion \cite{shen2020closedform,harkonen2020ganspace}. Second, we build on prior observations that latent space \textit{distances} are linearly correlated with semantic attribute strengths~\cite{nitzan2021large}. We can thus label the original training set by inverting all images into the latent space of the network and calculating their latent projections on the editing direction. Third, we convert the generator and the discriminator into conditional variants, where we condition the generation on the latent-space distance labels. Finally, we fine-tune the network with samples that are drawn uniformly according to their latent-distance labels.

By following this approach, the conditional model is penalized for ignoring the edges of the distribution, drawing it towards a more uniform output distribution. Moreover, the editing directions are baked into the model's architecture, allowing for better semantic control. Our method additionally enables multi-attribute editing, even for those cases where latent traversal often fails. We validate our approach through a set of experiments and demonstrate that it leads to more robust editing and better identity preservation.

In summary, our key contributions are:
\begin{itemize}
    \item A framework for enhancing existing editing linear methods through self-conditional learning.
    \item An approach to editing that tackles shortcomings in the existing latent space by unfreezing the generator and fine-tuning it towards a fairer representation.
\end{itemize}

\ifx\arxiv\undef
Our code and models will be made public.
\fi

\input{resources/figures/fig_overview}

%% file: resources/figures/teaser.tex
\begin{figure}[!hbt]
\vspace{-8pt}
    \centering
    \setlength{\belowcaptionskip}{-2.5pt}
    \setlength{\tabcolsep}{1pt}
    {
    
    \begin{tabular}{c c c c c c}
    
        \raisebox{0.055in}{\rotatebox{90}{StyleFlow}} &
        \includegraphics[width=0.18\linewidth]{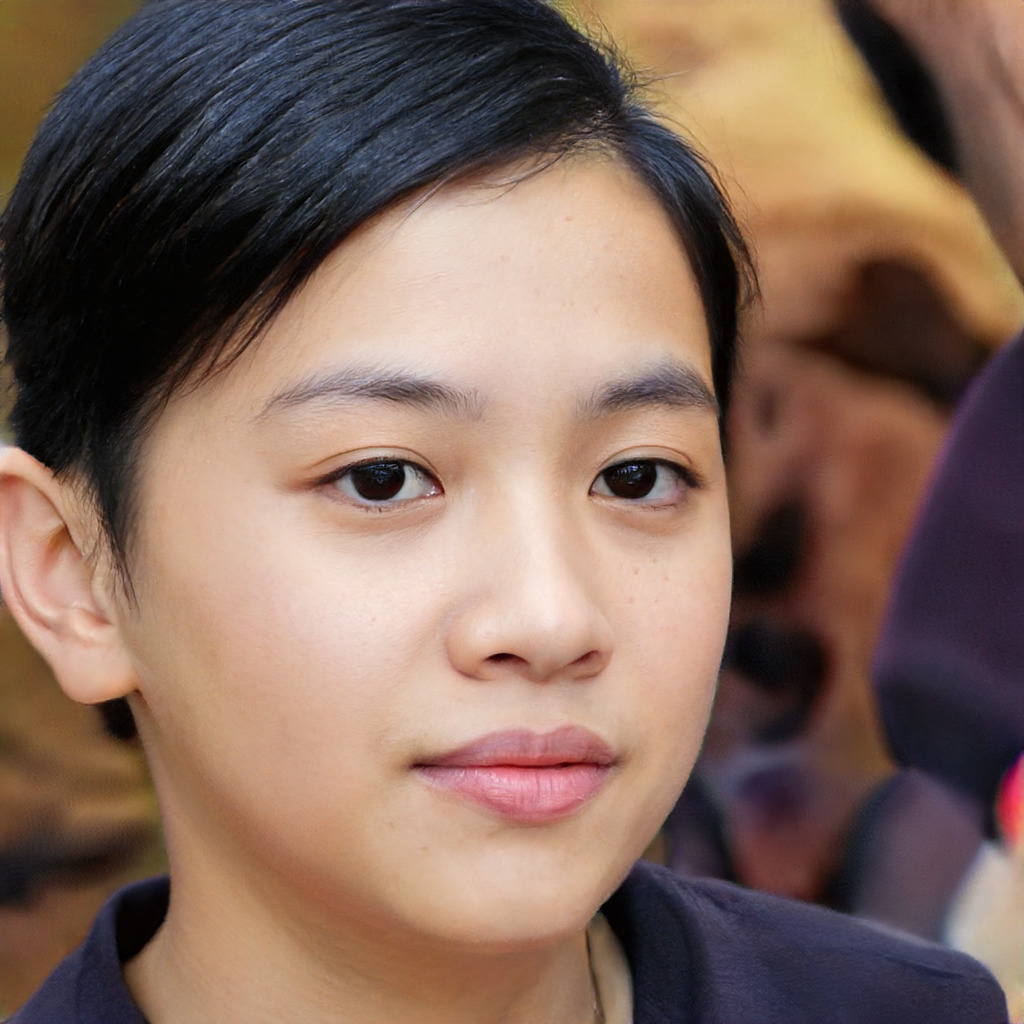} &
        \includegraphics[width=0.18\linewidth]{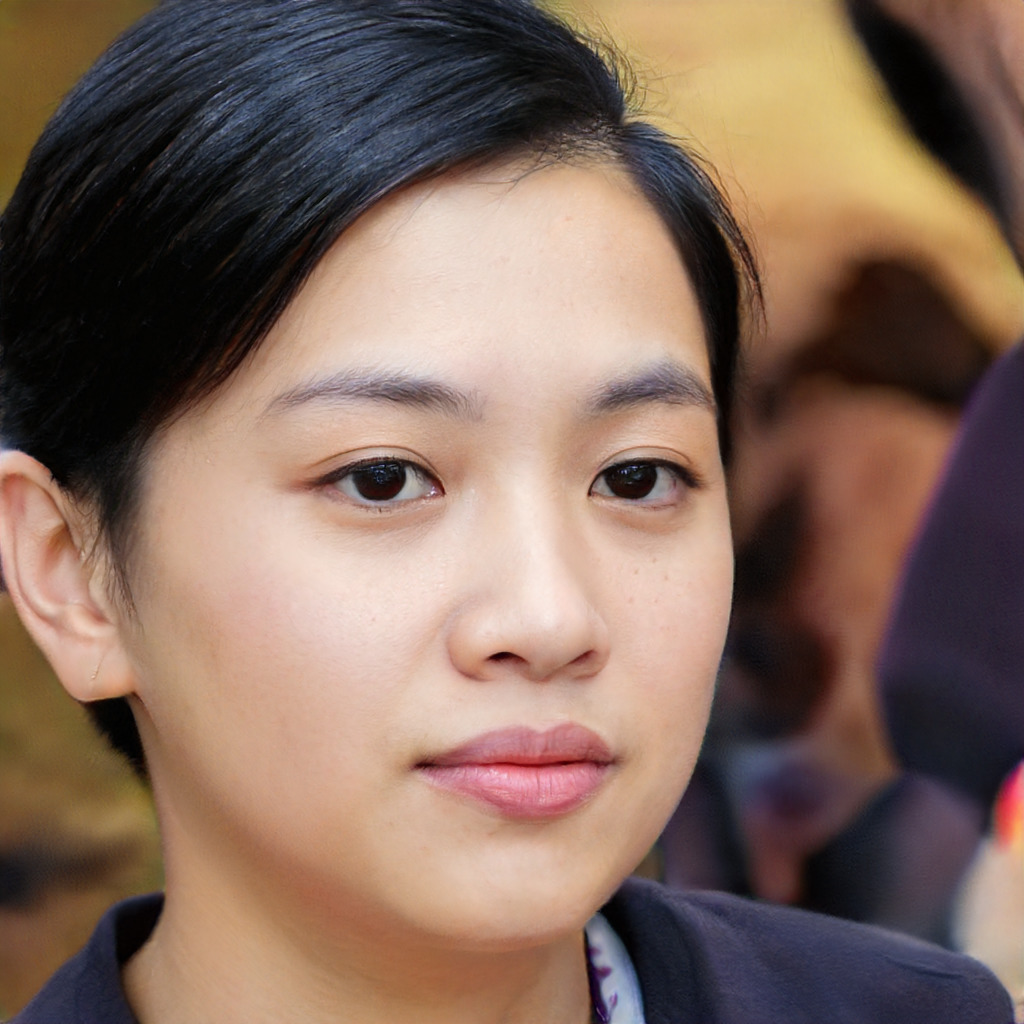} &
        \includegraphics[width=0.18\linewidth]{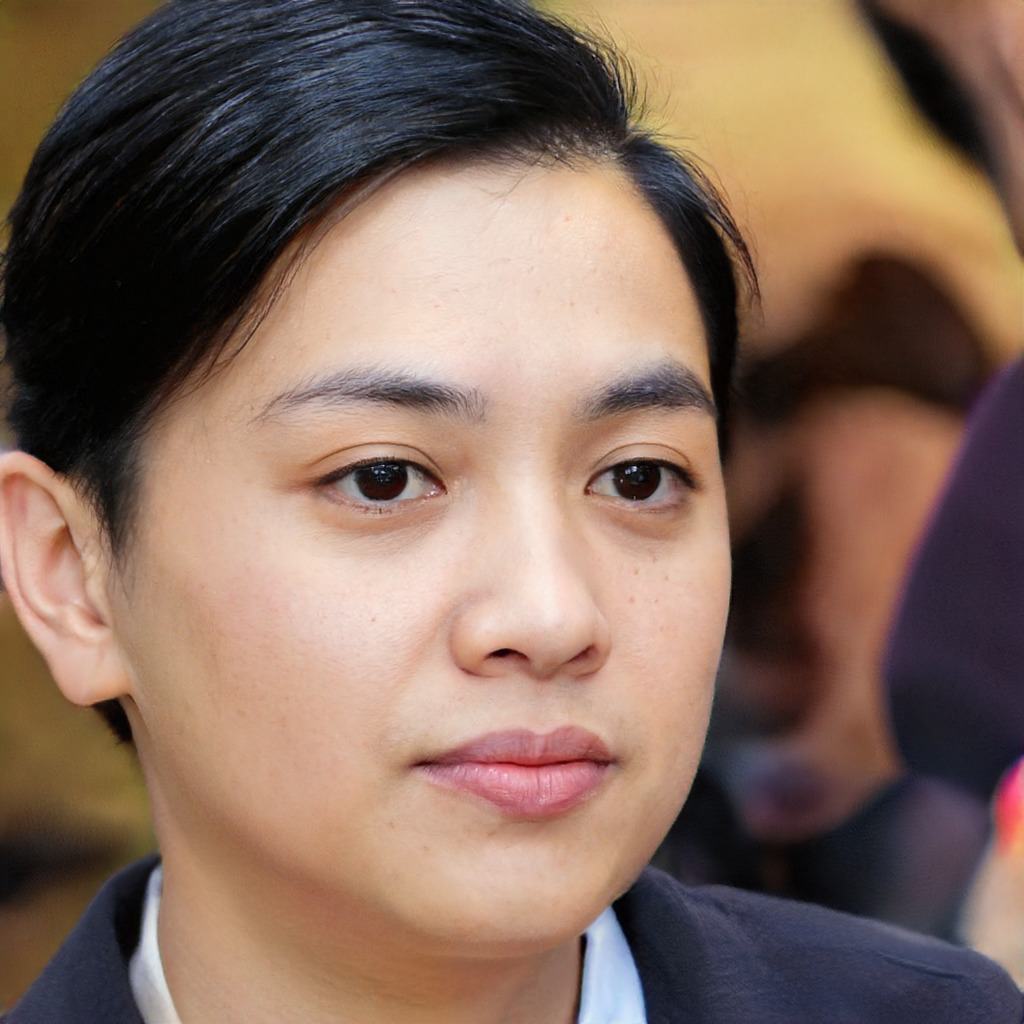} &
        \includegraphics[width=0.18\linewidth]{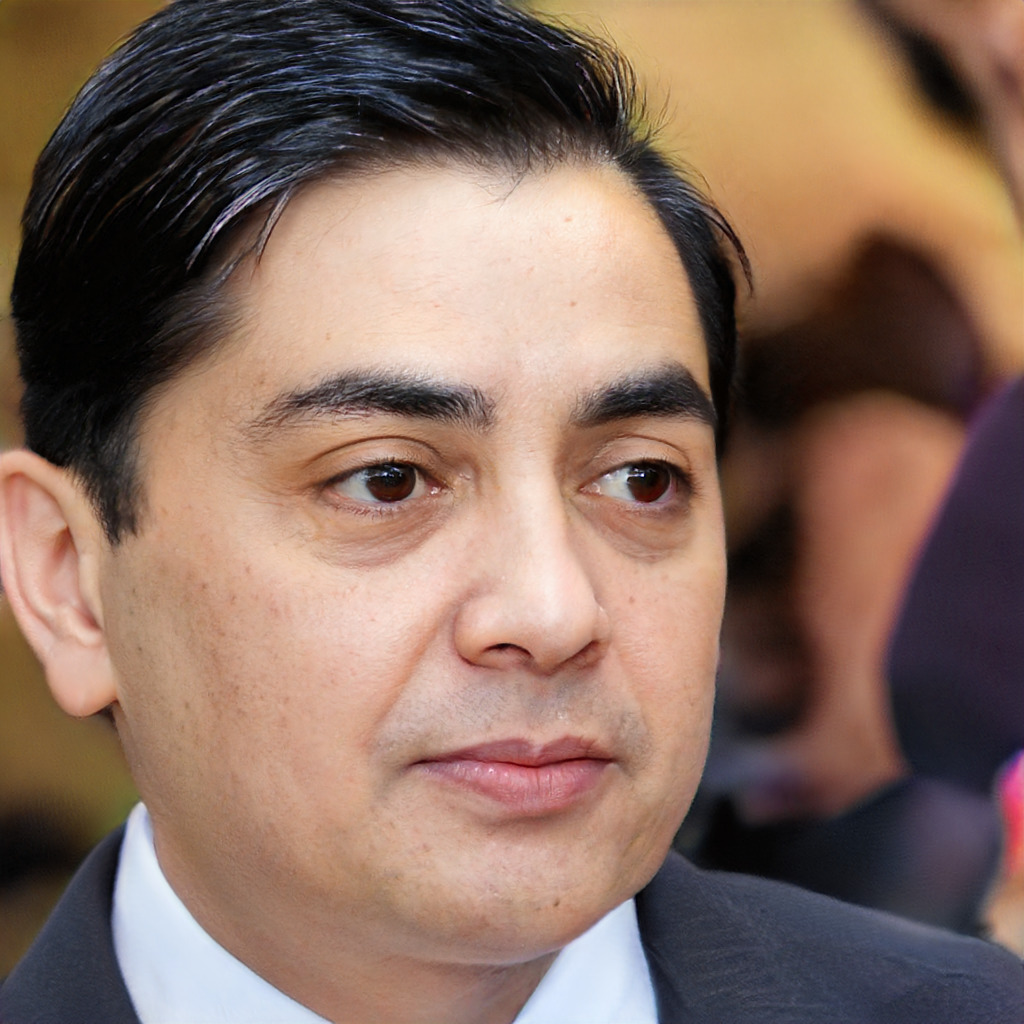} &
        \includegraphics[width=0.18\linewidth]{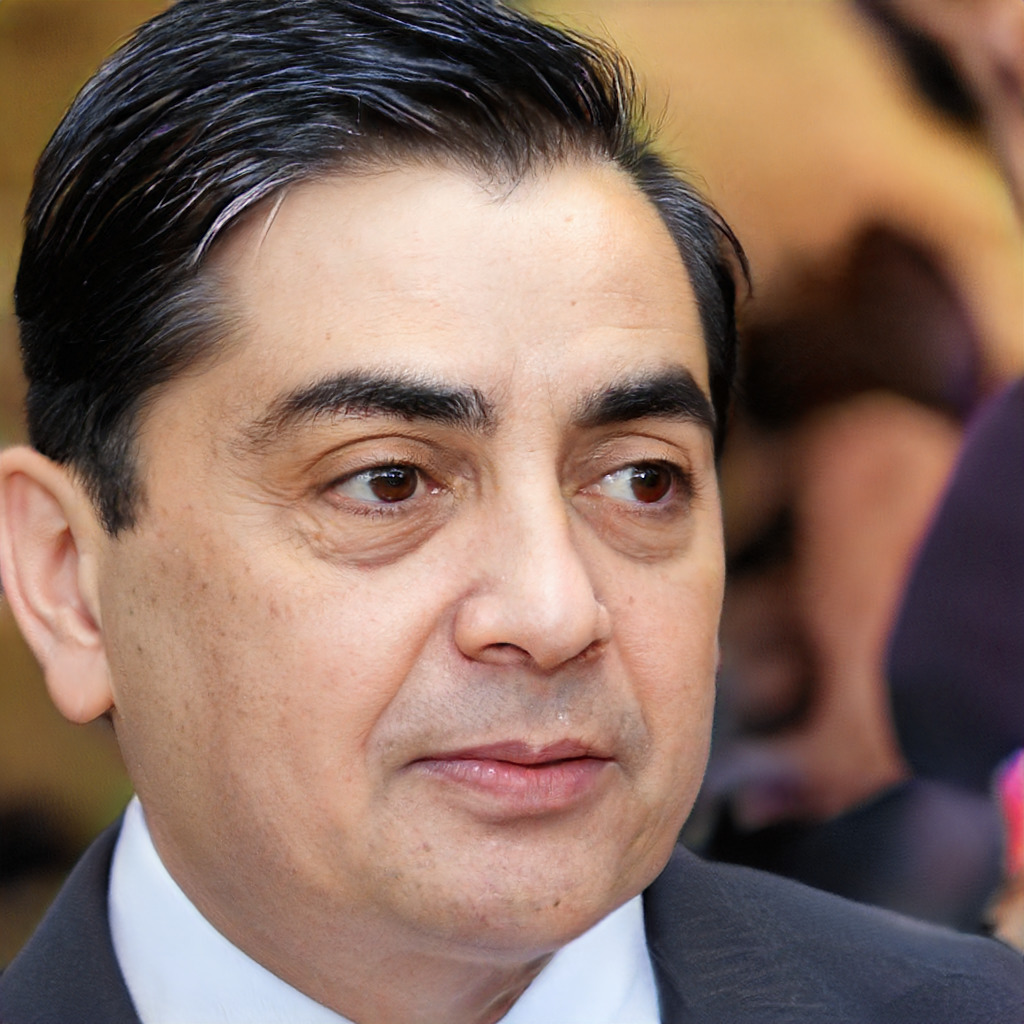} \\
        
        \raisebox{0.165in}{\rotatebox{90}{Ours}} & 
        \includegraphics[width=0.18\linewidth]{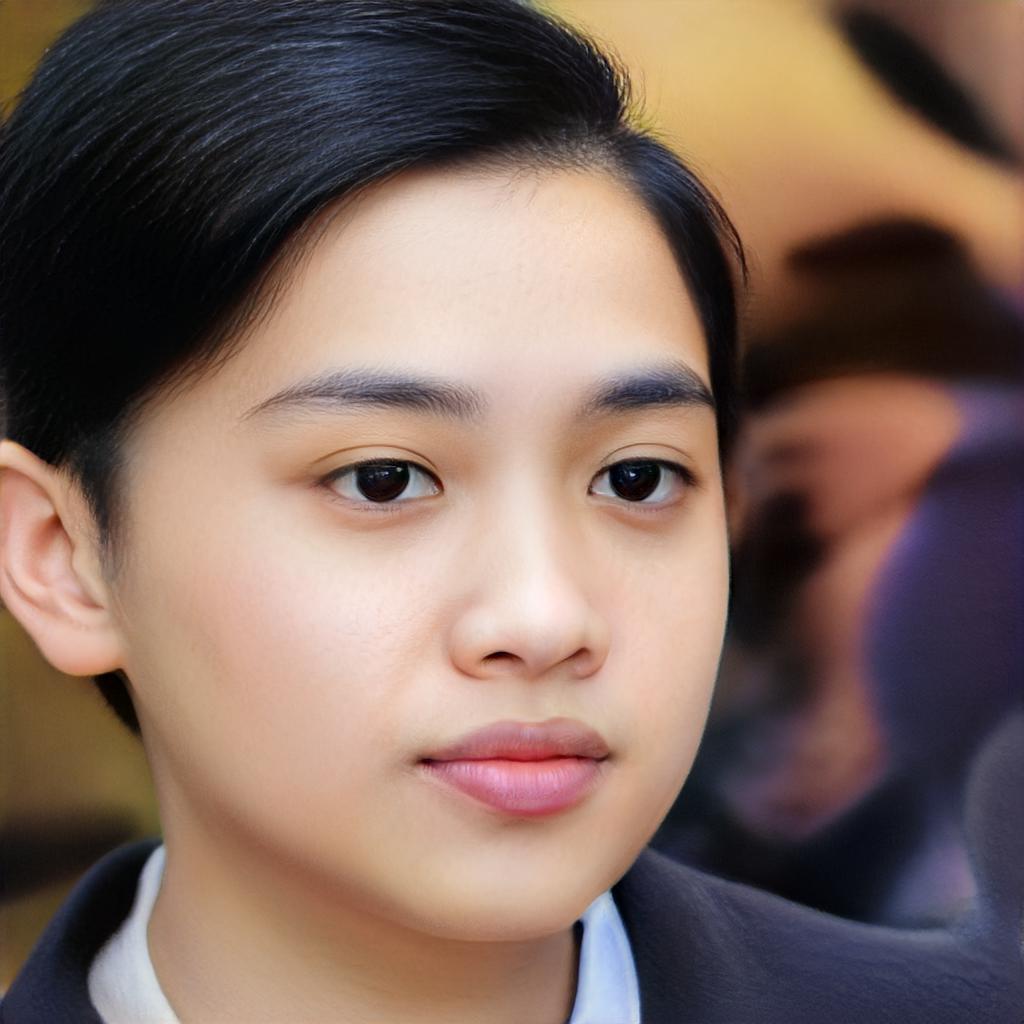} &
        \includegraphics[width=0.18\linewidth]{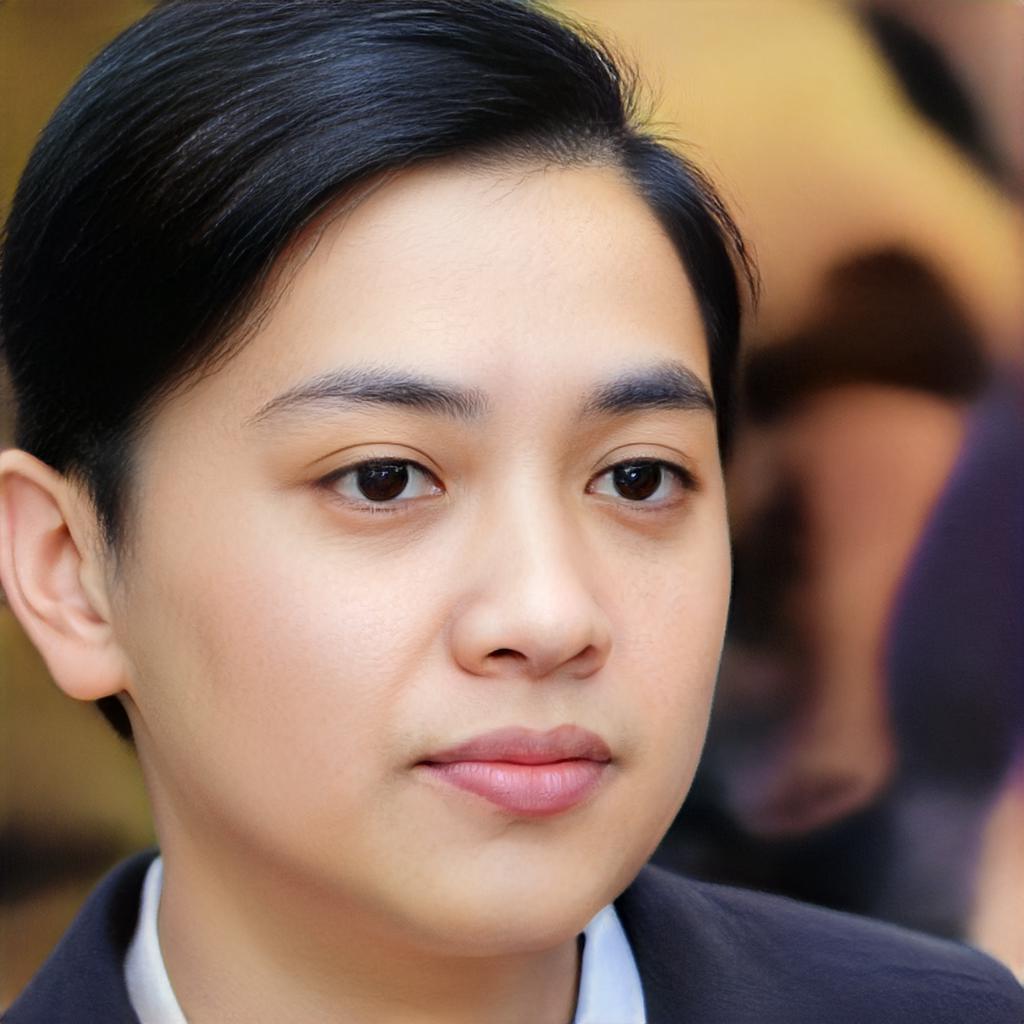} &
        \includegraphics[width=0.18\linewidth]{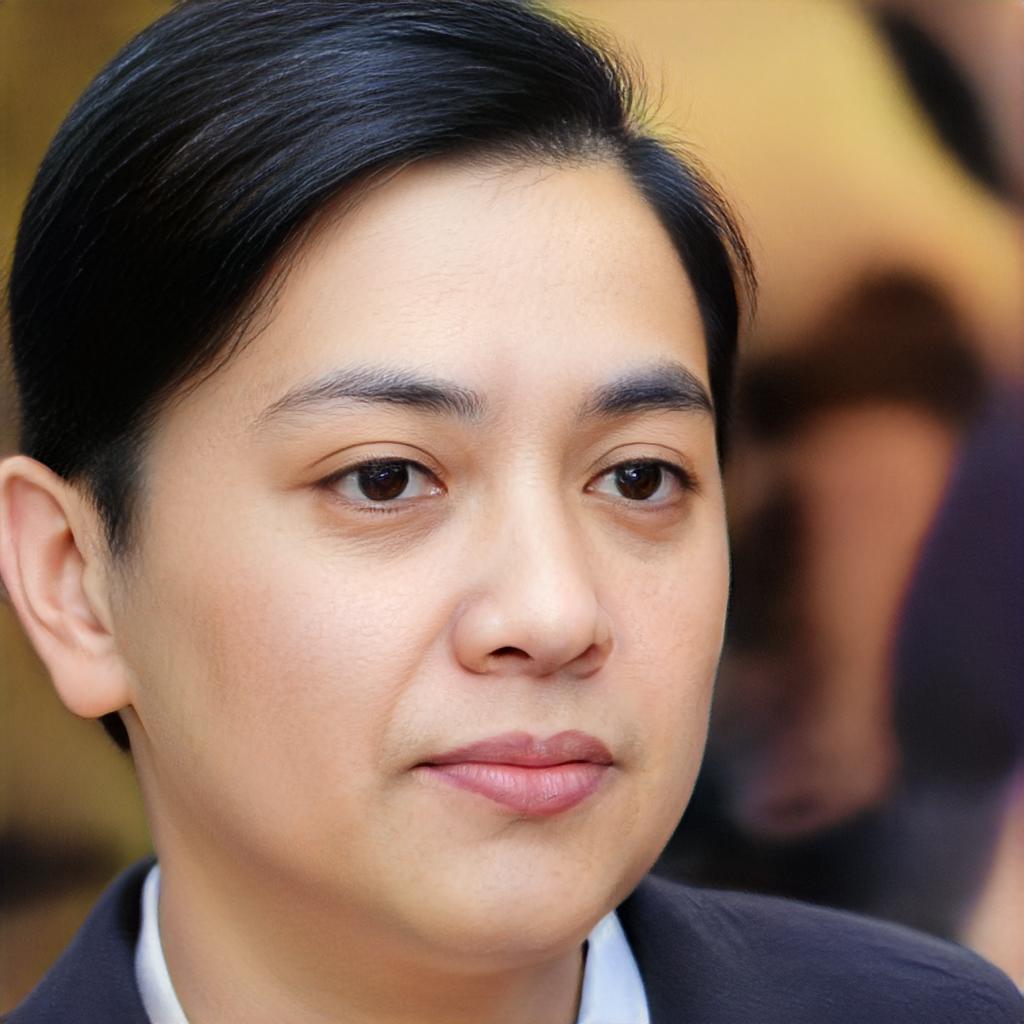} &
        \includegraphics[width=0.18\linewidth]{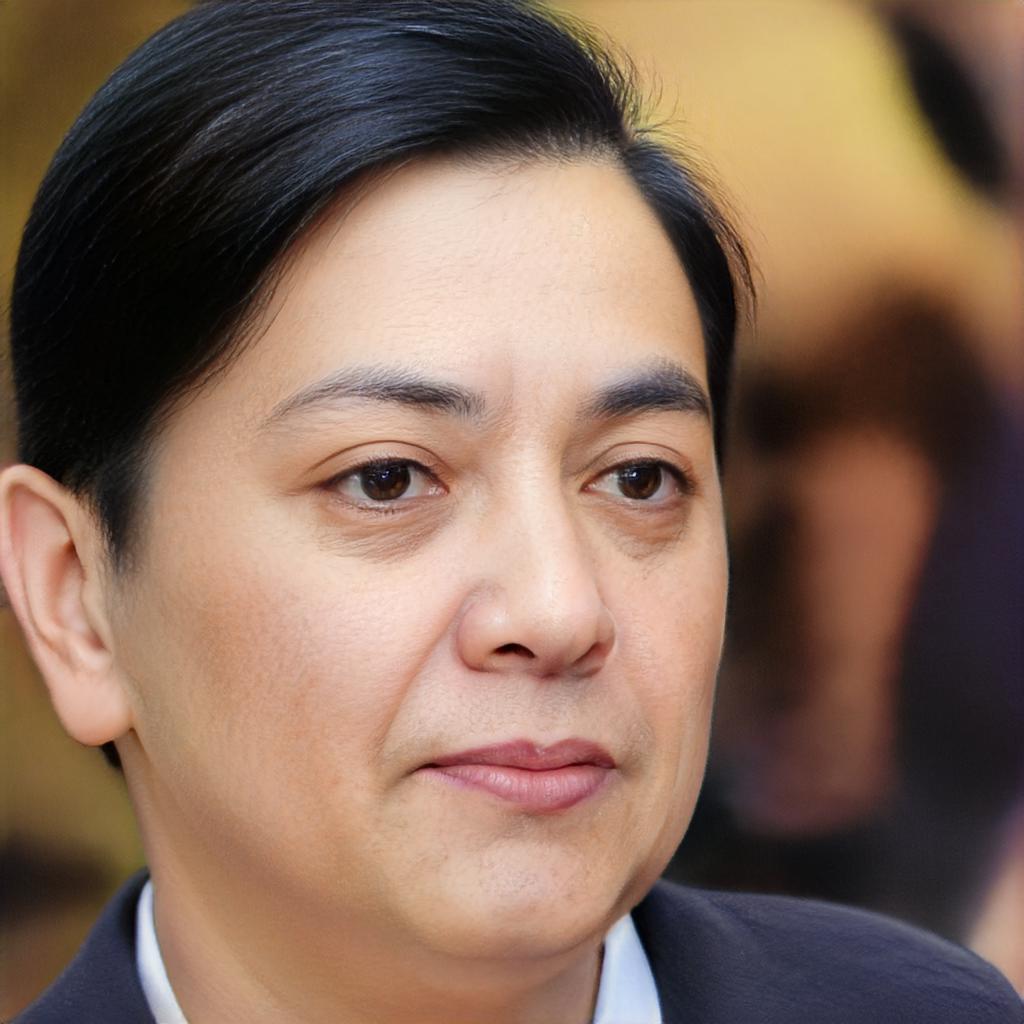} &
        \includegraphics[width=0.18\linewidth]{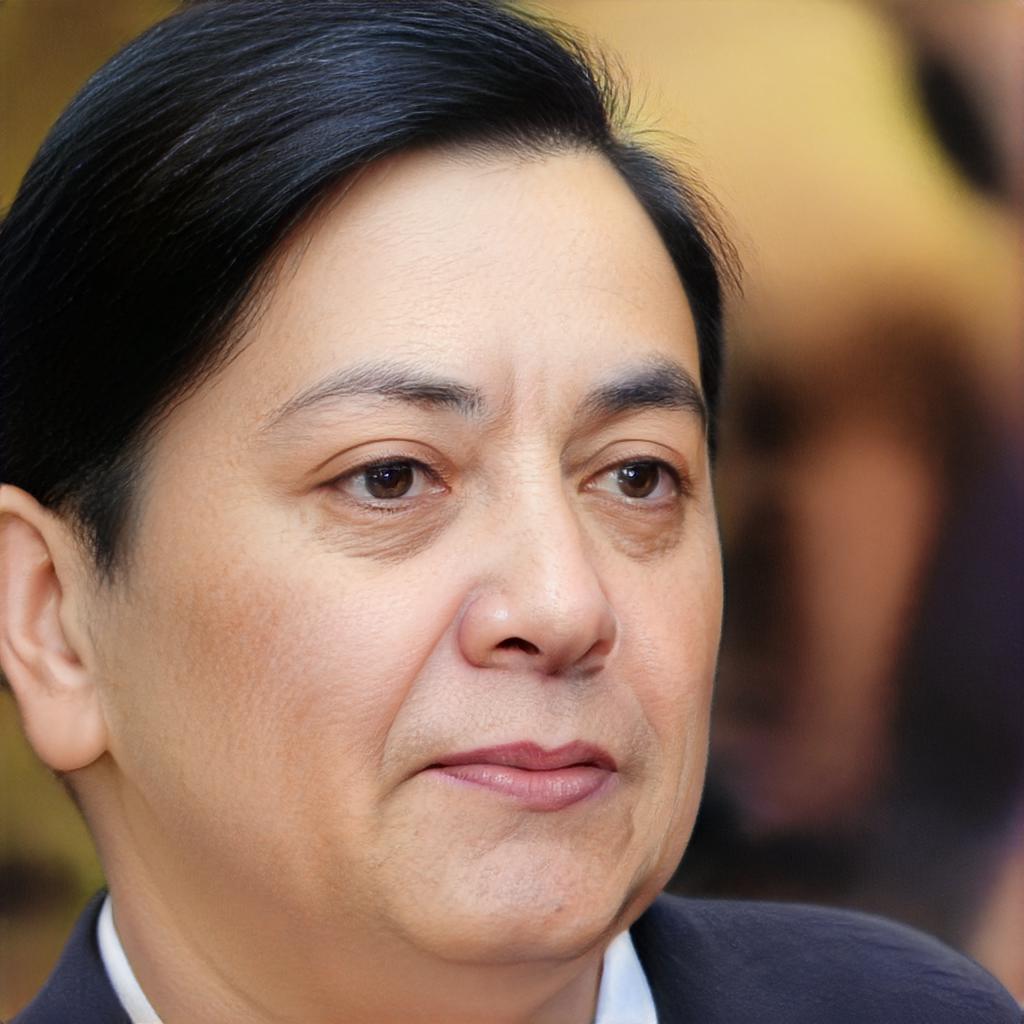} \\

        \raisebox{0.055in}{\rotatebox{90}{StyleFlow}} &
        \includegraphics[width=0.18\linewidth]{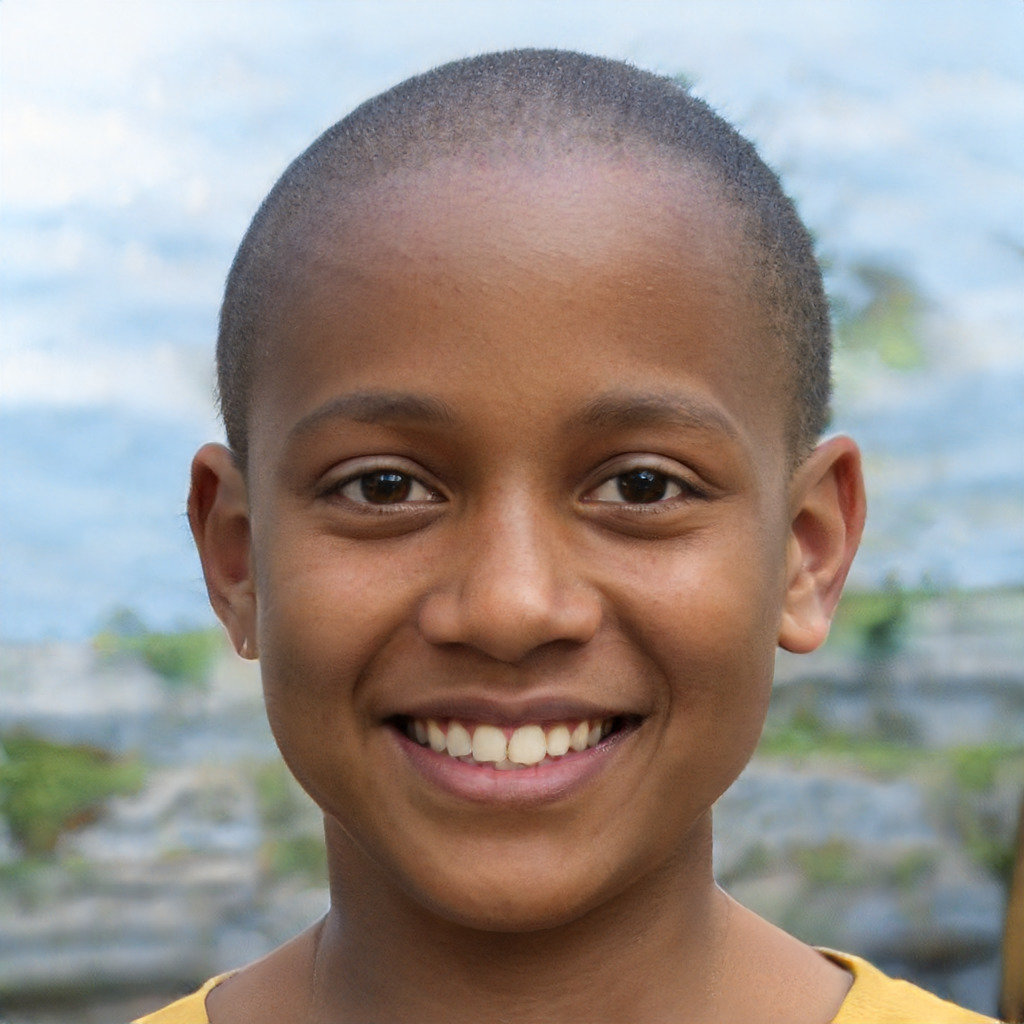} &
        \includegraphics[width=0.18\linewidth]{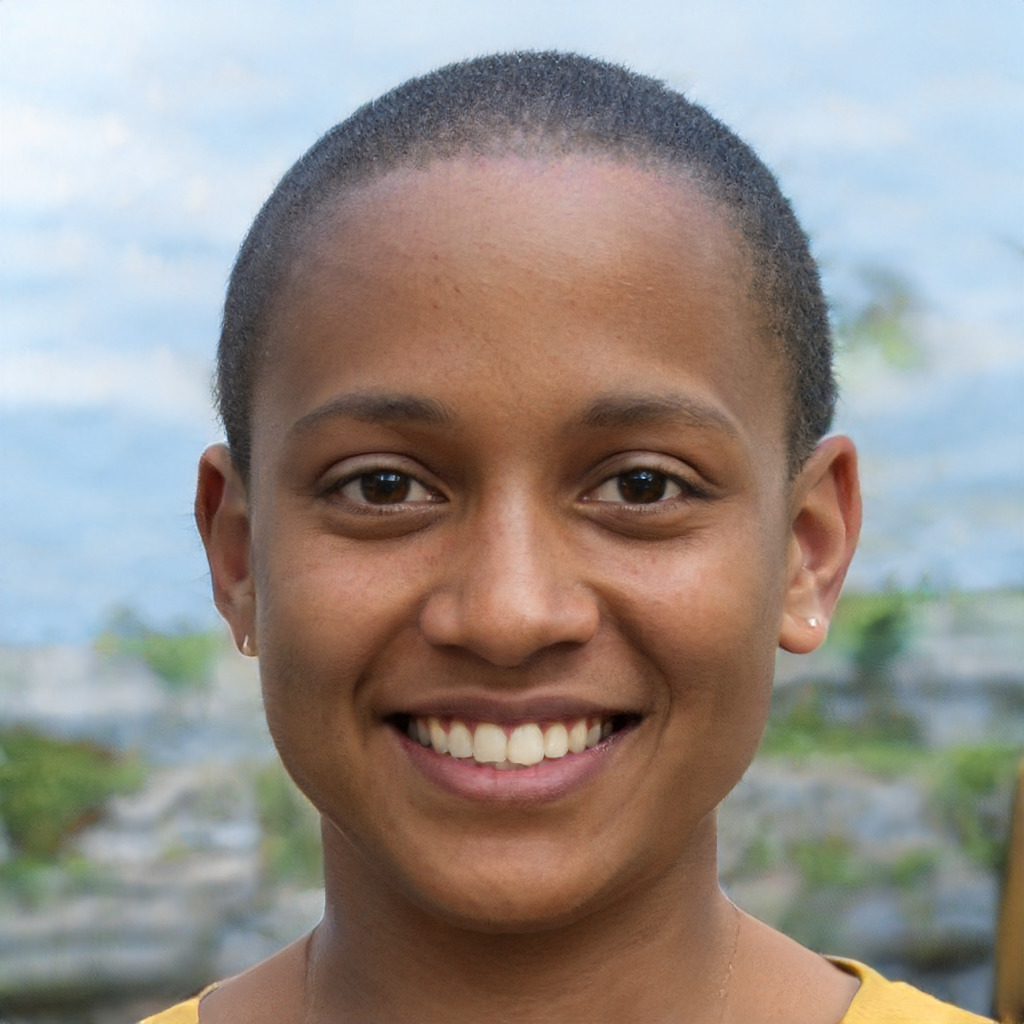} &
        \includegraphics[width=0.18\linewidth]{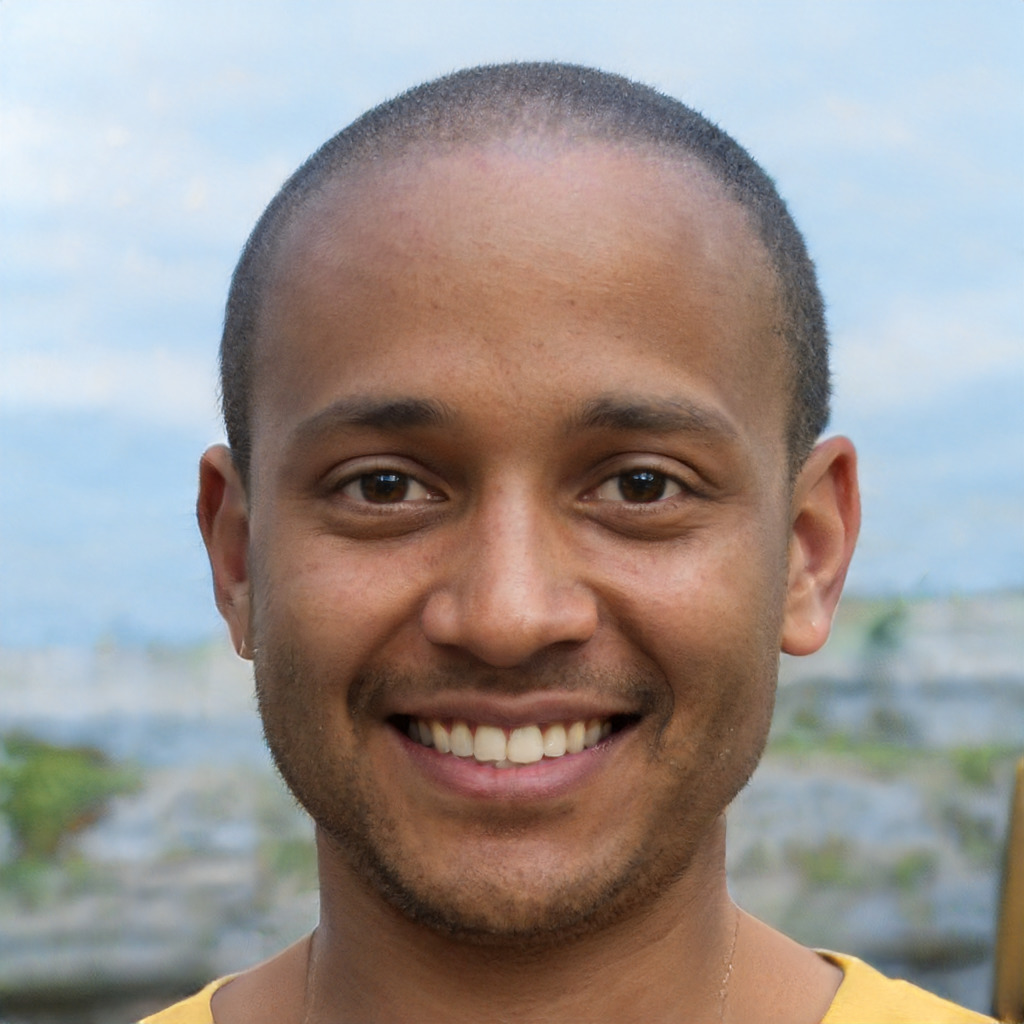} &
        \includegraphics[width=0.18\linewidth]{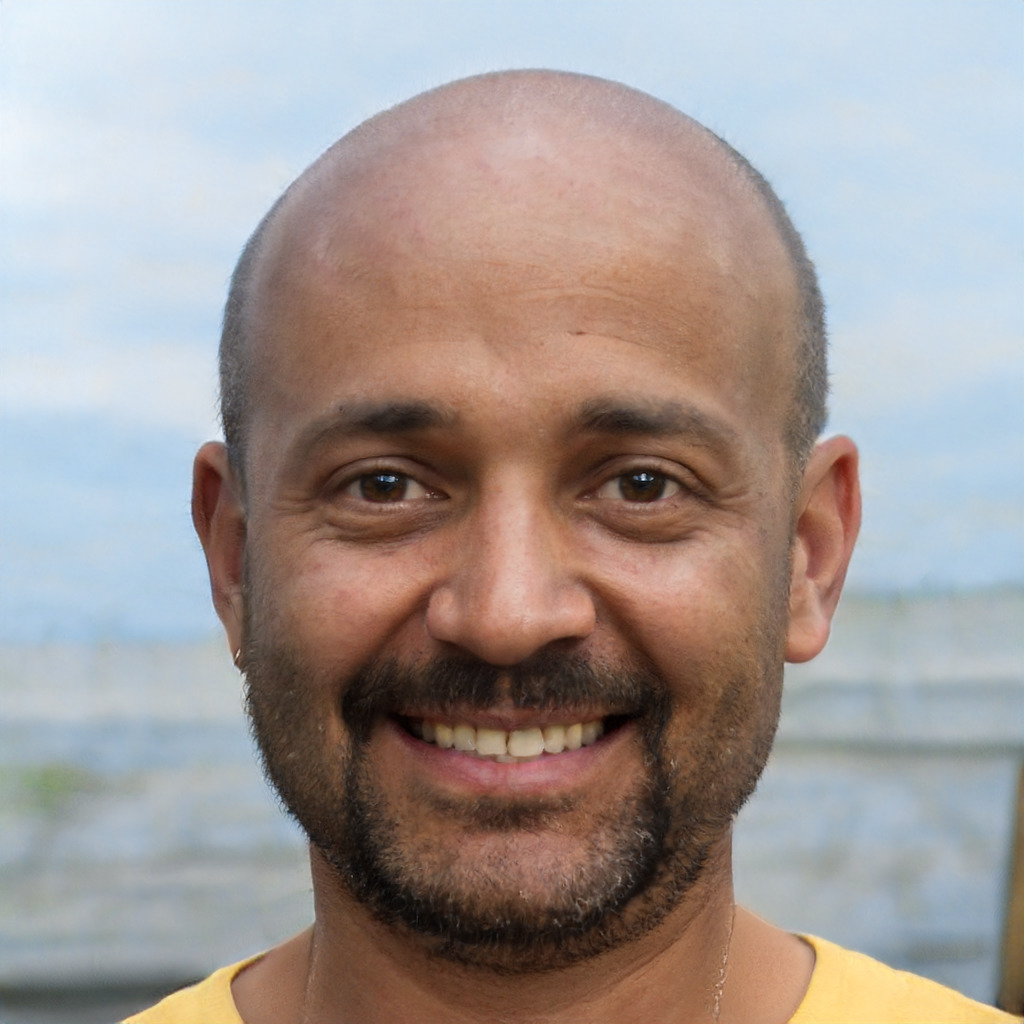} &
        \includegraphics[width=0.18\linewidth]{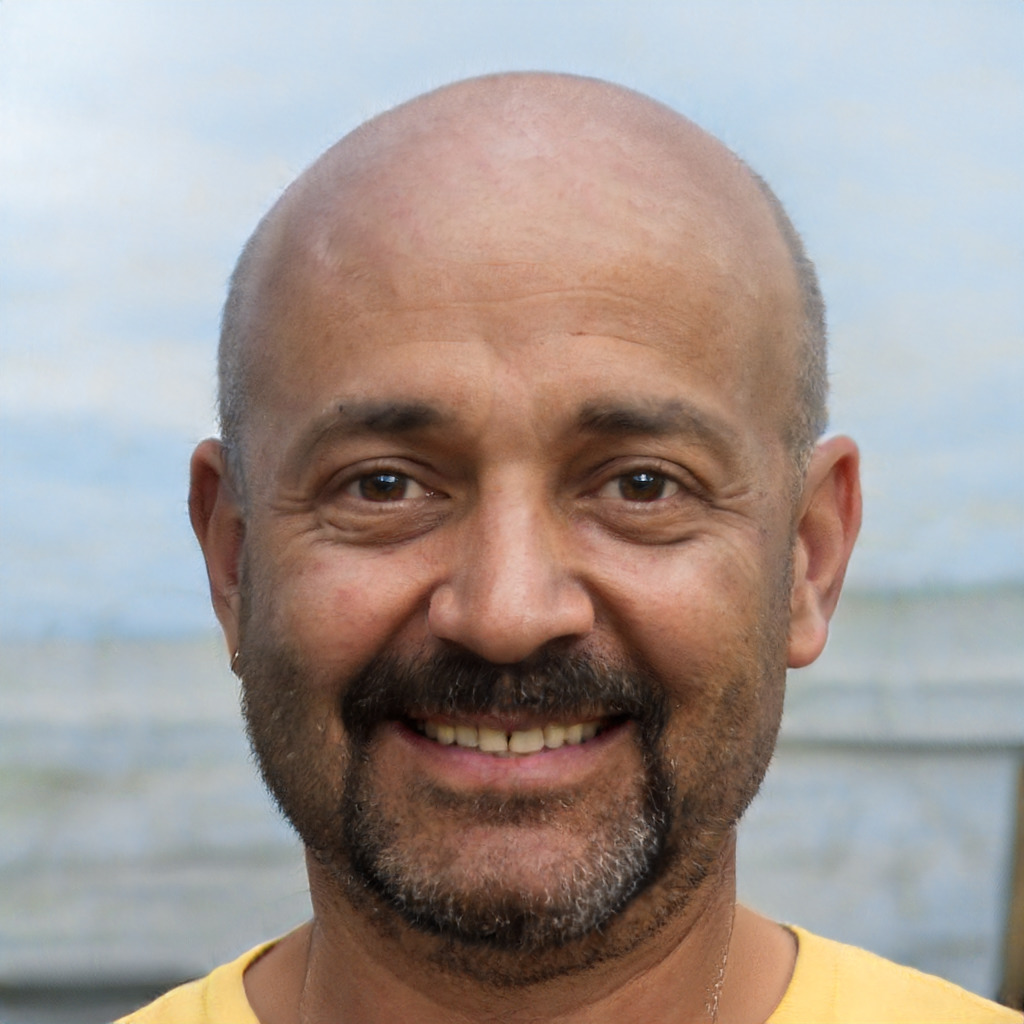} \\
        
        \raisebox{0.165in}{\rotatebox{90}{Ours}} & 
        \includegraphics[width=0.18\linewidth]{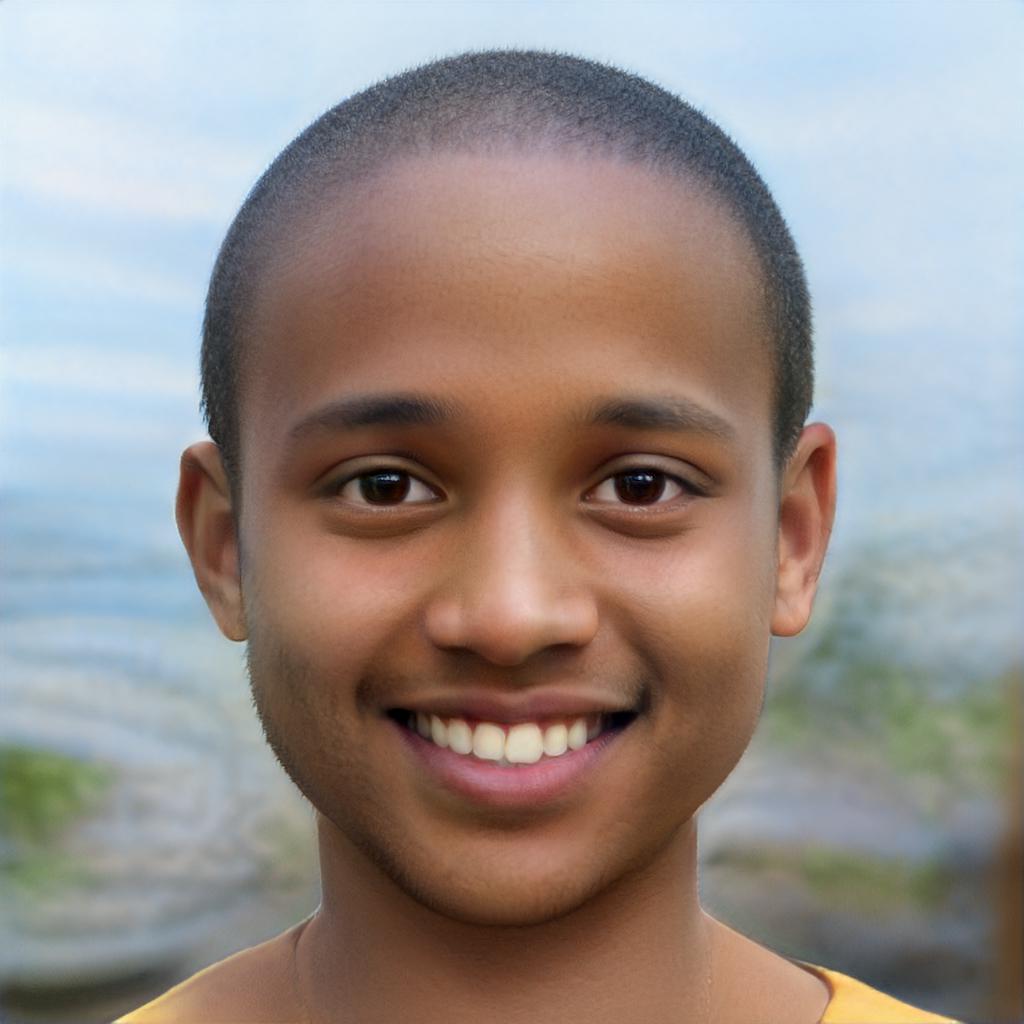} &
        \includegraphics[width=0.18\linewidth]{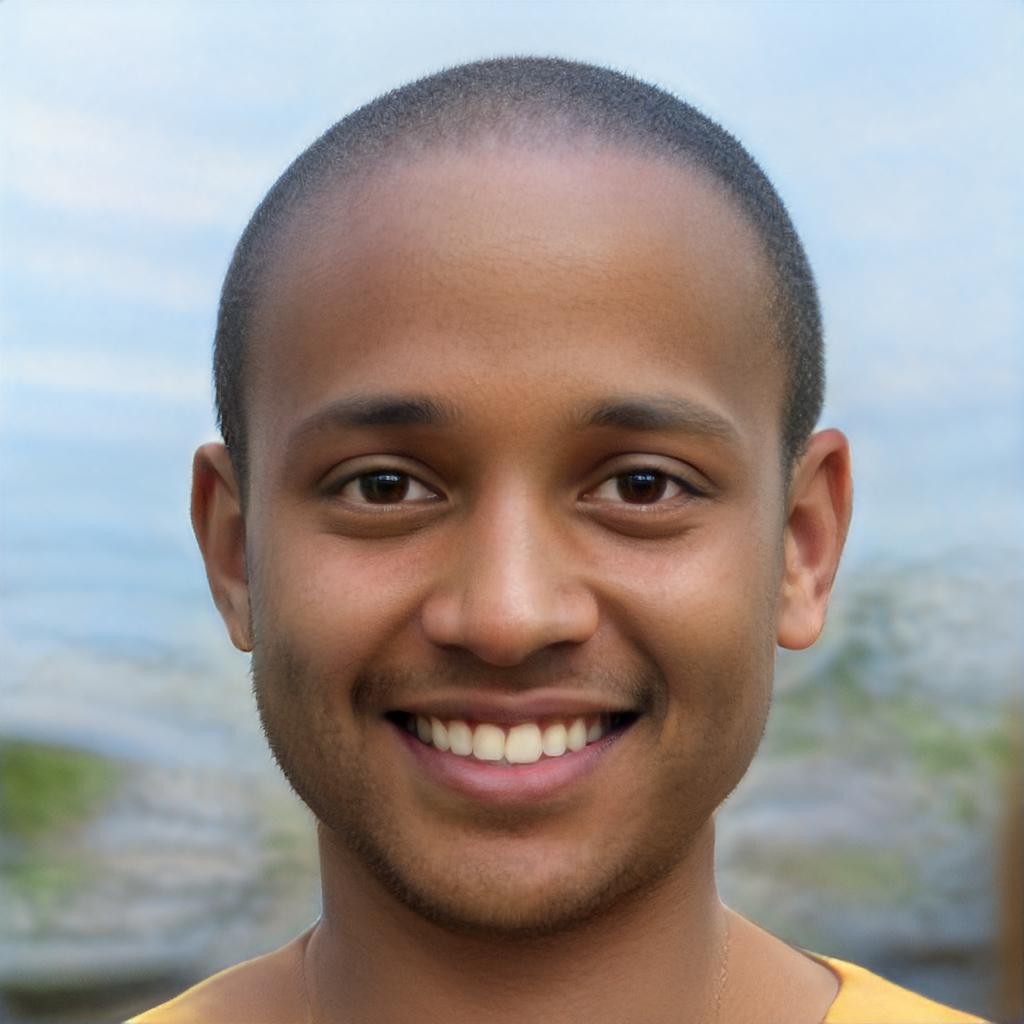} &
        \includegraphics[width=0.18\linewidth]{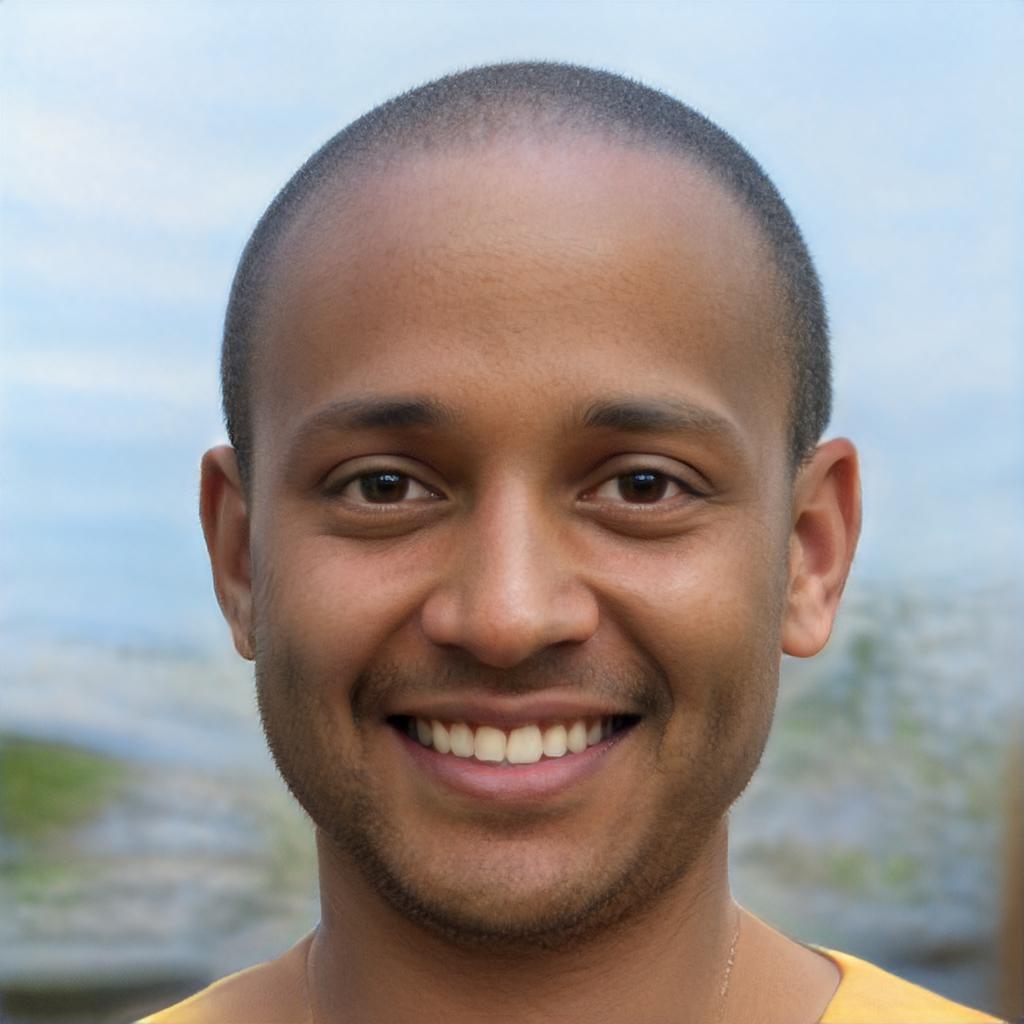} &
        \includegraphics[width=0.18\linewidth]{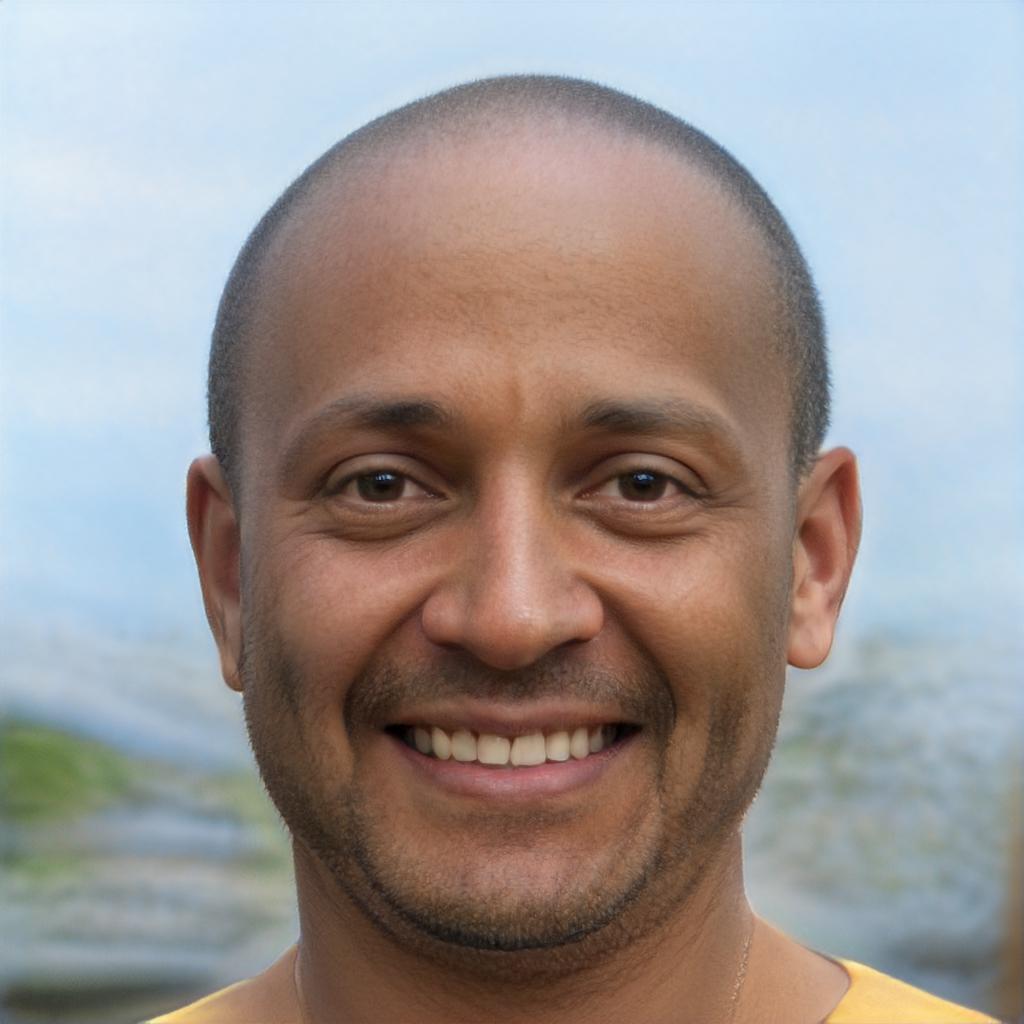} &
        \includegraphics[width=0.18\linewidth]{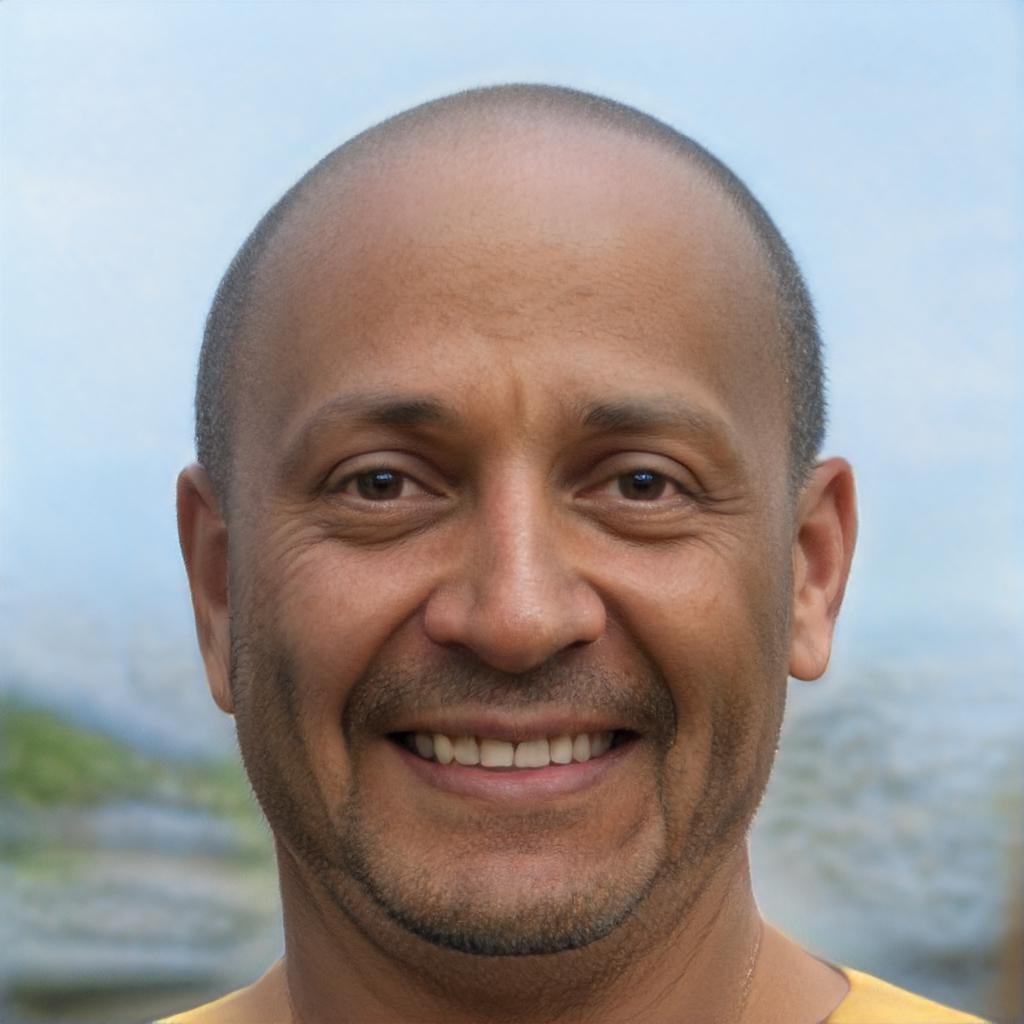}  \\
        & \multicolumn{5}{c}{$\myleftarrow$ Age $\myarrow$} 
        
    \end{tabular} 
    }
    \vspace{-6pt}
    \caption{Our method reduces the Generative Bias of neural networks, which stems from the training data \textit{and} the training scheme. Our self-conditioned training scheme induces robust editing in regions where data is scarce. For example, identity is better preserved during age manipulations, even for ethnicities underrepresented in the training data. }
    \label{fig:teaser} \vspace{-5pt}
\end{figure} 

%% file: resources/figures/fig_overview.tex
\vspace{-2pt}
\begin{figure*}[t]
\setlength{\tabcolsep}{1pt}
    \centering
    \includegraphics[width=0.9\linewidth]{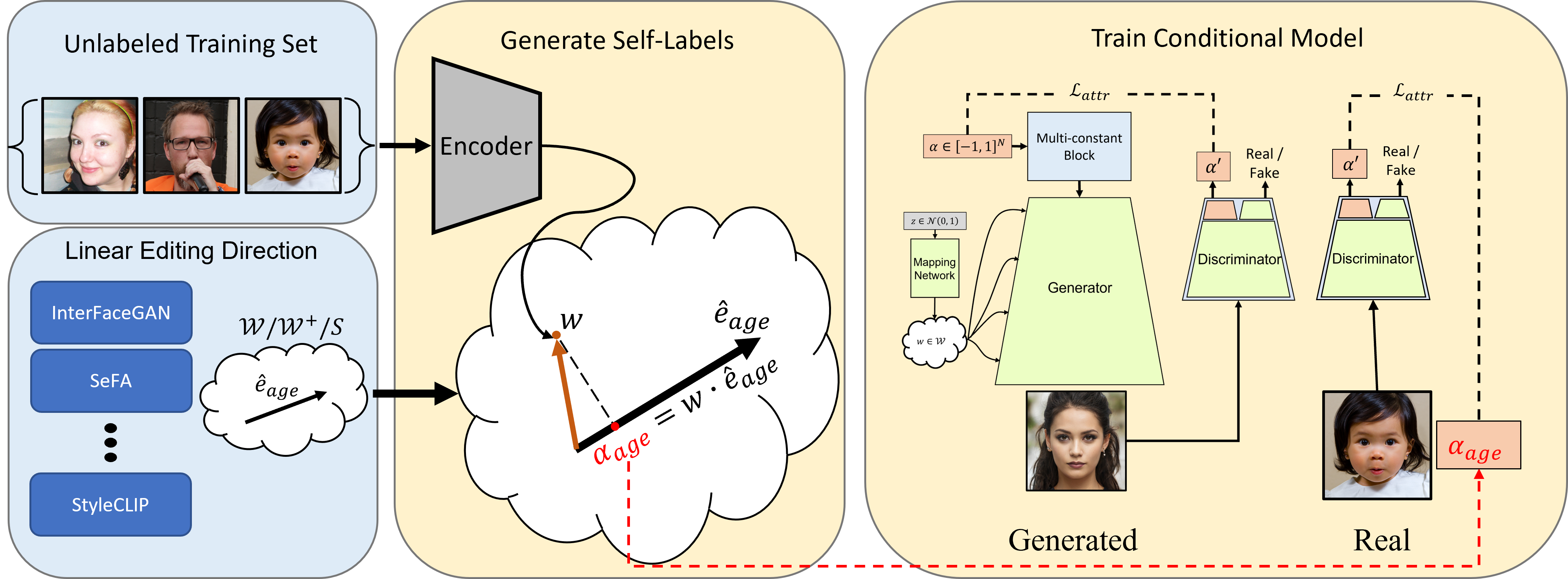}\vspace{-0pt}
    \caption{Overview of our training pipeline. Our model takes as an input a set of linear editing directions and an unlabeled training set. We label the training set by inverting each image into the latent space of the GAN, and finding its projection on each of the linear editing directions. These labels are then used to fine-tune a pre-trained StyleGAN model into a conditional version. }
    \label{fig:overview}
    \vspace{-7pt}
\end{figure*}

%% file: related.tex
\section{Related Work}
\label{sec:related}

\paragraph{Bias in Generative Networks}

Generative models have been extensively studied in the context of bias evaluation and mitigation. Early works focused on fairer generation with the aim of improving the performance of a downstream classifier \cite{xu2018fairgan,sattigeri2019fairness,grover2019bias,choi2020fair}. They employ importance re-weighting schemes \cite{grover2019bias}, leverage a reference data set \cite{choi2020fair}, or train a generator conditioned on the biased attributes \cite{xu2018fairgan,sattigeri2019fairness}. Others analyze the role of inductive biases in the generative process \cite{zhao2018bias} or mitigate biases without training through better sampling of latent codes \cite{tan2020improving}. Yu \etal \shortcite{yu2020inclusive} studied Generative Bias in greater depth. They demonstrated that when training GANs, minority modes are more likely to suffer from collapse.

In contrast to these works, we investigate the bias as a root cause for poor performance in editing. We employ a self-conditioned model and a re-sampling scheme to mitigate this bias, stave off mode collapse, and achieve greater editability for minority attributes.

\paragraph{Latent-Space Editing}

The unprecedented ability of StyleGAN to encode semantic properties within its latent space has spawned an impressive array of image manipulation methods \cite{shen2020interpreting, patashnik2021styleclip,shen2020closedform, harkonen2020ganspace,wu2021stylespace}. These methods typically aim to find linear directions in the latent space of the GAN, such that modifying a latent code along these directions will produce a change in a single semantic property of the generated image. These methods range widely in the level of supervision they require, from weak supervision in the form of binary attribute classification \cite{shen2020interpreting} through detailed 3D morphable face models \cite{Tewari2020StyleRigRS}. Others have proposed ways to identify such semantic directions in an entirely unsupervised manner \cite{shen2020closedform, harkonen2020ganspace} or in a zero-shot manner by leveraging models \cite{radford2021learning} that jointly encode image and text \cite{patashnik2021styleclip}.
Linear editing directions, however, typically suffer from entanglement or rapidly deteriorating performance when applying large changes. Recently, it has been suggested that these shortcomings can be tackled by discovering non-linear paths in the latent space~\cite{alaluf2021matter}. Typically these methods train a network to perform local manipulations on a given latent code~\cite{abdal2020styleflow,yao2021latent,li2021dystyle}. Others suggest to model the warped manifold of the GAN~\cite{Tzelepis_2021_ICCV} or traverse this manifold by finding a new local-basis at every step~\cite{choi2021escape}. While such methods have enjoyed relative success, we argue that they aim at solving the problem by attacking the symptom rather than its cause.
Instead of employing complex methods to find non-linear directions in the space of a pre-trained GAN, we propose to re-train the model and alleviate the bias which gives rise to these flaws.

\paragraph{Generator Unfreezing}
Typically, frameworks which utilize a pre-trained GAN for downstream tasks elect to keep the GAN frozen. Doing so is expected to bring stability and ensure higher quality. A recent line of works, however, challenges these assumption and demonstrates that some tasks can be better handled by modifying the generator itself~\cite{bojanowski2018optimizing}.

The intuition is that brief fine-tuning sessions, whether adversarial~\cite{wu2021stylealign} or non adversarial~\cite{gal2021stylegannada}, tend to keep the models well aligned and preserve most of the structure of the latent space. While this property is typically used for tasks that cannot be accomplished with the original generator, such as out-of-domain editing \cite{pinkney2020resolution, gal2021stylegannada}, it has been shown that modifying the generator can also dramatically improve inversions \cite{pan2020exploiting,roich2021pivotal,alaluf2021hyperstyle}, create new editing directions \cite{cherepkov2021navigating}, improve temporal consistency~\cite{tzaban2022stitch} or enable a user to re-write the synthesis rules of the network~\cite{bau2020rewriting}.

In contrast to these works, we propose that generator fine-tuning can also be used to improve on existing editing approaches. By converting implicitly learned latent directions into explicit conditioning codes, we force the generator to contend with minority attributes, reduce the Generative Bias, and enable better control of the modified attributes under more significant modifications.

%% file: method.tex
\section{Method}
\label{sec:method}

To tackle the Generative Bias and enable more robust editing, we propose a scheme for converting a pre-trained generator and an existing linear editing direction into a self-conditioned model. We do so by tackling three aspects of the training flow: First, we generate self-labeled data using distances in the latent-space of the generator. Second, we modify the generator's architecture in order to explicitly represent a multitude of rare modalities. Third, we employ a re-sampled fine-tuning session with broadened discriminator supervision. These modifications are outlined below. An overview of the different steps is provided in \cref{fig:overview}.

\ifx\arxiv\undefined
\subsection{Latent-Space Distances as Self-Conditional Labels}\label{sec:latent_labels}
\else
\subsection{Latent-Distances as Self-Conditional Labels}\label{sec:latent_labels}
\fi

In a recent work, Nitzan~\etal~\shortcite{nitzan2021large} demonstrated that the latent space distance of an image from a semantic editing hyperplane is linearly correlated with the magnitude of the semantic attribute in the image. Drawing on their insight, we propose to use these distances as labels for the original dataset. These labels could then be used to fine-tune the generator into an unbiased \textit{conditional} model.

To self-label the data, we begin by extracting an editing hyperplane for each property that we want to de-bias. The typical method for identifying such planes is through the use of InterfaceGAN \cite{shen2020interpreting}. However, any method which extracts linear editing directions in any of the GAN's multiple latent spaces is equally suitable. As we shall later demonstrate, our method can work equally well with directions extracted by a wide array of methods in multiple latent spaces, including: StyleCLIP \cite{patashnik2021styleclip} directions in $\mathcal{S}$~\cite{wu2021stylespace}, InterFaceGAN~\cite{shen2020interpreting} directions in \w, and SeFA~\cite{shen2020closedform} directions in \wplus.

Armed with these editing directions, we next turn to labeling our training data. We invert all training set images into the latent space of the network using a pre-trained e4e \cite{tov2021designing} model. We then calculate, for each image, the latent space distance between its inverted code and the editing hyperplane. We find the minimal and maximal distances in the set and re-scale all distance labels such that the range of possible values is in $[-1, 1]$. We use these distances as a set of continuous labels for each image.

In the case of methods which produce editing directions without an intercept, we arbitrarily set the intercept to $0$ (\ie we use the projection of the latent code on the editing direction). As distances are then normalized, this choice bears no effect on the results.

\vspace{-2pt}
\subsection{Conditional Multi-Constants}

\input{resources/figures/fig_arch}

Using the self-labeled data, we turn to converting our generator into a conditional model.
Our goal is to enhance control and improve representation of rare dataset modalities. We thus adopt the multi-constant approach proposed by Sendik \etal \cite{sendik2020unsupervised}. 

Multi-constant models expand StyleGAN's initial learned constant to a set of constants, each of which is expected to control a different modality present in the data. At inference time, a constant can be chosen either conditional on a label, or by augmenting StyleGAN's mapping network to output a weight vector which denotes an importance-weighting score for each constant. One can then choose the most dominant constant, or simply mix them together in proportion to their weights. Sendik \etal \shortcite{sendik2020unsupervised} demonstrated that such a setup allows the model to better encode the unique attributes relevant to each modality, while allowing the rest of the network to share information from all modalities, leading to higher quality synthesis. In our case, the information sharing aspect is crucial as we are interested in better treatment of rare modalities, which may not contain sufficient samples to train a high quality GAN.

In practice, we modify the multi-constant approach in the following way:
In addition to StyleGAN's original constant $C$, we add two new constants for each attribute we wish to control: $c_i^+, c_i^-$. The intuition here is that many attributes may not be symmetric around the average image (\eg young faces differ significantly from old faces) and by introducing a constant for each editing direction, we enable the network to model this discrepancy. When synthesizing a new image, we provide the network with a score for each attribute $\alpha_i \in [-1, 1]$. We then use a 'mixed' constant that contains the information about our desired attribute strengths and directions:
\begin{equation}
    C' = C + \Sigma_i |\alpha_i| * c_i^{sign\left(\alpha_i\right)} ~.
\end{equation}
In contrast to Sendik \etal \shortcite{sendik2020unsupervised}, our method uses additive constants for each attribute. This allows us to learn a model which can simultaneously control multiple attributes, without having to learn a constant for each attribute combination. Moreover, this motivates the constants to focus only the information which differs between the modalities, a simpler task which can be tackled with fewer data.

Typically, conditional StyleGAN models inject the conditioning code through the mapping network~\cite{Karras2020ada}. Adding such conditioning to a pre-trained model through fine-tuning is non trivial, owing to the increased dimensionality of the latent codes. In such a scenario, the mapping network has to be entirely replaced, or part of the latent code has to be re-purposed for the conditioning. Additive constants, meanwhile, can be used to modulate the existing network without any re-learning of the core generative path. In \cref{sec:experiments} we investigate the option of replacing the mapping network with a conditional one and find that this loss of information leads to poor editing control and demeaned performance.

On the discriminator front, we employ the original StyleGAN2 \cite{karras2020analyzing} discriminator and augment it with a new prediction head tasked with regressing the mixture of modalities, $\alpha_i$, used in the image synthesis. In order to help the network encode differences between asymmetric modalities such as adding hats or glasses (where the ideal negative direction constant may simply be a tensor of zeros), we once again separate positive and negative editing directions. We do so by defining a 3-entry score vector for each attribute we want to control: $[|\alpha_i|, 1 - |\alpha_i|, 0]$ for $\alpha_i < 0$ and $[0, 1 - |\alpha_i|, |\alpha_i|]$ for $\alpha_i > 0$. We employ a soft cross-entropy loss between the discriminator's predictions and these soft labels. For real images, we substitute $\alpha_i$ with the normalized distance-labels for each attribute in the image.

By combining these modifications, we create a conditional GAN architecture where conditioning codes, $\alpha_i$, are injected through linear interpolation of learned constants. Our discriminator is further incentivized to pay attention to rarer modalities, making mode collapse less likely. The full network architecture is outlined in \cref{fig:arch}.

\subsection{Re-sampled Tuning}
With the image labels and architecture modifications at hand, we turn to fine-tuning the GAN. We continue training using the original, now labeled, dataset. However, rather than sampling a random image at every iteration, we uniformly sample a random score for each attribute and draw the nearest-neighbor image in attribute space. By doing so, we further ensure that the discriminator sees an unbiased distribution. 

\subsection{Image Inversion}
An essential requirement for latent space editing methods is the ability to modify real images, inverted into the latent space of the GAN~\cite{abdal2019image2stylegan,alaluf2021restyle,richardson2020encoding}.
As our method extracts particular semantics out of the latent space and into a set of explicit conditional labels, we find that codes in the original model are no longer tied to the same identities in the new model. We are thus unable to employ off-the-shelf encoders for use with our model. However, such models can be easily adapted to our conditional setting. We do so by fine-tuning a pre-trained e4e \cite{tov2021designing} model. The encoder's function is unchanged. It receives an image, and produces a latent code in \wplus. This code is then fed into our conditional generator, along with a set of latent-space distance labels derived from the same input image in the original source model. These distances can be efficiently approximated by simply passing the image through the discriminator's mode prediction head. The encoder is fine-tuned using the same optimization goal and loss terms as in the original e4e model.
A similar approach can be used to integrate our model with PTI~\cite{roich2021pivotal}, enabling accurate and highly editable reconstructions of real images.

\subsection{Training Details}
We train our models using the StyleGAN2-Pytorch implementation. We use the official FFHQ~\cite{karras2019style} 1024x1024 and AFHQ-Cat~\cite{choi2020starganv2} 512x512 checkpoints. FFHQ models were fine-tuned using $30k$ iterations and a batch size of $4$. The AFHQ model was fine tuned with $30k$ iterations and a batch size of $8$. e4e models were fine-tuned using $50k$ iterations and a batch size of $2$.
When integrating with PTI, we use the latent provided by our fine-tuned e4e model as a pivot. The generator optimization is performed over $350$ iterations.
Learning rates, relative loss weights and all other parameters were left unmodified from the original models.

%% file: resources/figures/fig_arch.tex
\vspace{-2pt}
\begin{figure}[t]
\setlength{\tabcolsep}{1pt}
    \centering
    \begin{tabular}{c}
    \includegraphics[width=0.90\linewidth]{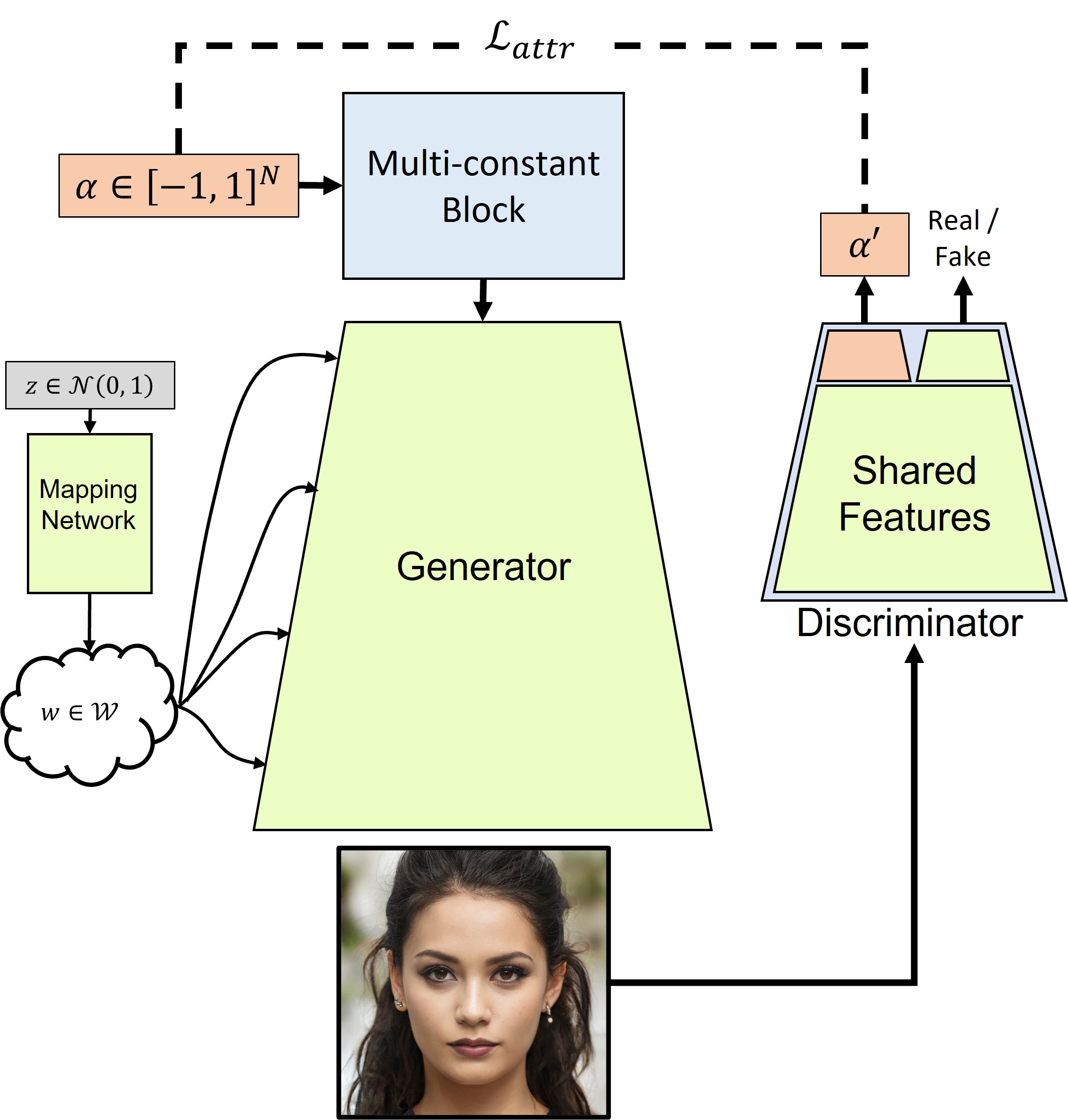}\vspace{2pt} \\ \midrule
    \includegraphics[width=0.8\linewidth]{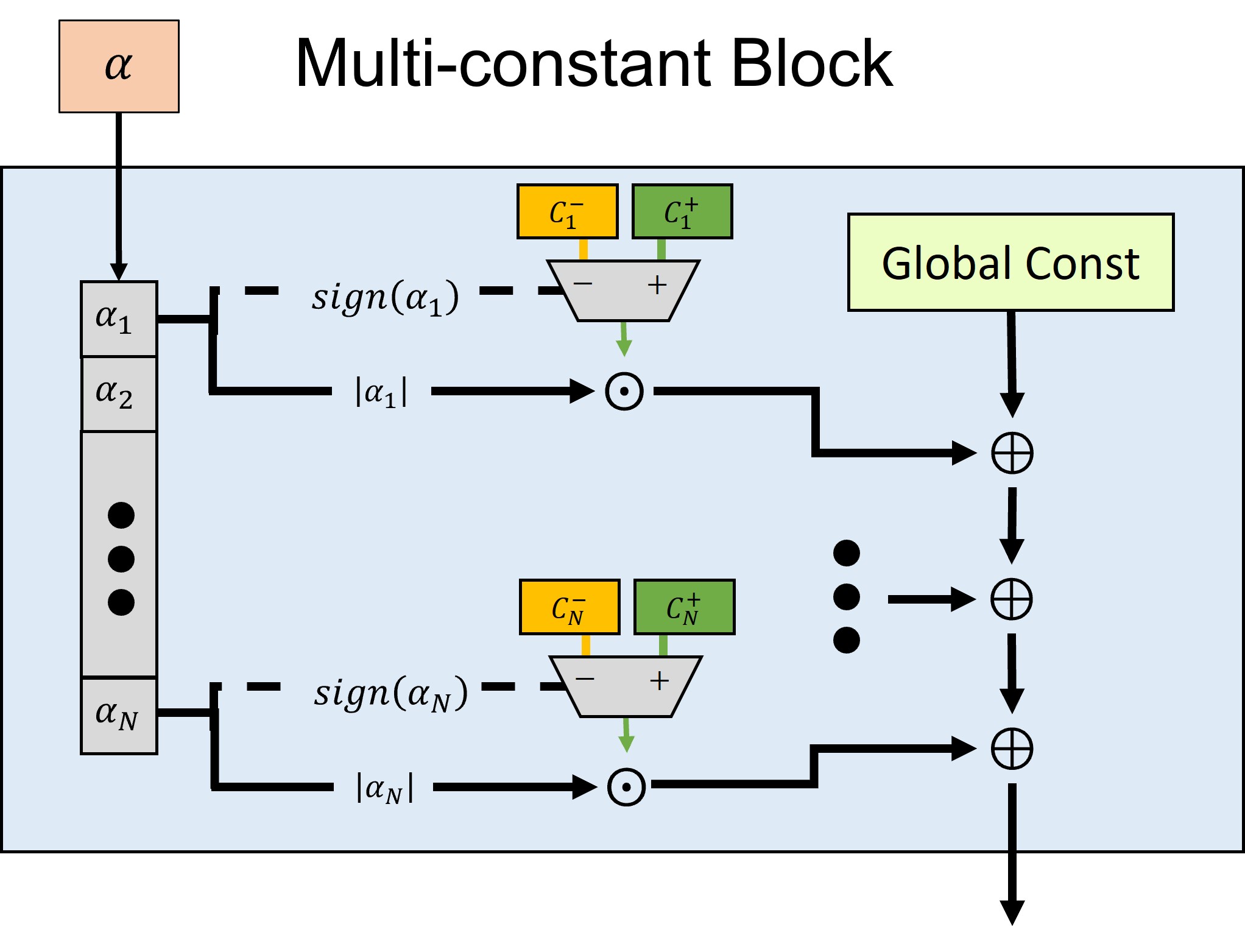}\vspace{-7pt}
    \end{tabular}
    \caption{Overview of our multi-constant conditional generator. We treat the controlled attributes as different data modalities, introduced through modulations of the learned constant. Each modality has two additive constants, denoting a positive and negative attribute change with respect to the mean image. The direction and strength of the attributes is controlled through a vector $\vec{\alpha}$, where each entry $\alpha_i$ controls a different image property. }
    \label{fig:arch}
    \vspace{-5pt}
\end{figure}

%% file: experiments.tex
\section{Experiments}
\label{sec:experiments}

\subsection{Qualitative Comparison}
\input{resources/figures/ours_multi_edit}

\ifx\arxiv\undef
\input{resources/figures/edit_comparisons_inversions}
\input{resources/figures/non_linear_comparisons}
\else
\input{resources/figures/edit_comparisons_inversions_arxiv}
\input{resources/figures/non_linear_comparisons_arxiv}
\fi

We begin with a qualitative comparison of our method to existing editing techniques.
In \cref{fig:multi_editing} we show sequential editing of synthesized images using our method. Our model is capable of successfully controlling multiple attributes, across large modifications, without significantly compromising the identity.

In \cref{fig:editing_comparisons} we compare our editing capabilities on real images to a range of linear editing methods. Real images were inverted into our model and into the official FFHQ 1024x1024 model, using PTI~\cite{roich2021pivotal}. For each identity, we compare the editing performance of our model to the baseline linear editing direction which was used to extract the conditioning labels. Our method consistently outperforms these baselines. The performance gap widens as we move closer to the edges of the distribution, such as when considering large poses or old faces. When considering asymmetric attributes such as glasses, our model maintains the same image even as we keep moving in the no-glasses direction. The same operation with the StyleCLIP baseline, meanwhile, leads to an increase in age. These results demonstrate that editing performance can be improved, with no additional supervision, simply by addressing the Generative Bias and allowing the generator to be fine-tuned.

Finally, in \cref{fig:non_linear_comparisons} we compare our performance to non-linear alternatives: Local Basis~\cite{choi2021escape} and StyleFlow~\cite{abdal2020styleflow}. Recall that our method does not maintain the identities of the original model. To facilitate comparisons in spite of this limitation, we use the alternatives to edit images synthesized by the original model, and project the same images into our models using PTI.

Despite the more challenging (inverted image) setup, our method still displays more robust editing performance, maintaining better identity through age changes and reducing deterioration for large poses. Importantly, unlike the supervised StyleFlow, our method relies only on the weak (left / right pose labels) or CLIP-based supervision used to find the initial editing directions (InterFaceGAN for pose, StyleCLIP for age).

As our experiments demonstrate, self-conditioned GANs are capable of consistent manipulations over larger spans than existing methods, even when compared to complex, non-linear approaches. 

\subsection{Quantitative Comparisons}\label{sec:quant_comp}

We quantitatively evaluate the performance of our method by considering identity preservation for large modifications.

We follow \cite{roich2021pivotal} and measure the cosine similarity between the embeddings of a pre-trained identity recognition network (ArcFace \cite{Deng_2019_CVPR}) using pairs of pre- and post-editing images. When considering continuous attributes, we ensure a similar magnitude of change by employing pre-trained pose \cite{ruiz2018pose} and age detection \cite{yusuke2018age} networks. Images are edited until they reach a desired level of change, \eg $+20$ years. For binary attributes, we observe that for any fixed step size, there exist a portion of images where the manipulation fails (\eg glasses are not added), and a portion of images where the manipulation is too strong, and identity is lost. Increasing the step size leads to an increase in both successful manipulations, and in identity loss.
To facilitate fairer comparisons, we thus test each method along a wide range of step sizes and report the identity preservation scores as a function of the percent of images that were successfully manipulated. A manipulation is considered successful if it causes an off-the-shelf classifier~\cite{karras2019style} to change its result.

The results are shown in \cref{fig:id_comps}. Our model maintains a higher degree of identity similarity for rare attributes (glasses) or in regions where data is sparse and the baseline generator tends towards mode-collapse (large age, pose), and performs on par with non-linear methods where data is abundant (smiles).

\input{resources/figures/id_comp}

\subsection{Ablation study}

\input{resources/figures/ablation}

We evaluate different aspects of our proposed method by conducting a qualitative ablation study. Specifically, we investigate our choice of a multi-constant setup, the use of latent-space labels, the importance of uniform sampling, and the benefits of unfreezing the generator.

In the first scenario, rather than spreading our modalities across constants, we build on the conditional setup of \cite{Karras2020ada} and extend the latent code with a conditioning code that consists of one entry per property, with values $\alpha_i \in \left[-1, 1\right]$. For a model with control over $n$ attributes, the latent code therefore takes the form $z' = z \oplus \alpha_1 \oplus \alpha_2 \oplus ... \oplus \alpha_n$. The discriminator is similarly tasked with predicting the set $\{\alpha_i\}_{i=1}^n$. For real images, the $\alpha$ values are given by the self-labeling scores described in \cref{sec:latent_labels}.

In the second scenario, we replace the latent-space distance labels with confidence scores derived from a binary attribute classifier - the same classifier used to generate InterFaceGAN directions.

In the third scenario, we fine-tune the model with images sampled randomly from the FFHQ set, with no regard to their labels.

Finally, in the fourth scenario, we learn new constants without modifying any of the pre-trained generator's weights.

In \cref{fig:ablation} we show the results of each experiment. When replacing the constants with a conditional mapping network, we observe a severely decreased range of control. We hypothesize that, as the mapping network needs to be trained from scratch, the optimization prefers to devote most efforts to re-aligning the generated distribution rather than to the attribute control. If this is the case, a change of hyper-parameters might enable successful conditional editing, but we were unable to find such parameters.

Using classifier scores similarly leads to a reduced editing range. This is likely a result of the quick saturation in classifier scores.

When we do not perform uniform sampling, editing range is similar to that of our full model, but the quality of results near the edges of the distribution deteriorates, demonstrating that the data bias contributes directly to the performance in these regions.

Lastly, if the generator remains frozen, it fails to adapt to the changes in constants. Attempts to manipulate the image through constant interpolations lead to minor changes at best, and more commonly to severe quality deterioration.

%% file: resources/figures/ours_multi_edit.tex
\begin{figure}[!hbt]
    \centering
    \setlength{\belowcaptionskip}{-2.5pt}
    \setlength{\tabcolsep}{1.2pt}
    {
    \begin{tabular}{c c c c}
    
        \includegraphics[width=0.22\linewidth]{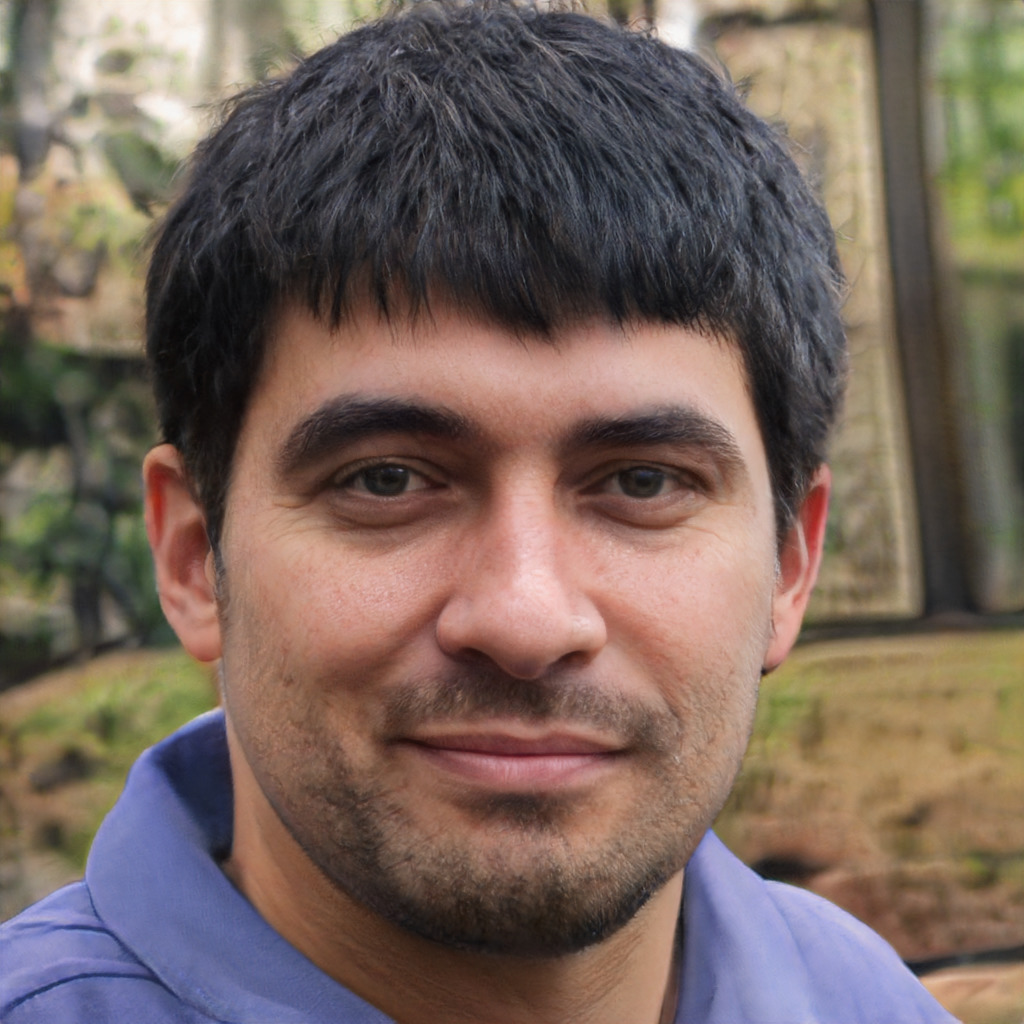} &
        \includegraphics[width=0.22\linewidth]{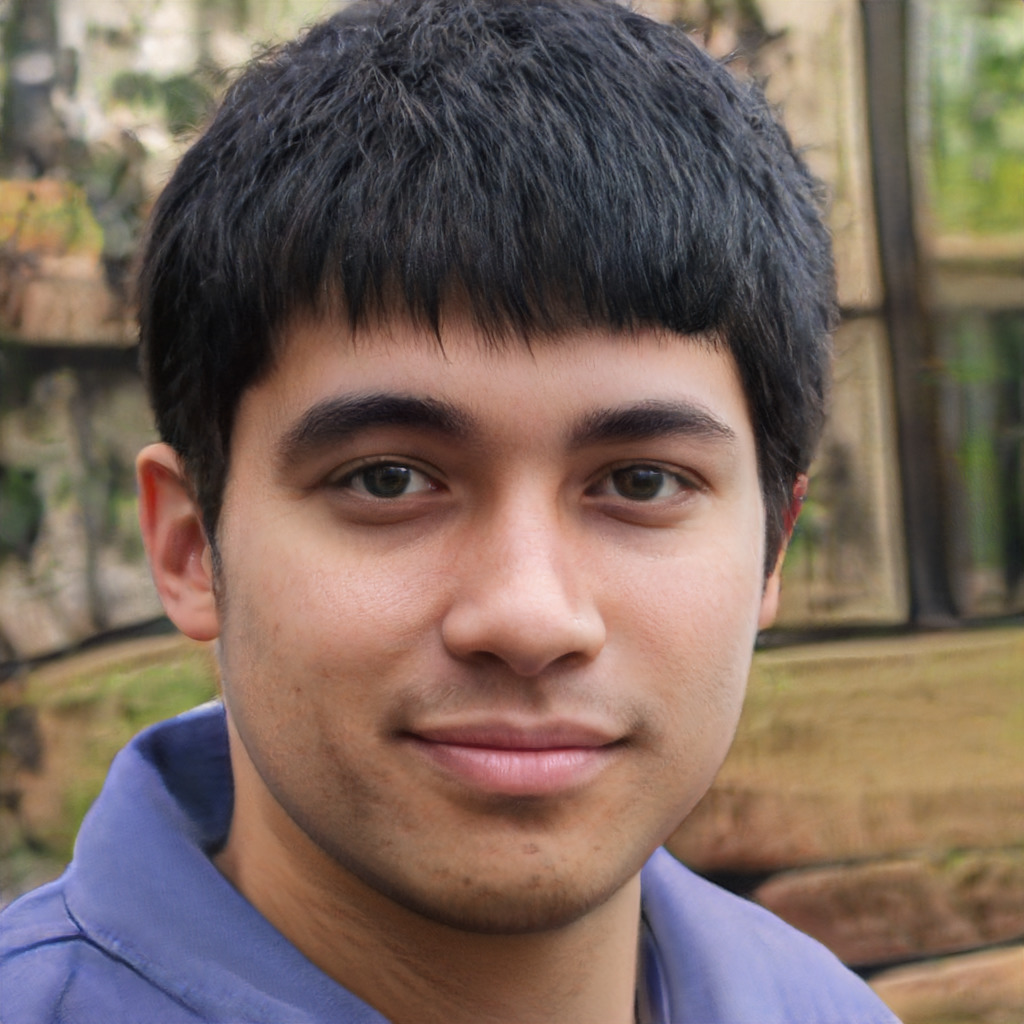} &
        \includegraphics[width=0.22\linewidth]{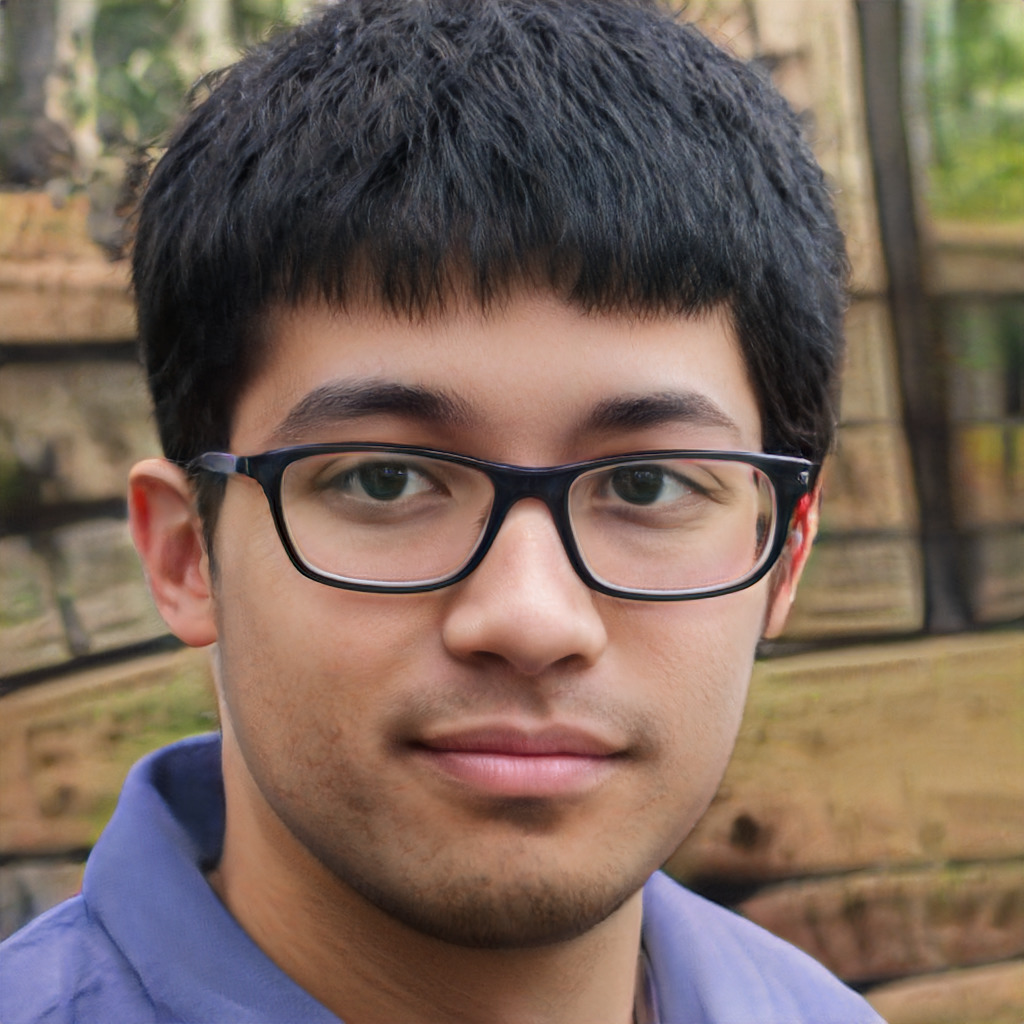} &
        \vspace{-2pt}
        \includegraphics[width=0.22\linewidth]{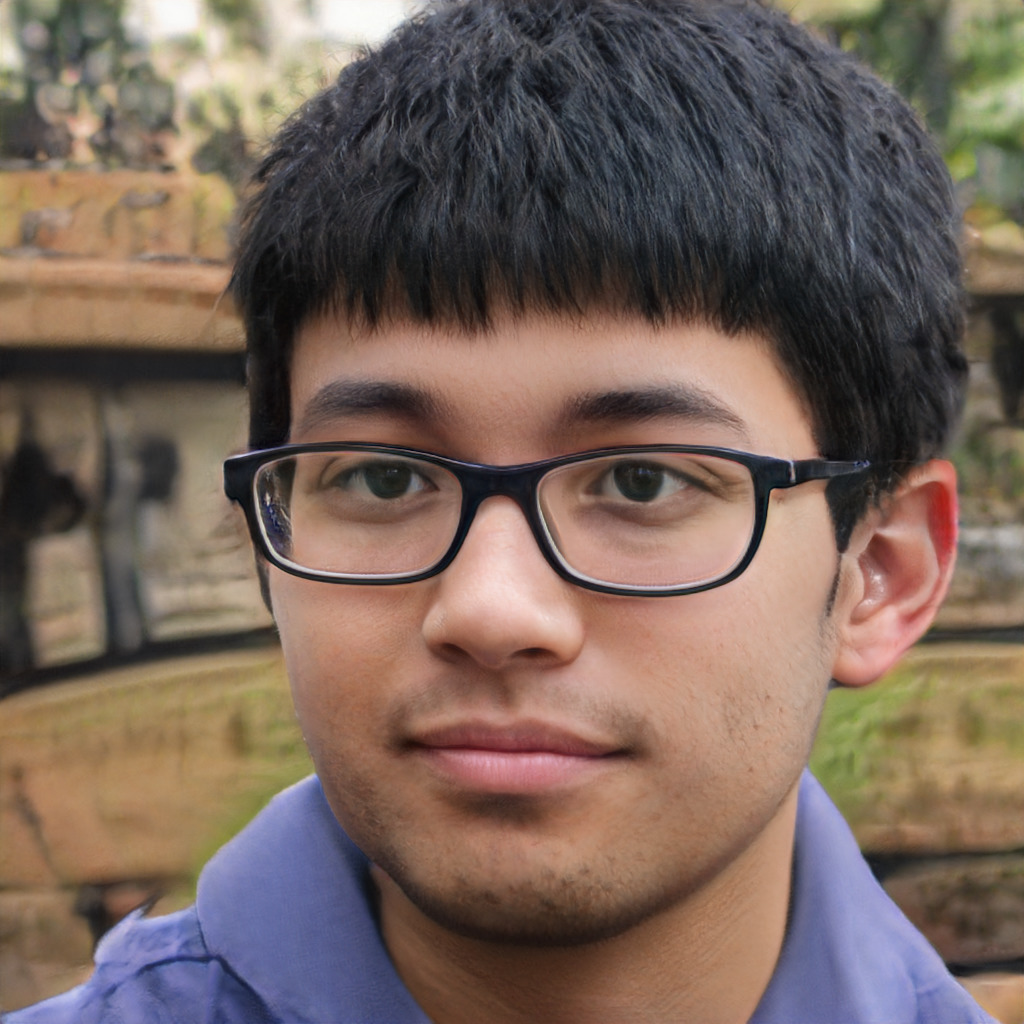} \\
        Original & $+$Age & $+$Glasses & $+$Pose \\
        
        \includegraphics[width=0.22\linewidth]{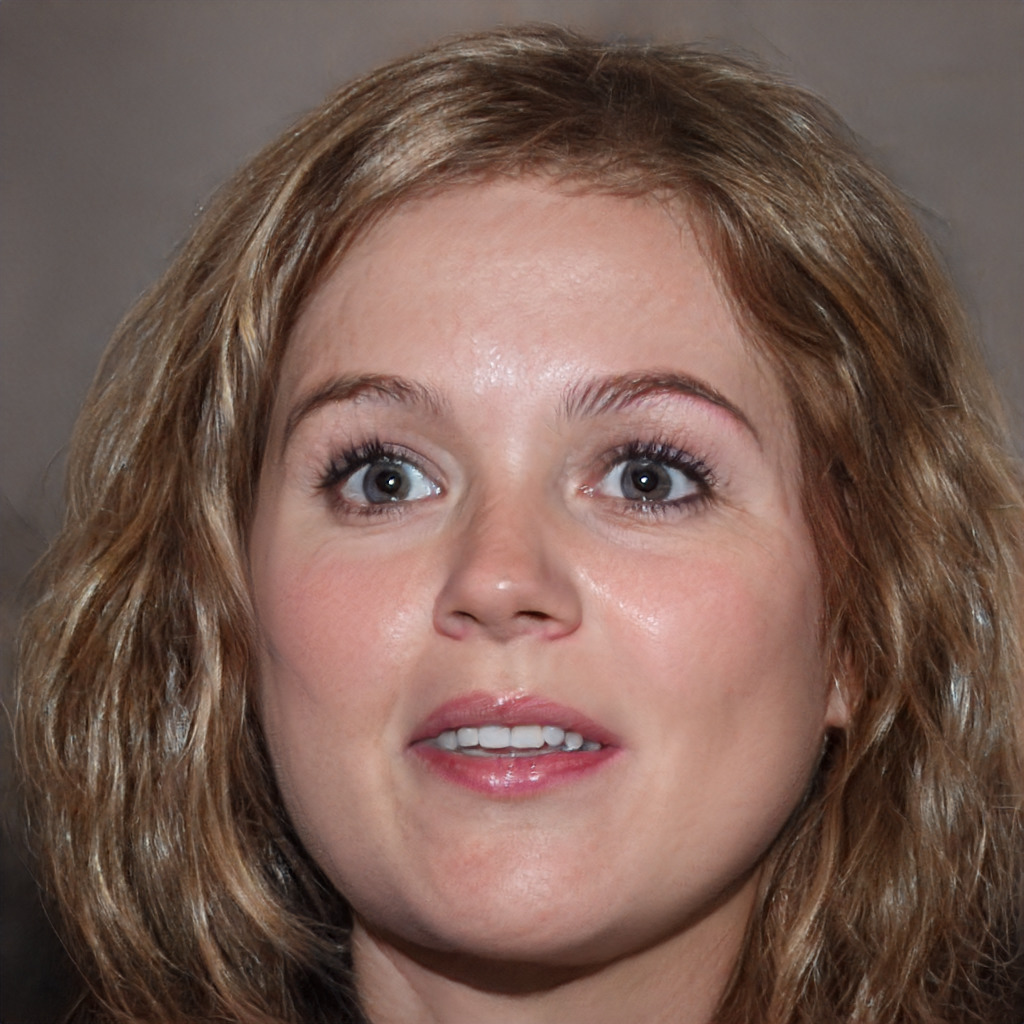} &
        \includegraphics[width=0.22\linewidth]{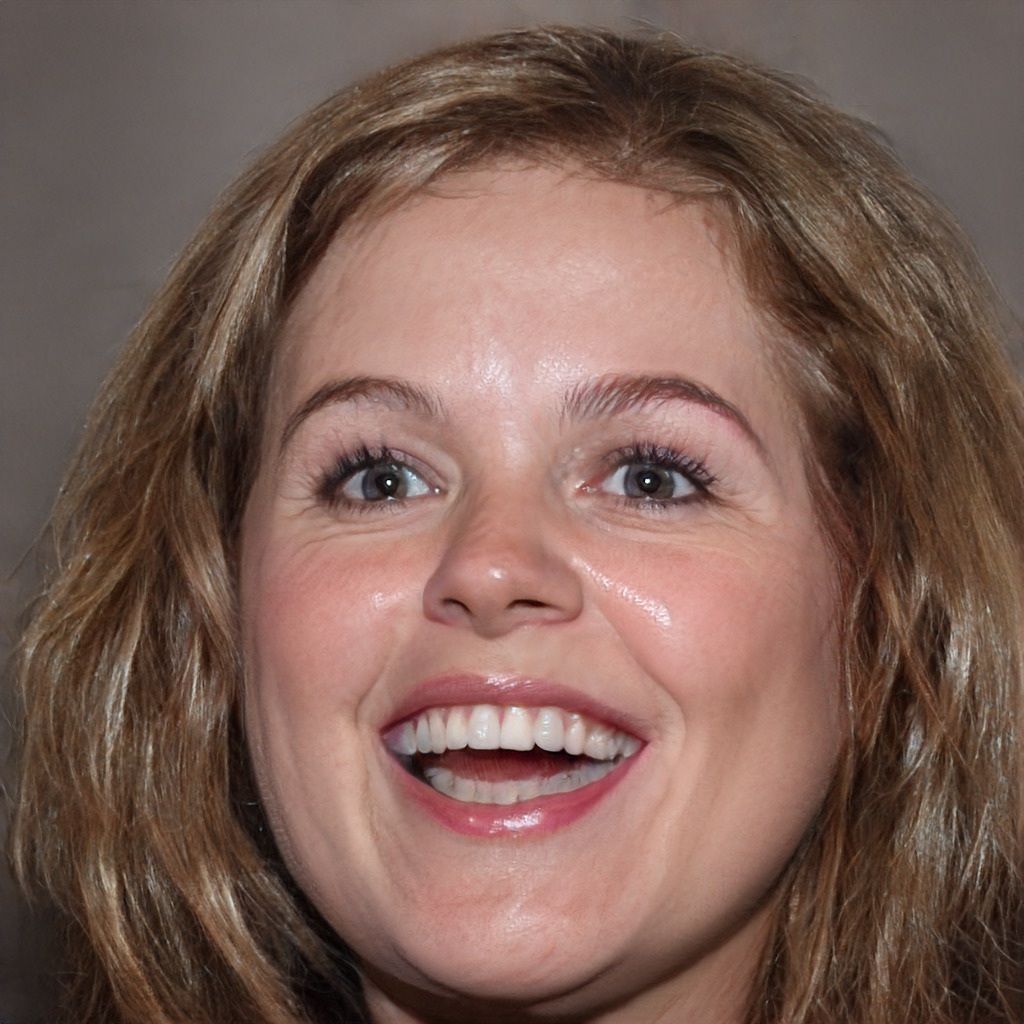} &
        \includegraphics[width=0.22\linewidth]{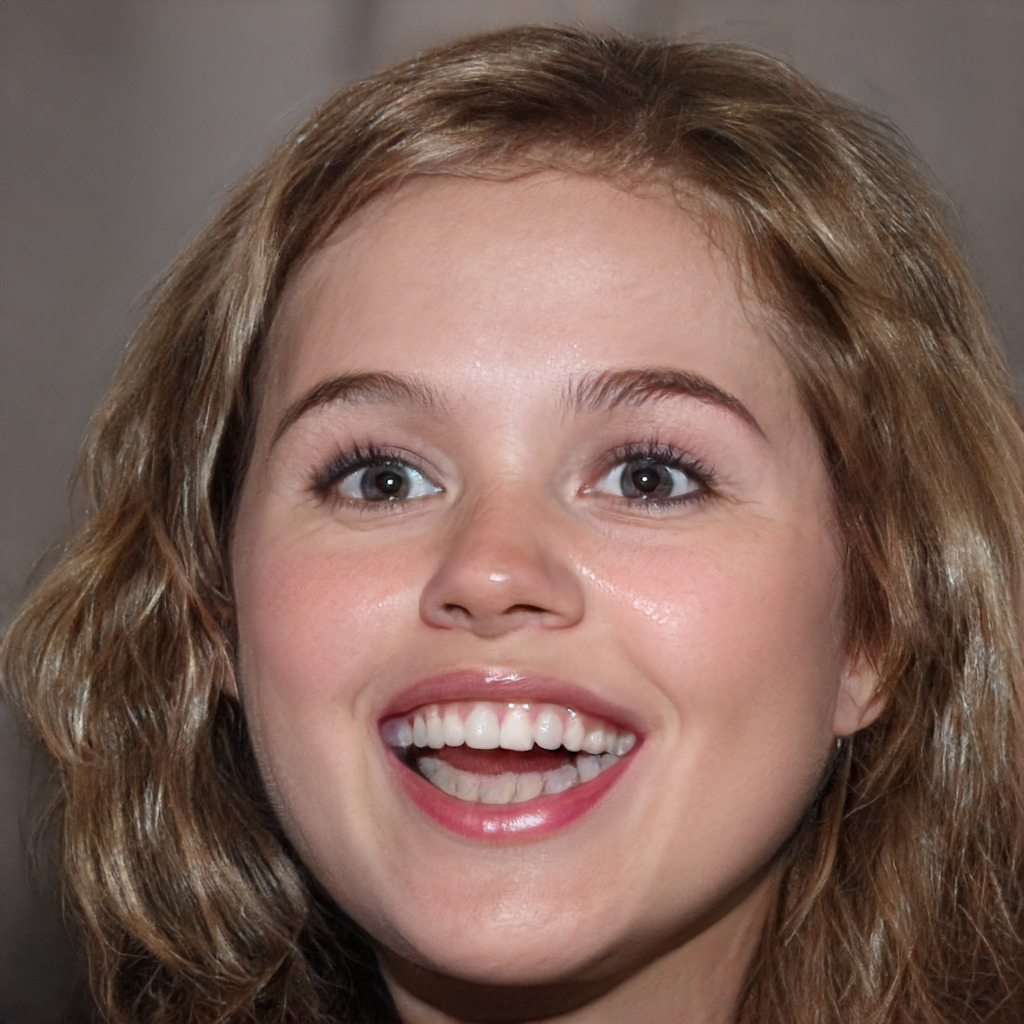} & \vspace{-2pt}
        \includegraphics[width=0.22\linewidth]{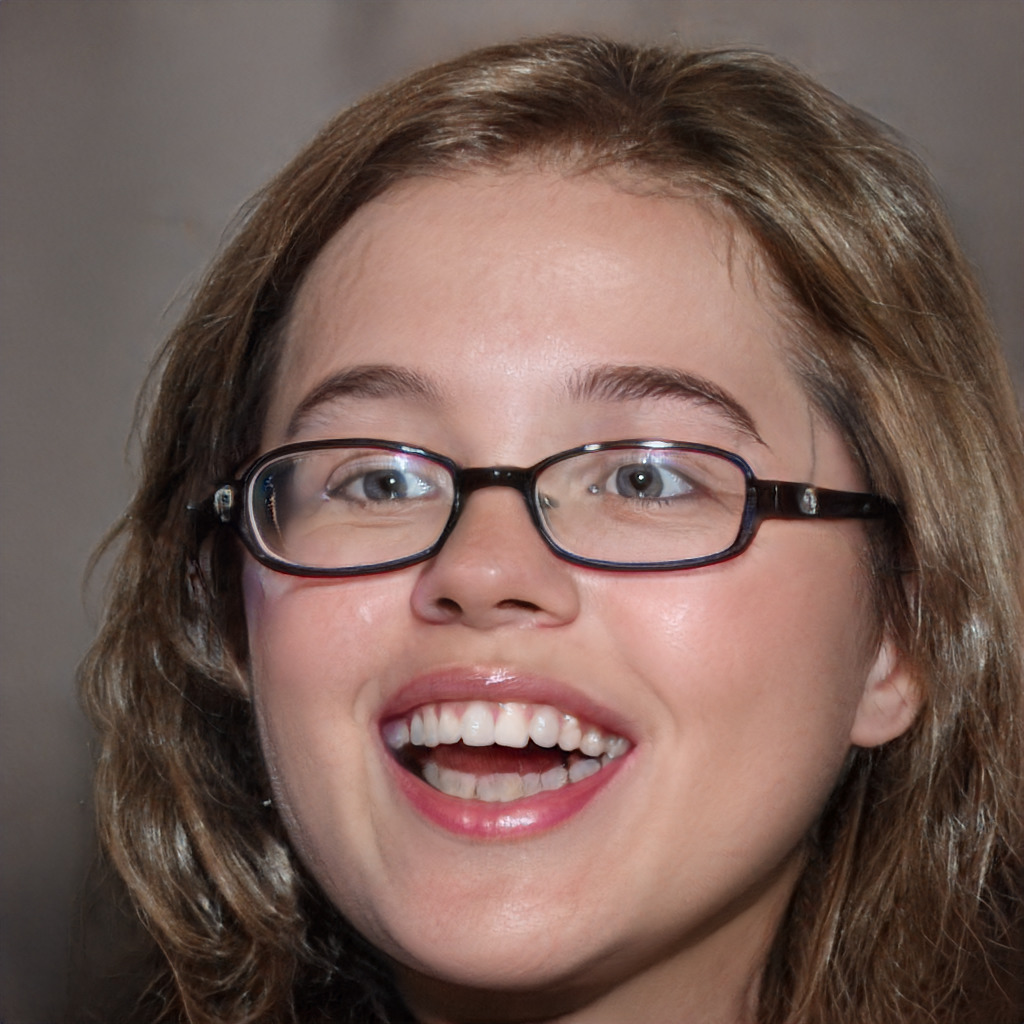} \\
        Original & $+$Smile & $+$Age & $+$Glasses \\
        
        \includegraphics[width=0.22\linewidth]{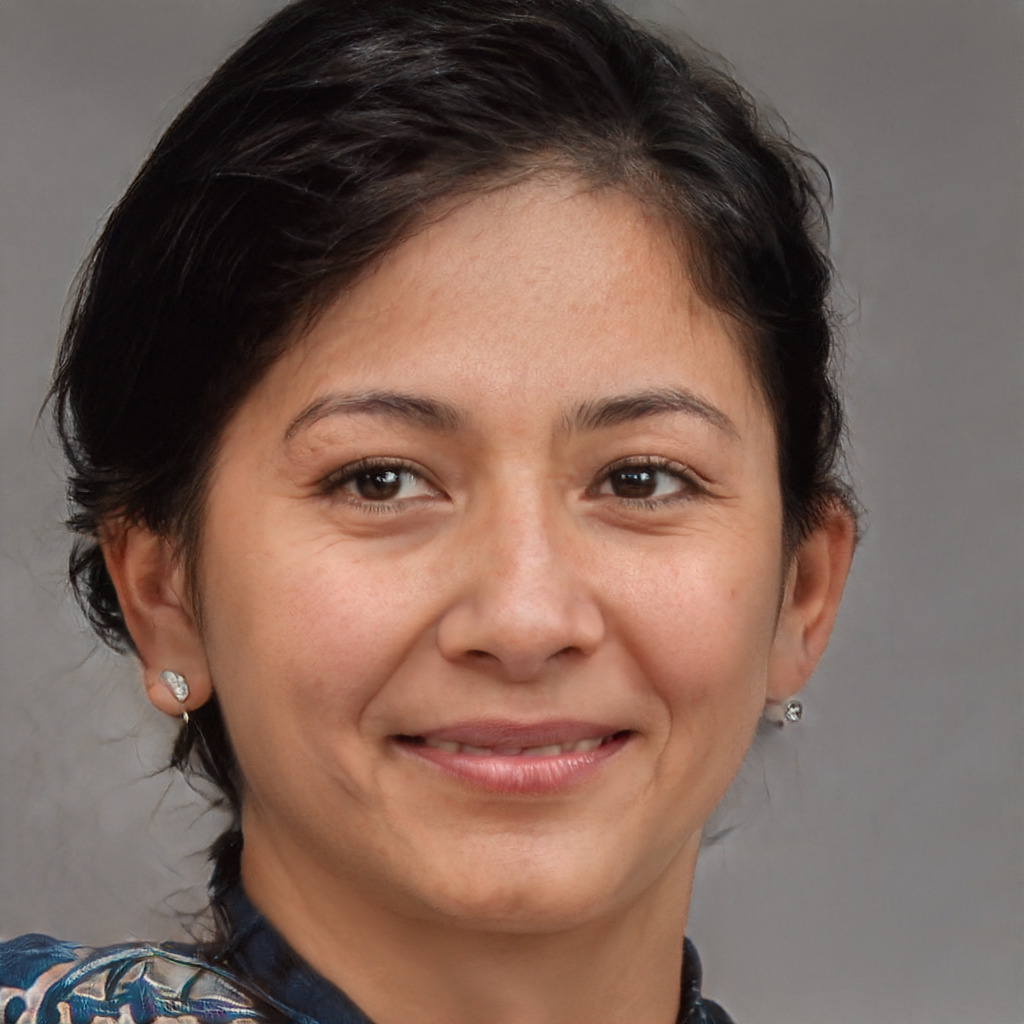} &
        \includegraphics[width=0.22\linewidth]{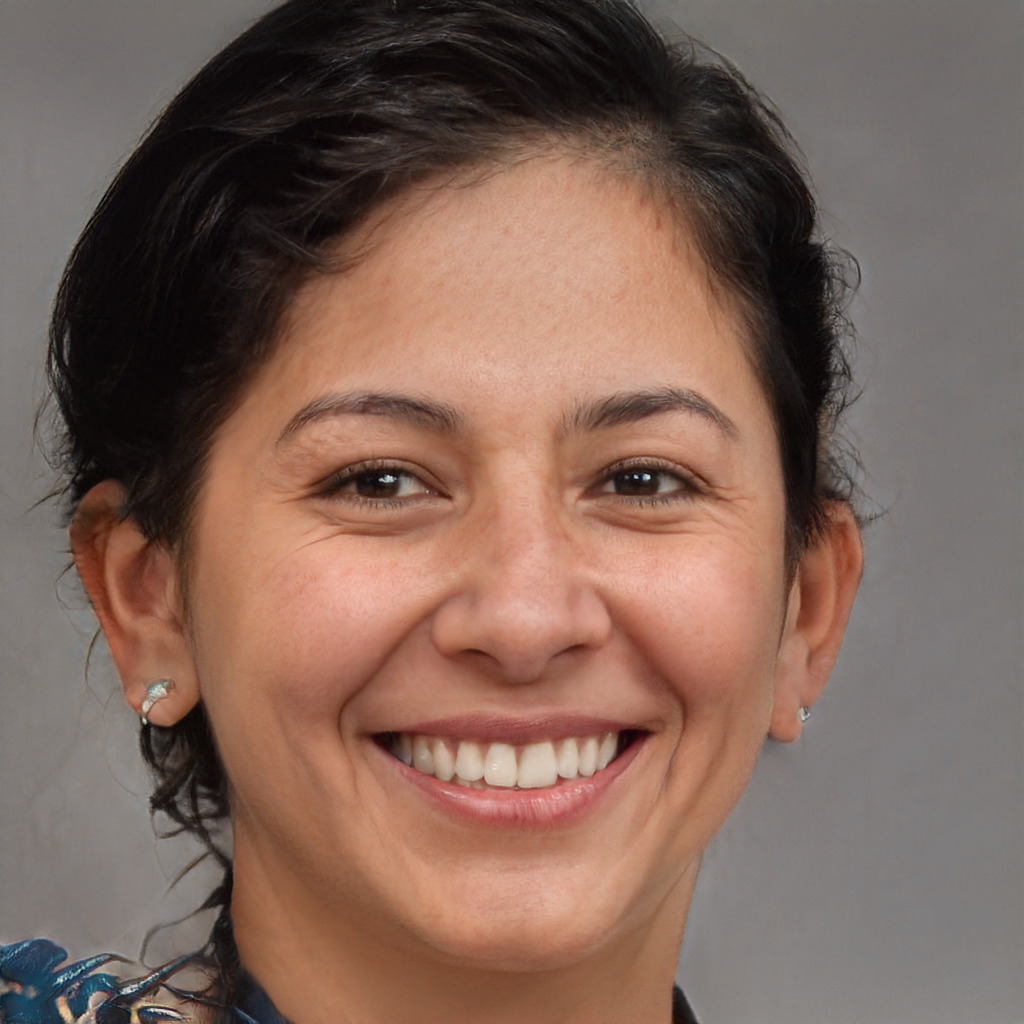} &
        \includegraphics[width=0.22\linewidth]{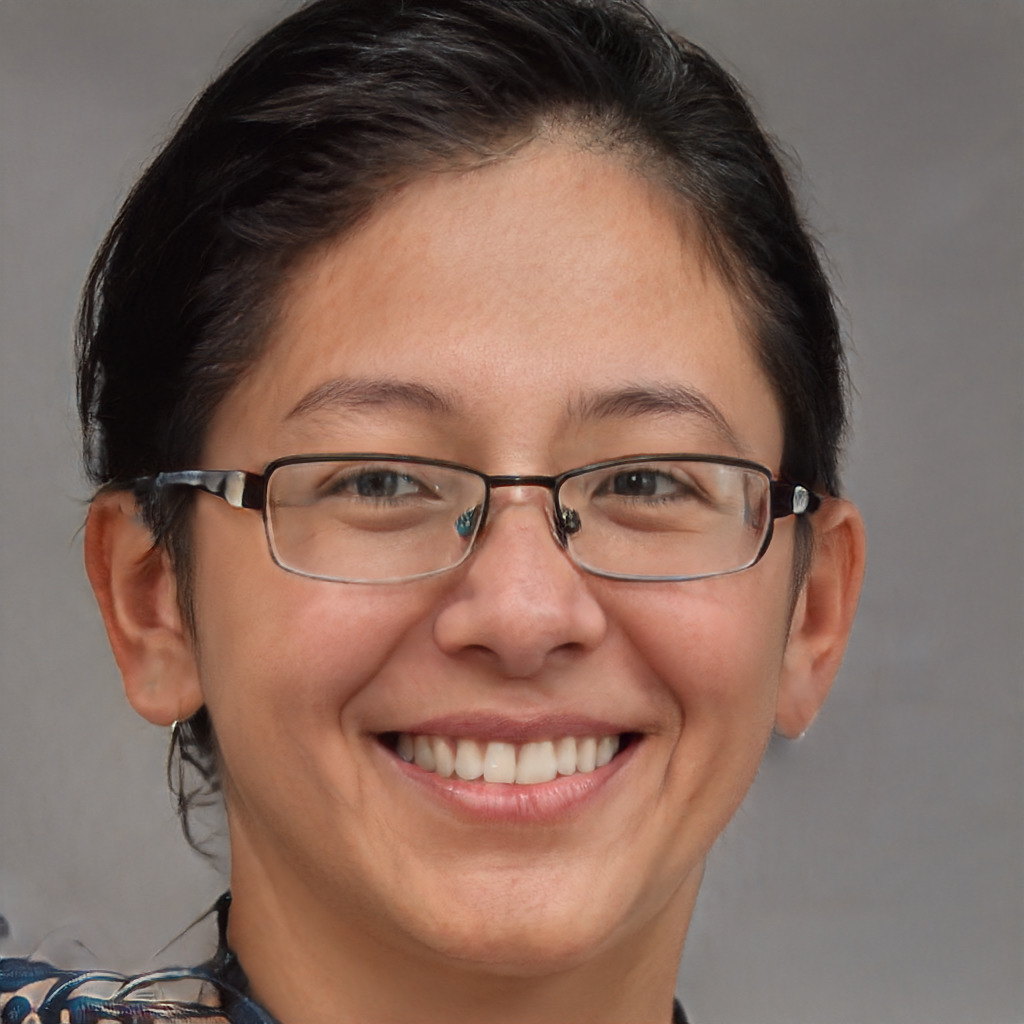} & \vspace{-2pt}
        \includegraphics[width=0.22\linewidth]{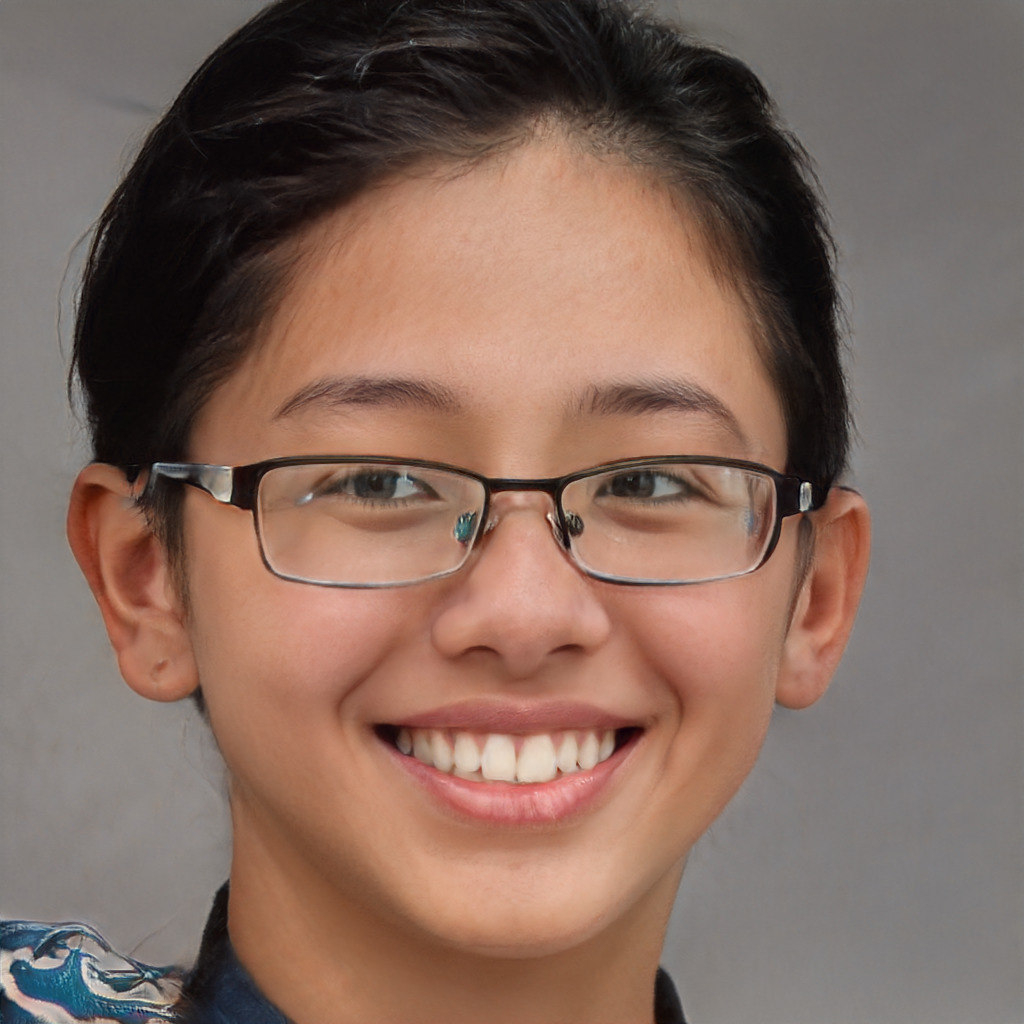} \\
        Original & $+$Smile & $+$Glasses & $+$Age \\
        
        \includegraphics[width=0.22\linewidth]{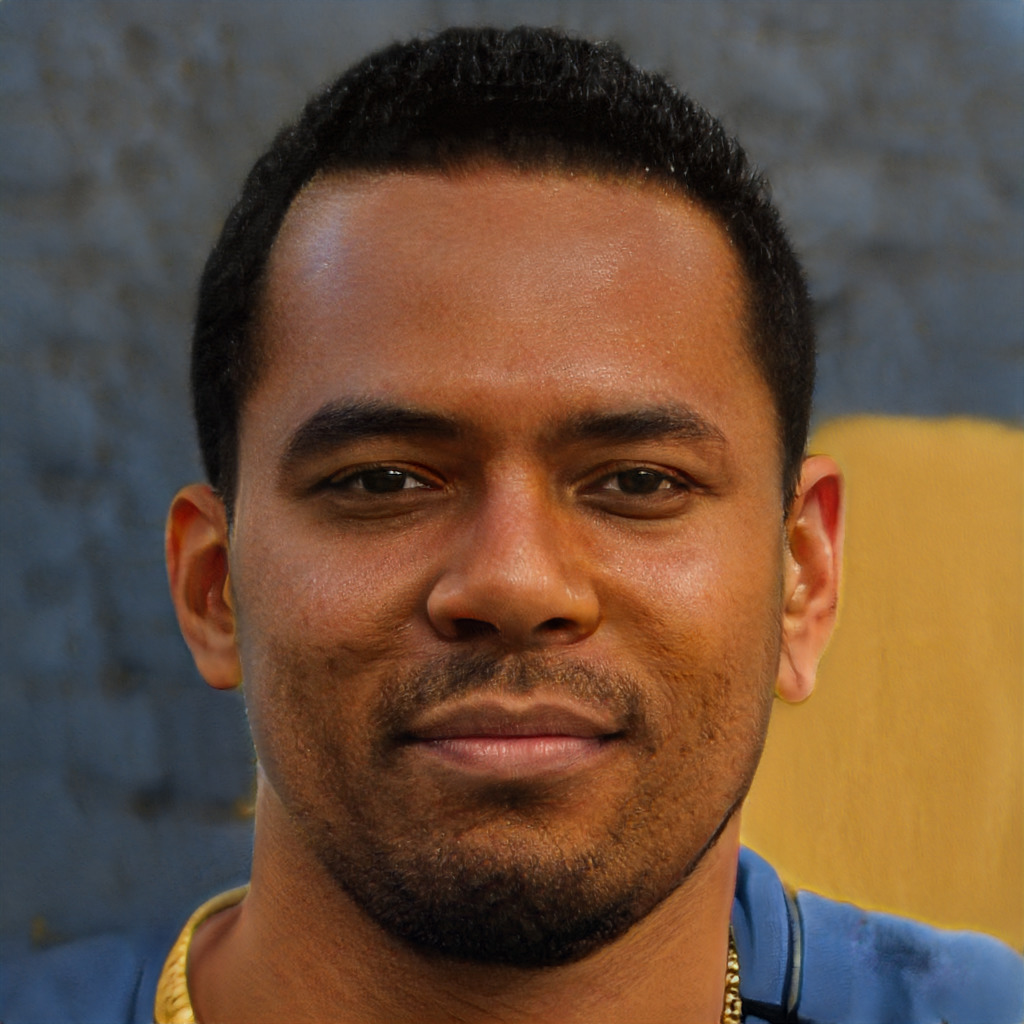} &
        \includegraphics[width=0.22\linewidth]{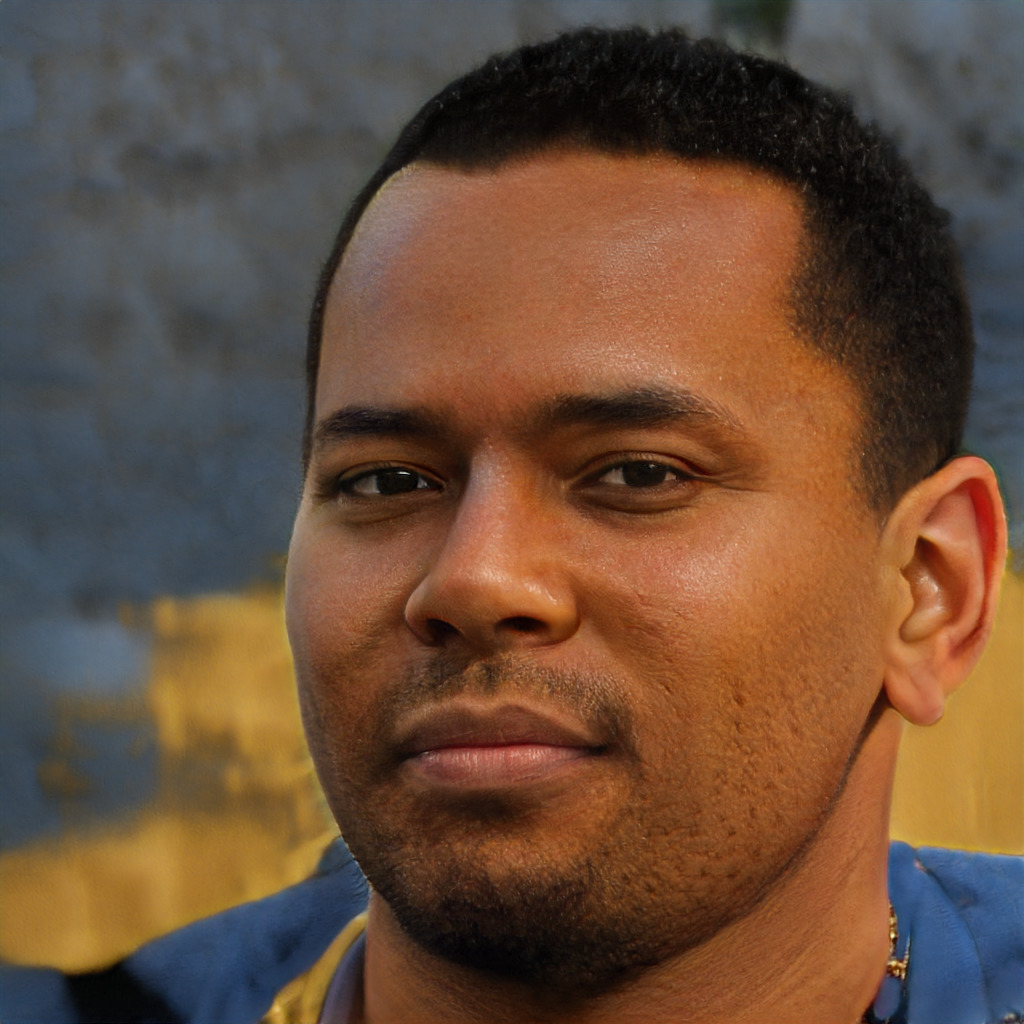} &
        \includegraphics[width=0.22\linewidth]{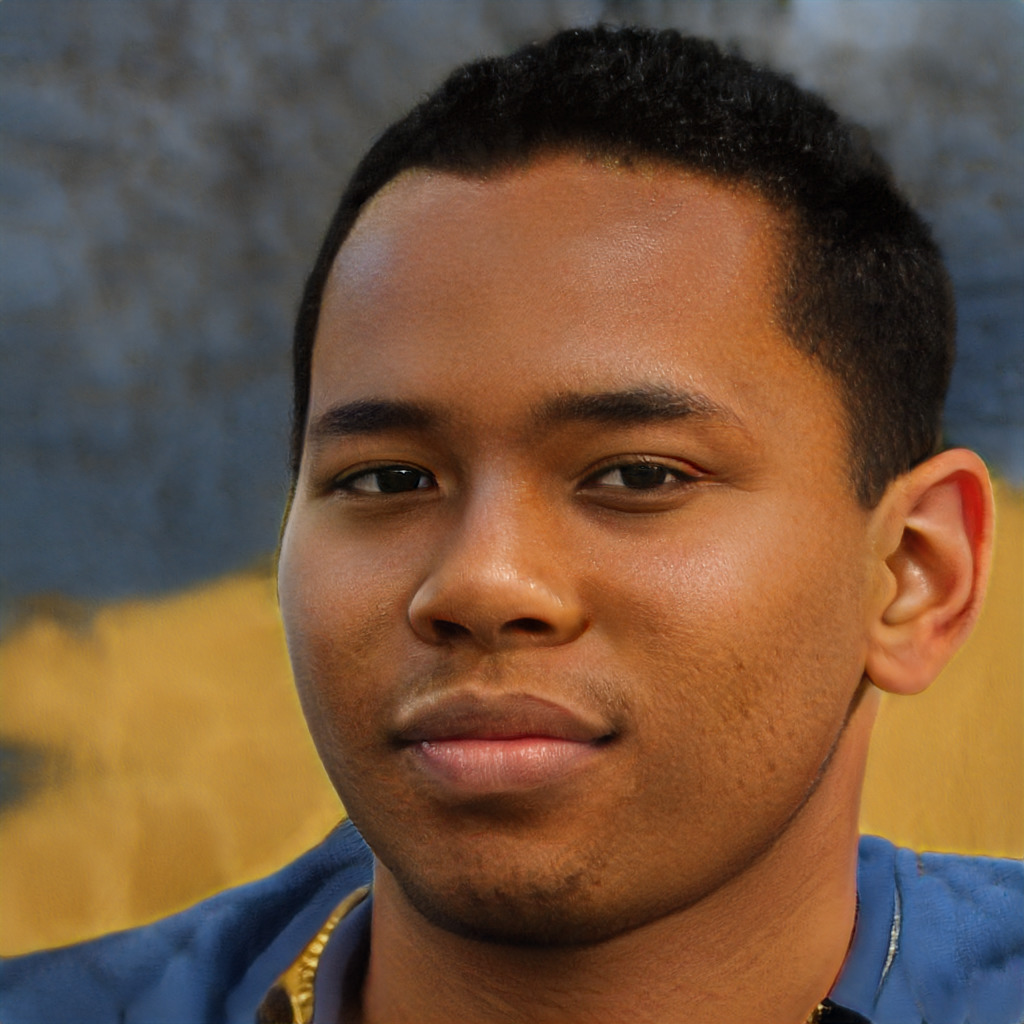} & \vspace{-2pt}
        \includegraphics[width=0.22\linewidth]{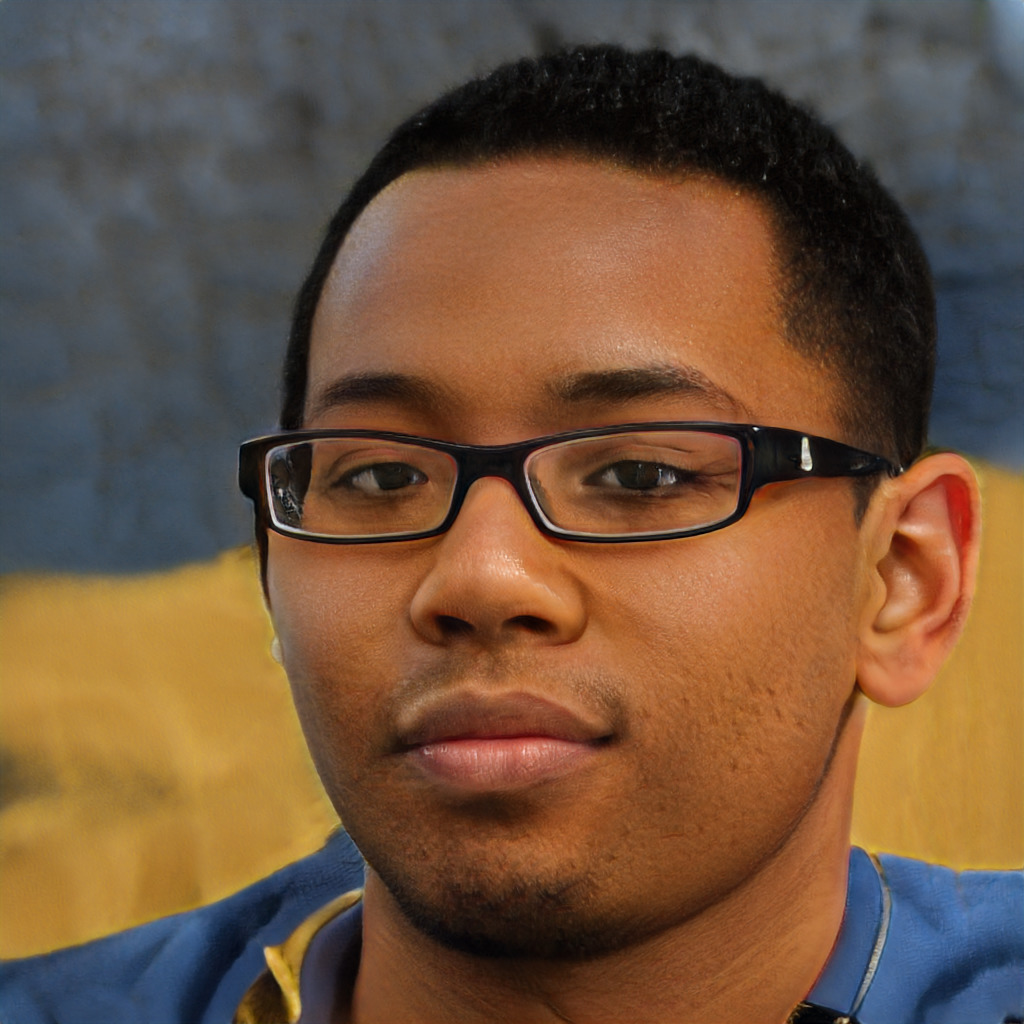} \\
        Original & $+$Pose & $+$Age & $+$Glasses \\
        
        \includegraphics[width=0.22\linewidth]{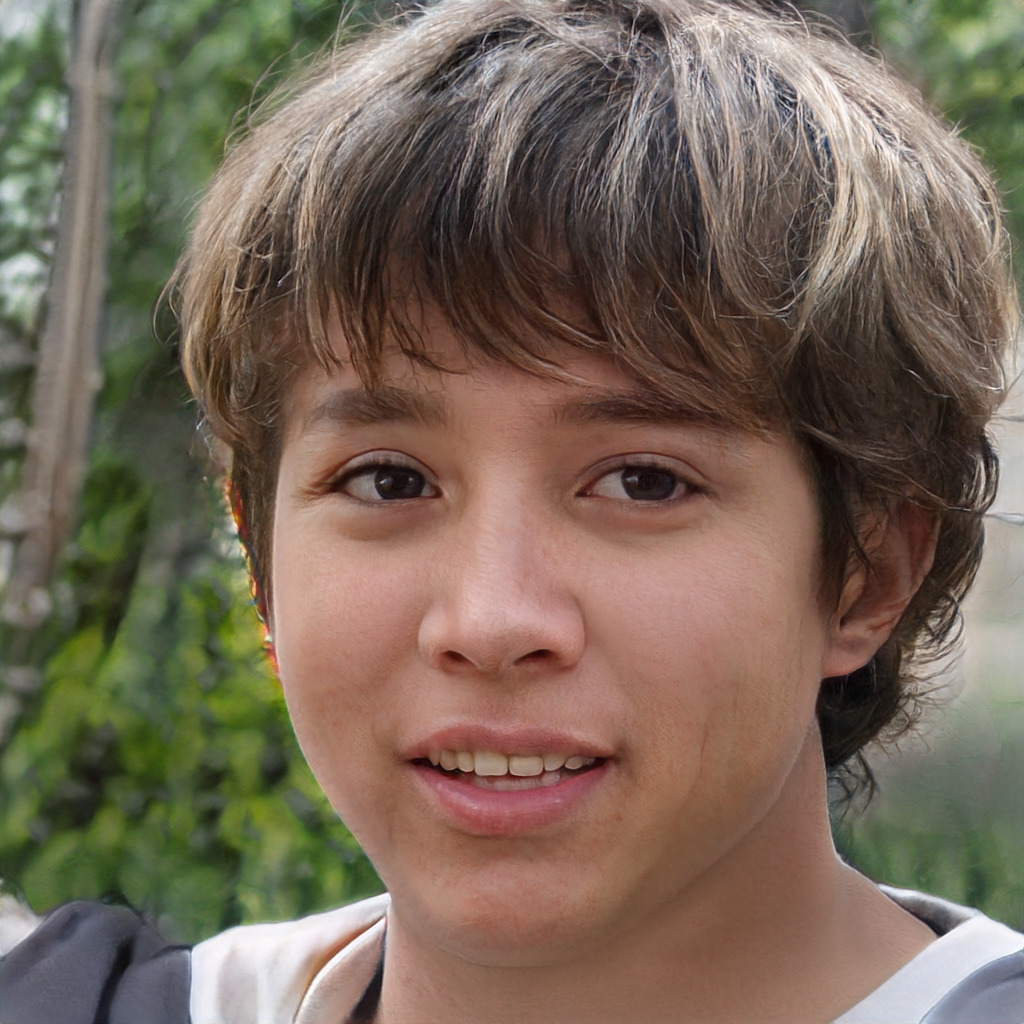} &
        \includegraphics[width=0.22\linewidth]{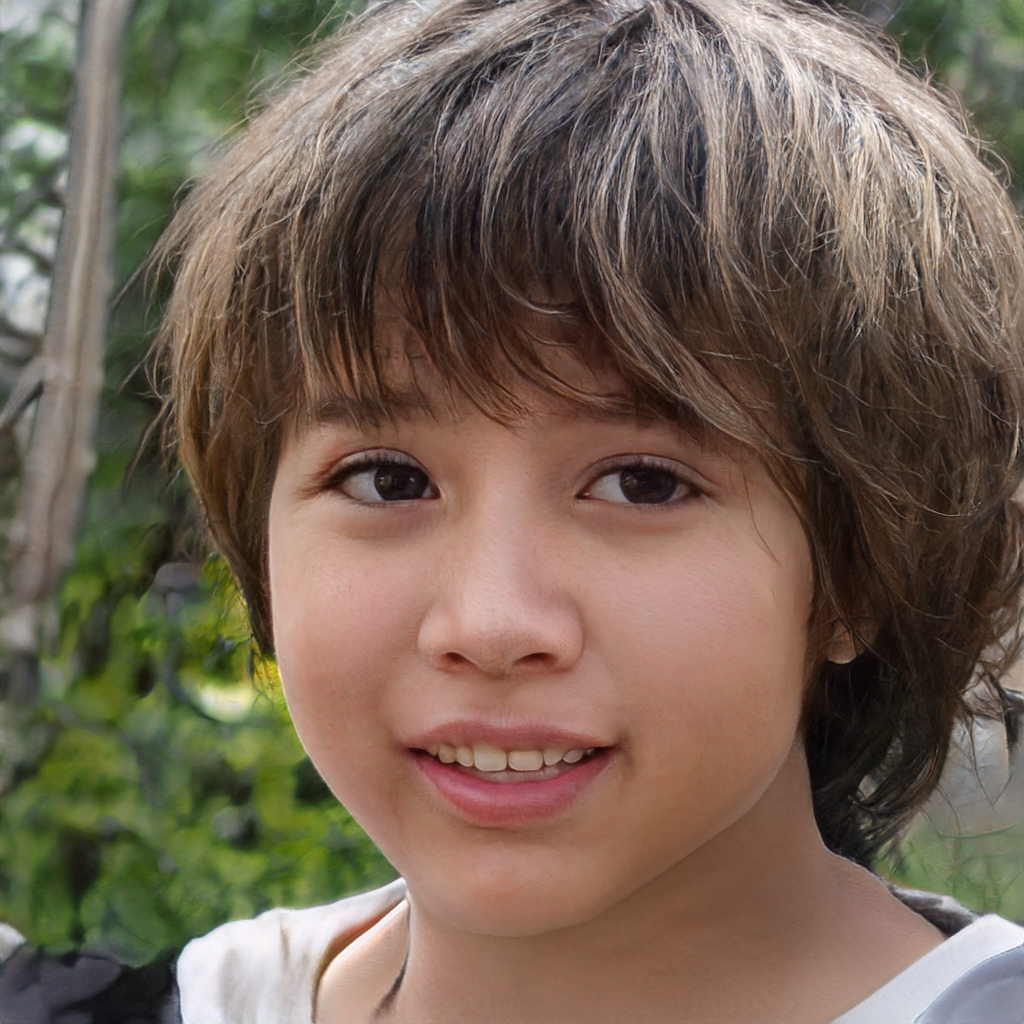} &
        \includegraphics[width=0.22\linewidth]{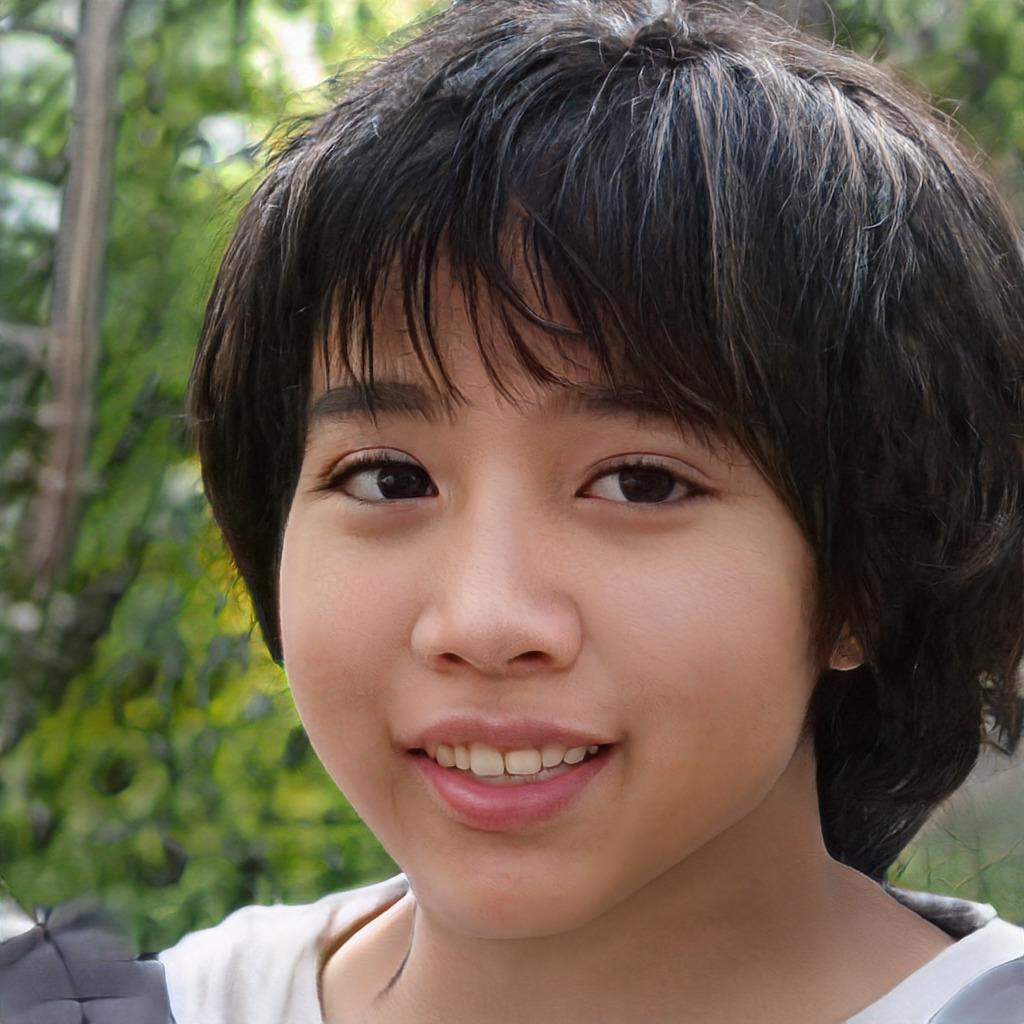} & \vspace{-2pt}
        \includegraphics[width=0.22\linewidth]{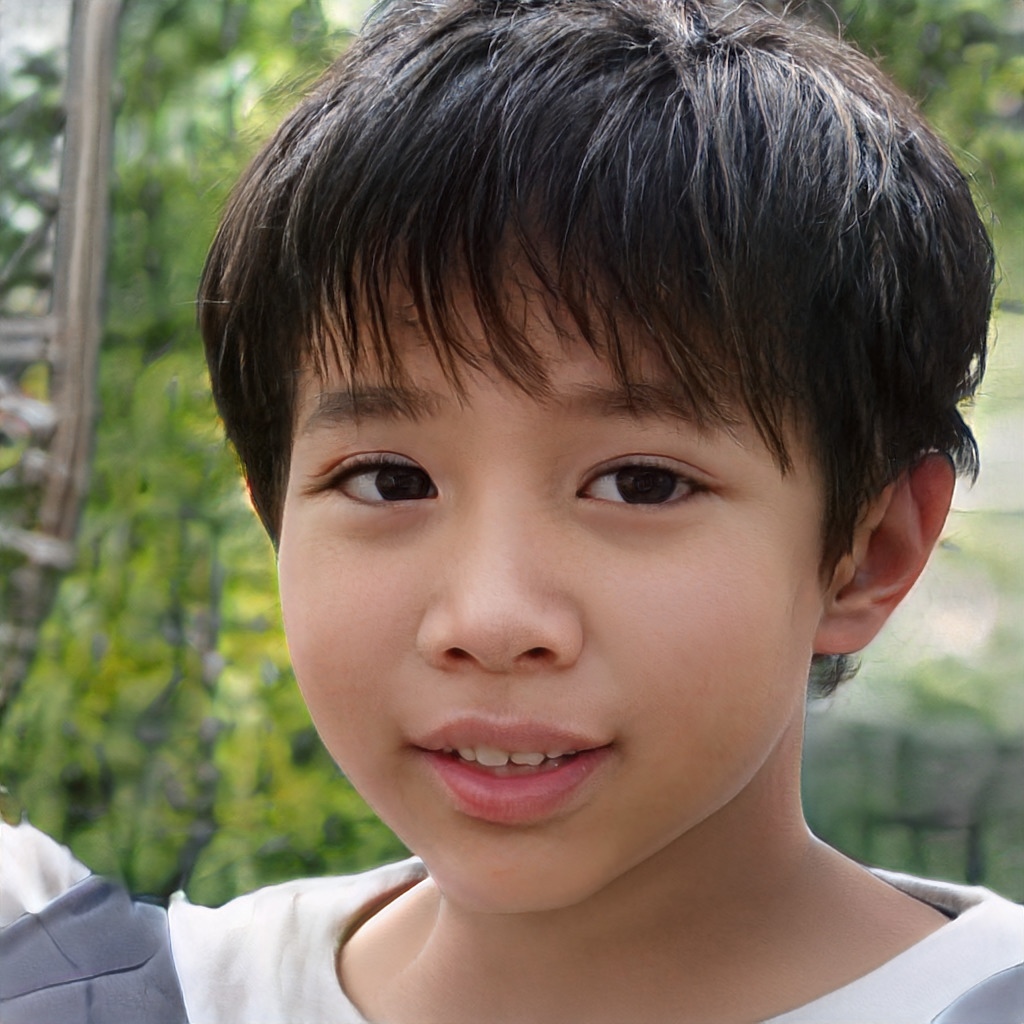} \\
        \vspace{-7pt}
        Original & $+$Age & $+$Hair Color & $+$Hair Length
        
    \end{tabular}}
    \vspace{-2pt}
    \caption{Our framework enables sequential editing of multiple attributes, even when conditioned on latent distances derived from methods which do not. The self-conditioned model is insensitive to the order of editing operations, and can successfully synthesize rare combinations such as young people with glasses and a large pose.}
    
    \label{fig:multi_editing} \vspace{-5pt}
\end{figure} 

%% file: resources/figures/edit_comparisons_inversions.tex
\begin{figure*}
    \centering
    \setlength{\belowcaptionskip}{-8pt}
    \setlength{\tabcolsep}{3.25pt}
    {\small
    \begin{tabular}{c c | c | c c c c c }

        \raisebox{0.1in}{\rotatebox{90}{\tiny InterFaceGAN}} &
        \includegraphics[width=0.086\textwidth]{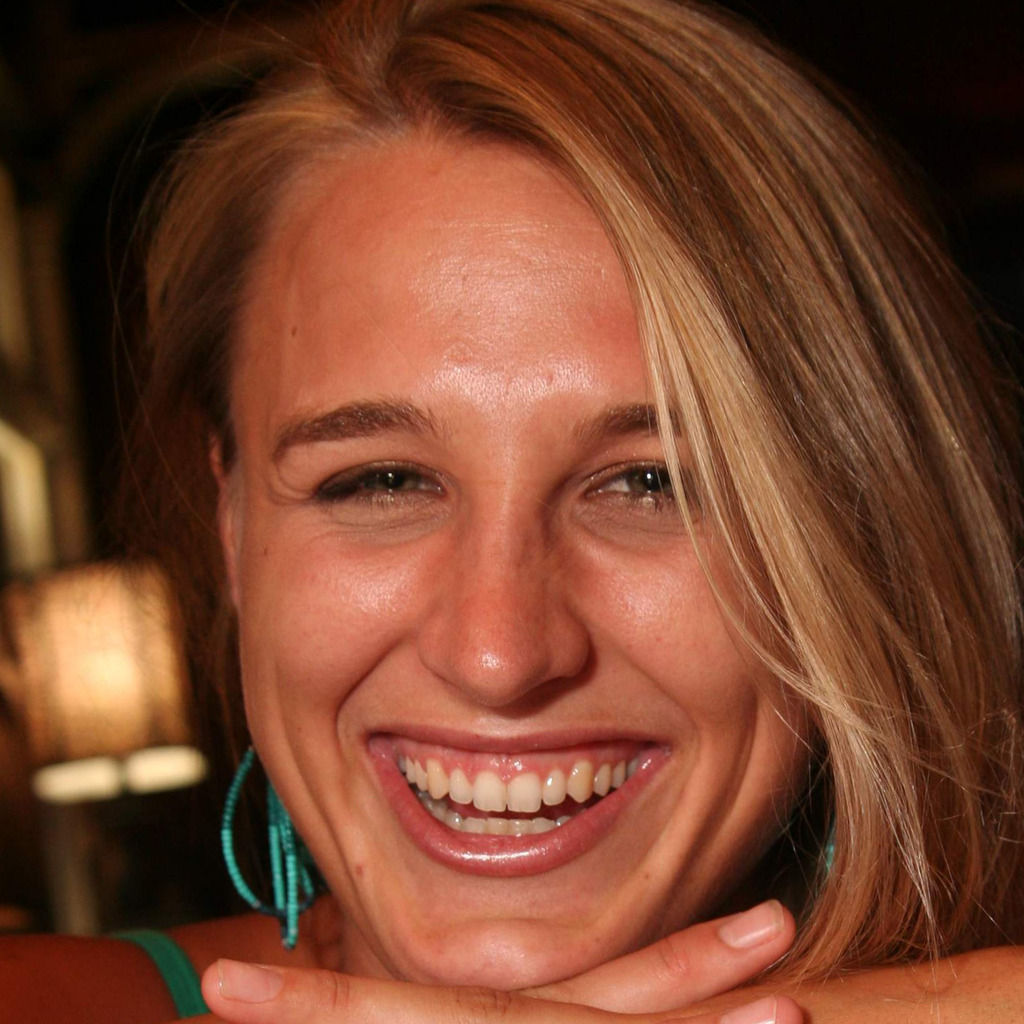} &
        \includegraphics[width=0.086\textwidth]{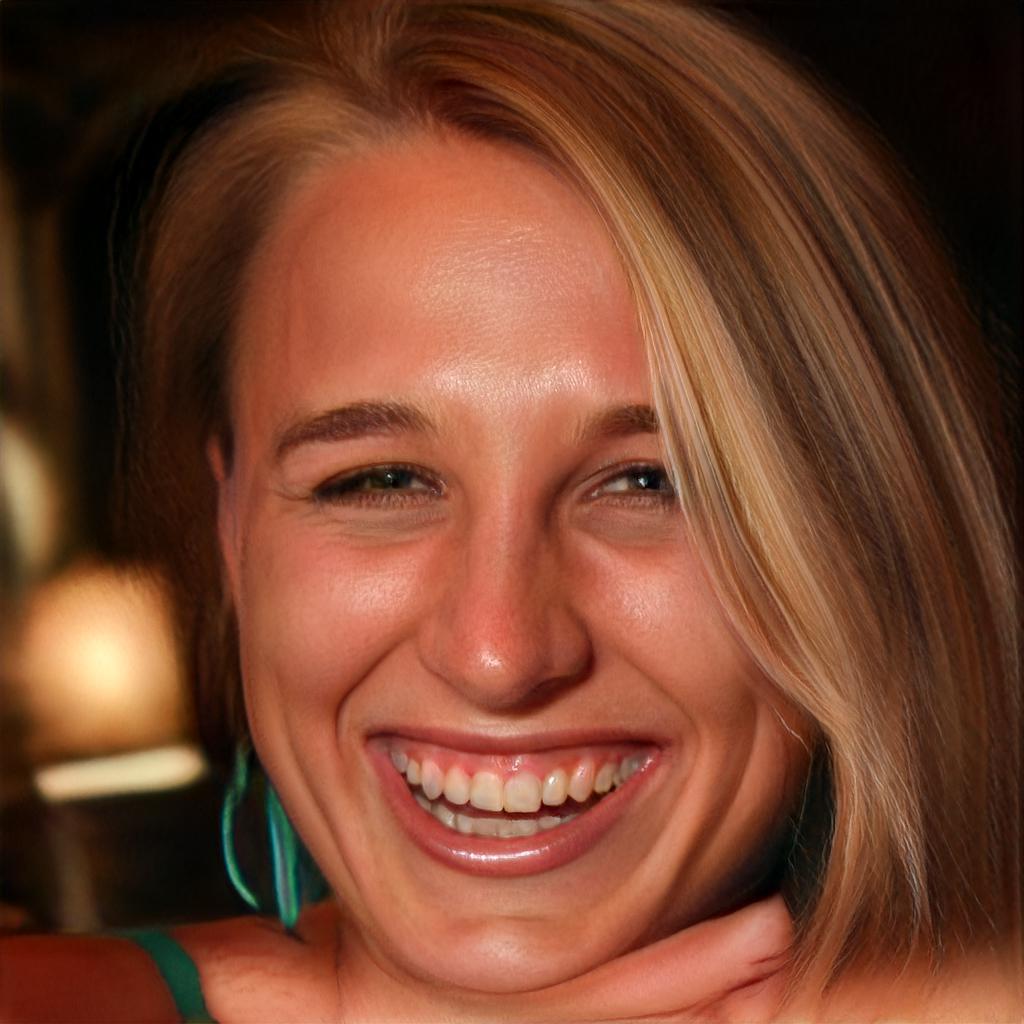} &
        \includegraphics[width=0.086\textwidth]{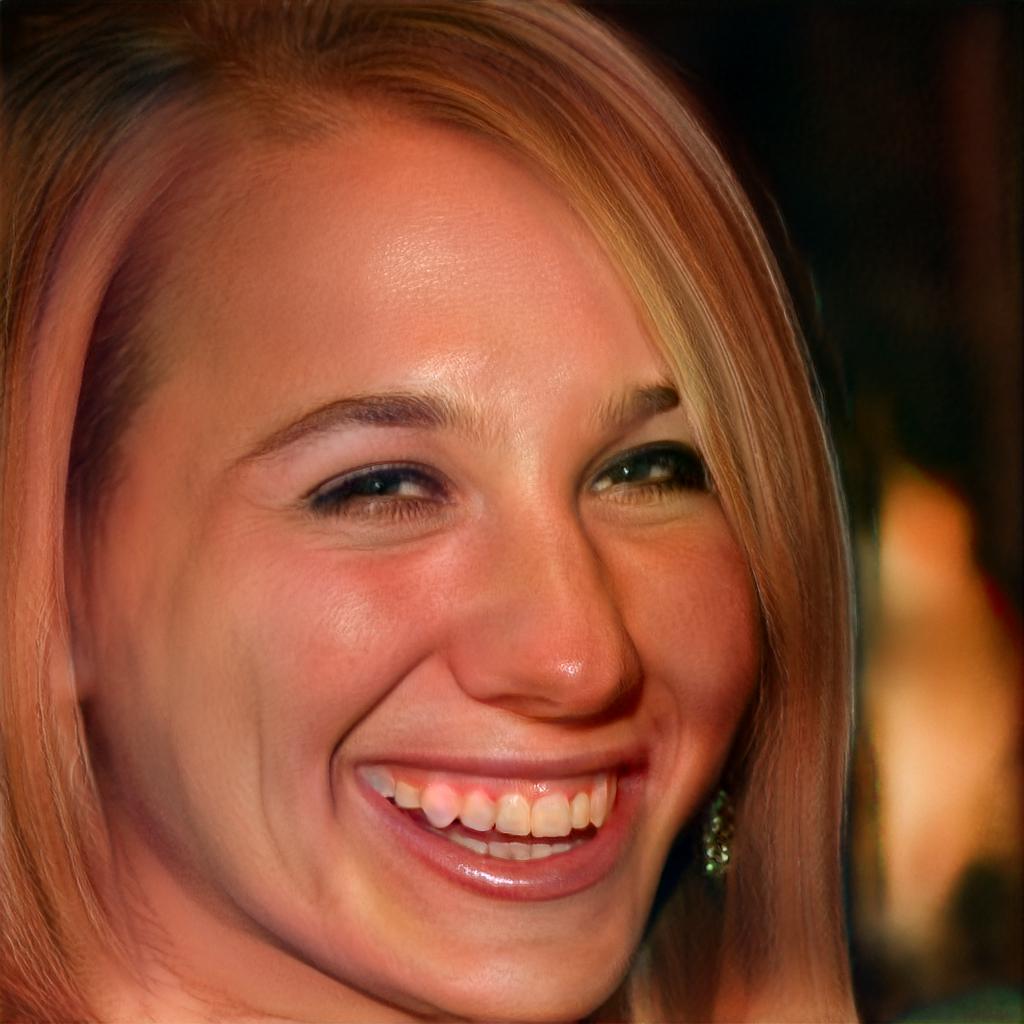} &
        \includegraphics[width=0.086\textwidth]{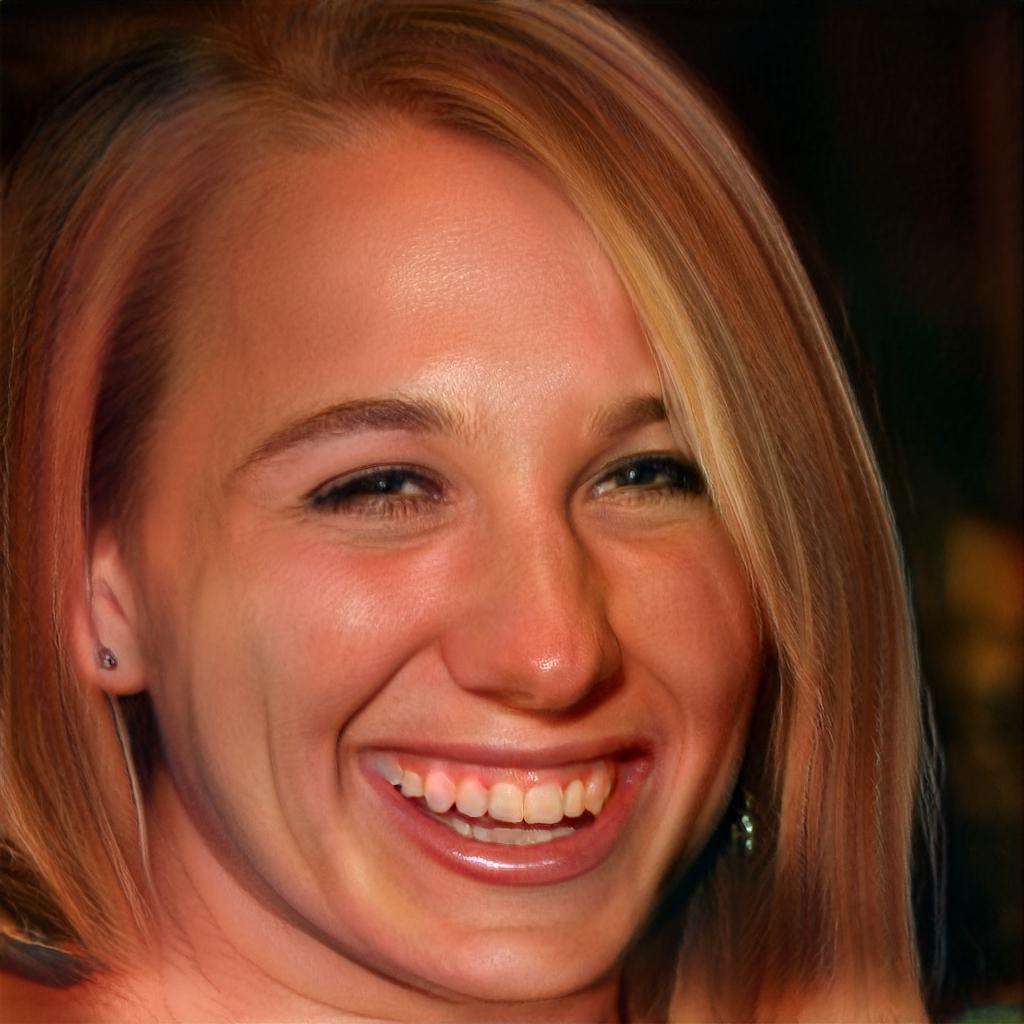} &
        \includegraphics[width=0.086\textwidth]{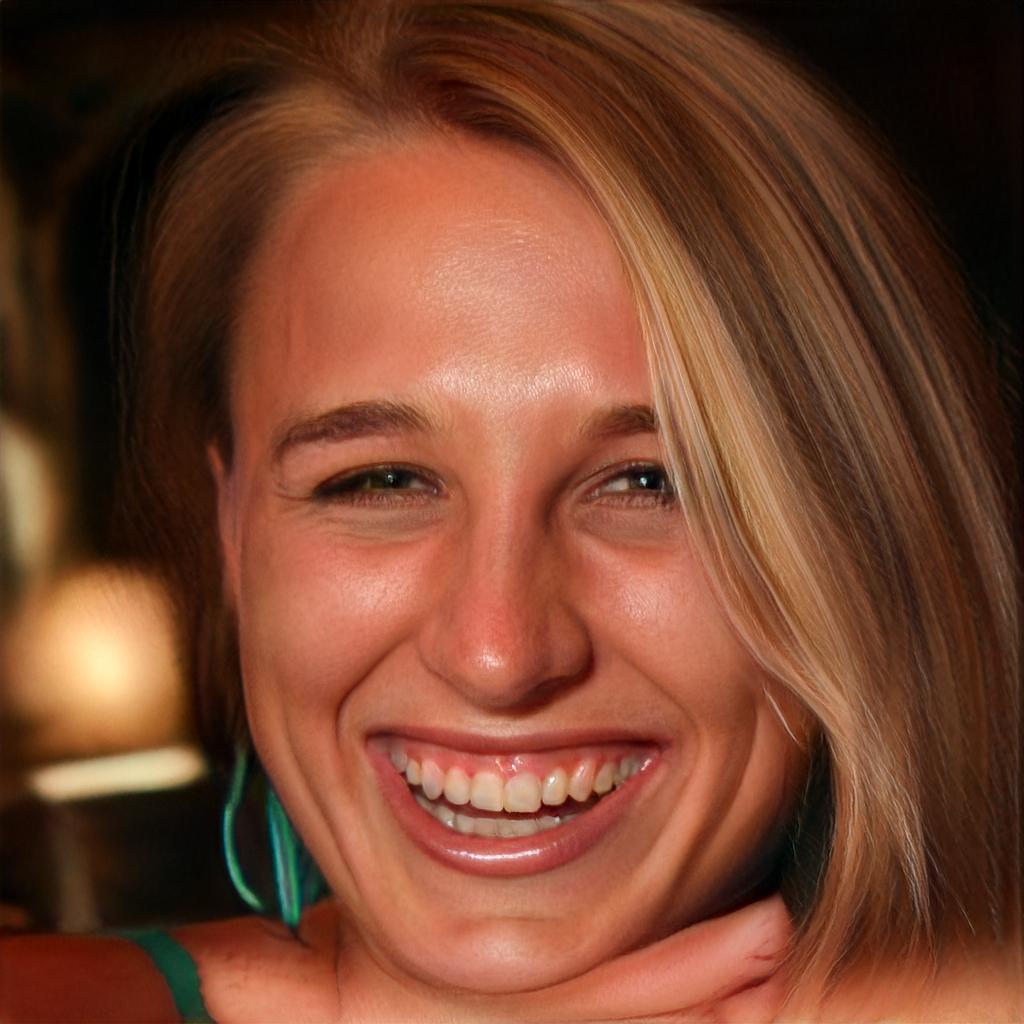} &
        \includegraphics[width=0.086\textwidth]{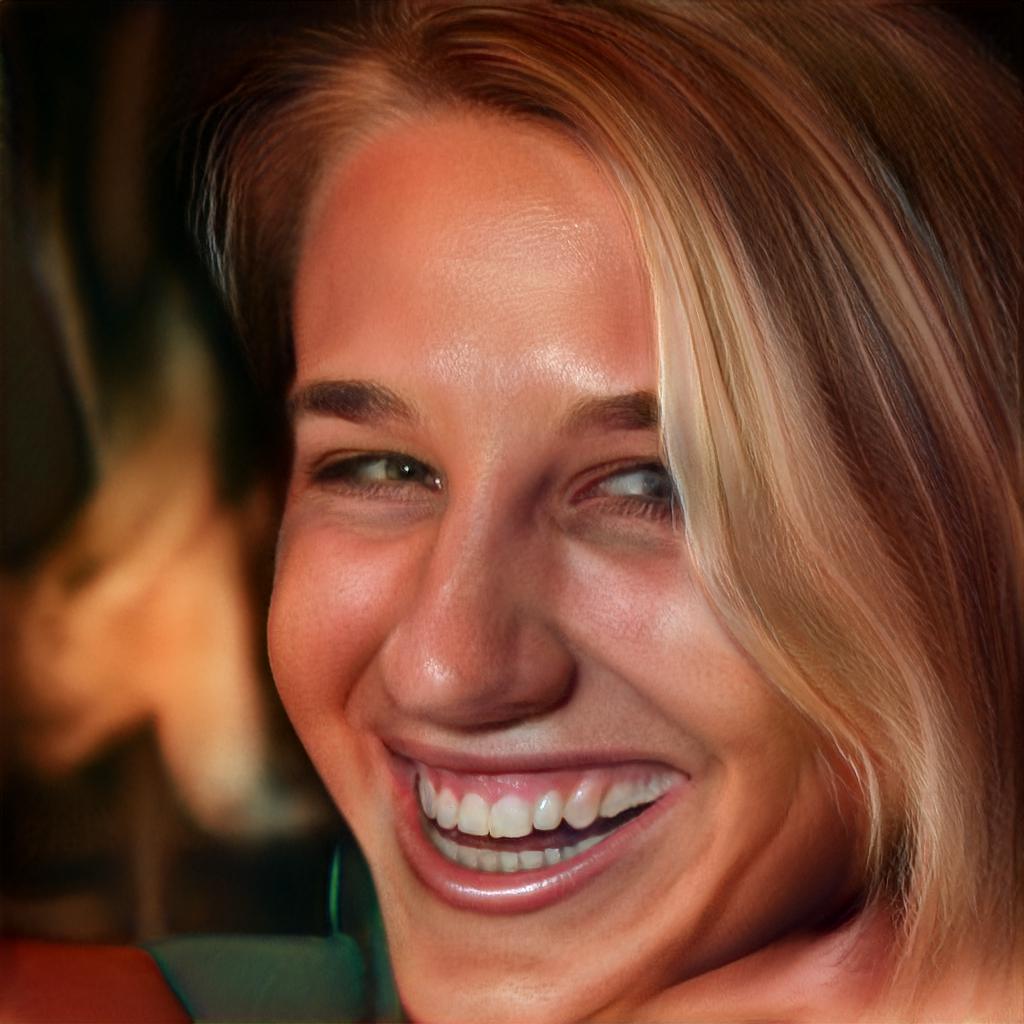} &
        \includegraphics[width=0.086\textwidth]{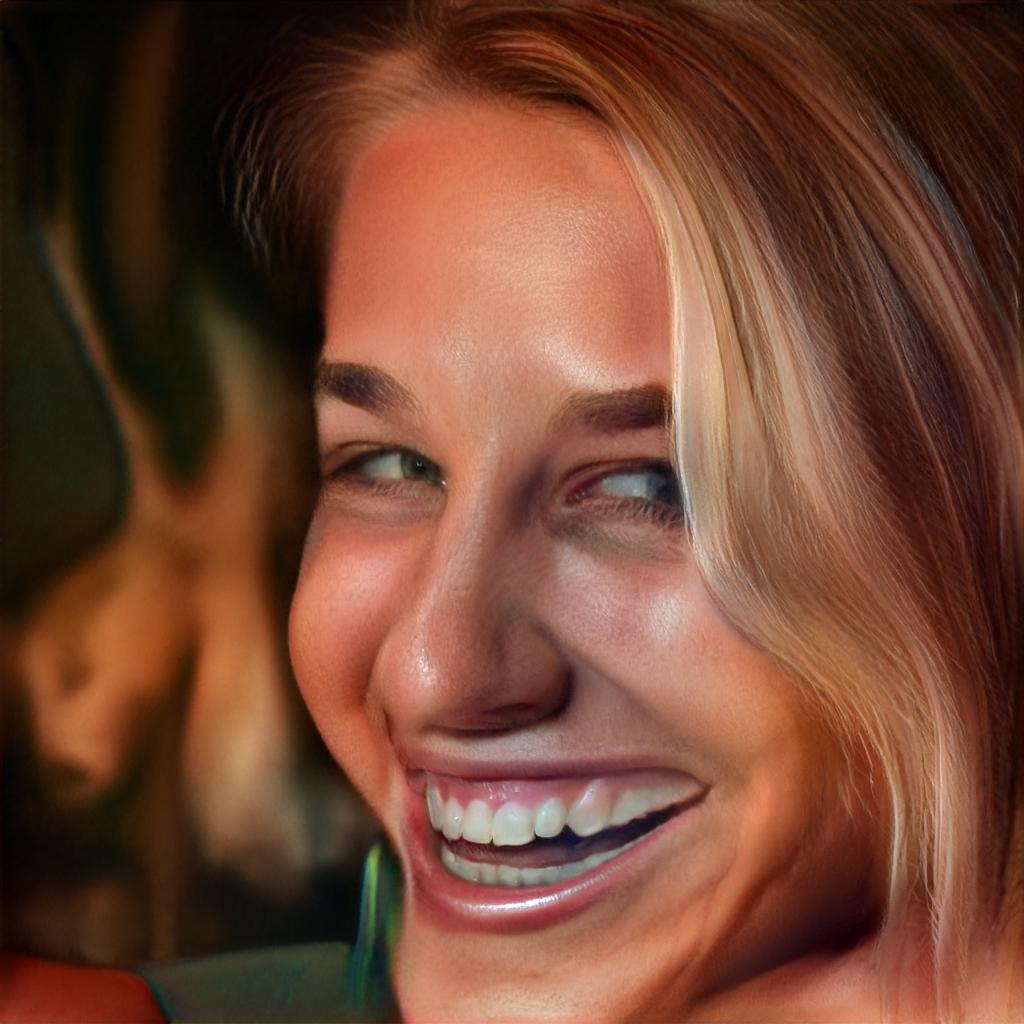} \\
        
        \raisebox{0.265in}{\rotatebox{90}{Ours}} &
        \includegraphics[width=0.086\textwidth]{resources/images/comparisons/faces/unedited/1/1.jpg} &
        \includegraphics[width=0.086\textwidth]{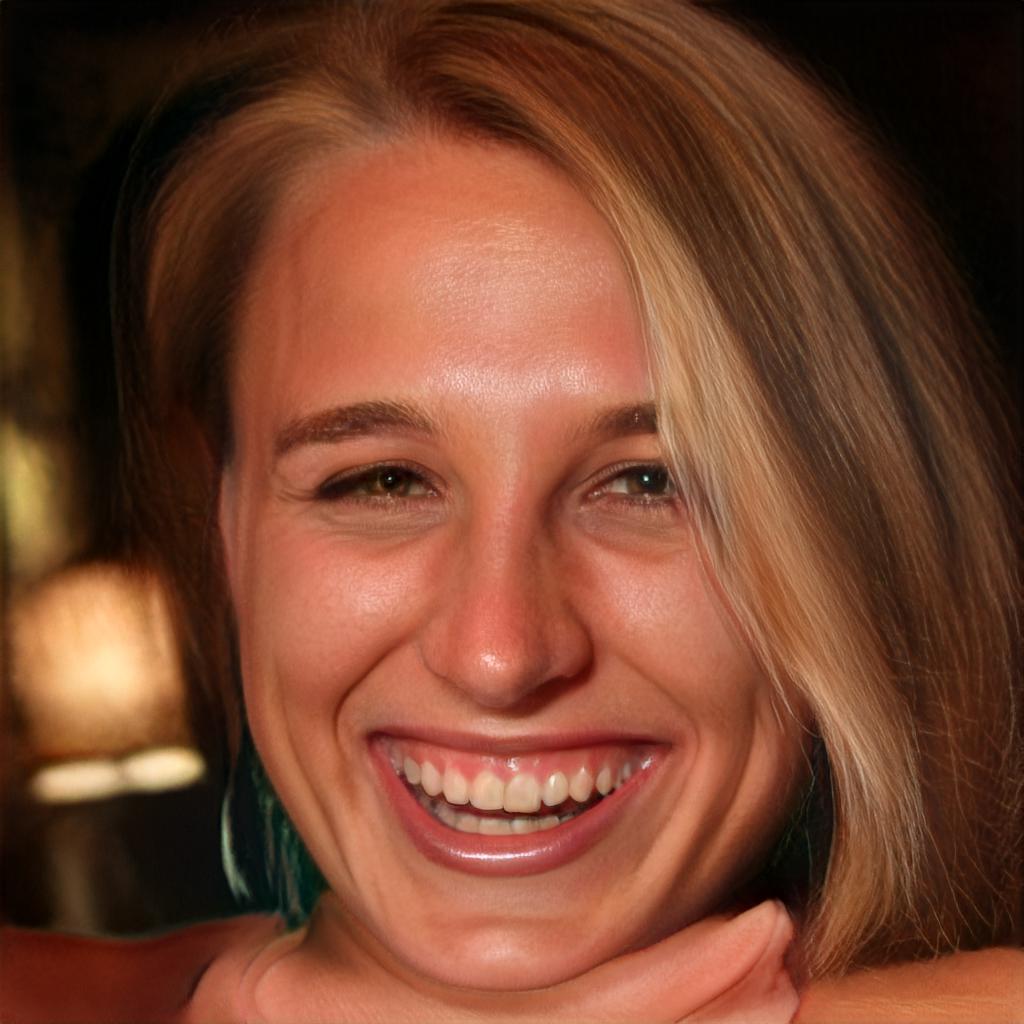} &
        \includegraphics[width=0.086\textwidth]{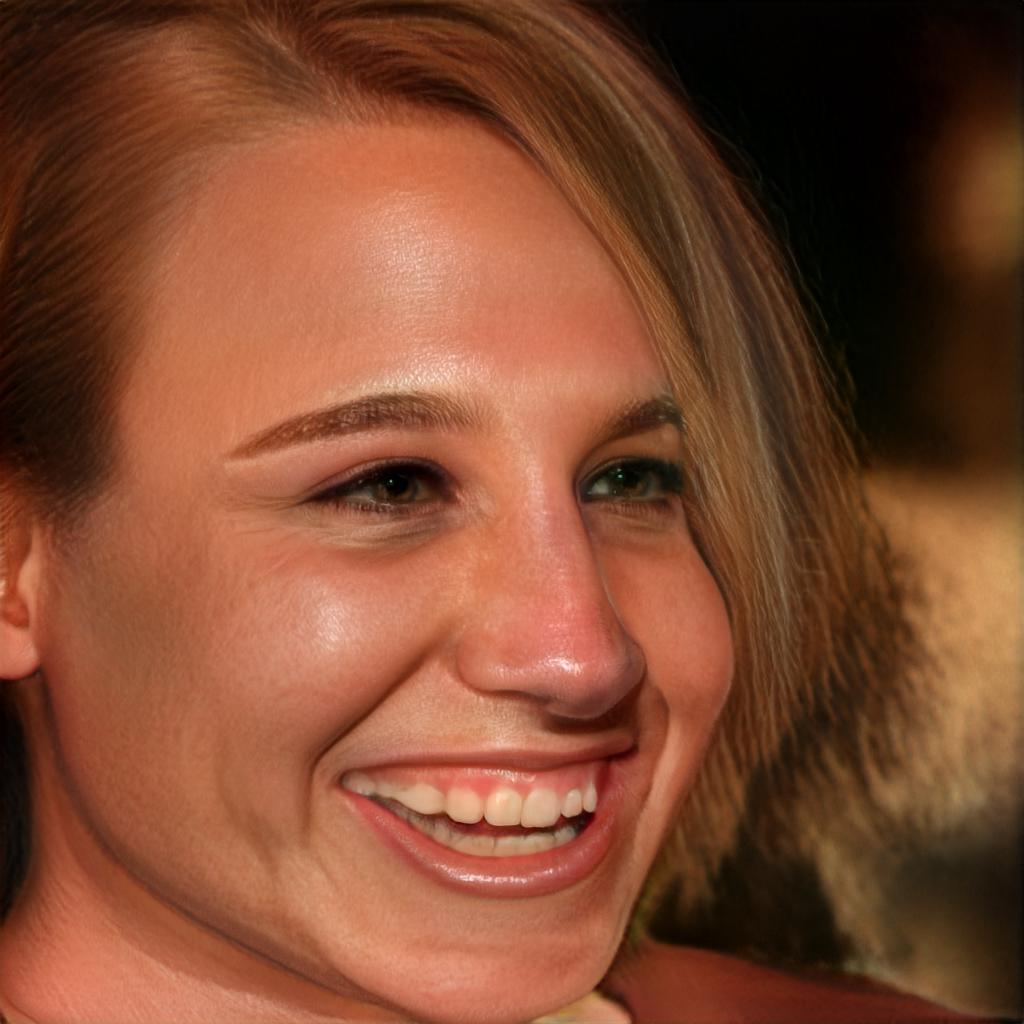} &
        \includegraphics[width=0.086\textwidth]{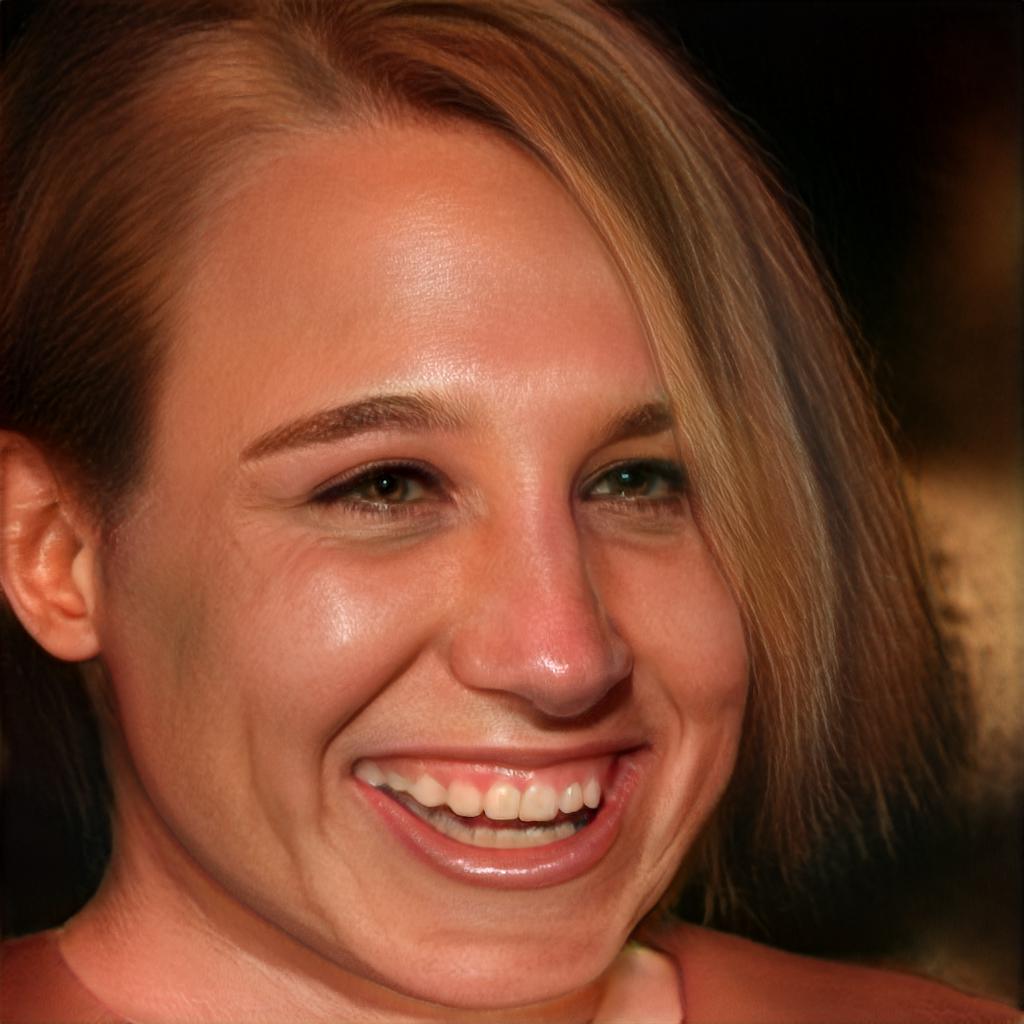} &
        \includegraphics[width=0.086\textwidth]{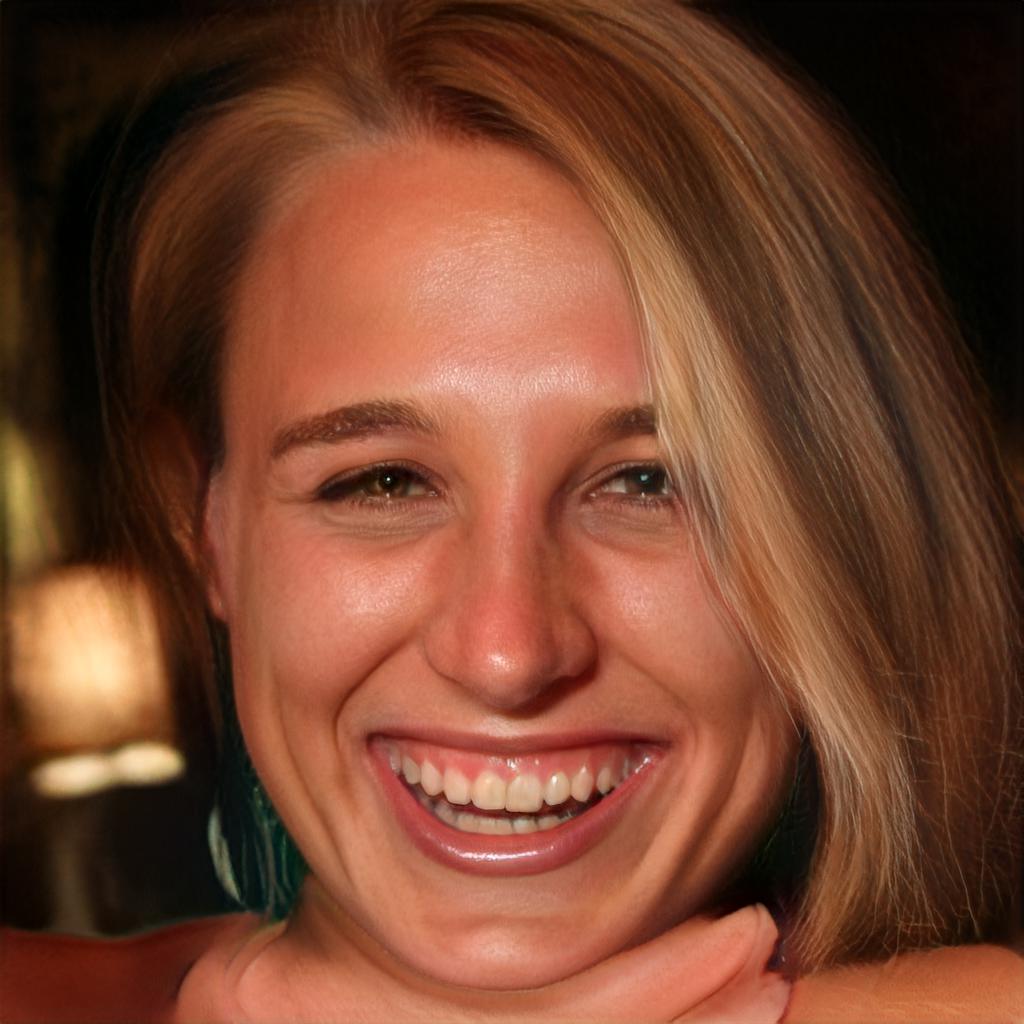} &
        \includegraphics[width=0.086\textwidth]{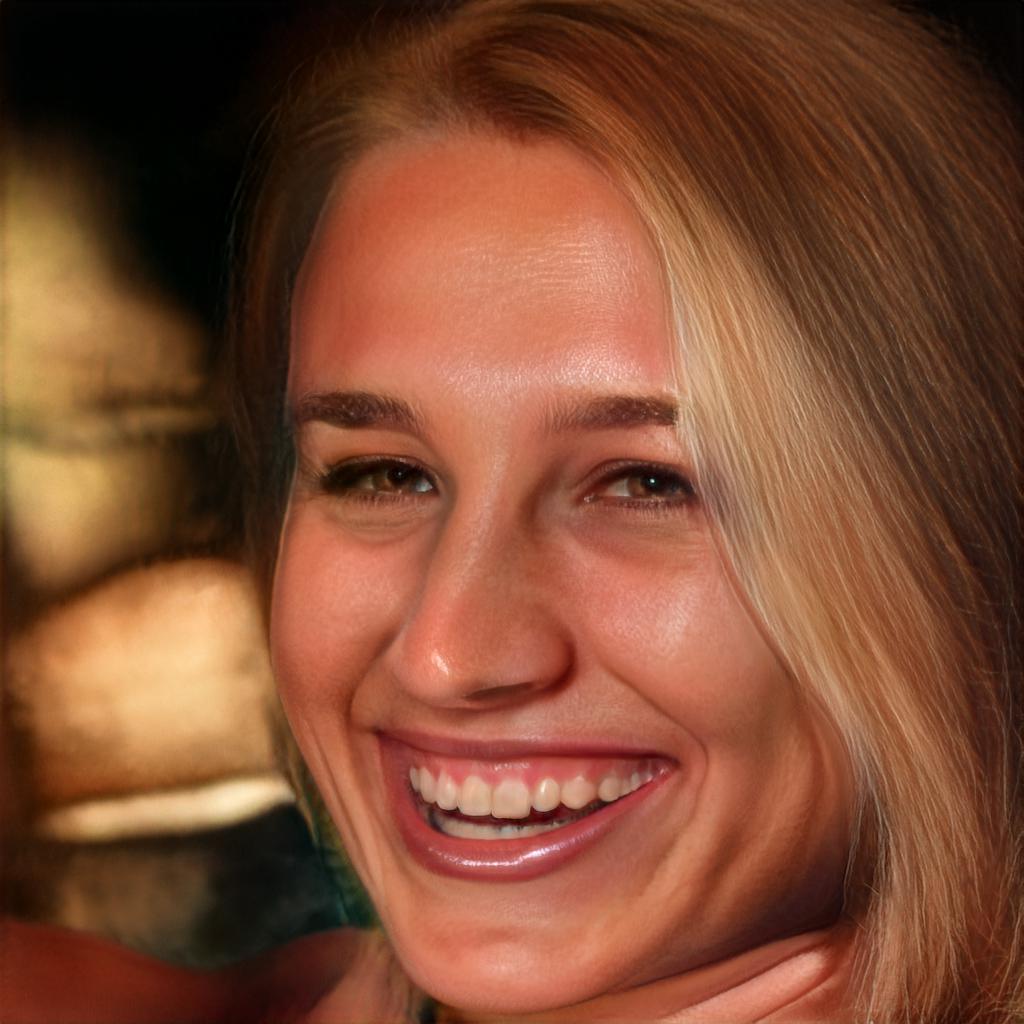} &
        \includegraphics[width=0.086\textwidth]{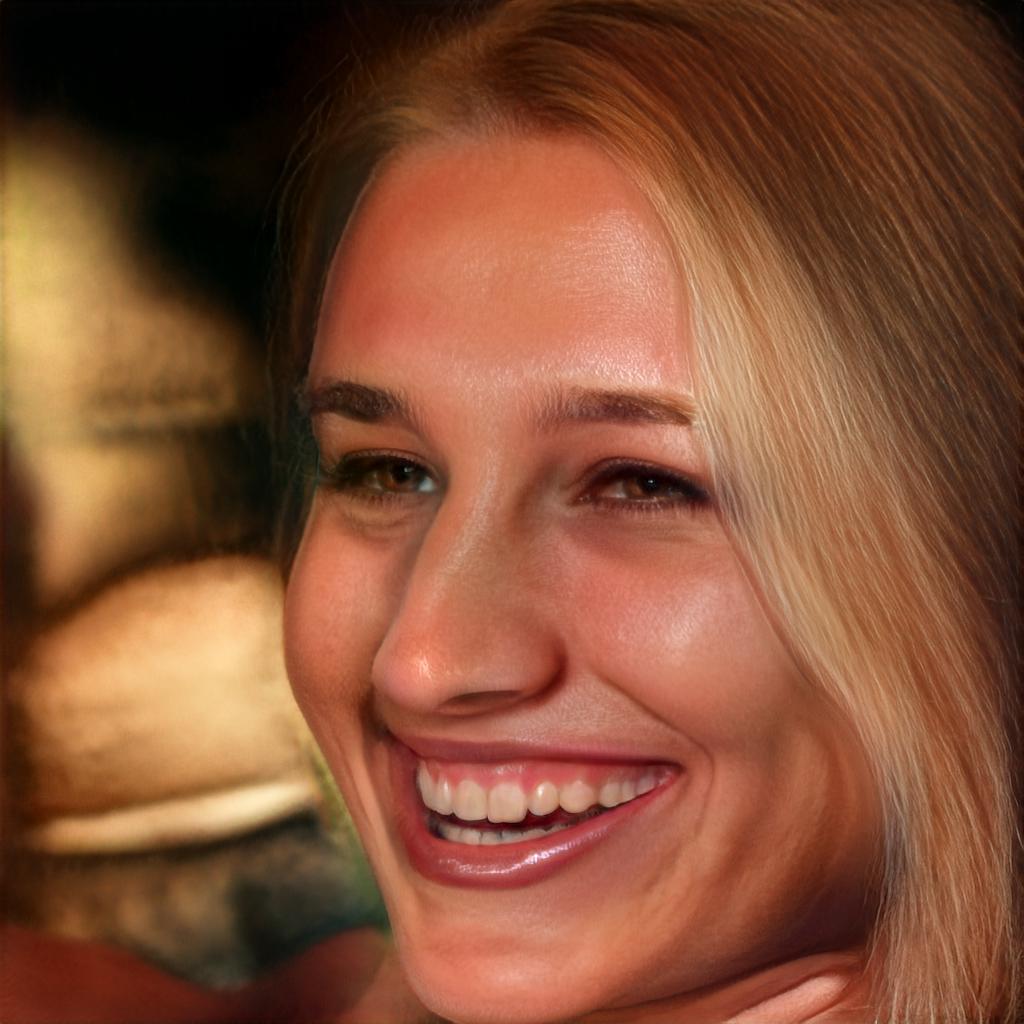} 

        \\

        \raisebox{0.265in}{\rotatebox{90}{\footnotesize SeFa}} &
        \includegraphics[width=0.086\textwidth]{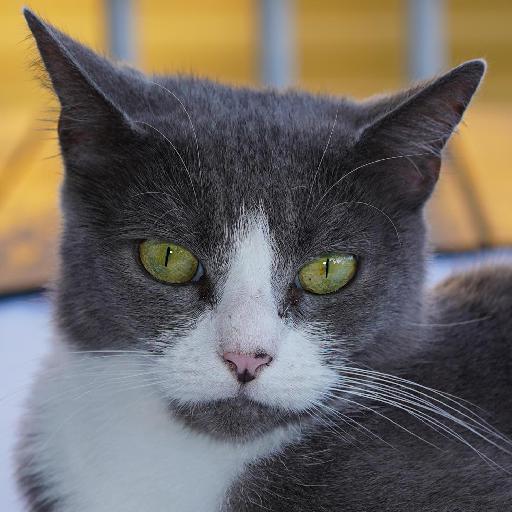} &
        \includegraphics[width=0.086\textwidth]{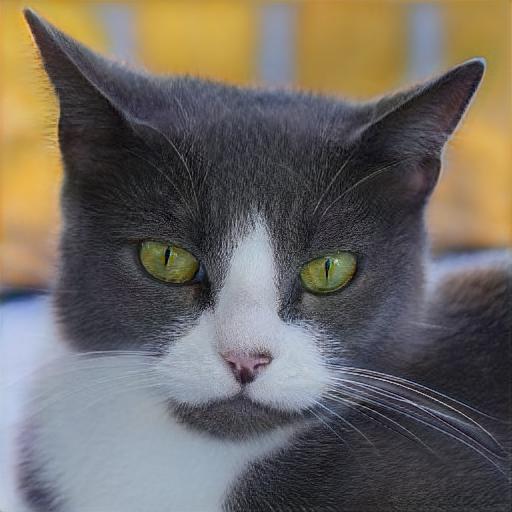} &
        \includegraphics[width=0.086\textwidth]{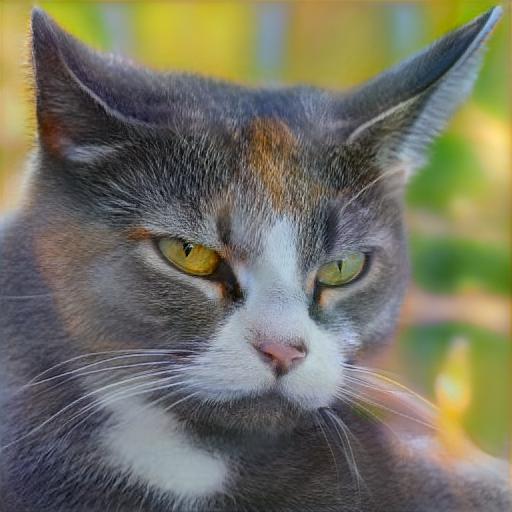} &
        \includegraphics[width=0.086\textwidth]{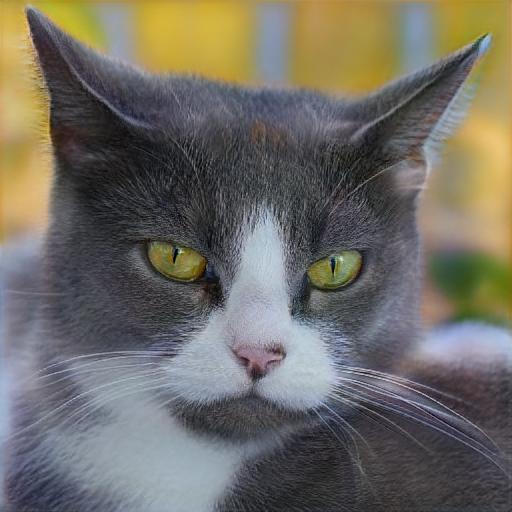} &
        \includegraphics[width=0.086\textwidth]{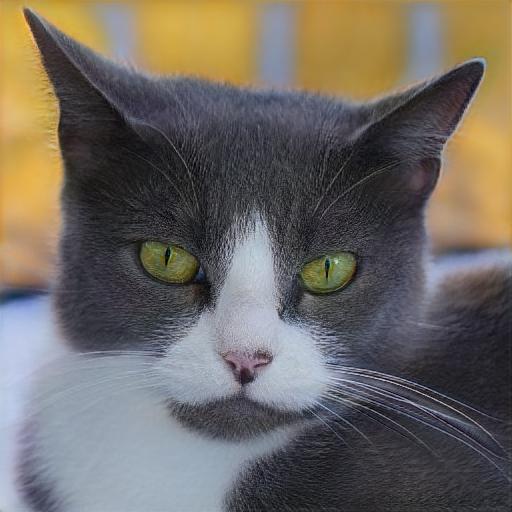} &
        \includegraphics[width=0.086\textwidth]{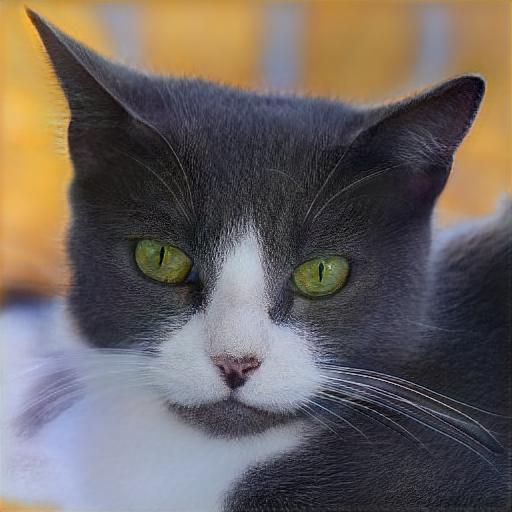} &
        \includegraphics[width=0.086\textwidth]{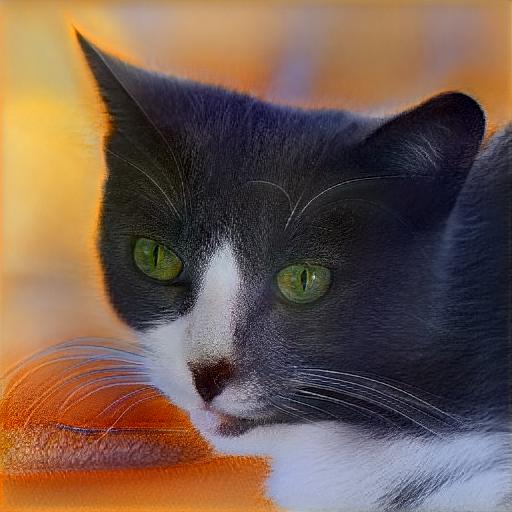} \\
        
        \raisebox{0.265in}{\rotatebox{90}{Ours}} &
        \includegraphics[width=0.086\textwidth]{resources/images/comparisons/cats/unedited/4/4.jpg} &
        \includegraphics[width=0.086\textwidth]{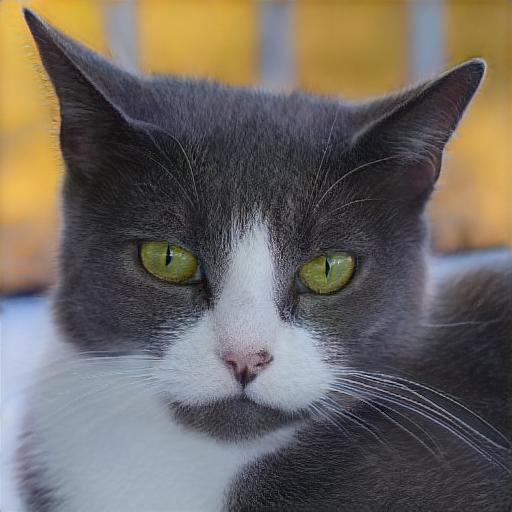} &
        \includegraphics[width=0.086\textwidth]{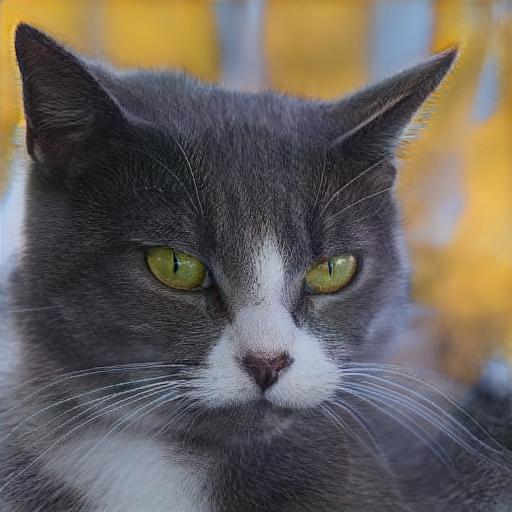} &
        \includegraphics[width=0.086\textwidth]{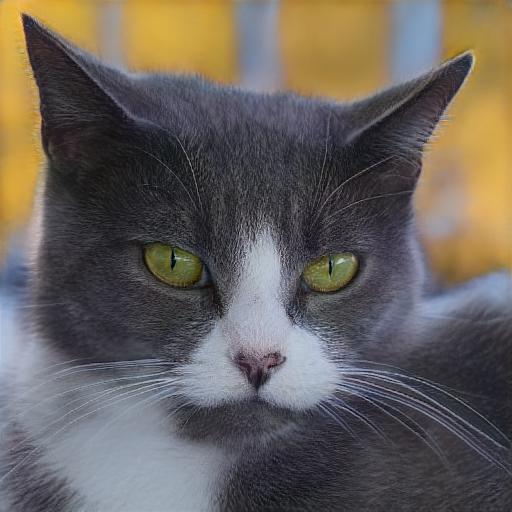} &
        \includegraphics[width=0.086\textwidth]{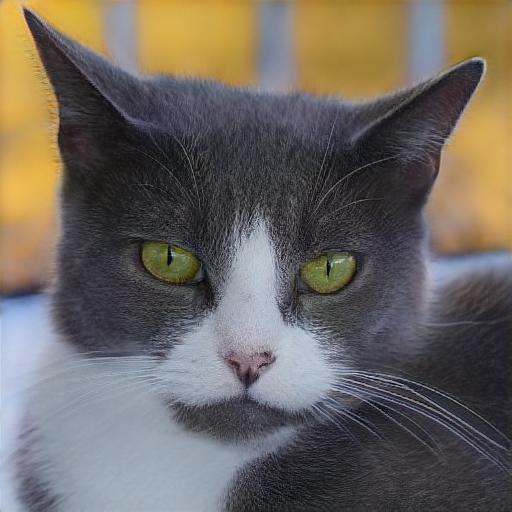} &
        \includegraphics[width=0.086\textwidth]{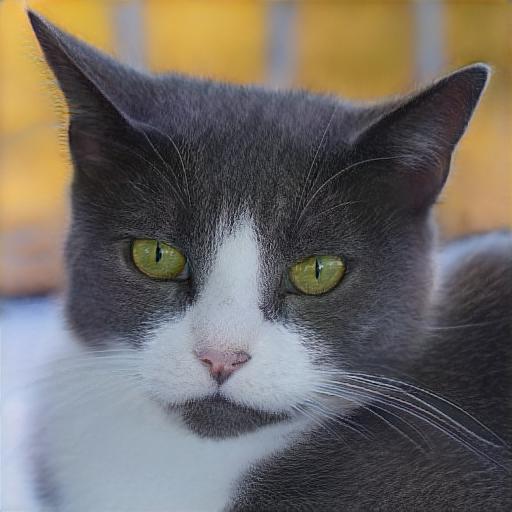} &
        \includegraphics[width=0.086\textwidth]{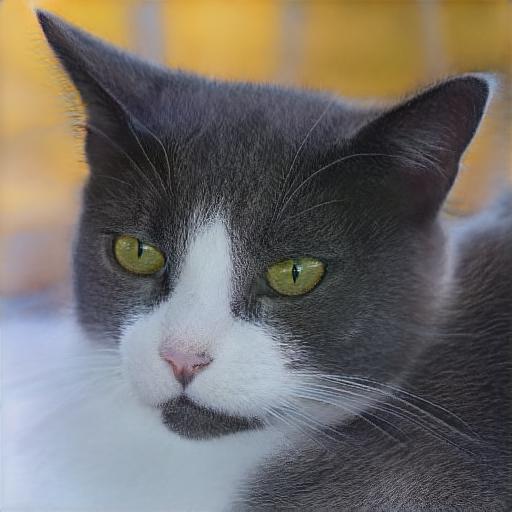}

        \\ & Input & Inversion & \multicolumn{5}{c}{$\myleftarrow$~Pose~$\myarrow$ } \\

        \raisebox{0.15in}{\rotatebox{90}{\footnotesize StyleCLIP}} &
        \includegraphics[width=0.086\textwidth]{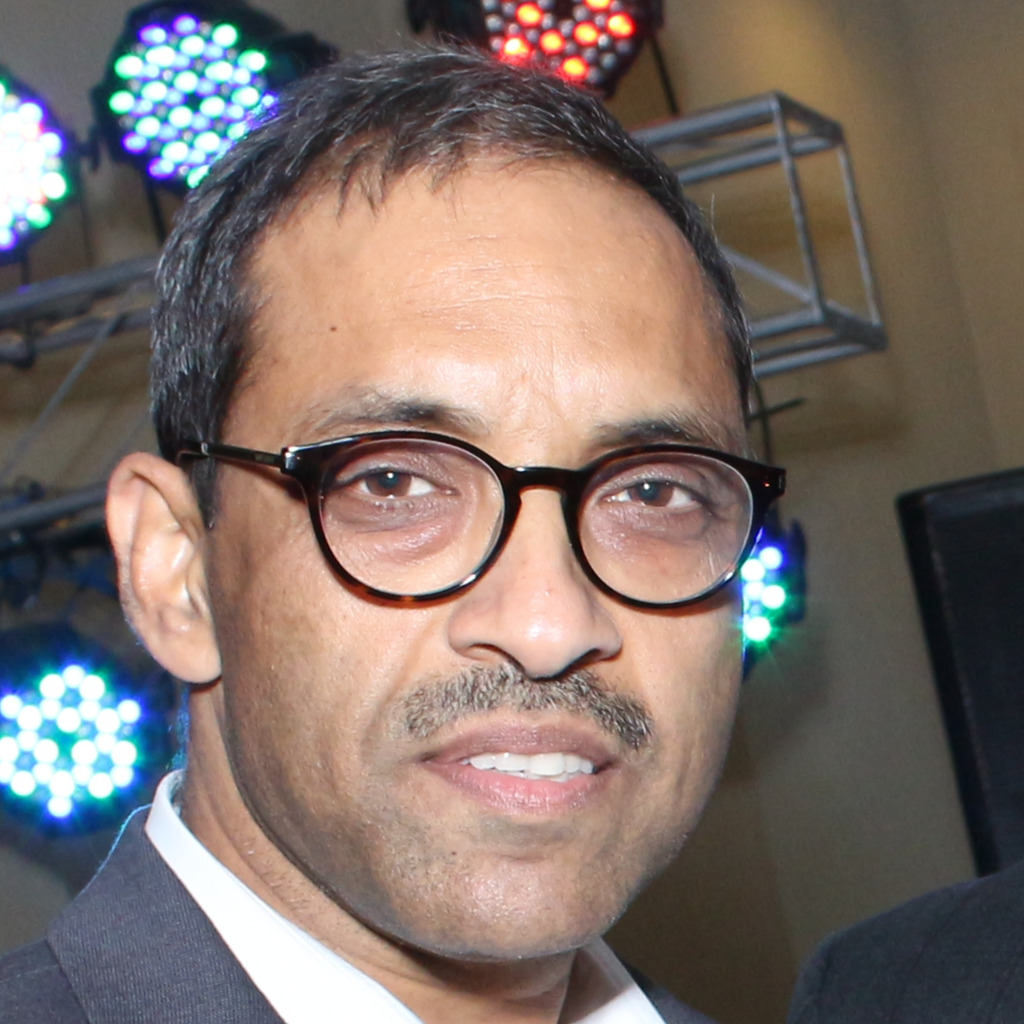} &
        \includegraphics[width=0.086\textwidth]{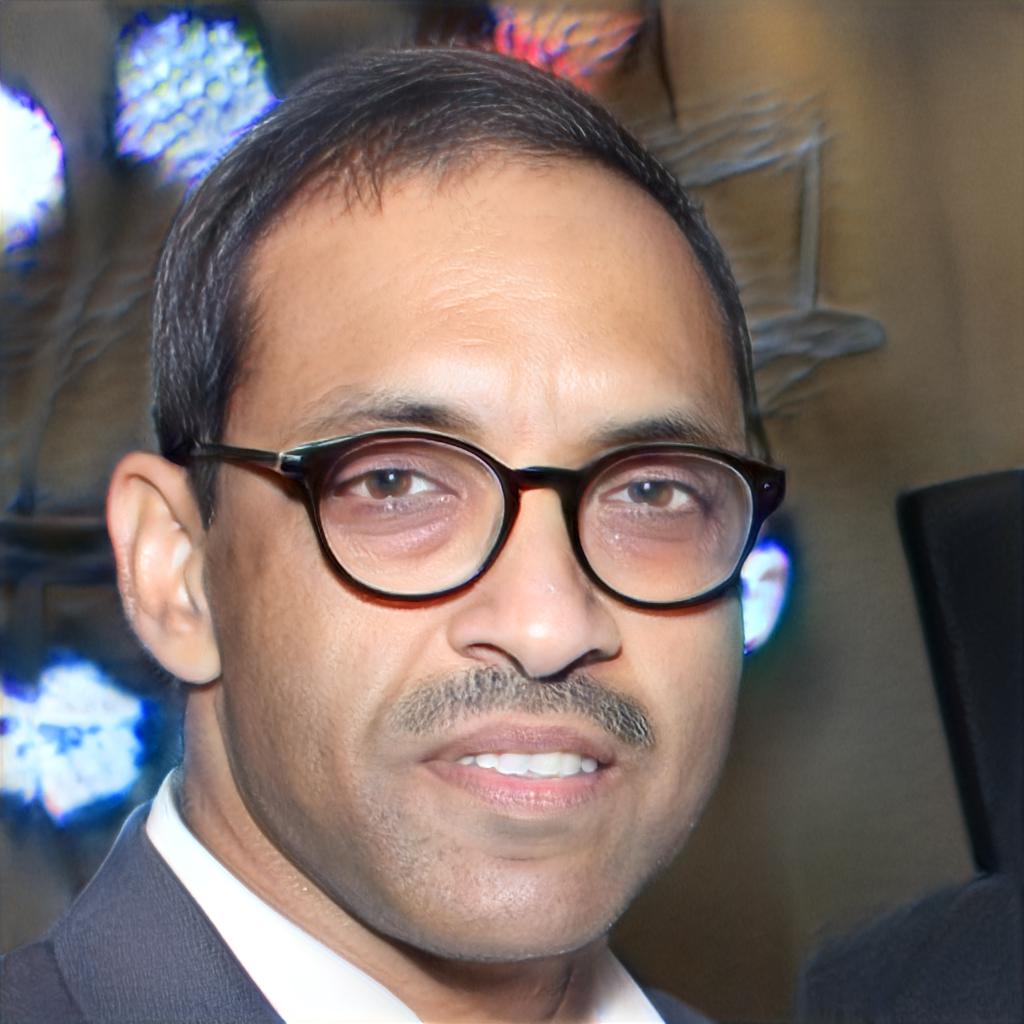} &
        \includegraphics[width=0.086\textwidth]{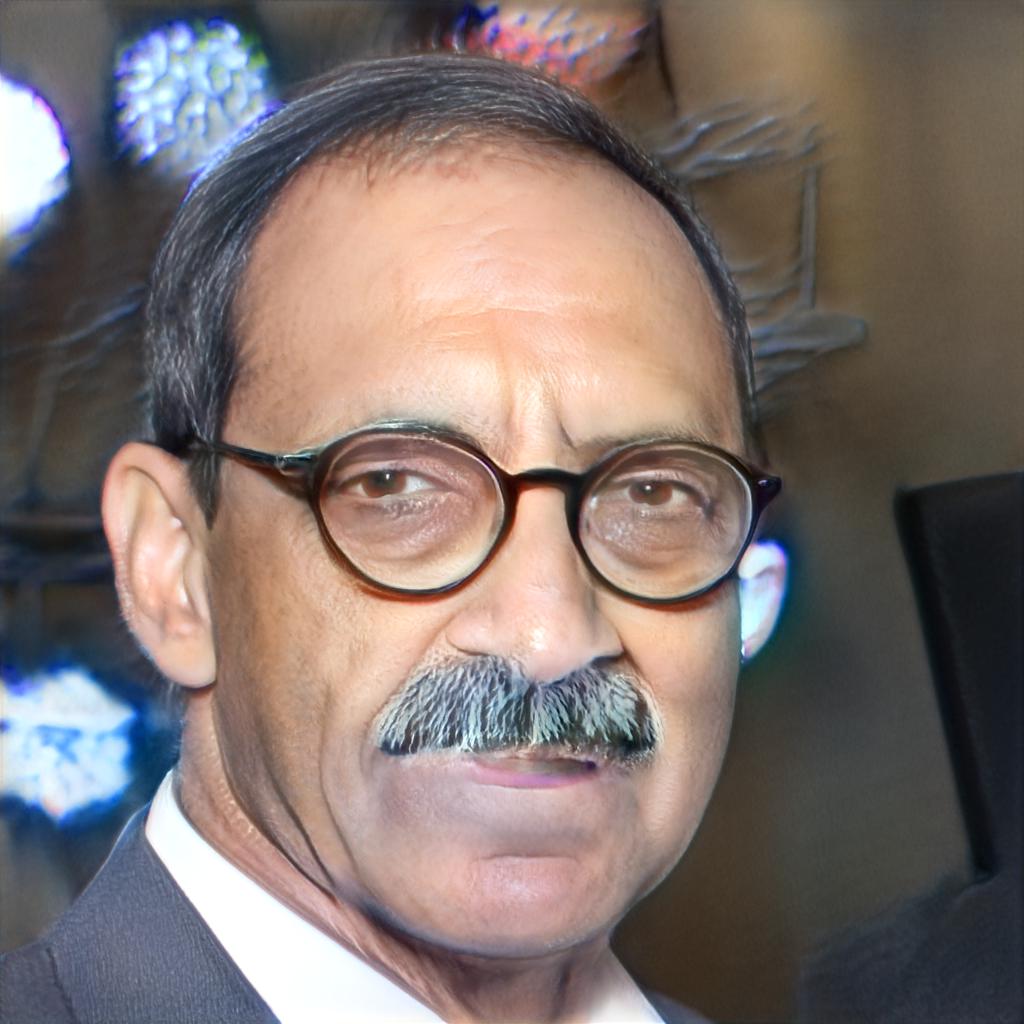} &
        \includegraphics[width=0.086\textwidth]{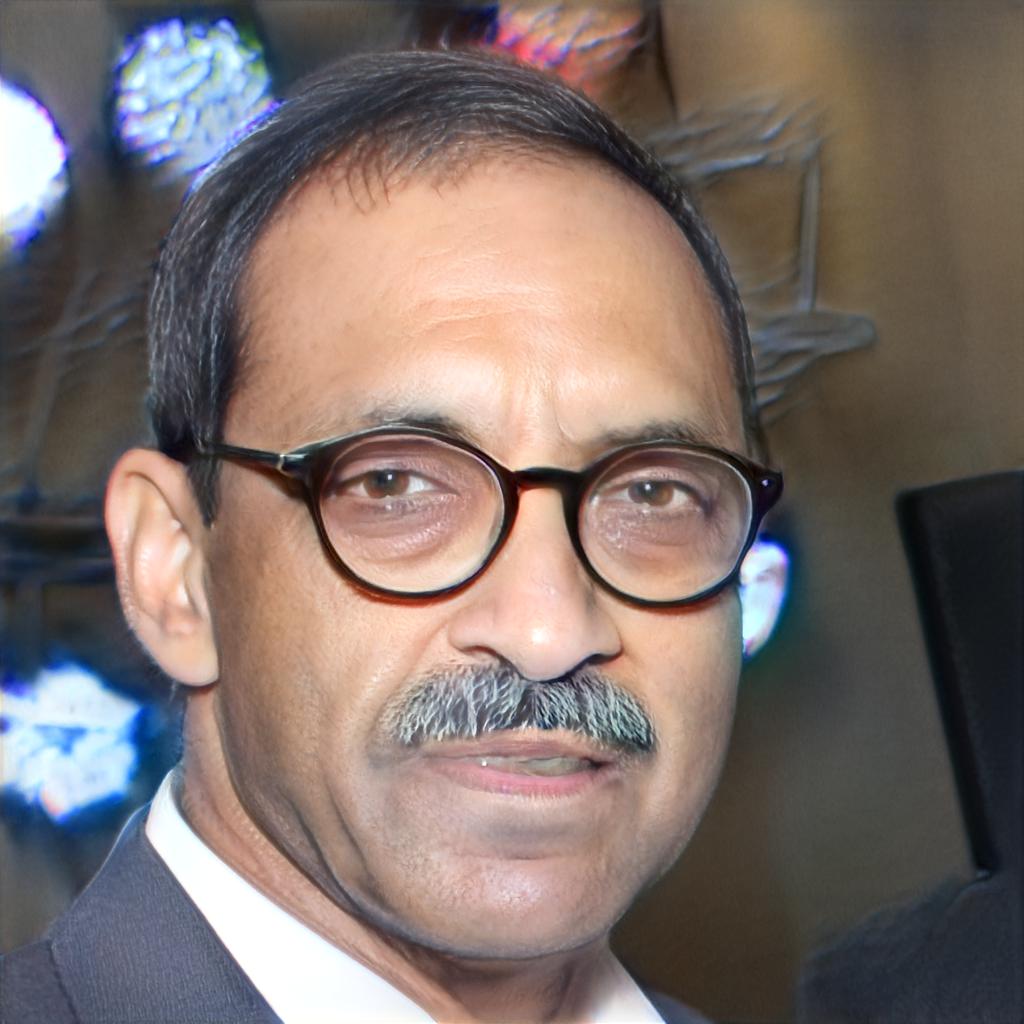} &
        \includegraphics[width=0.086\textwidth]{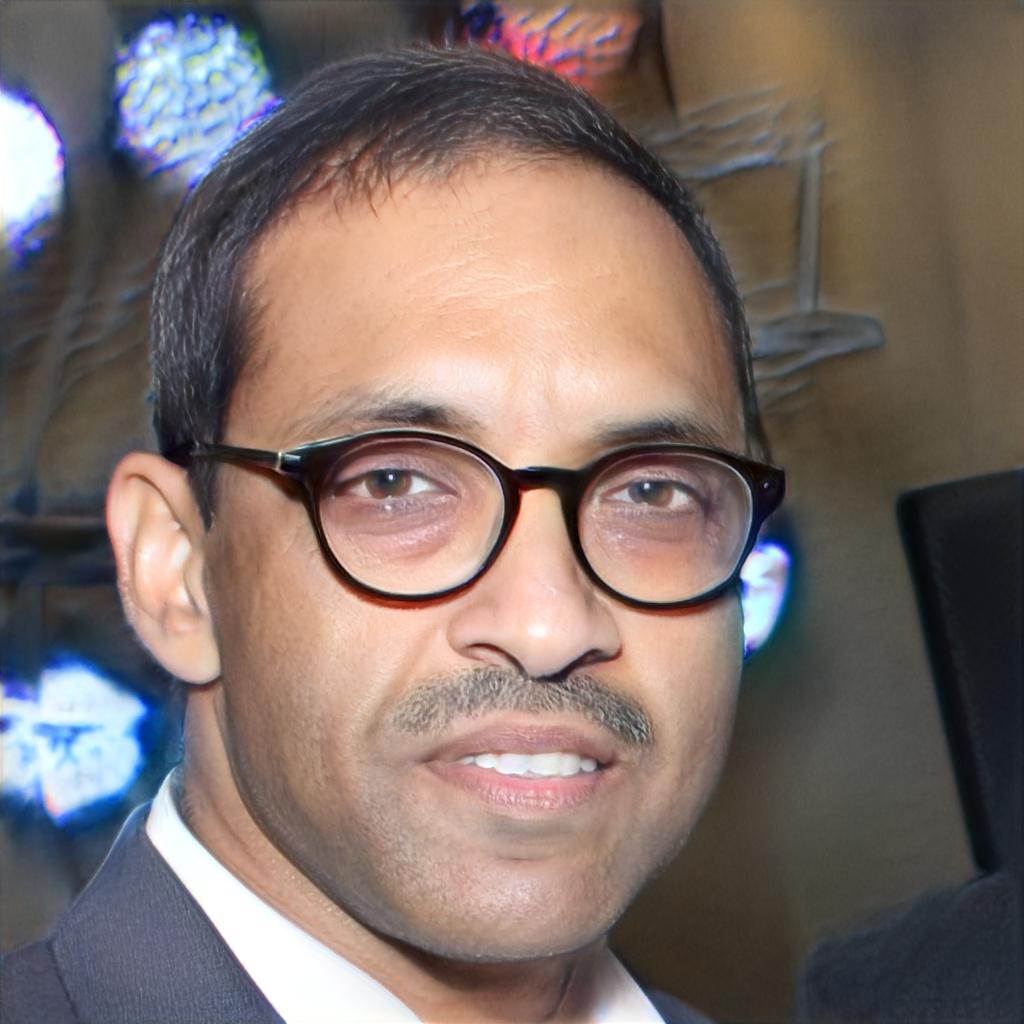} &
        \includegraphics[width=0.086\textwidth]{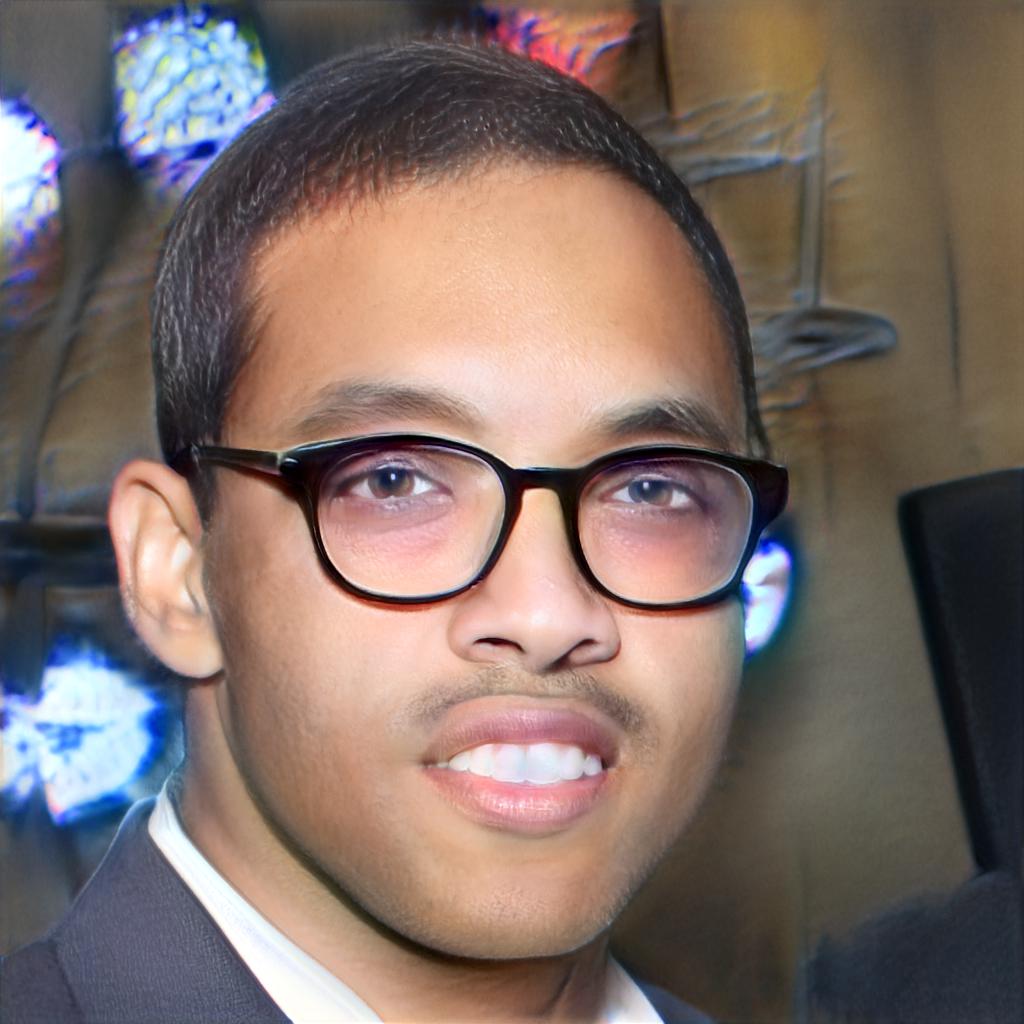} &
        \includegraphics[width=0.086\textwidth]{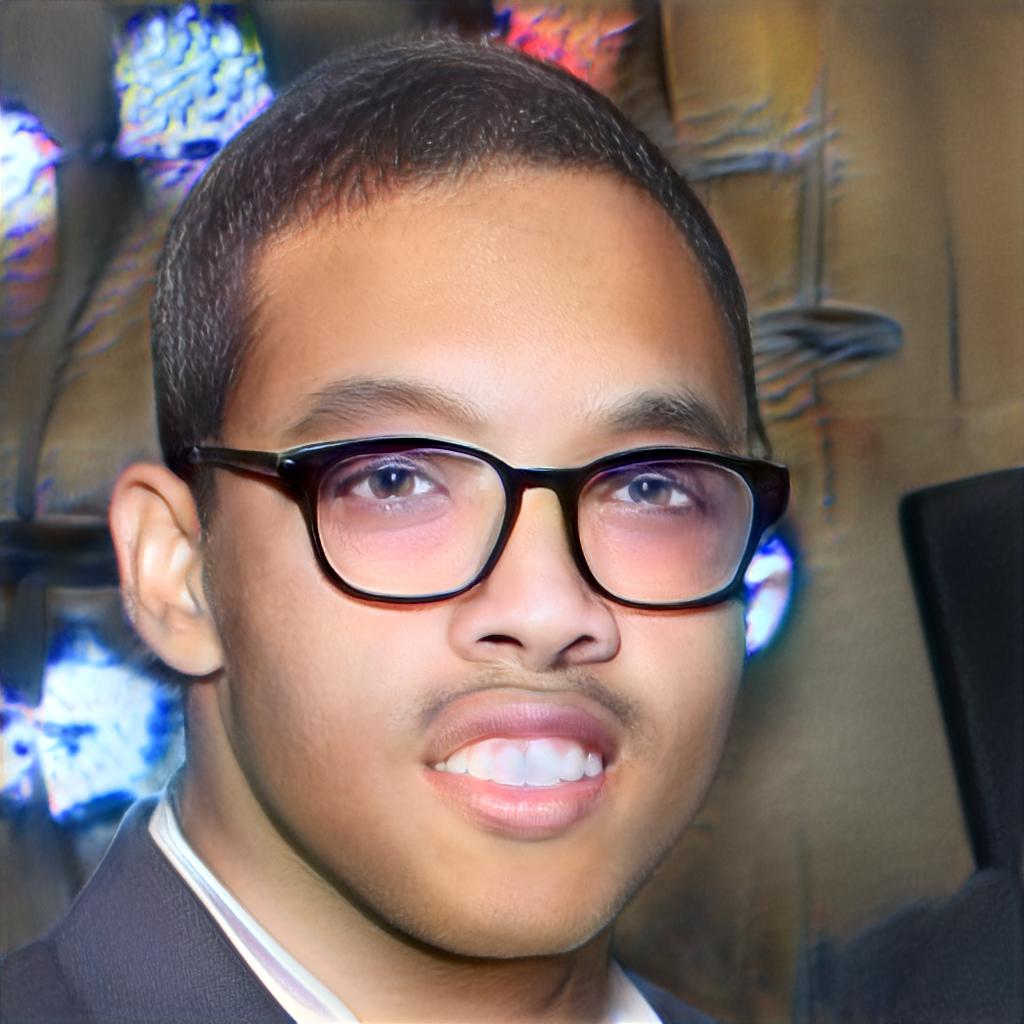} \\
        
        \raisebox{0.265in}{\rotatebox{90}{Ours}} &
        \includegraphics[width=0.086\textwidth]{resources/images/comparisons/faces/unedited/52/52.jpg} &
        \includegraphics[width=0.086\textwidth]{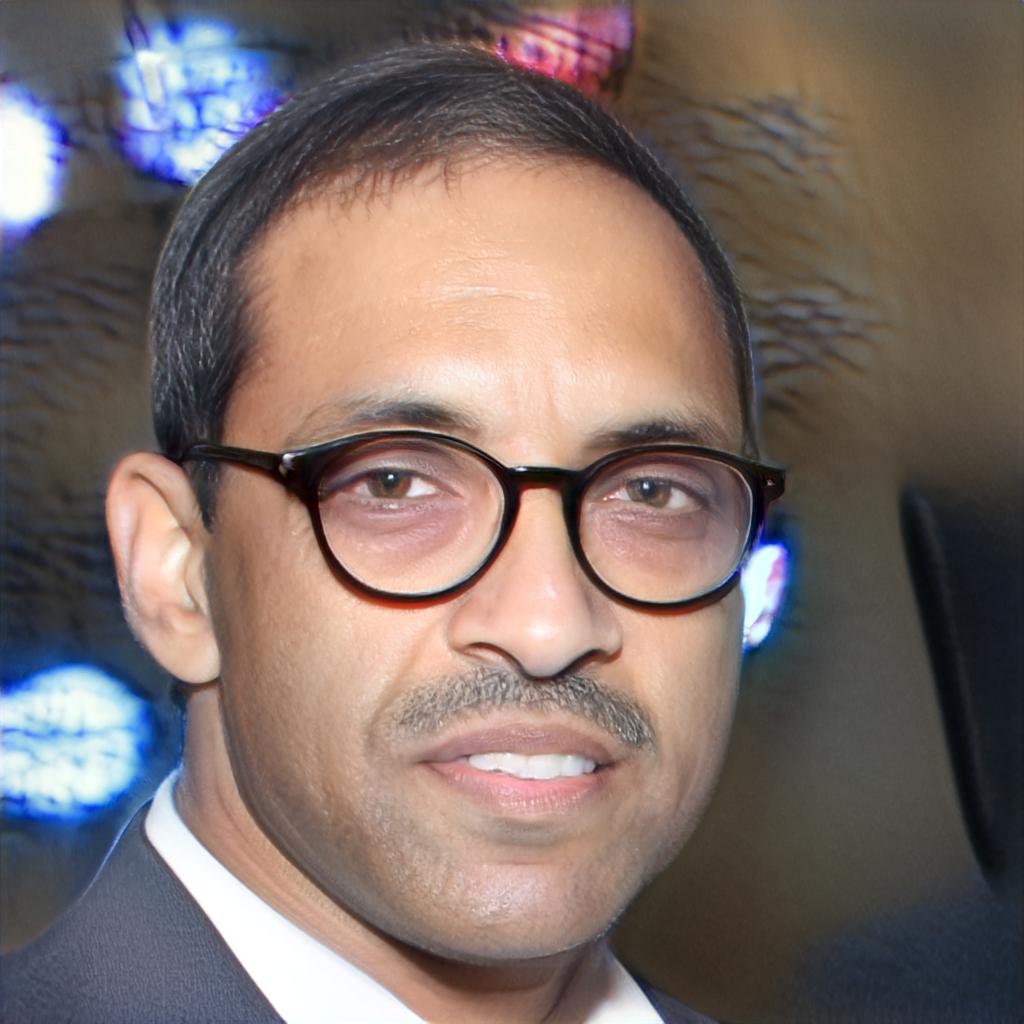} &
        \includegraphics[width=0.086\textwidth]{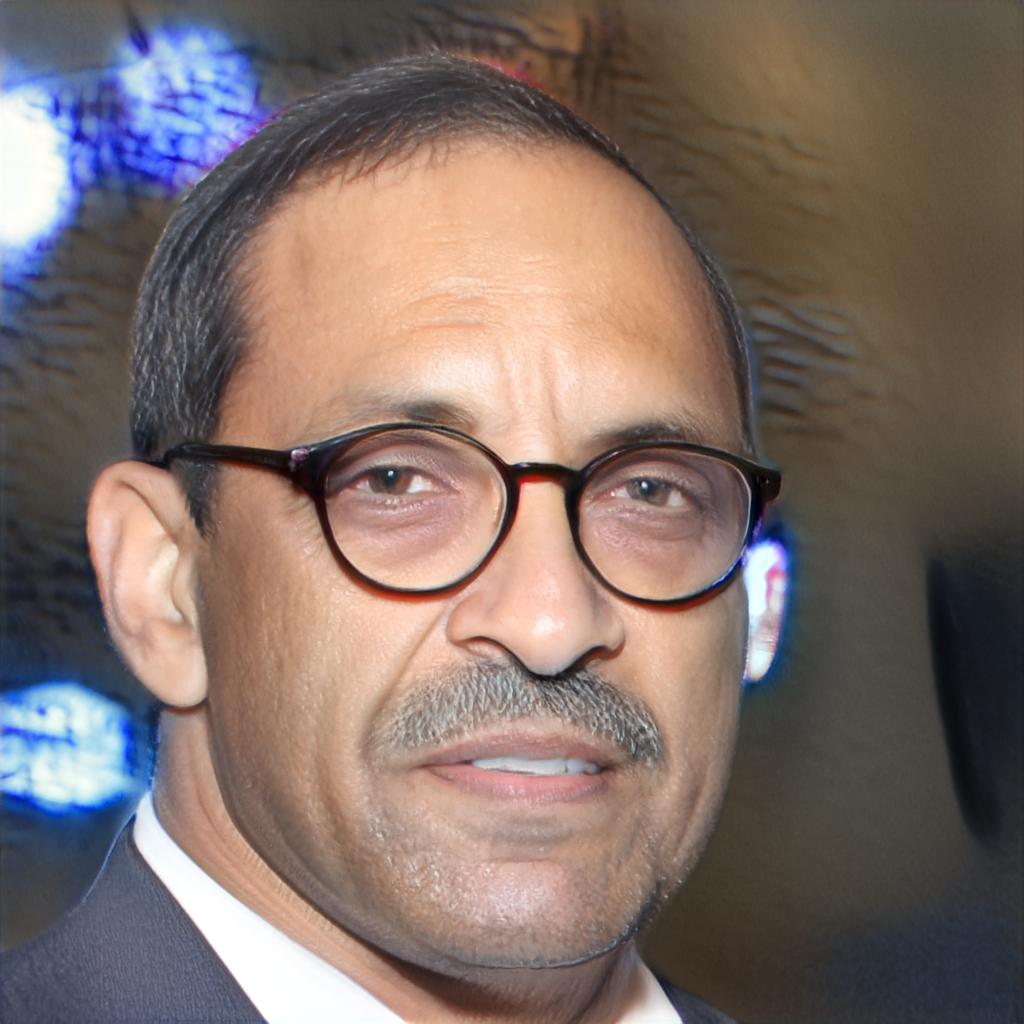} &
        \includegraphics[width=0.086\textwidth]{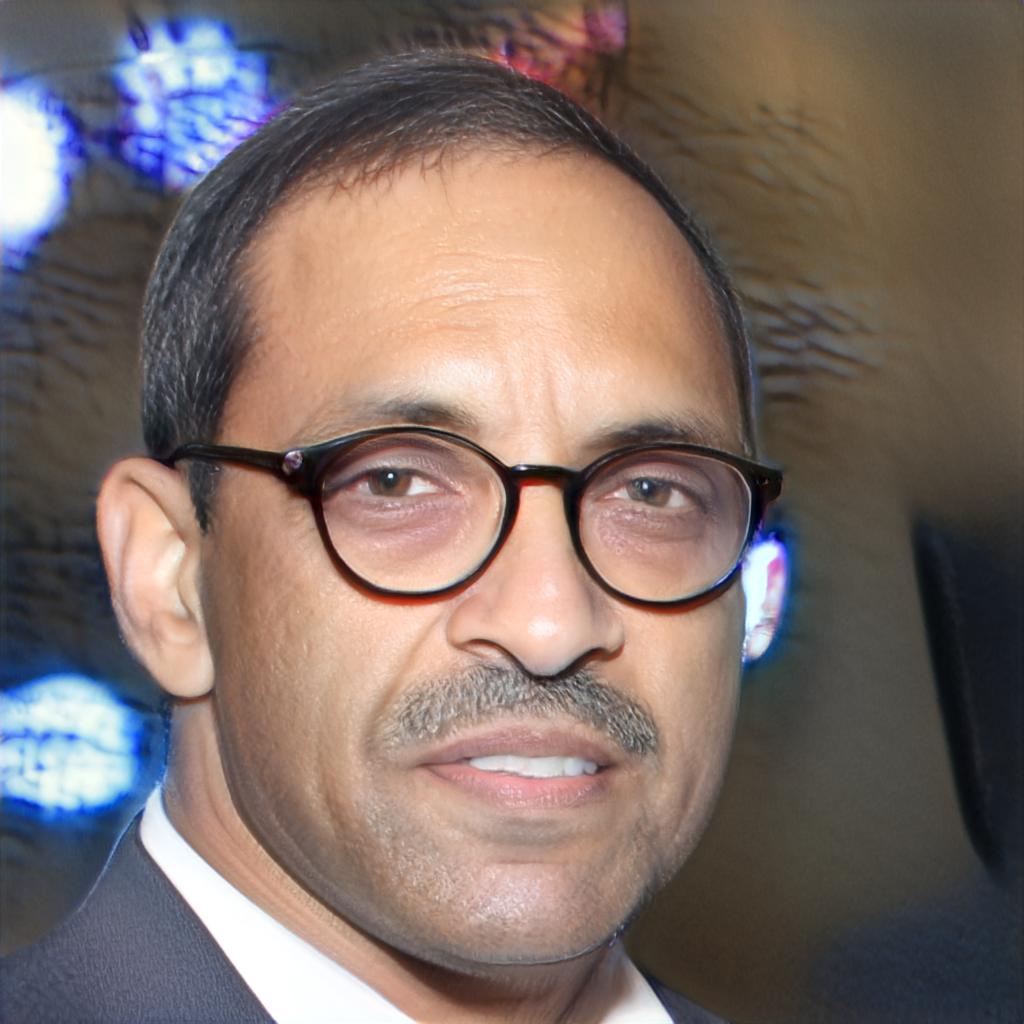} &
        \includegraphics[width=0.086\textwidth]{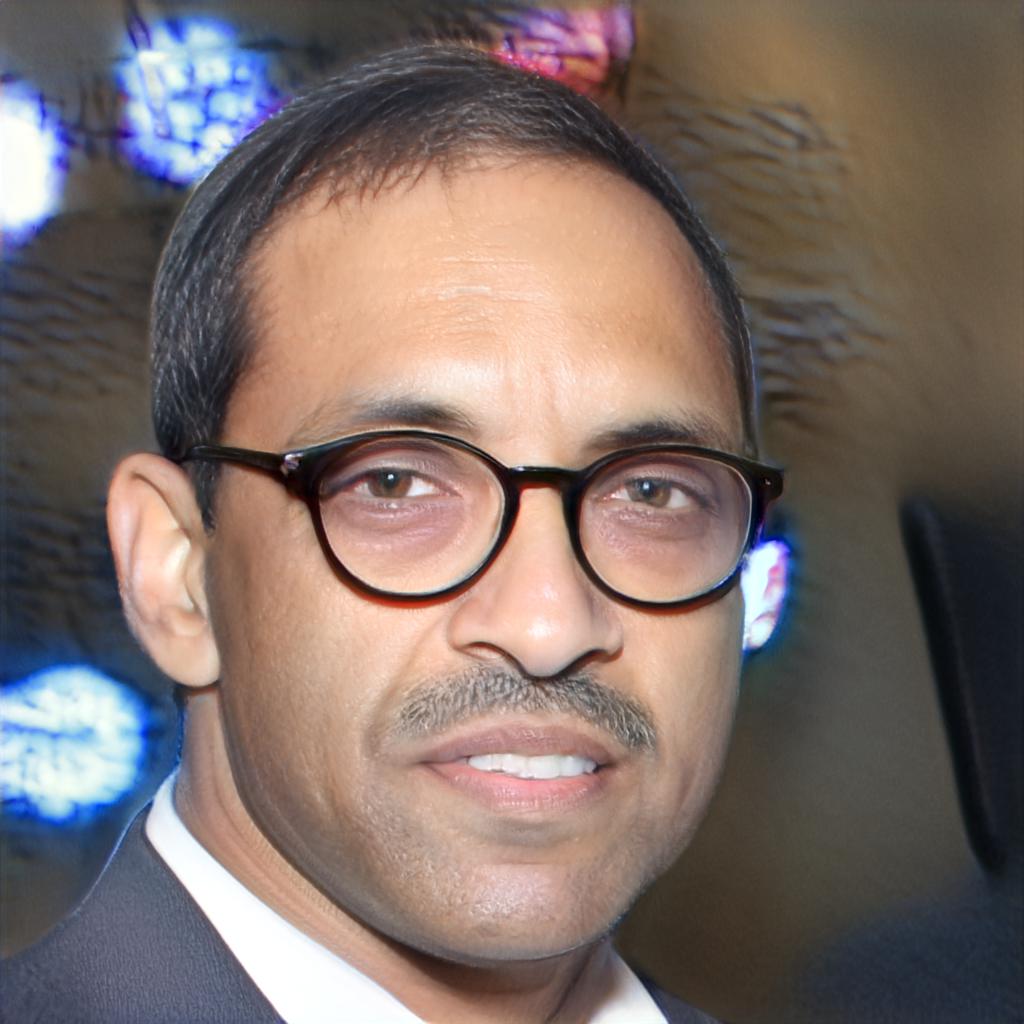} &
        \includegraphics[width=0.086\textwidth]{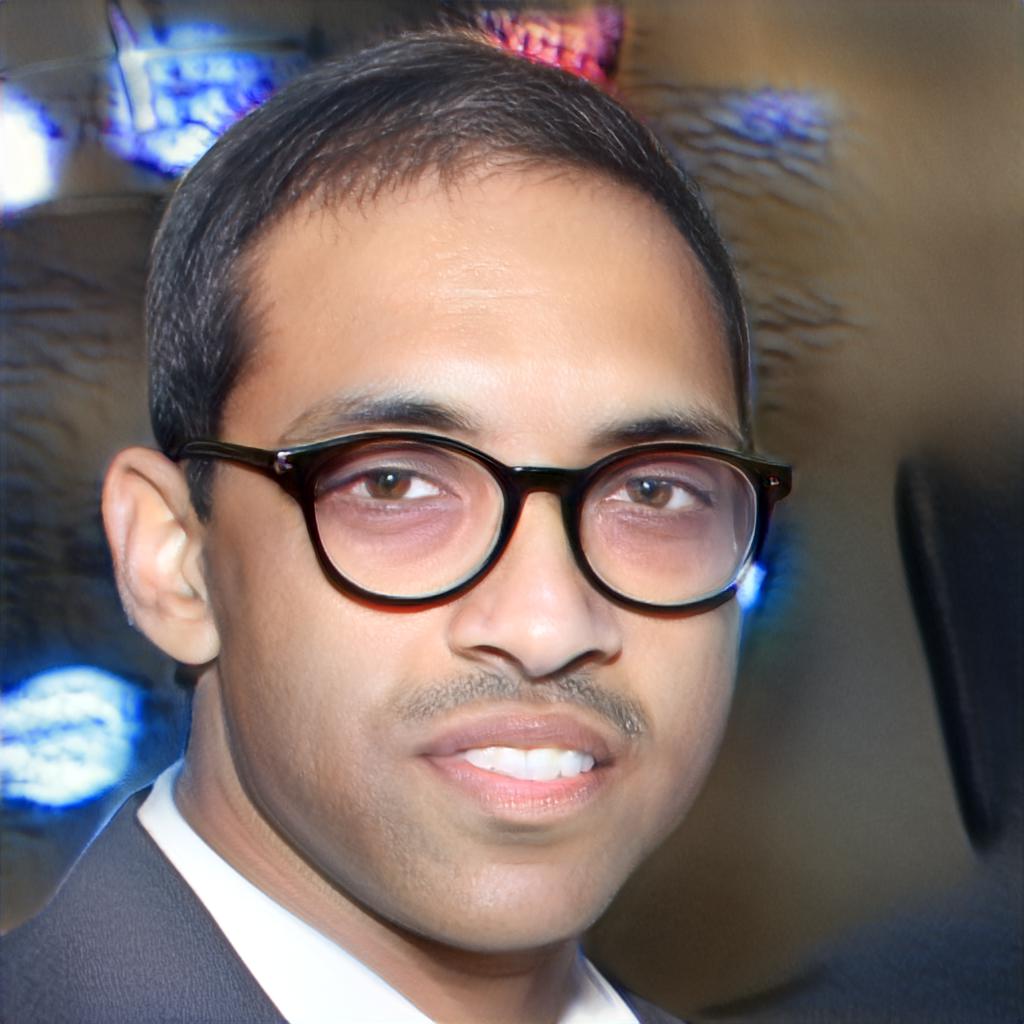} &
        \includegraphics[width=0.086\textwidth]{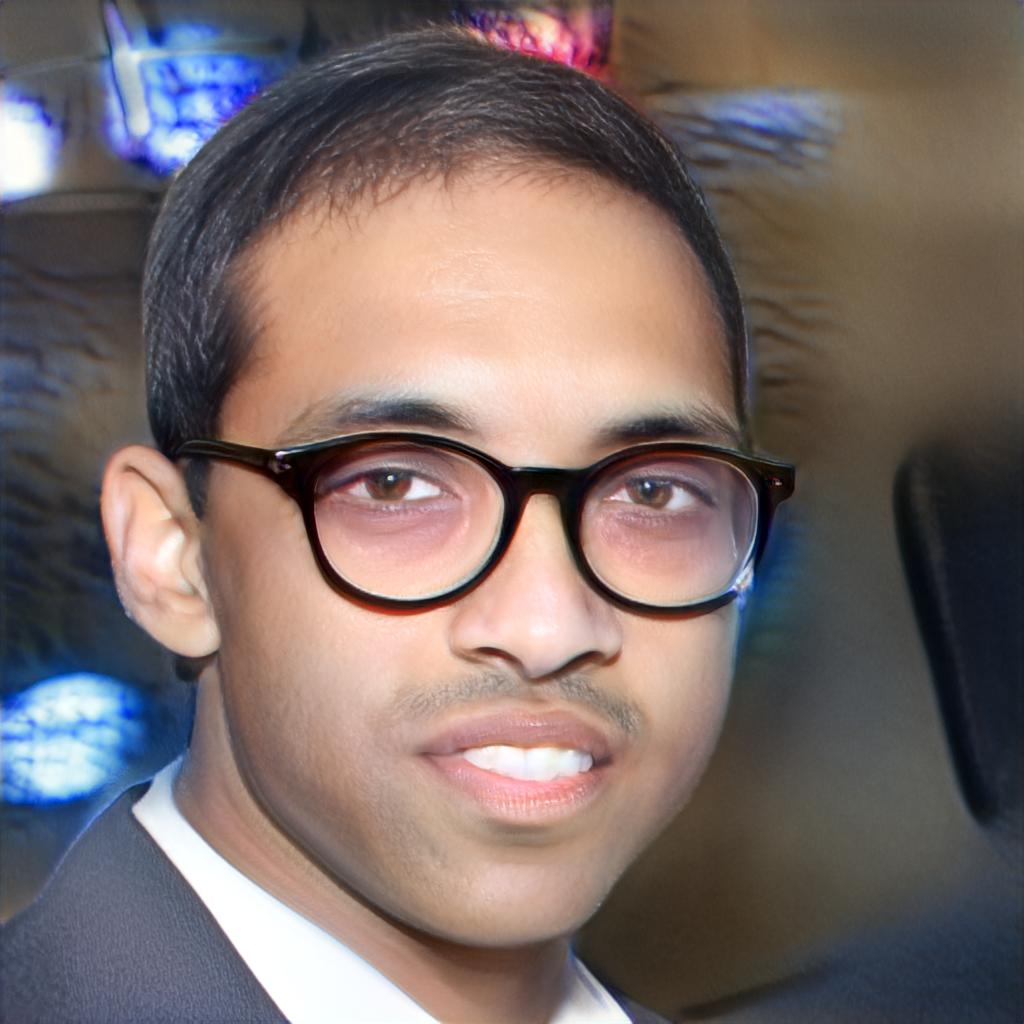} \\
        
        \raisebox{0.265in}{\rotatebox{90}{\footnotesize SeFa}} &
        \includegraphics[width=0.086\textwidth]{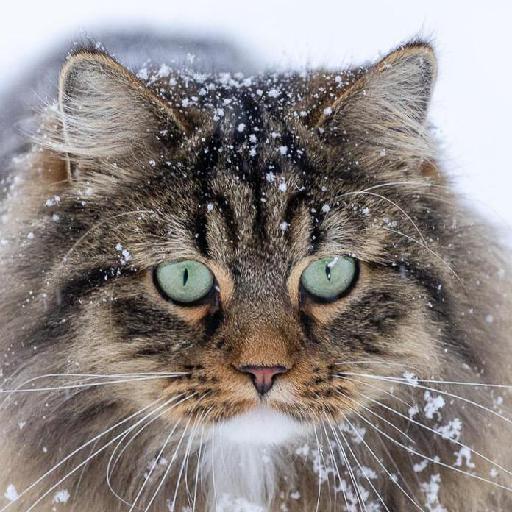} &
        \includegraphics[width=0.086\textwidth]{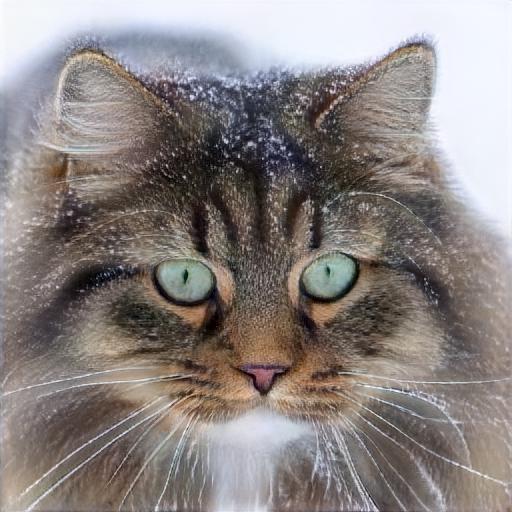} &
        \includegraphics[width=0.086\textwidth]{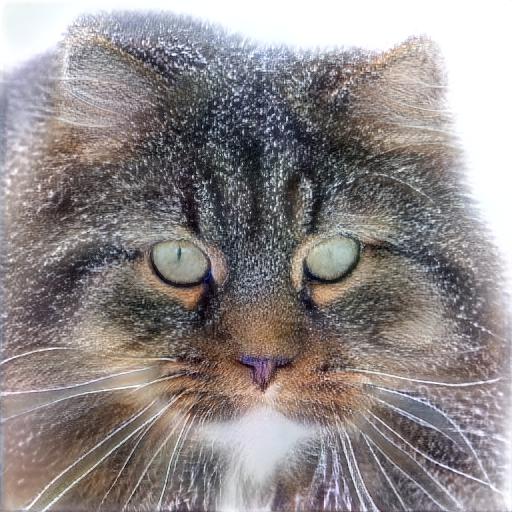} &
        \includegraphics[width=0.086\textwidth]{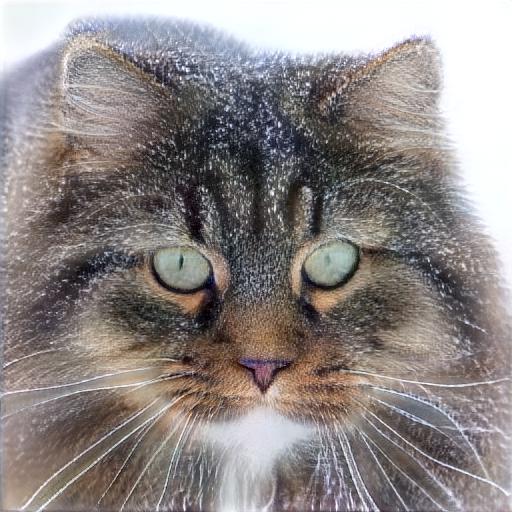} &
        \includegraphics[width=0.086\textwidth]{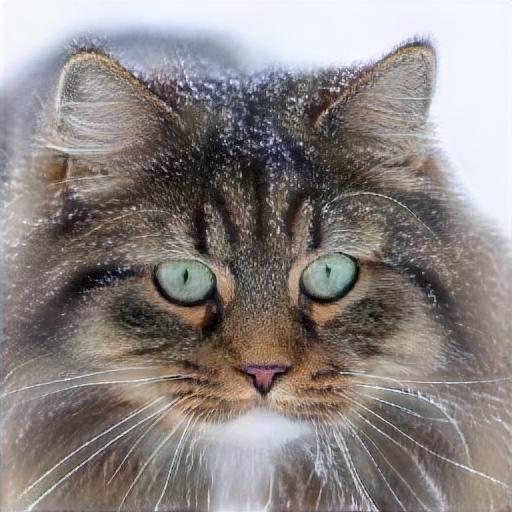} &
        \includegraphics[width=0.086\textwidth]{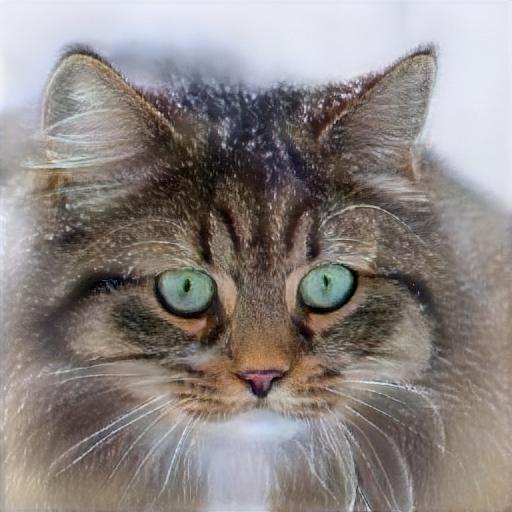} &
        \includegraphics[width=0.086\textwidth]{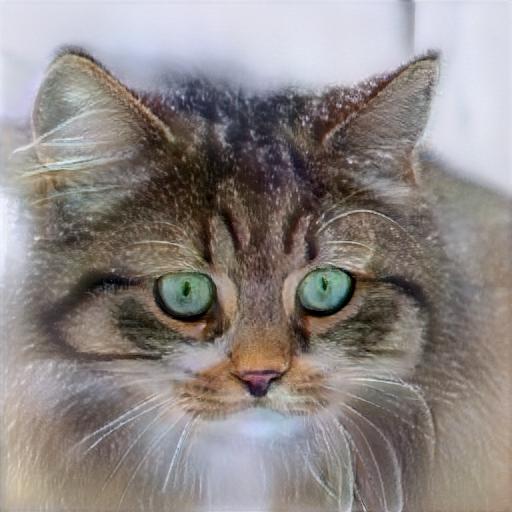} \\
        
        \raisebox{0.265in}{\rotatebox{90}{Ours}} &
        \includegraphics[width=0.086\textwidth]{resources/images/comparisons/cats/unedited/15/15.jpg} &
        \includegraphics[width=0.086\textwidth]{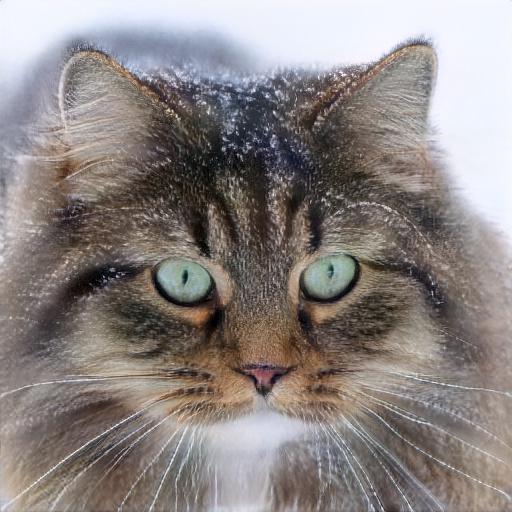} &
        \includegraphics[width=0.086\textwidth]{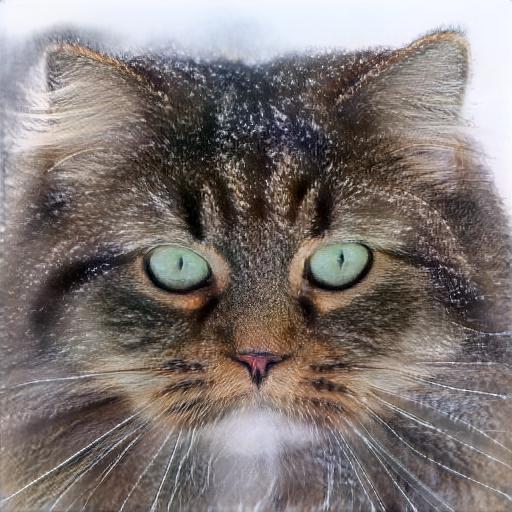} &
        \includegraphics[width=0.086\textwidth]{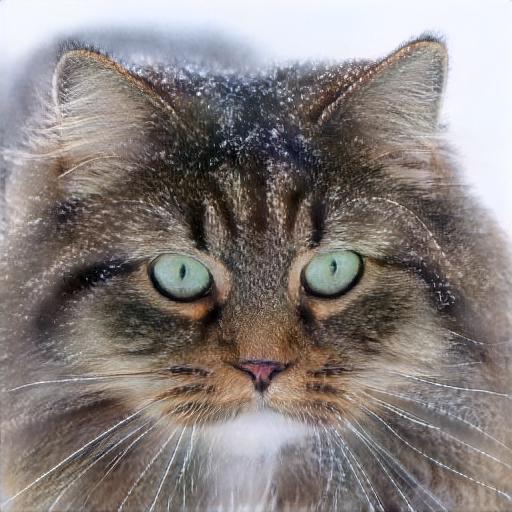} &
        \includegraphics[width=0.086\textwidth]{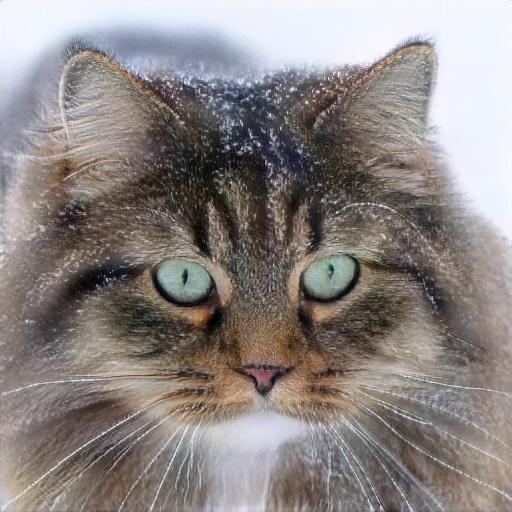} &
        \includegraphics[width=0.086\textwidth]{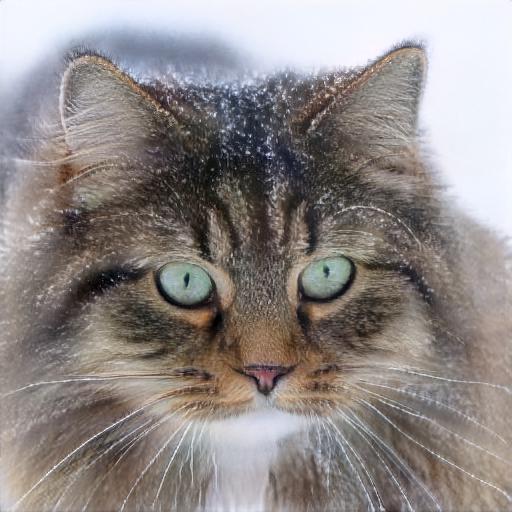} &
        \includegraphics[width=0.086\textwidth]{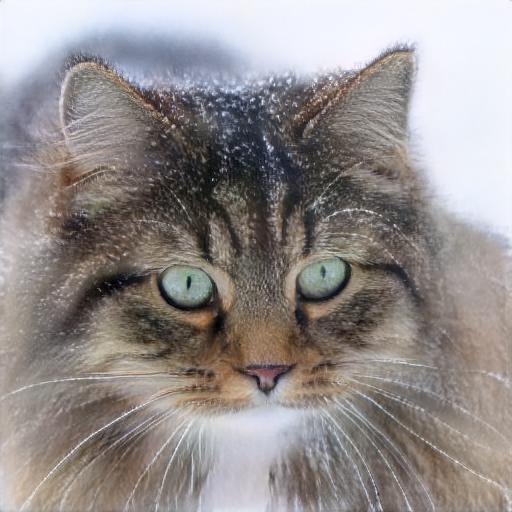} 

        \\ & Input & Inversion & \multicolumn{5}{c}{$\myleftarrow$~Age~$\myarrow$ } \\
        
        \raisebox{0.15in}{\rotatebox{90}{\footnotesize StyleCLIP}} &
        \includegraphics[width=0.086\textwidth]{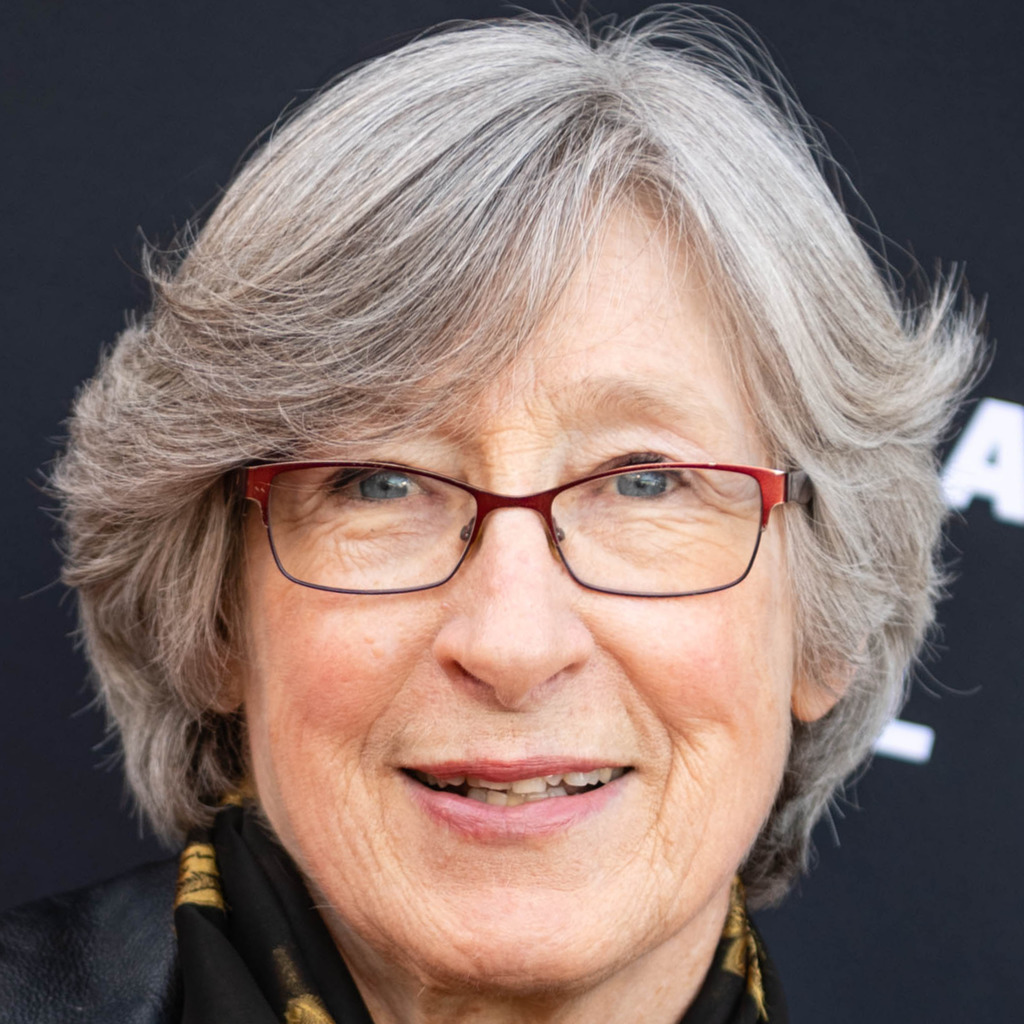} &
        \includegraphics[width=0.086\textwidth]{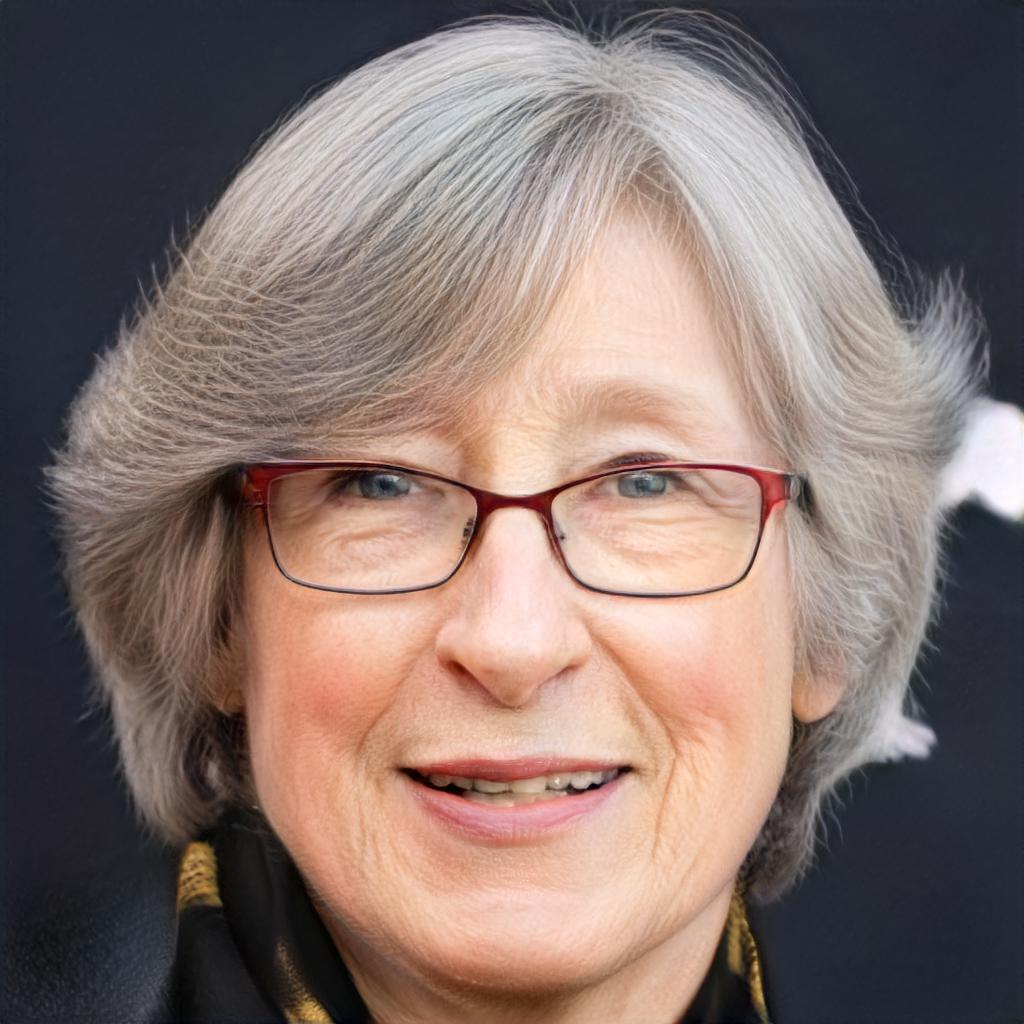} &
        \includegraphics[width=0.086\textwidth]{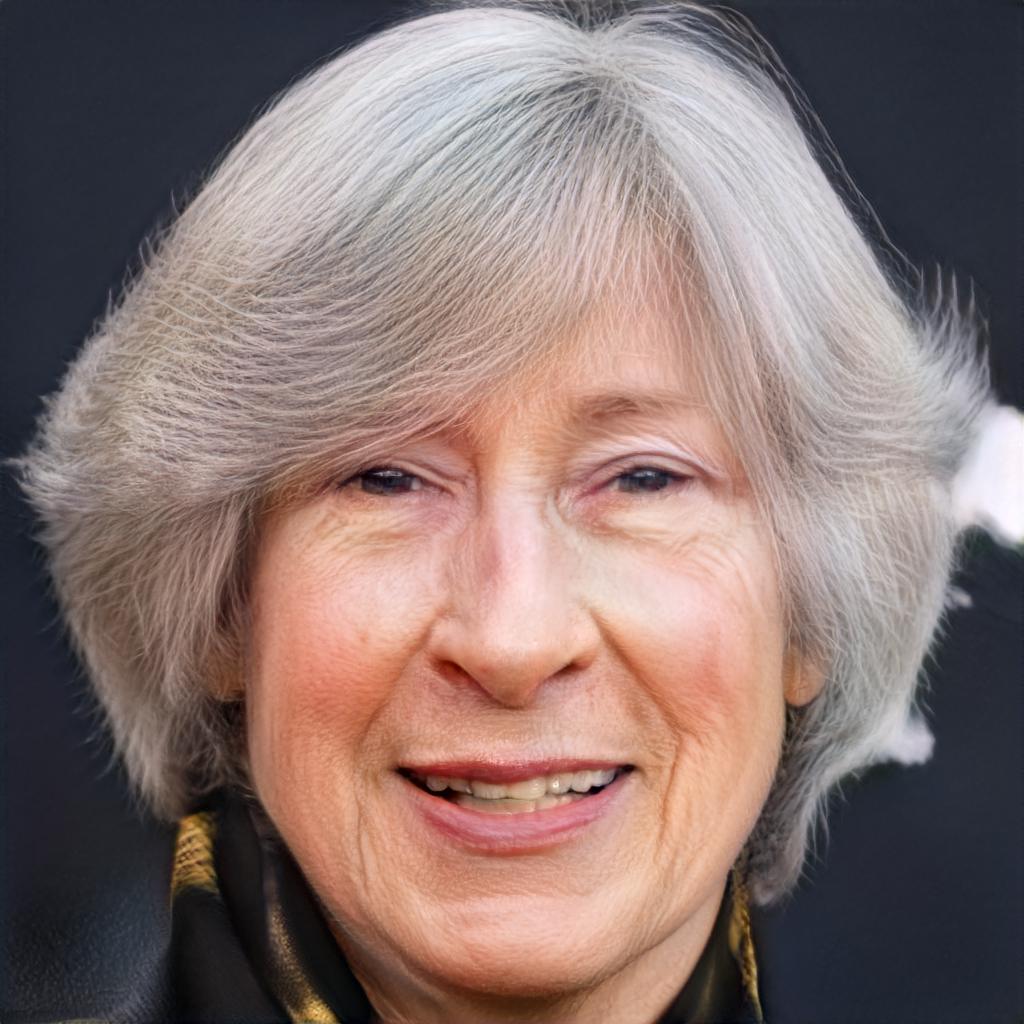} &
        \includegraphics[width=0.086\textwidth]{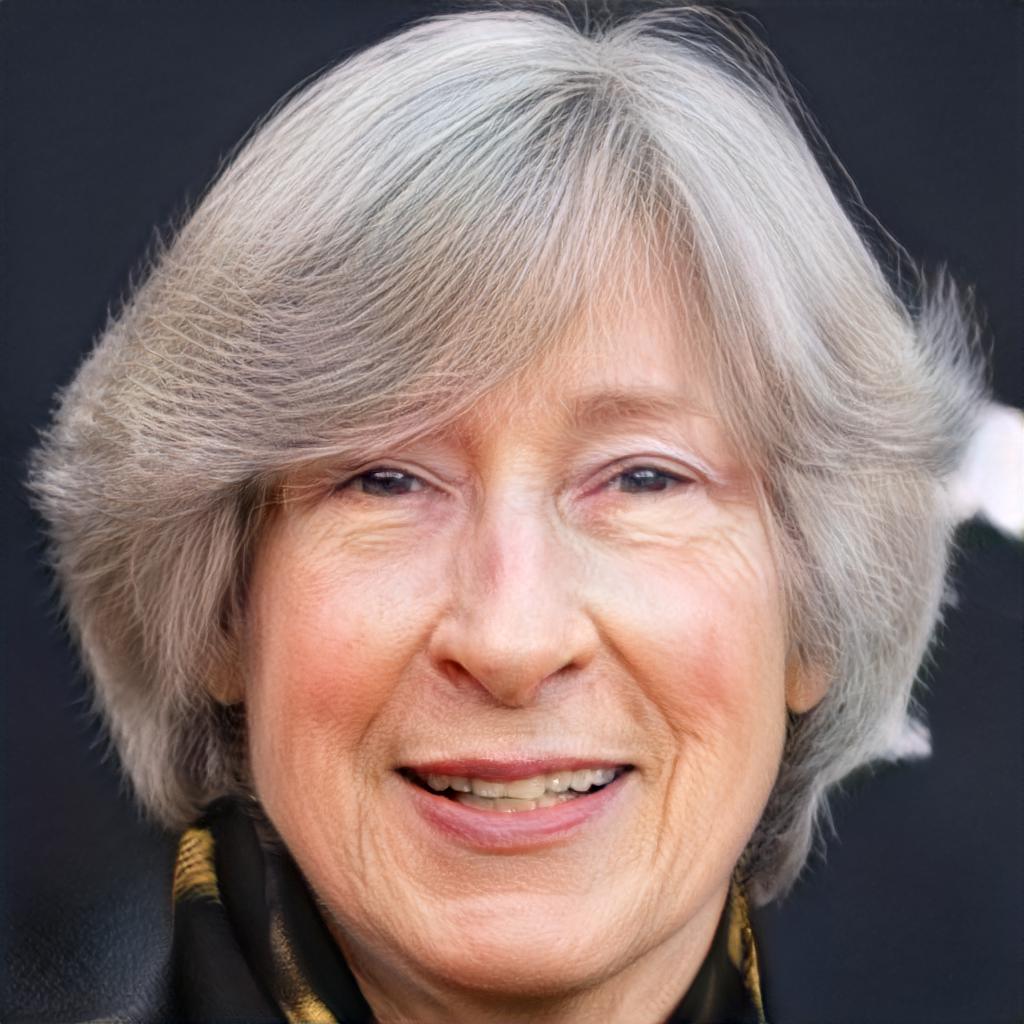} &
        \includegraphics[width=0.086\textwidth]{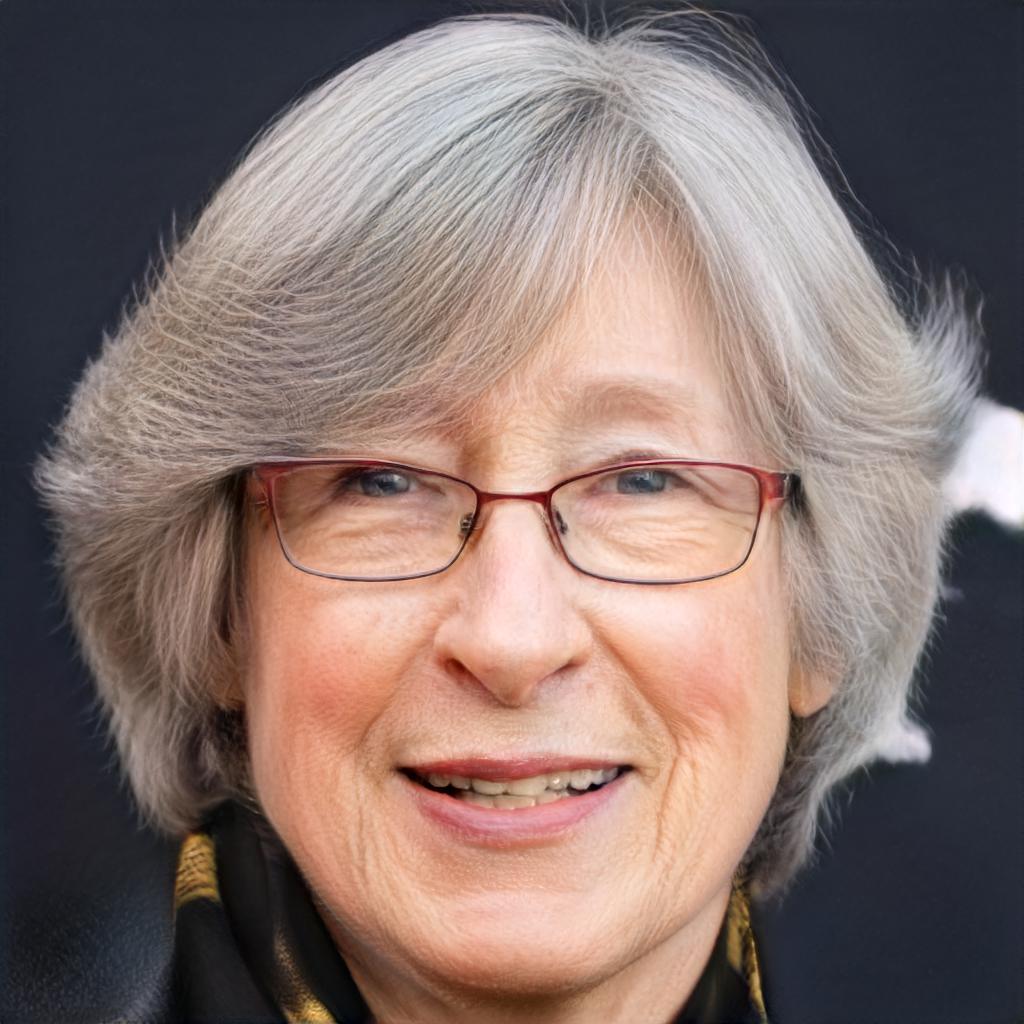} &
        \includegraphics[width=0.086\textwidth]{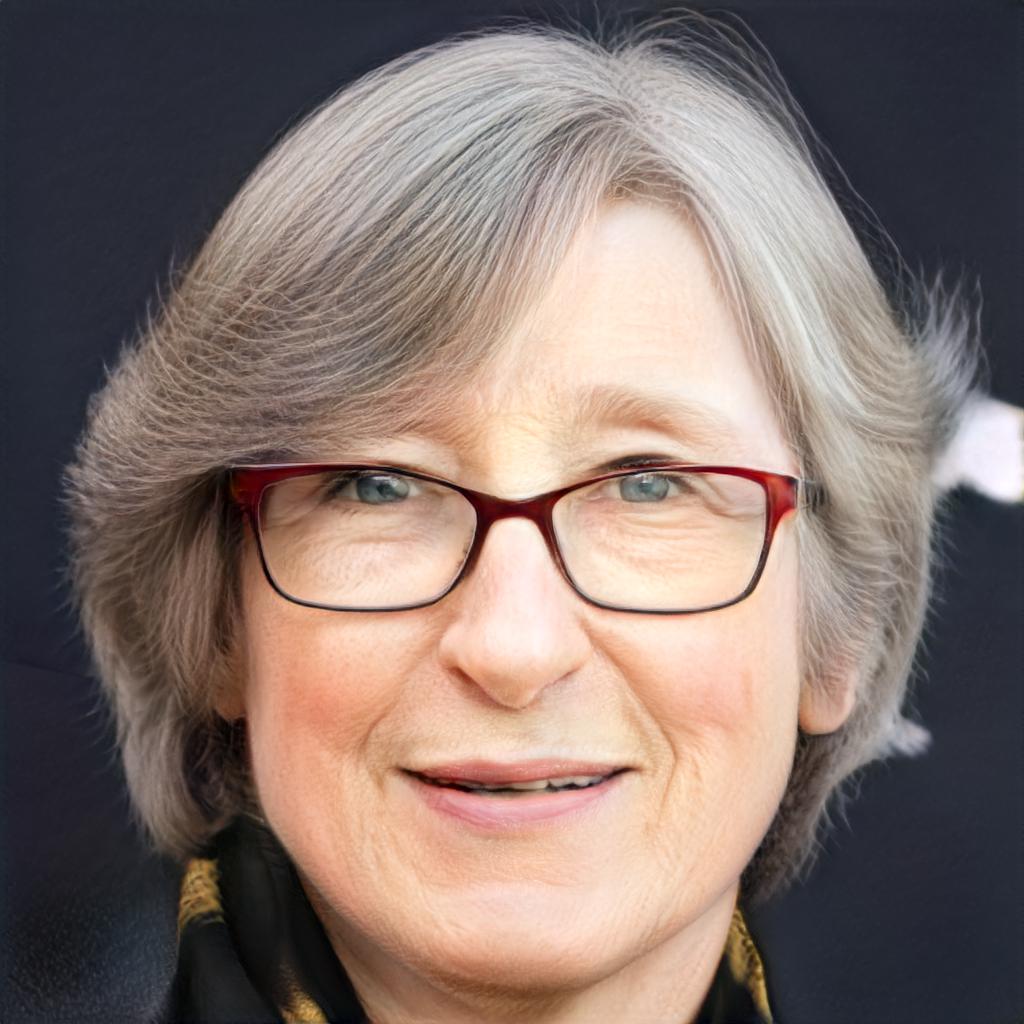} &
        \includegraphics[width=0.086\textwidth]{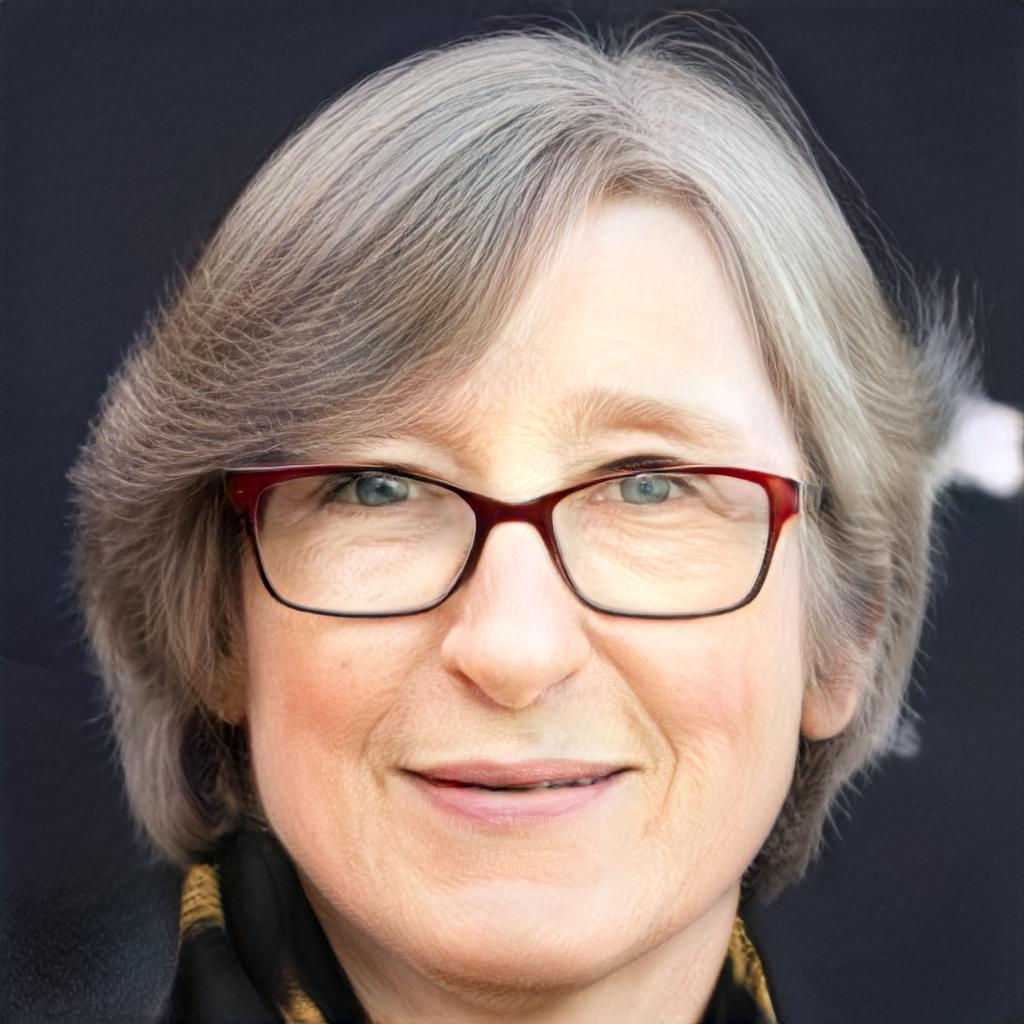} \\
        
        \raisebox{0.265in}{\rotatebox{90}{Ours}} &
        \includegraphics[width=0.086\textwidth]{resources/images/comparisons/faces/unedited/79/79.jpg} &
        \includegraphics[width=0.086\textwidth]{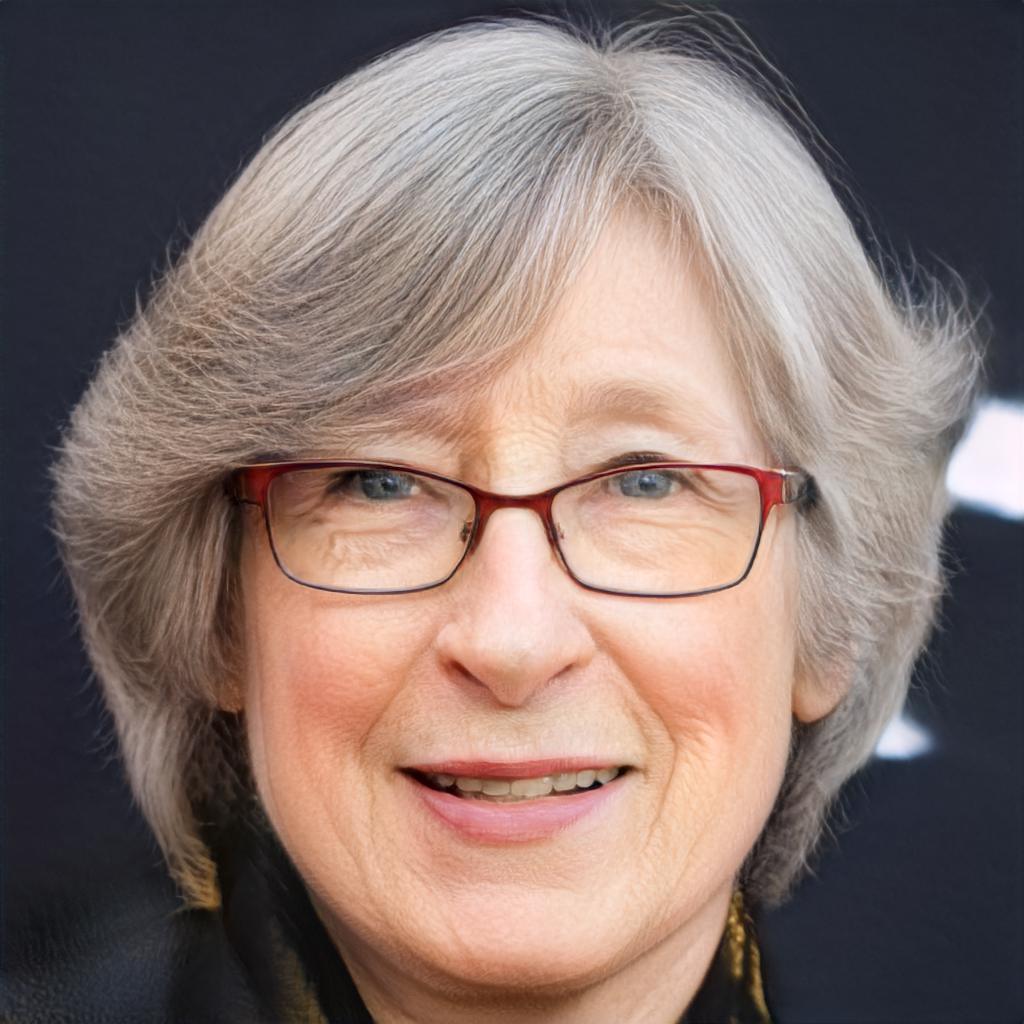} &
        \includegraphics[width=0.086\textwidth]{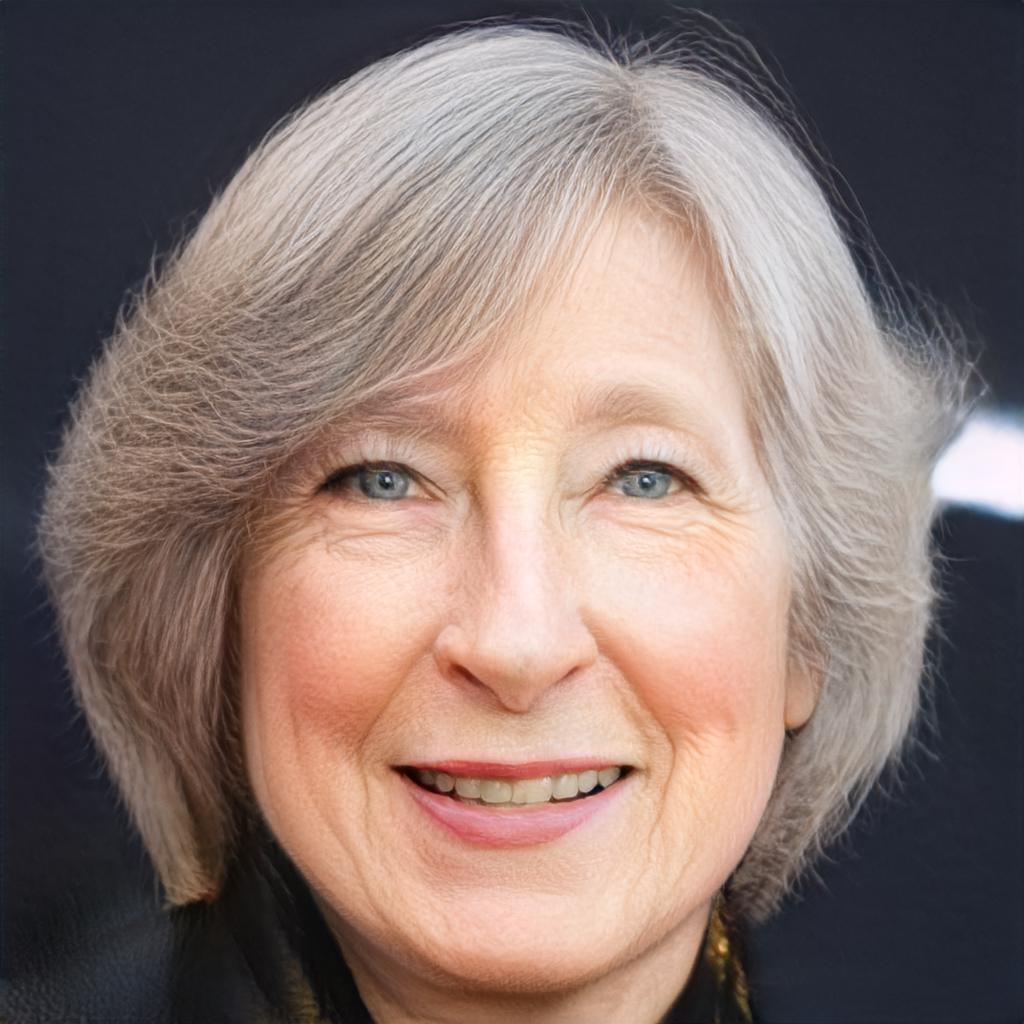} &
        \includegraphics[width=0.086\textwidth]{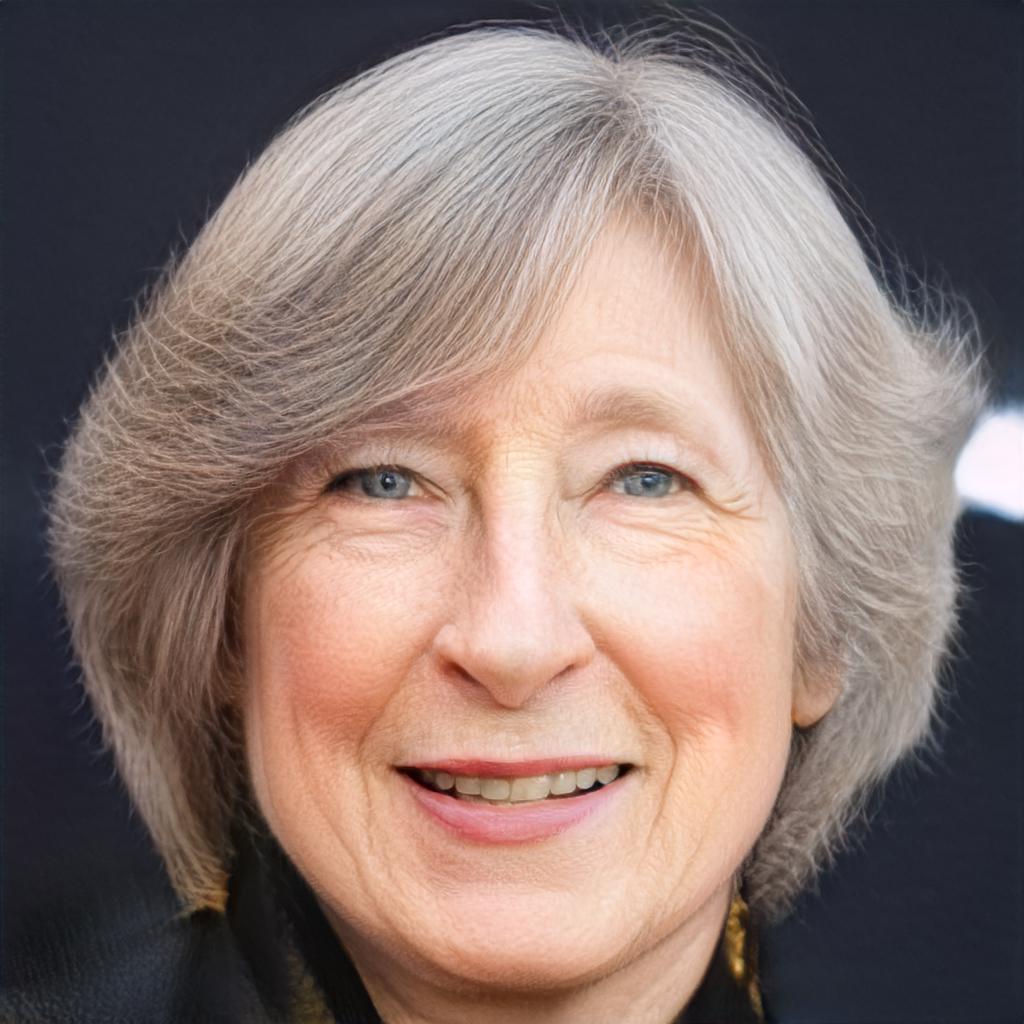} &
        \includegraphics[width=0.086\textwidth]{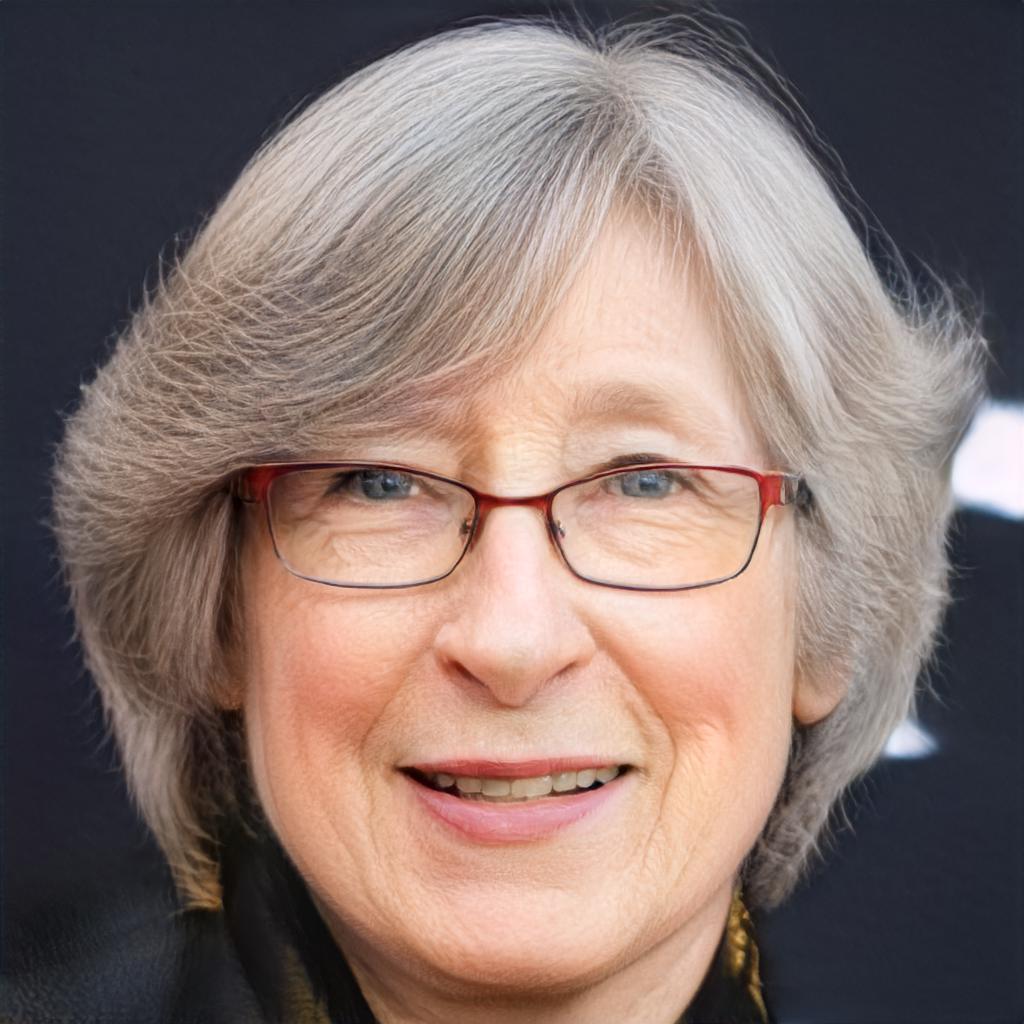} &
        \includegraphics[width=0.086\textwidth]{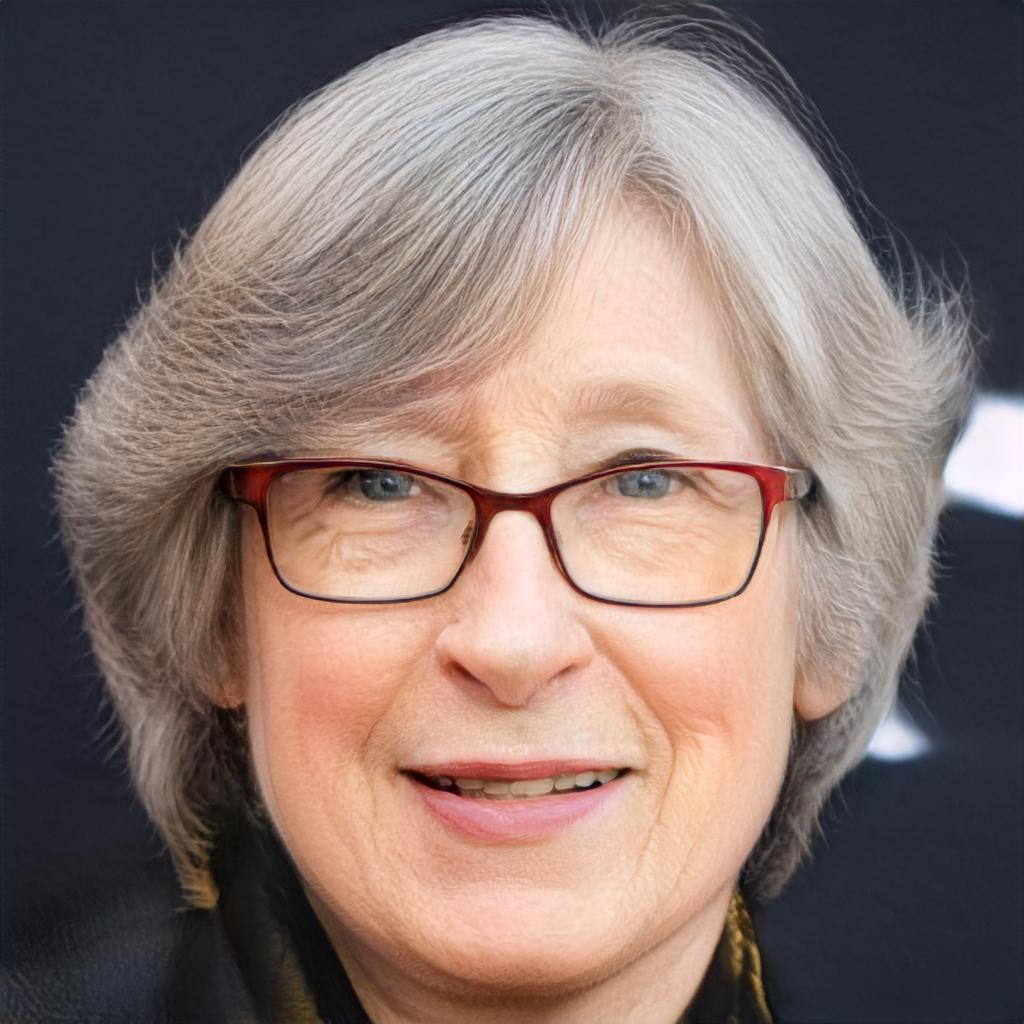} &
        \includegraphics[width=0.086\textwidth]{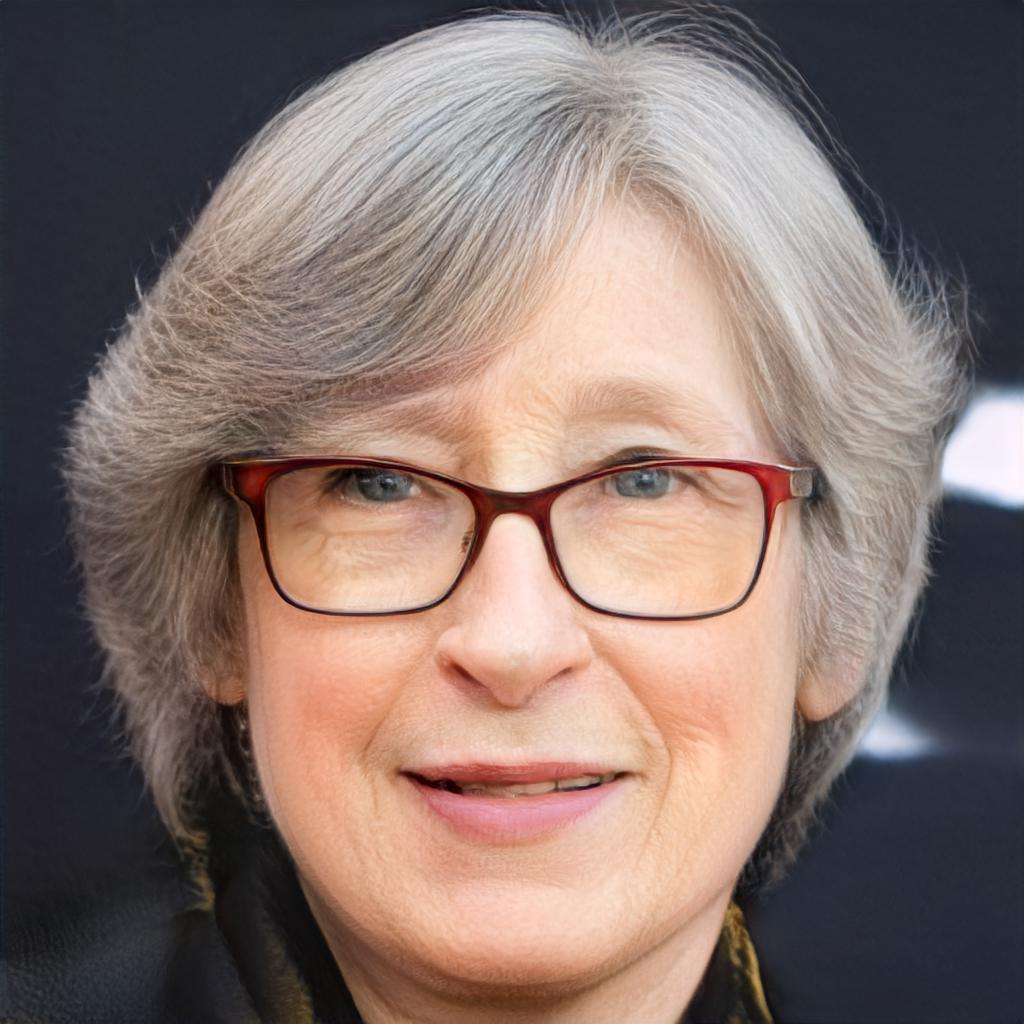} 
        
        \\ & Input & Inversion & \multicolumn{5}{c}{ $\myleftarrow$~Glasses~$\myarrow$ } \\

    \end{tabular}
    
    }
    \vspace{-0.21cm}
    \caption{Linear editing comparisons on real images. In each pair of rows, we compare our method against the initial latent direction which was used to extract our self-conditioning labels. In all cases, our method better preserves the identity and allows for more significant manipulations before image quality suffers drastically. In the glasses example (bottom), continued movement in the negative direction on an image without glasses leads to an increase in age. Our model avoids this problem and produces the same identity.}
    \vspace{-0.17cm}
    \label{fig:editing_comparisons}
\end{figure*}

%% file: resources/figures/non_linear_comparisons.tex
\begin{figure*}
    \centering
    \setlength{\belowcaptionskip}{-8pt}
    \setlength{\tabcolsep}{2.0pt}
    {\small
    \begin{tabular}{c c}
    
    \begin{tabular}{c c c c c c}

        \raisebox{0.00in}{\rotatebox{90}{\tiny \cite{choi2021escape}}} &
        \includegraphics[width=0.075\textwidth]{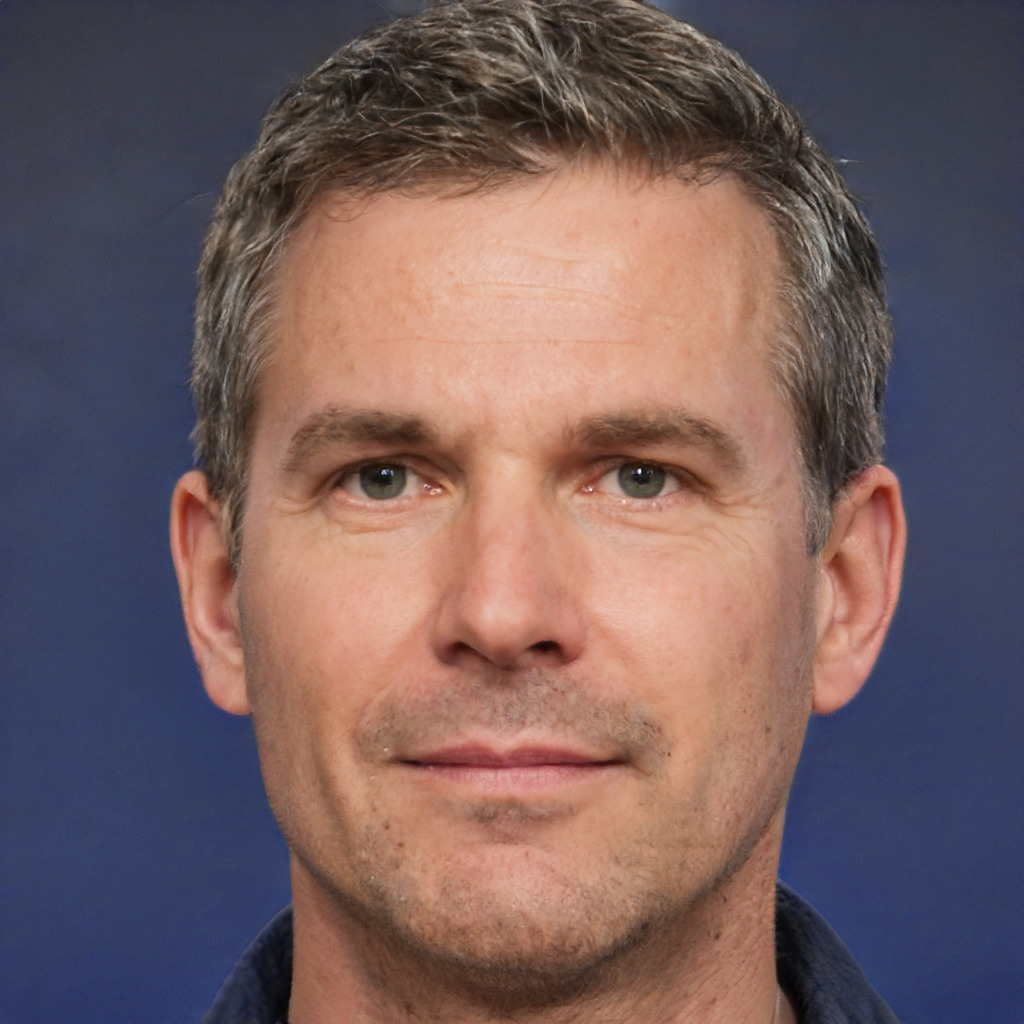} &
        \includegraphics[width=0.075\textwidth]{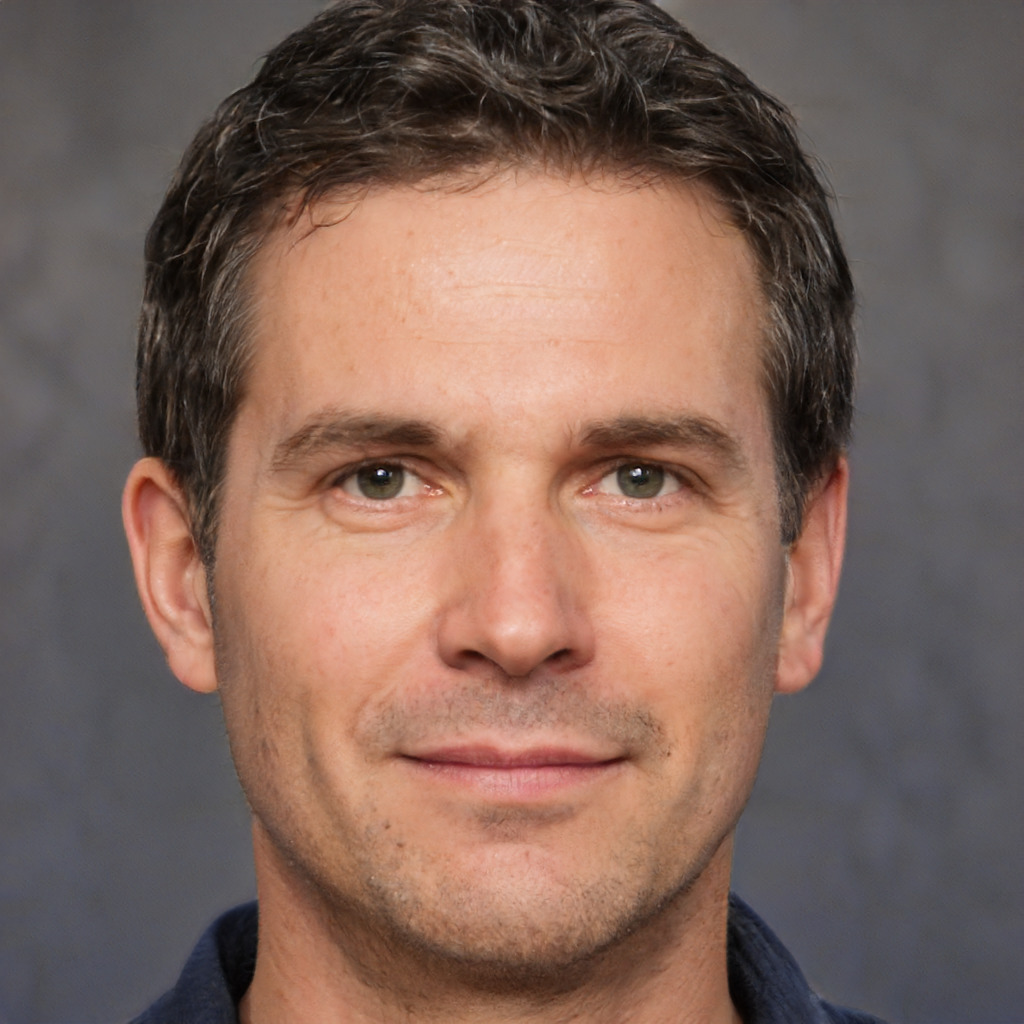} &
        \includegraphics[width=0.075\textwidth]{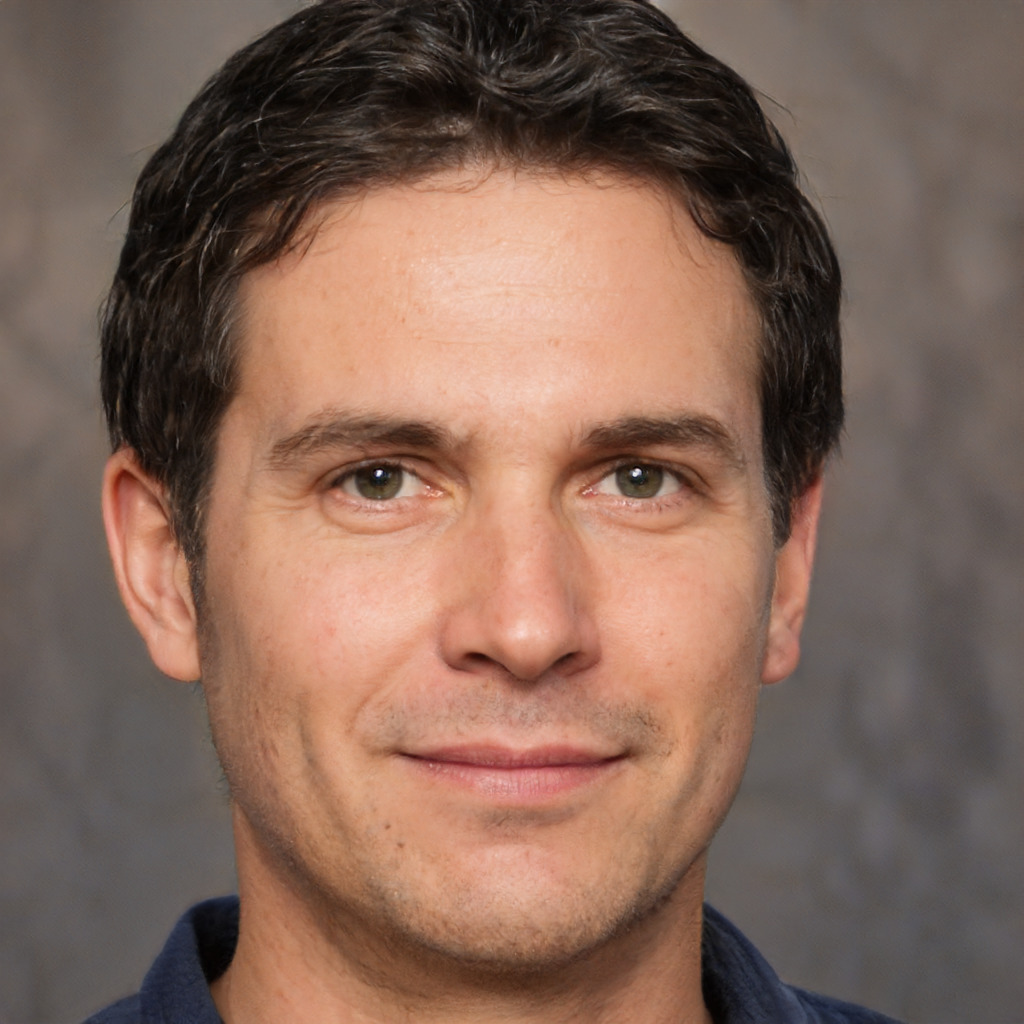} &
        \includegraphics[width=0.075\textwidth]{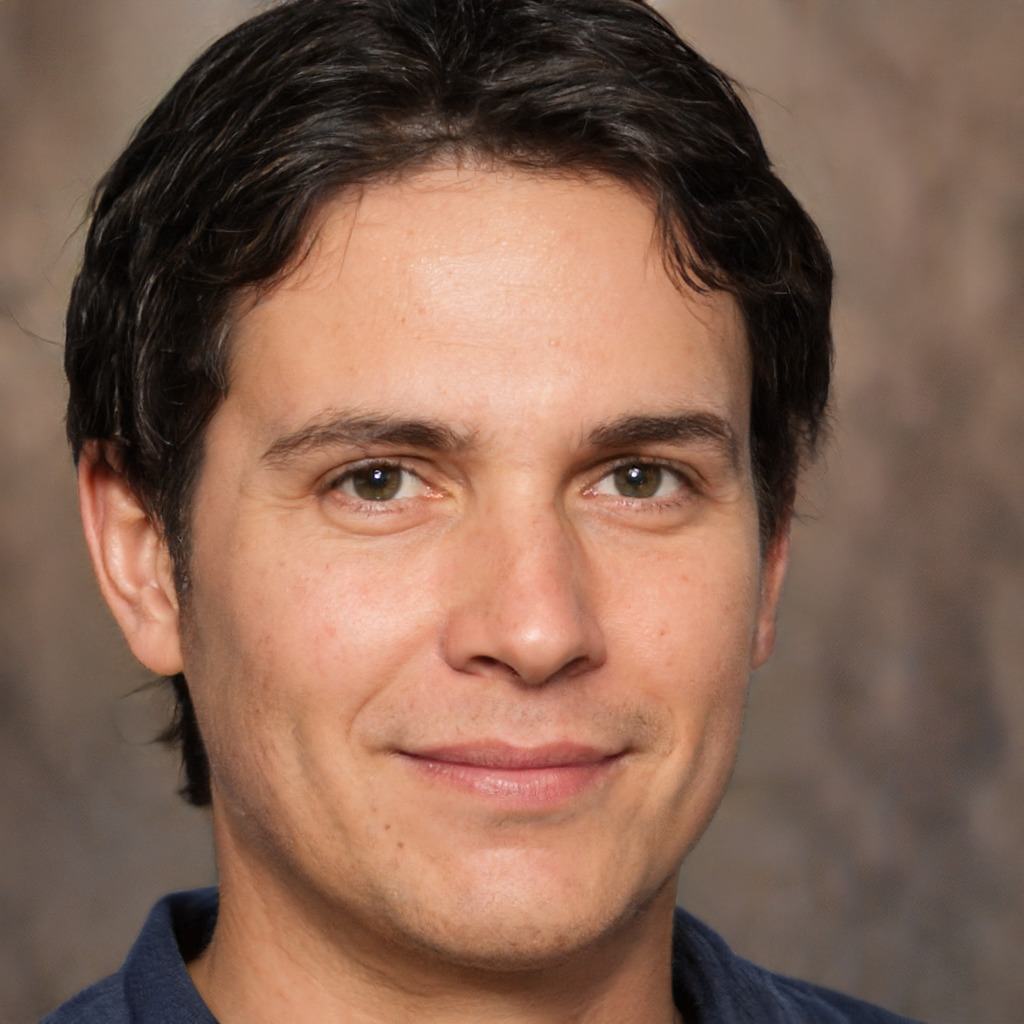} &
        \includegraphics[width=0.075\textwidth]{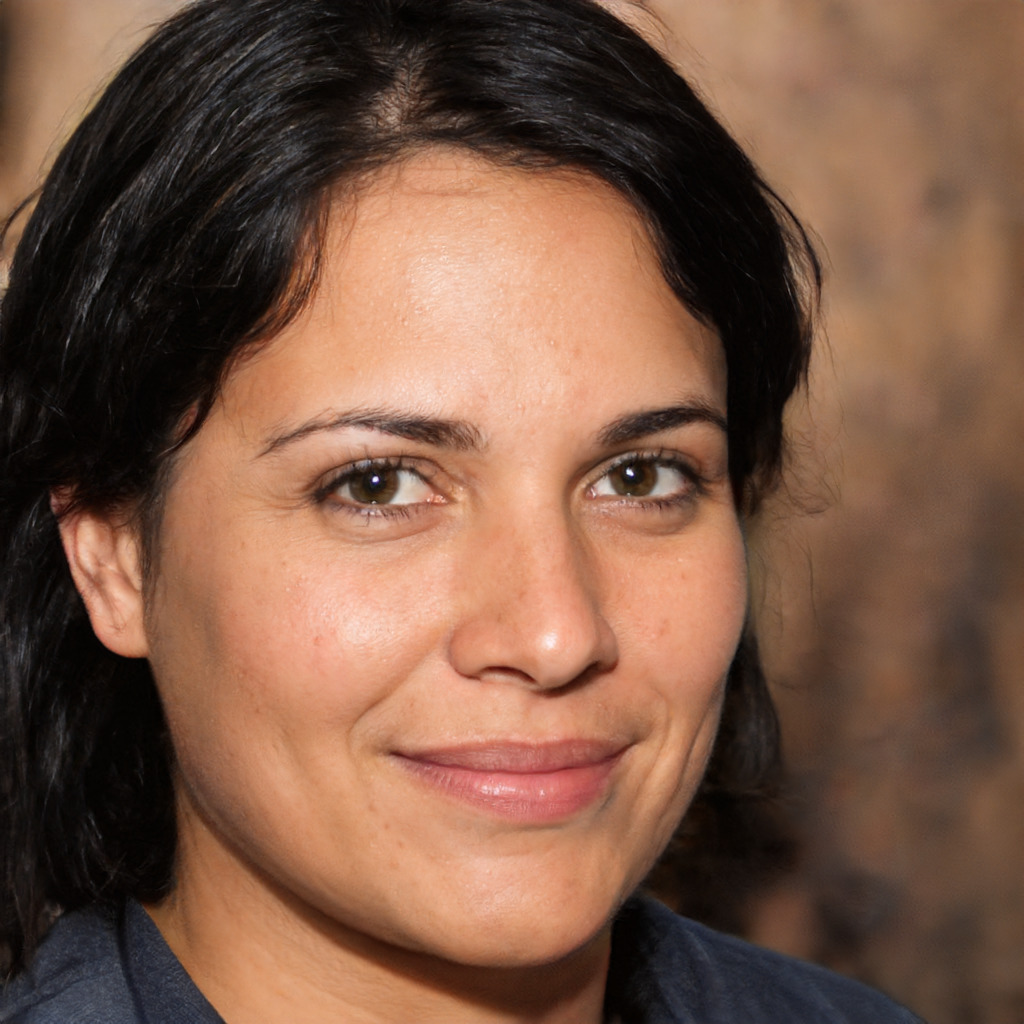} \\
        
        \raisebox{0.155in}{\rotatebox{90}{Ours}} &
        \includegraphics[width=0.075\textwidth]{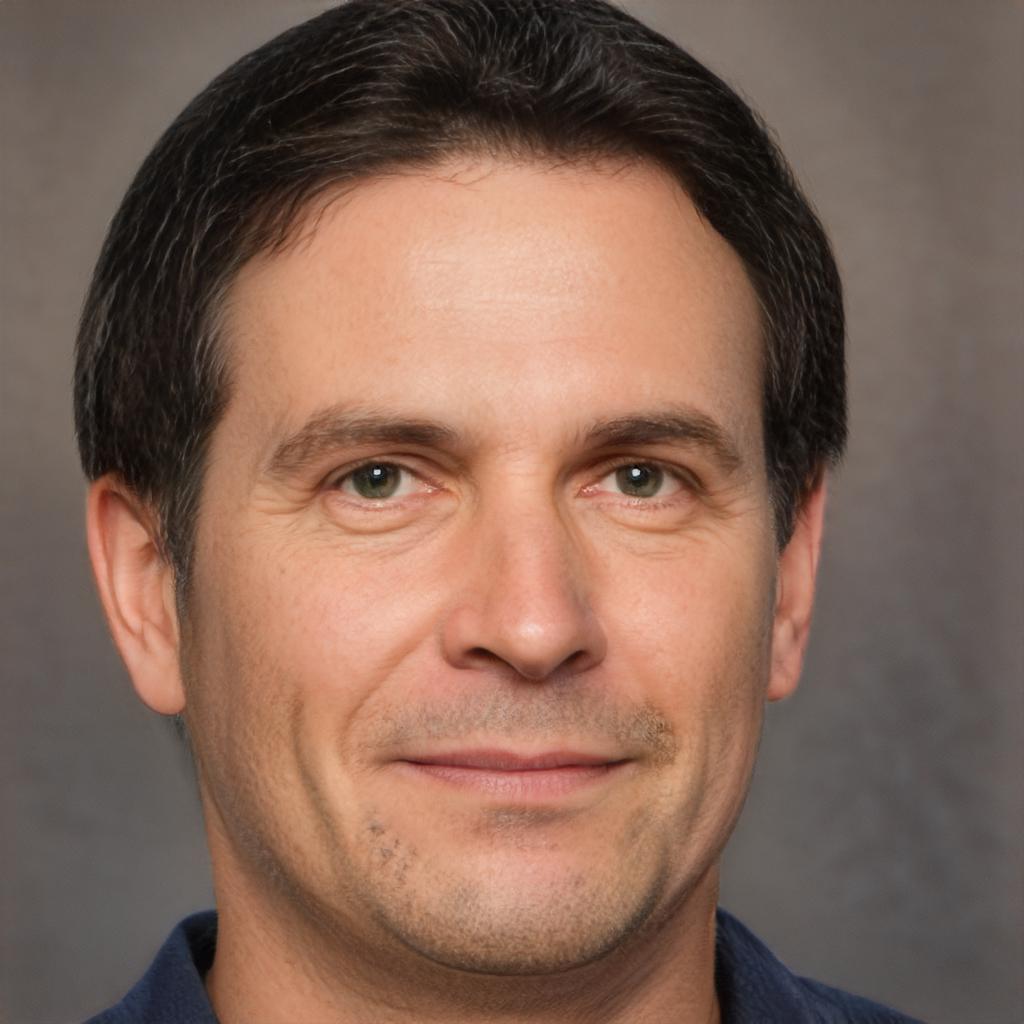} &
        \includegraphics[width=0.075\textwidth]{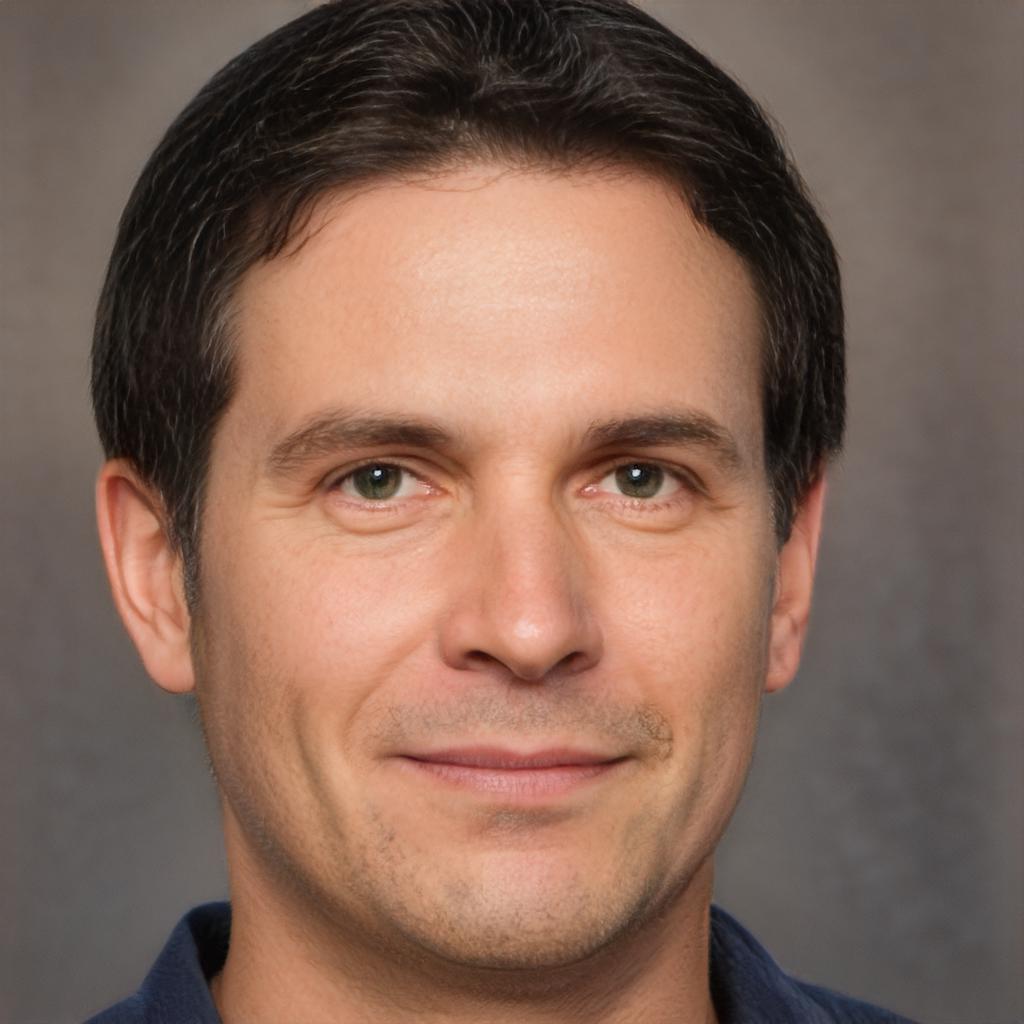} &
        \includegraphics[width=0.075\textwidth]{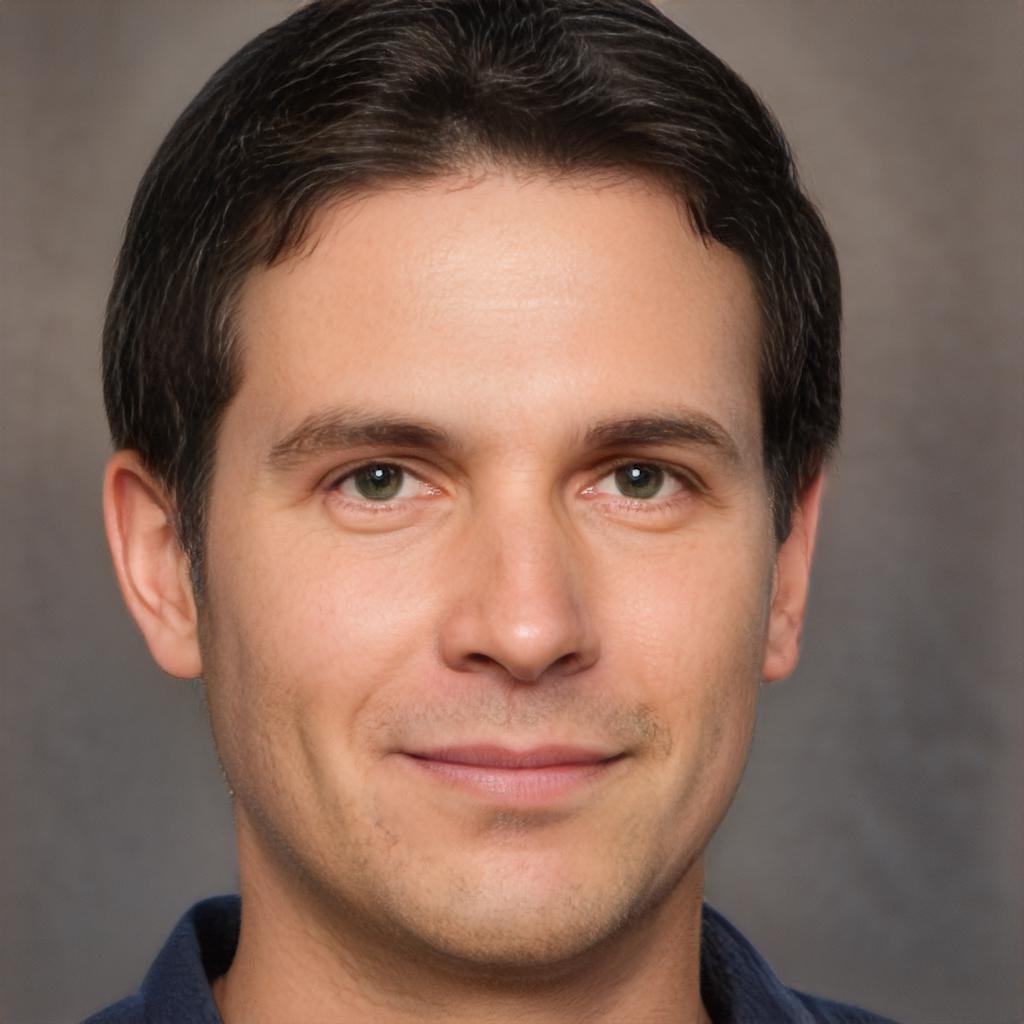} &
        \includegraphics[width=0.075\textwidth]{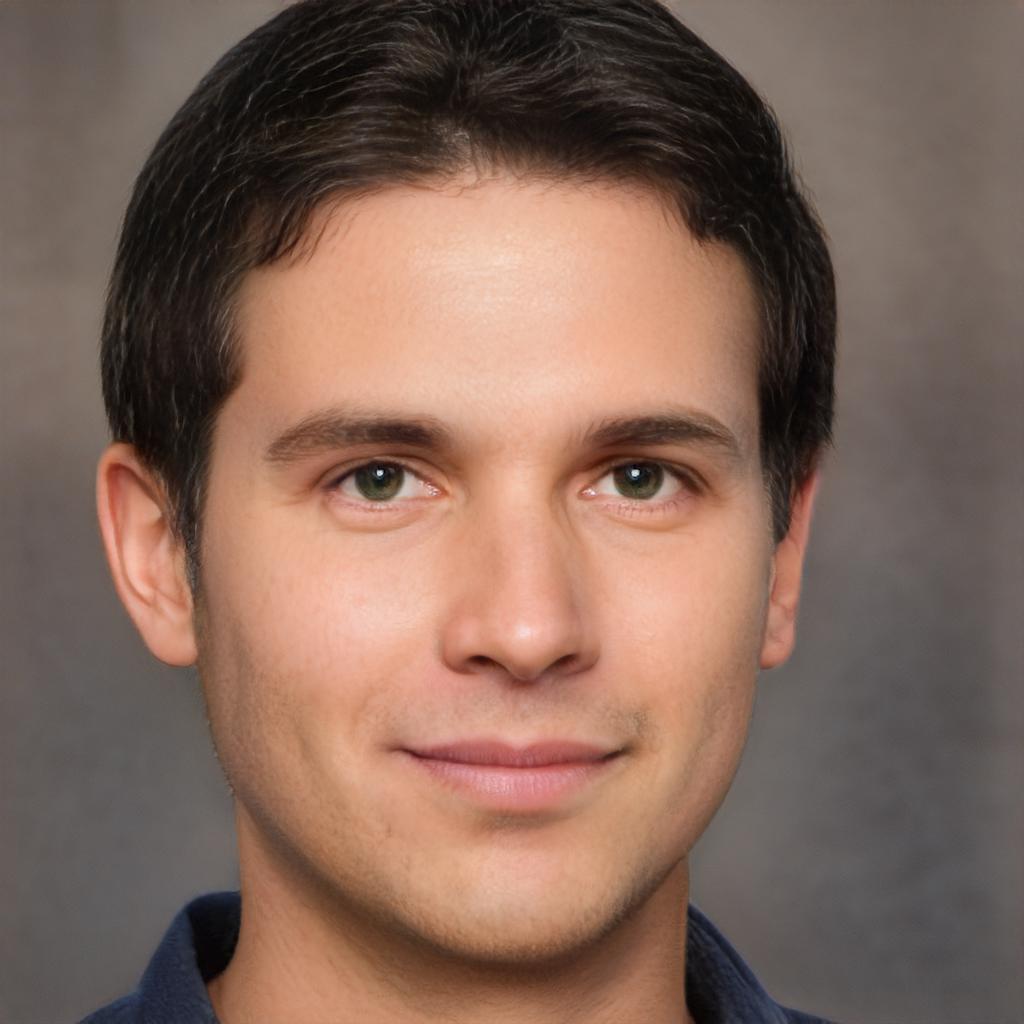} &
        \includegraphics[width=0.075\textwidth]{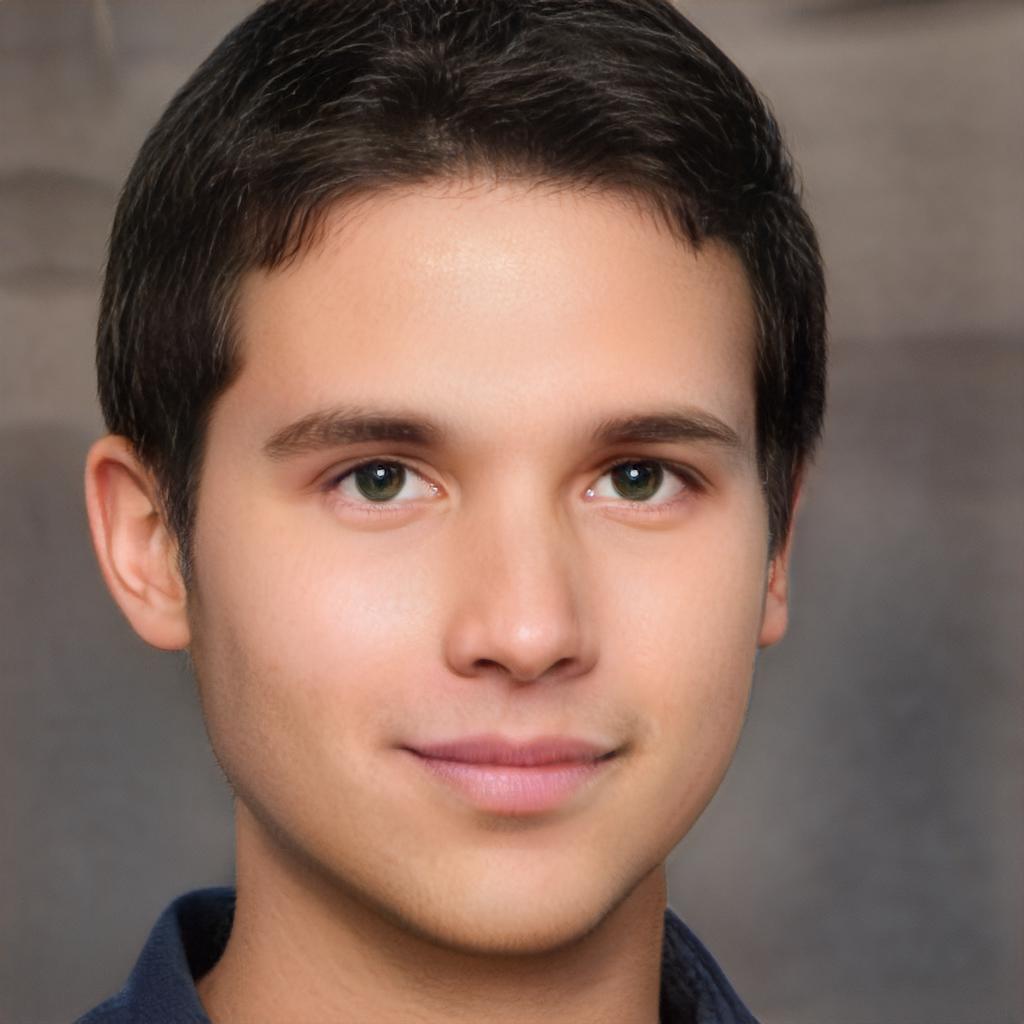} \\
        
        & \multicolumn{5}{c}{ Age }
        
    \end{tabular} 
    
    &
    
        \begin{tabular}{c c c c c c}

        \raisebox{0.00in}{\rotatebox{90}{\tiny \cite{choi2021escape}}} &
        \includegraphics[width=0.075\textwidth]{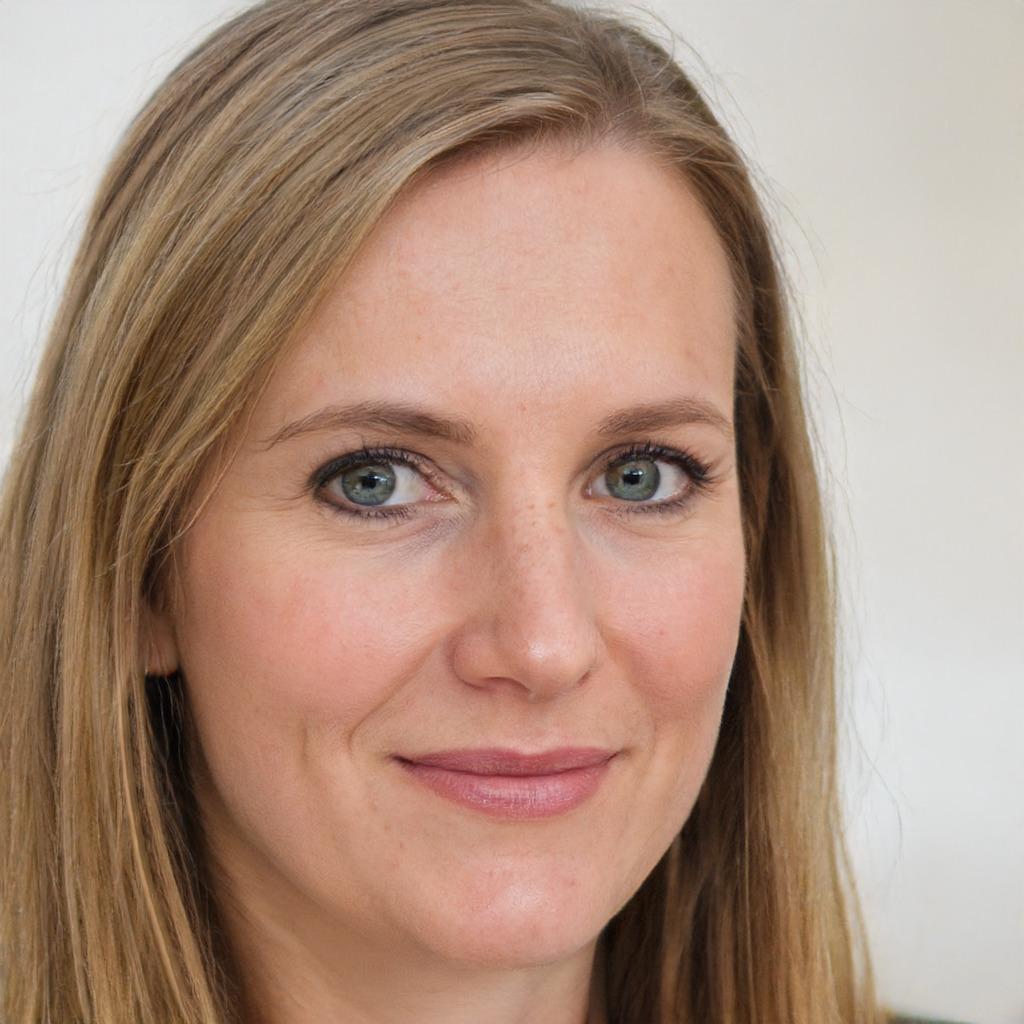} &
        \includegraphics[width=0.075\textwidth]{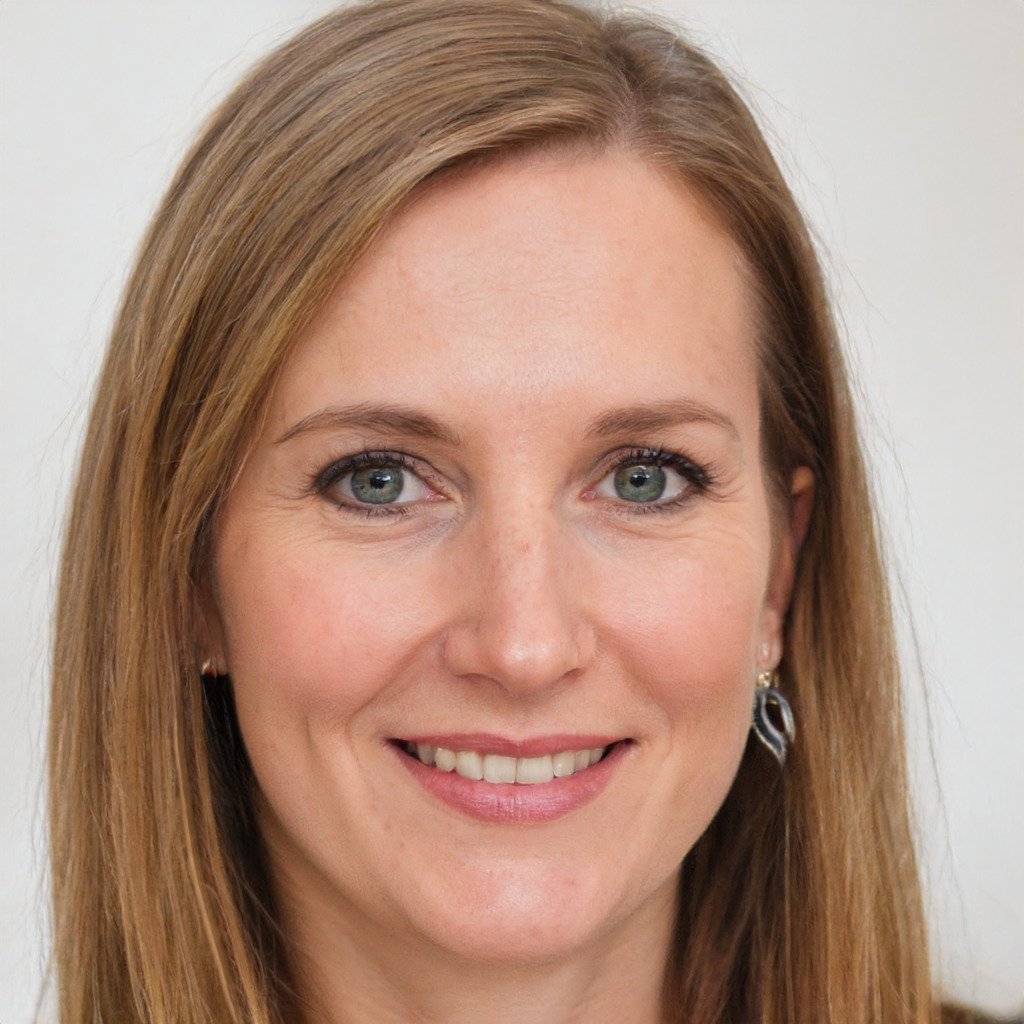} &
        \includegraphics[width=0.075\textwidth]{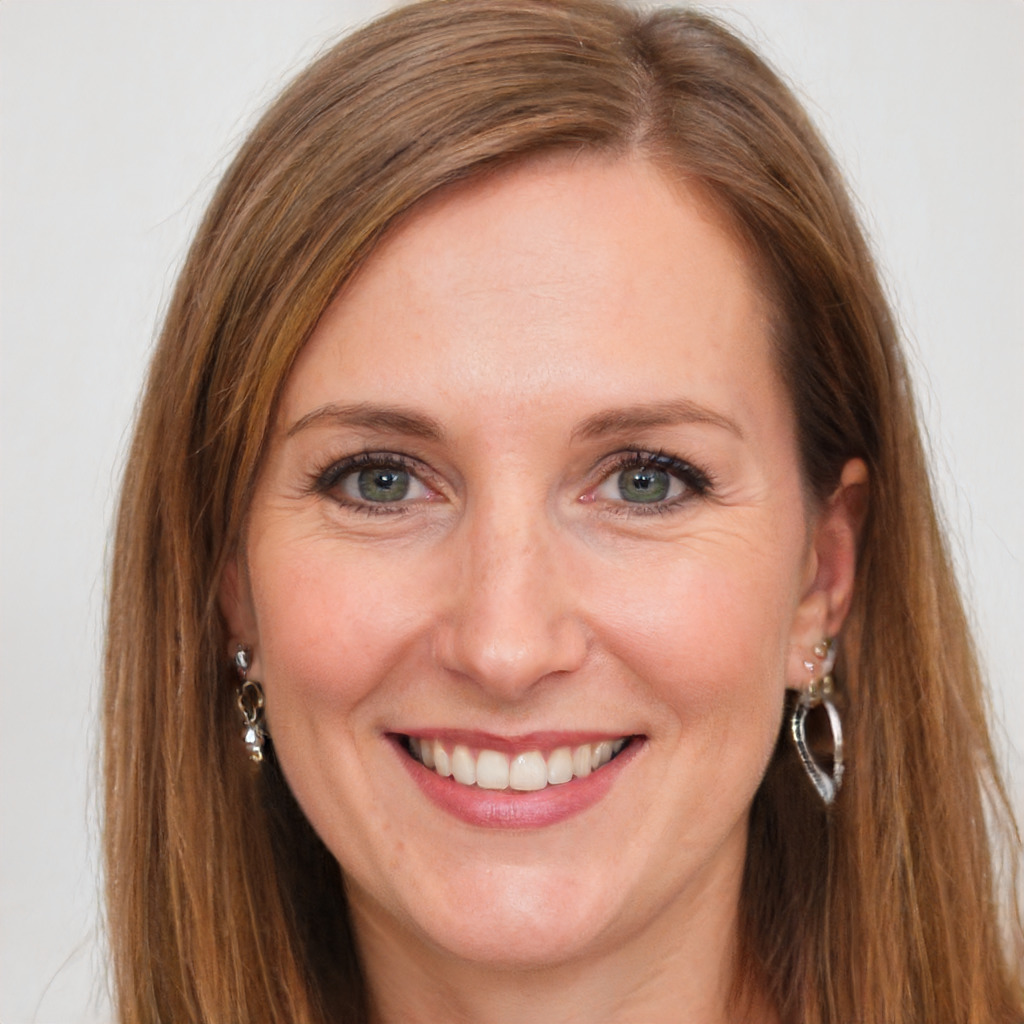} &
        \includegraphics[width=0.075\textwidth]{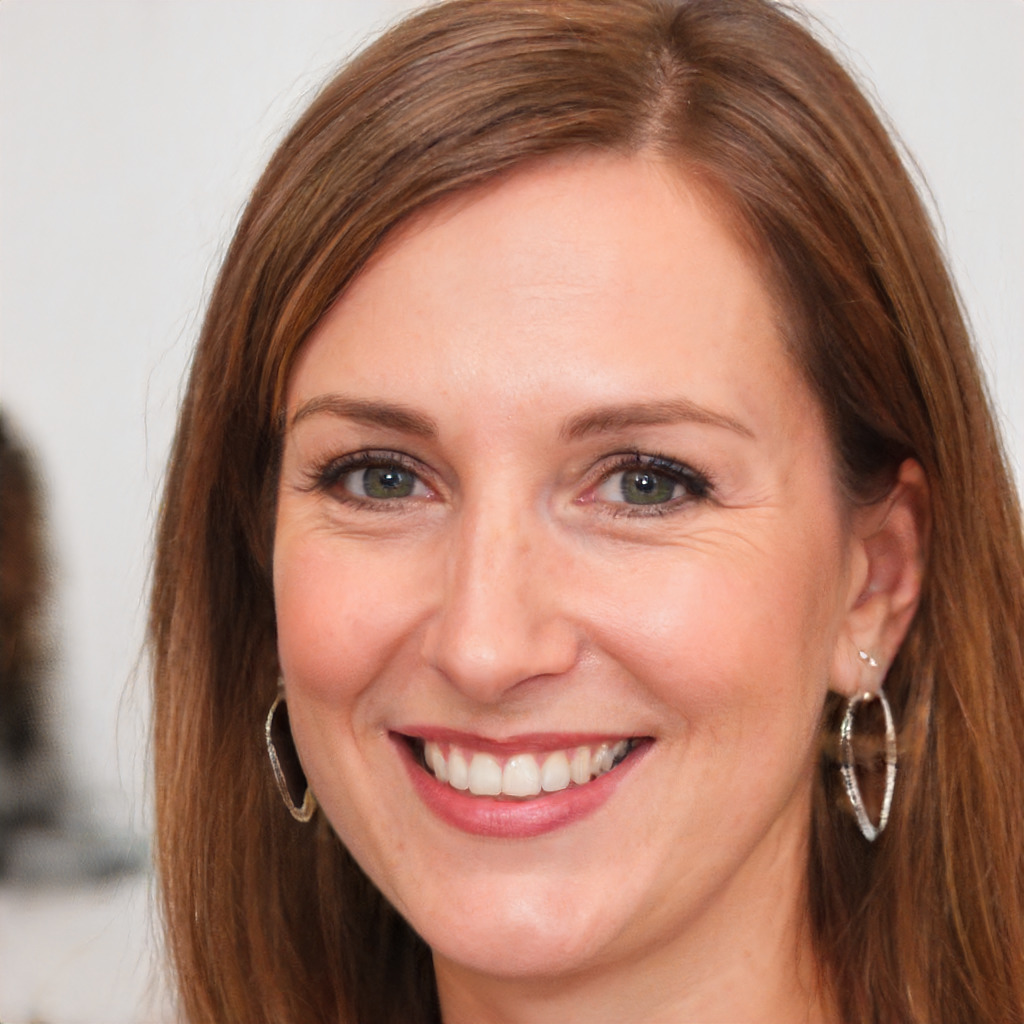} &
        \includegraphics[width=0.075\textwidth]{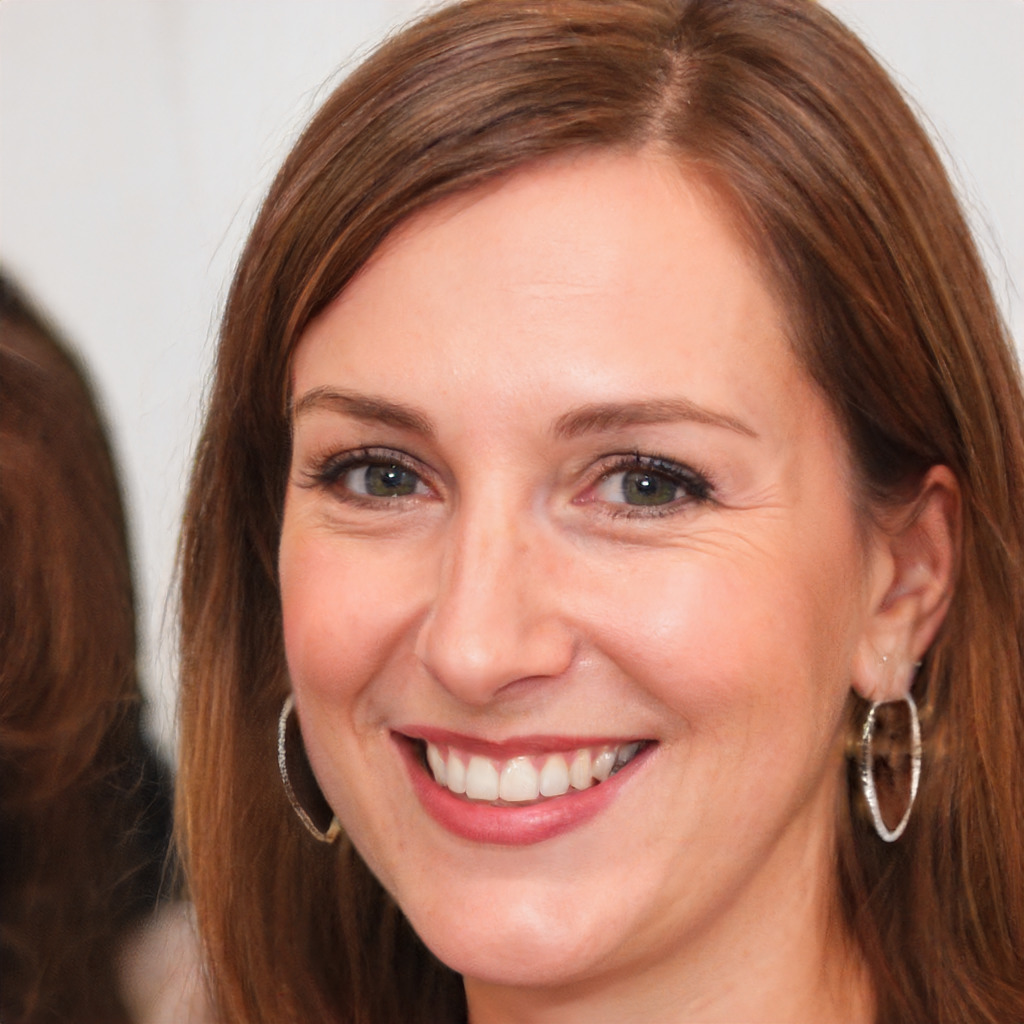} \\
        
        \raisebox{0.155in}{\rotatebox{90}{Ours}} &
        \includegraphics[width=0.075\textwidth]{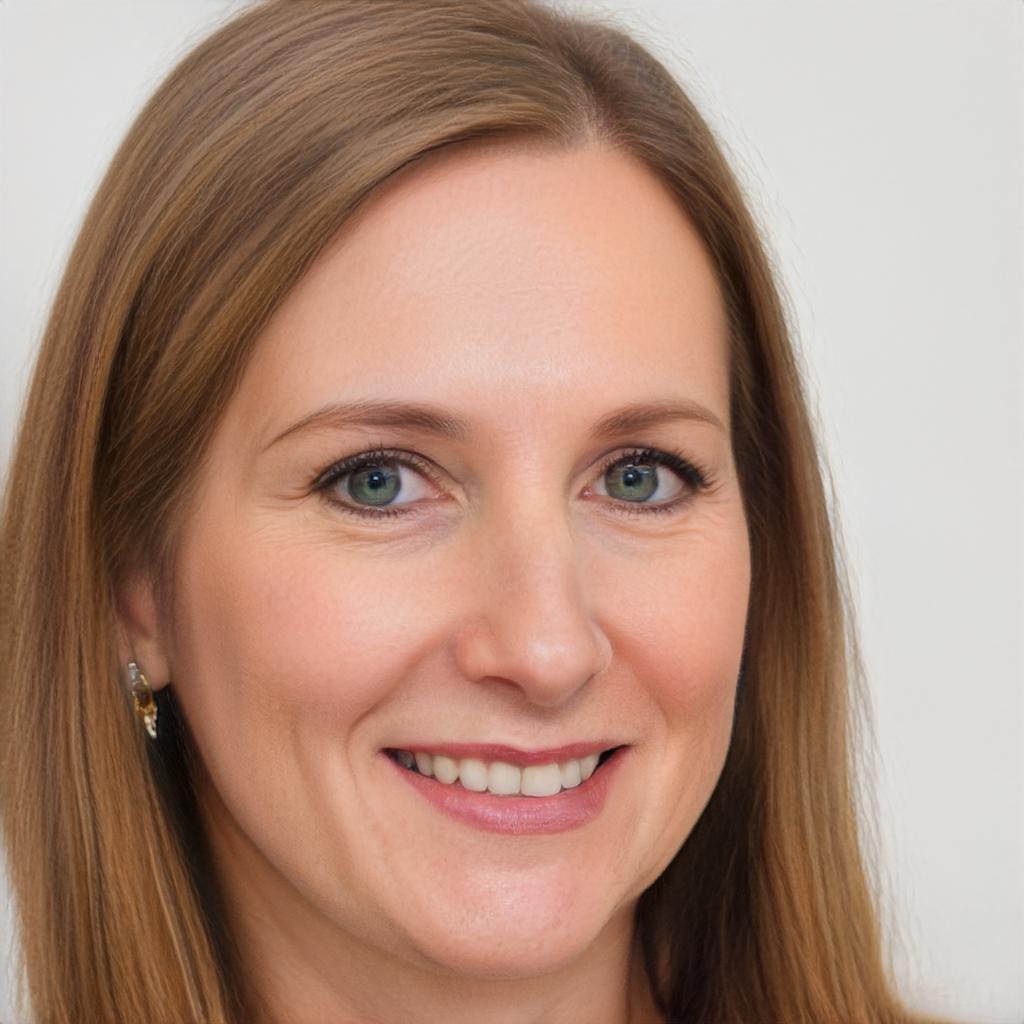} &
        \includegraphics[width=0.075\textwidth]{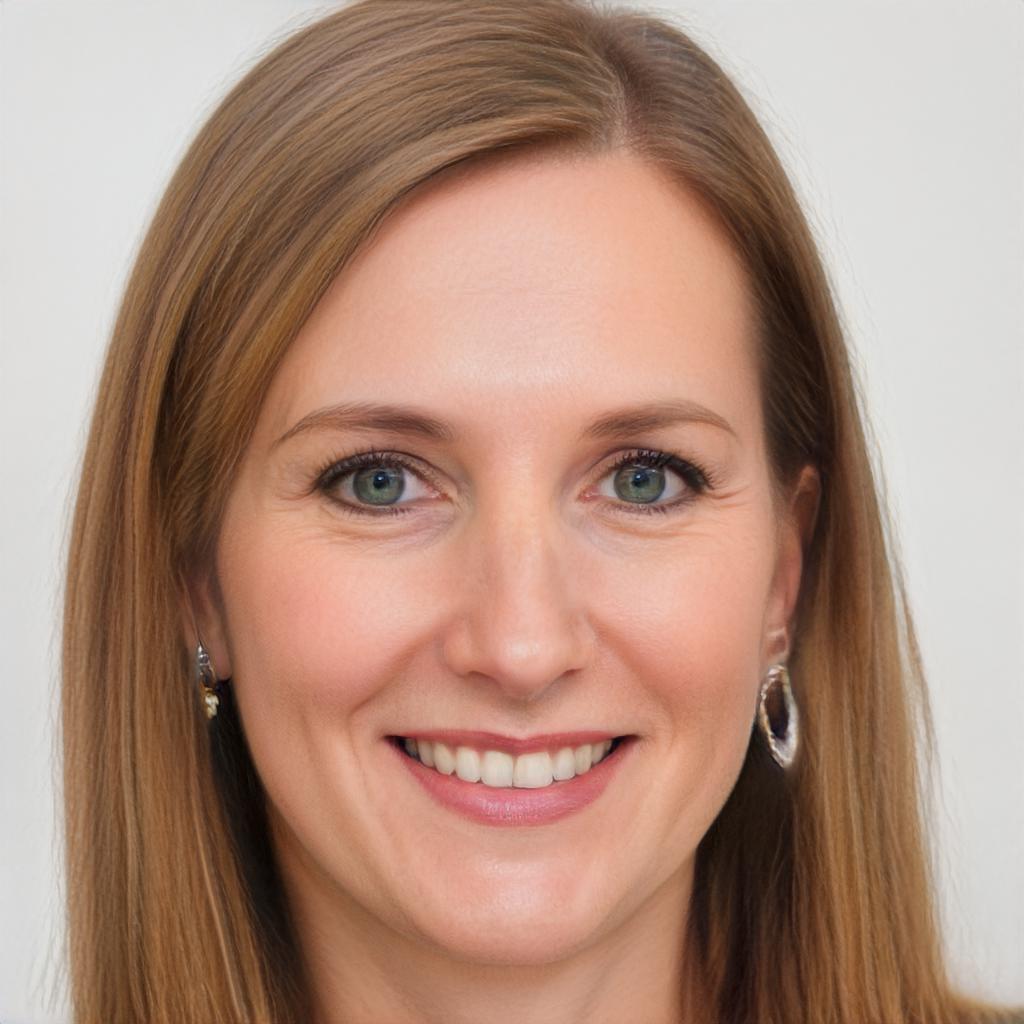} &
        \includegraphics[width=0.075\textwidth]{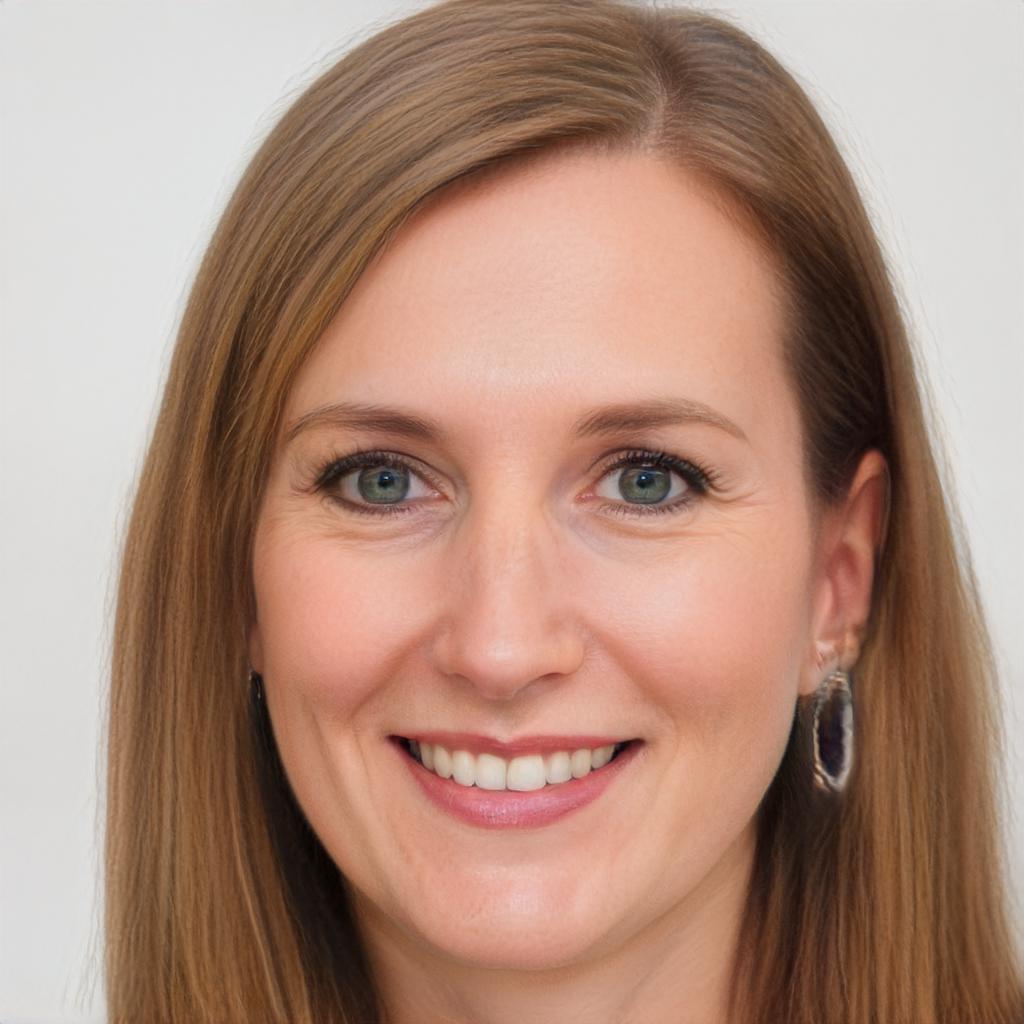} &
        \includegraphics[width=0.075\textwidth]{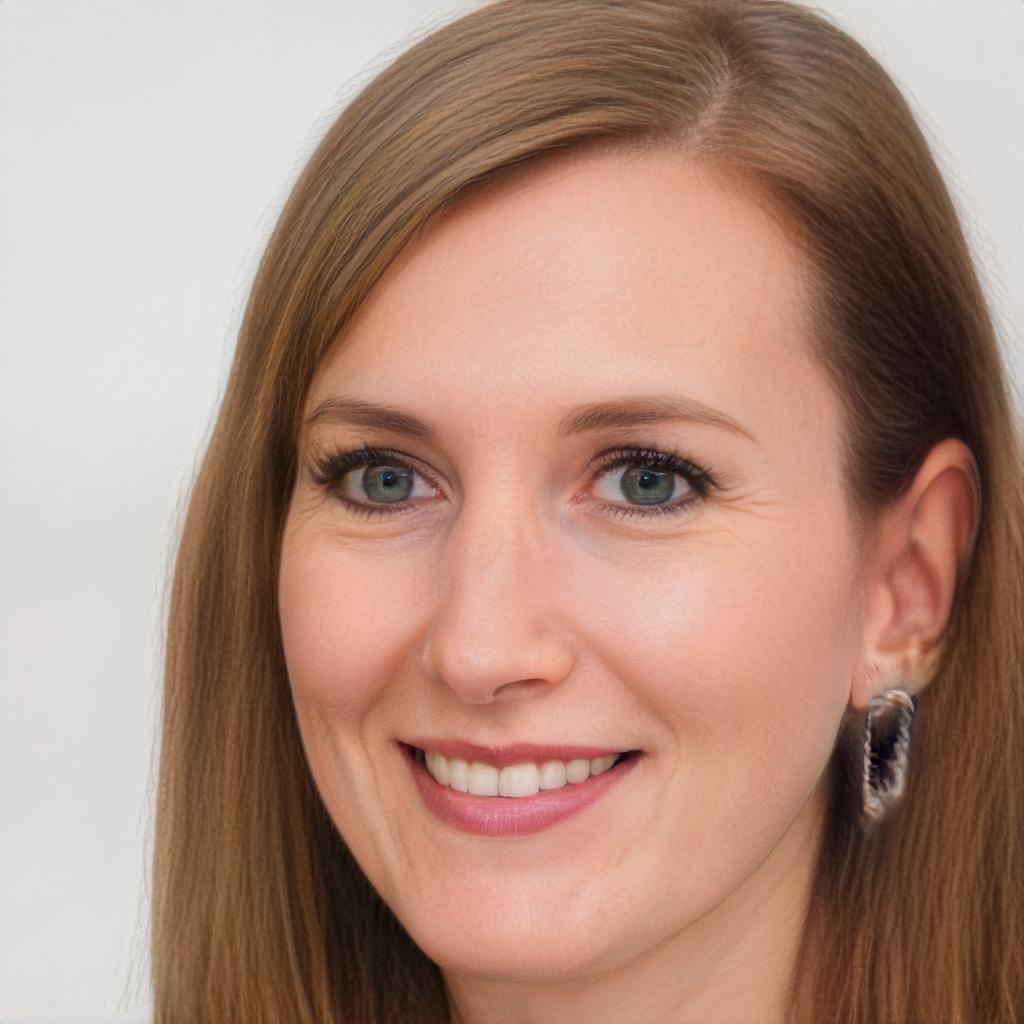} &
        \includegraphics[width=0.075\textwidth]{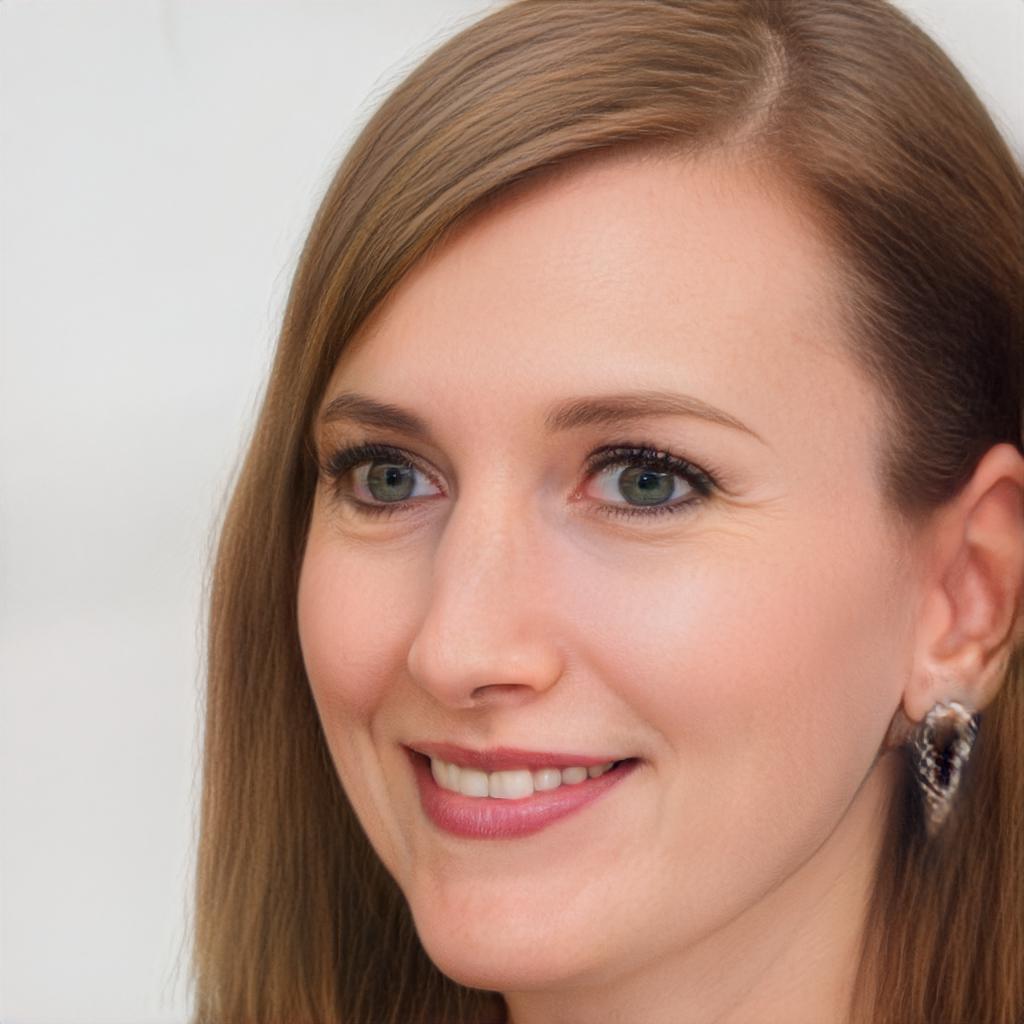} \\
        
        & \multicolumn{5}{c}{ Pose }
        
    \end{tabular}
    
    \\
    
        \begin{tabular}{c c c c c c}

        \raisebox{0.08in}{\rotatebox{90}{\footnotesize StyleFlow}} &
        \includegraphics[width=0.075\textwidth]{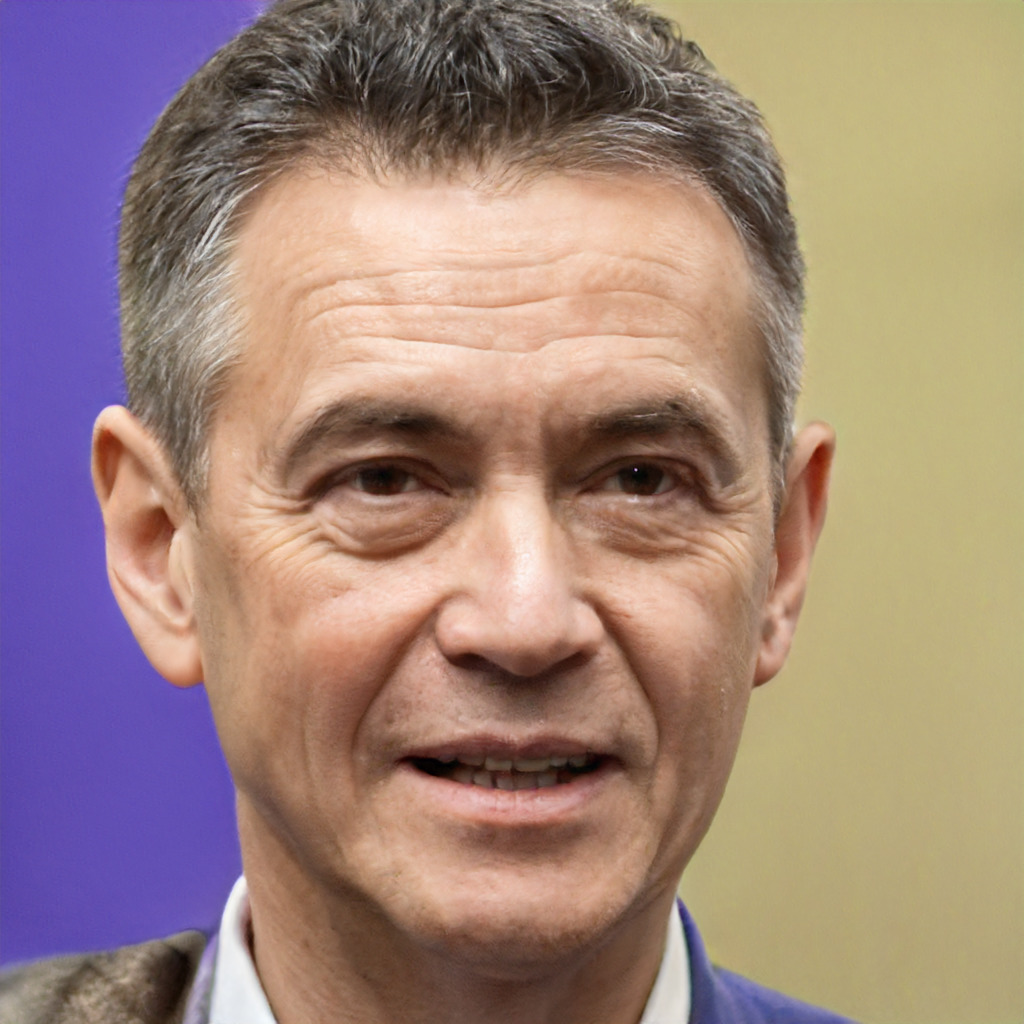} &
        \includegraphics[width=0.075\textwidth]{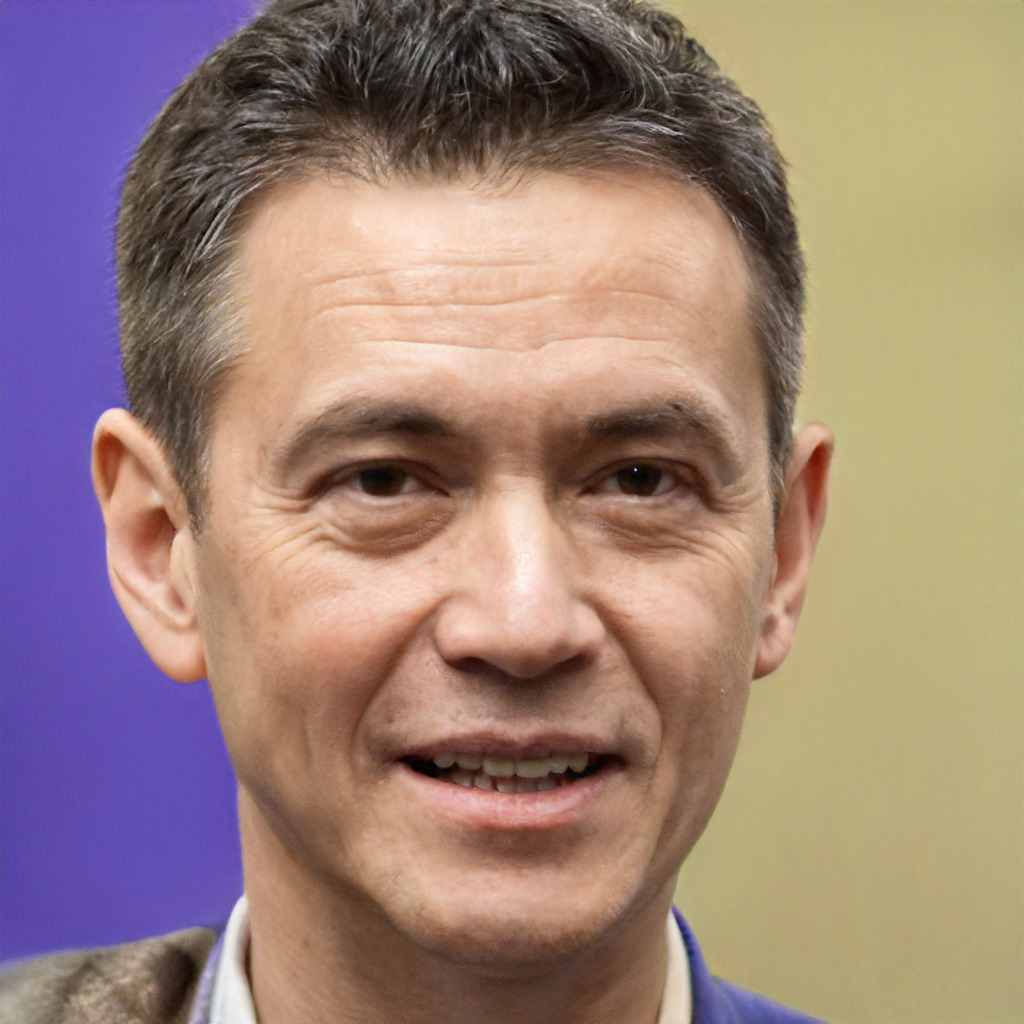} &
        \includegraphics[width=0.075\textwidth]{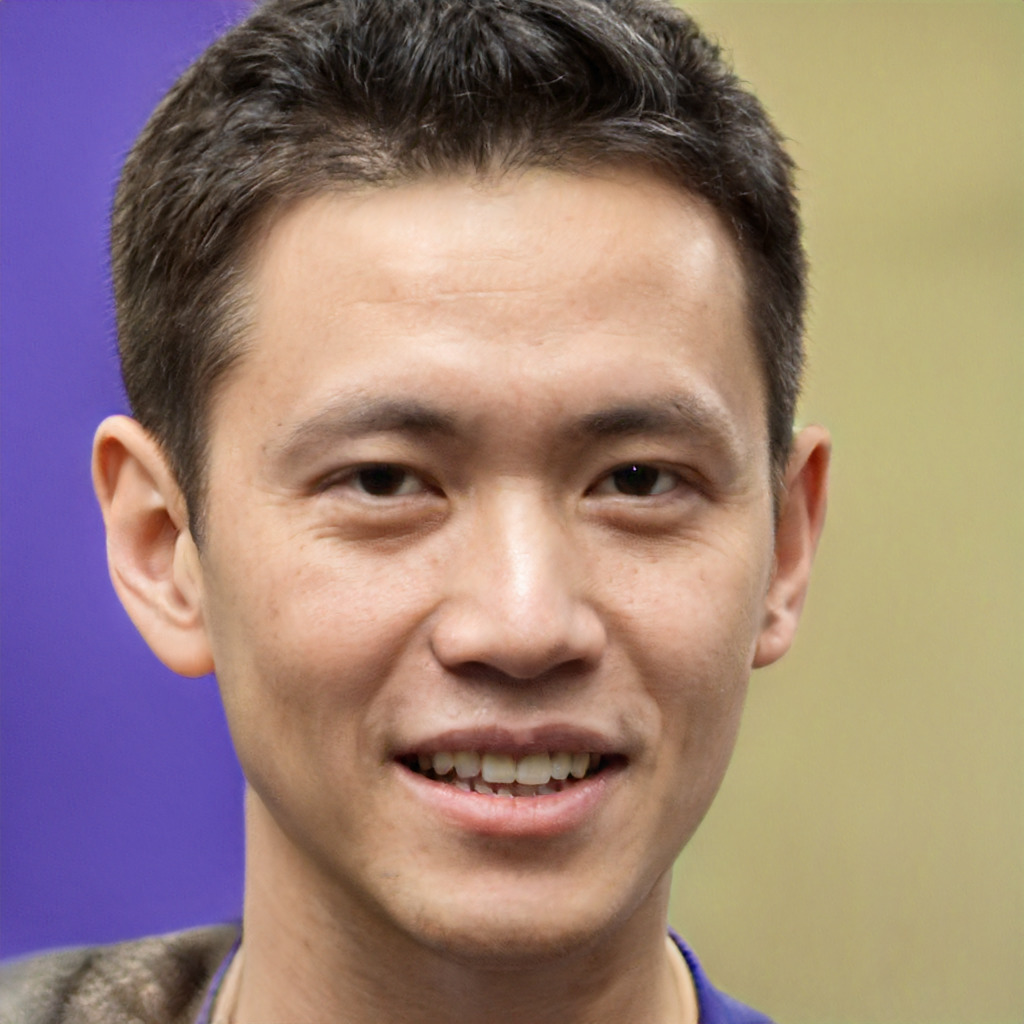} &
        \includegraphics[width=0.075\textwidth]{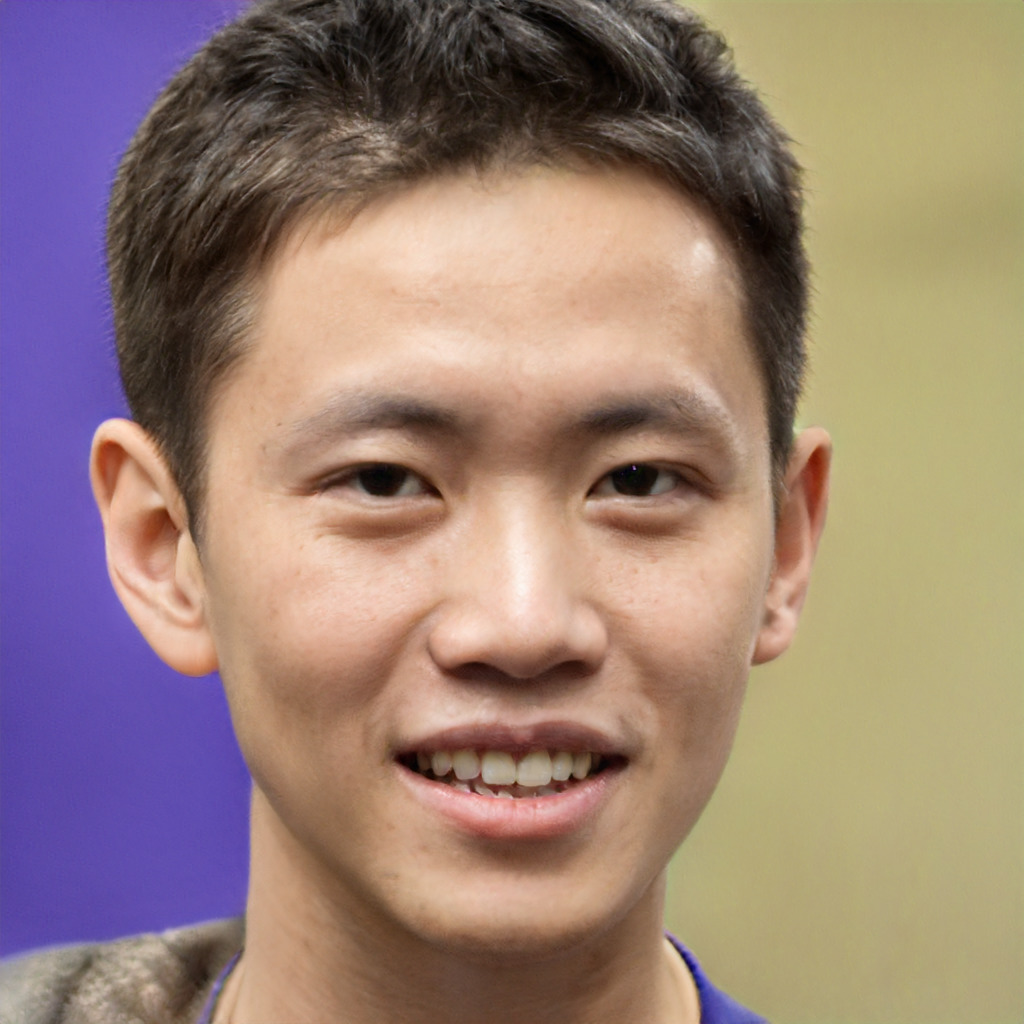} &
        \includegraphics[width=0.075\textwidth]{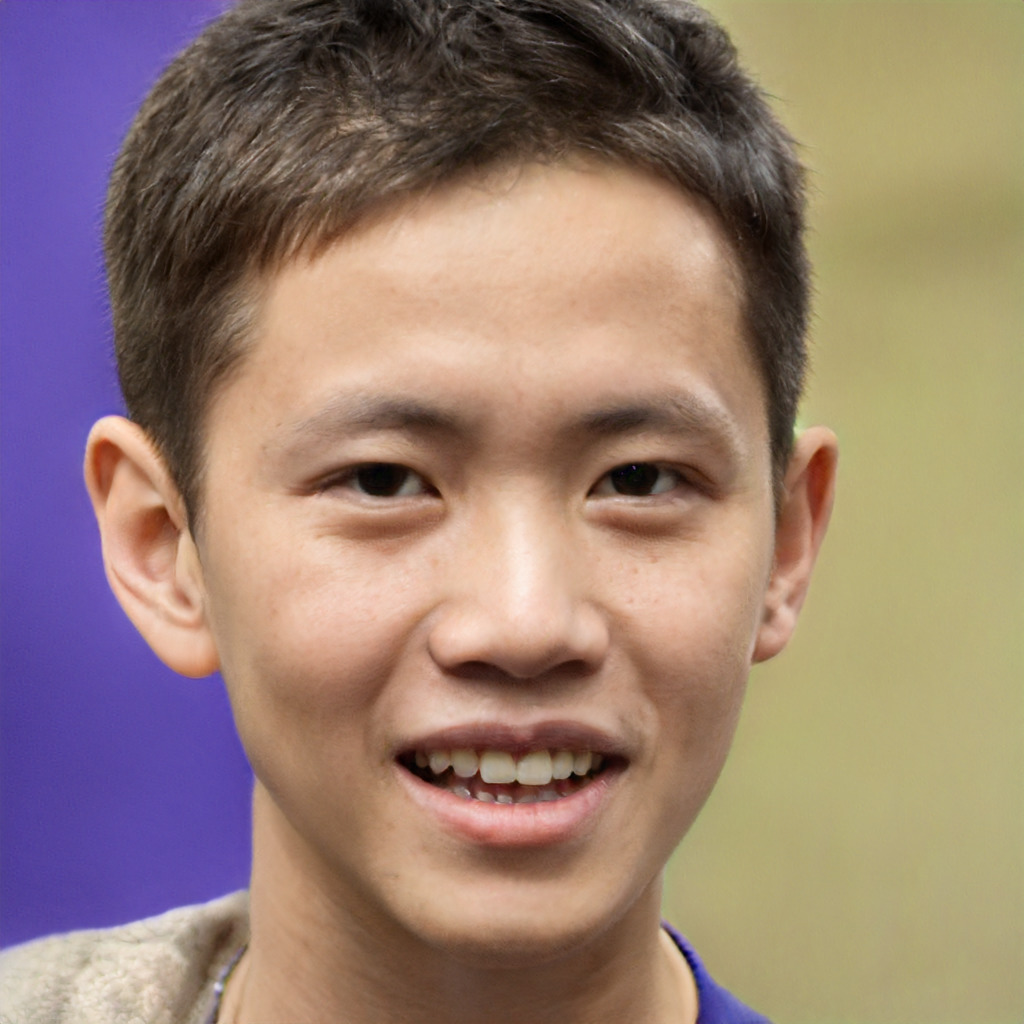} \\
        
        \raisebox{0.155in}{\rotatebox{90}{Ours}} &
        \includegraphics[width=0.075\textwidth]{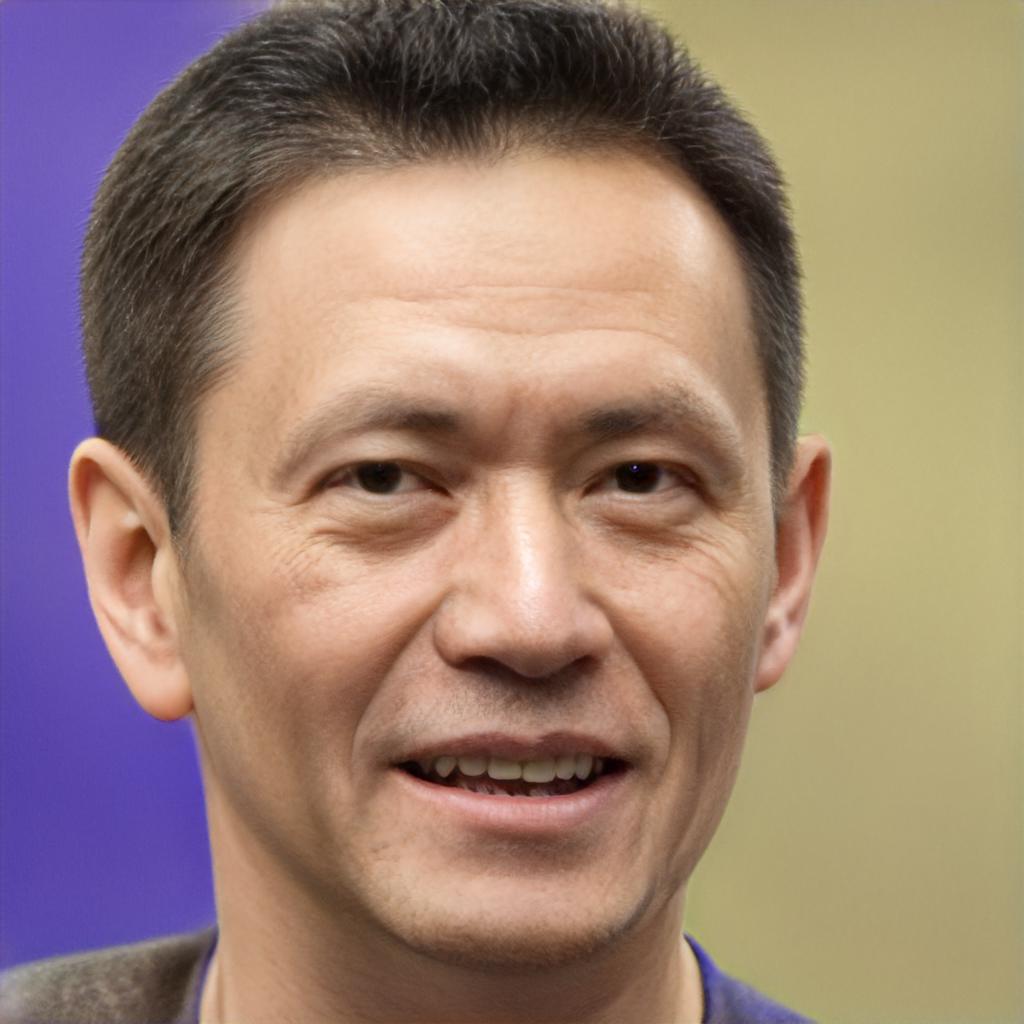} &
        \includegraphics[width=0.075\textwidth]{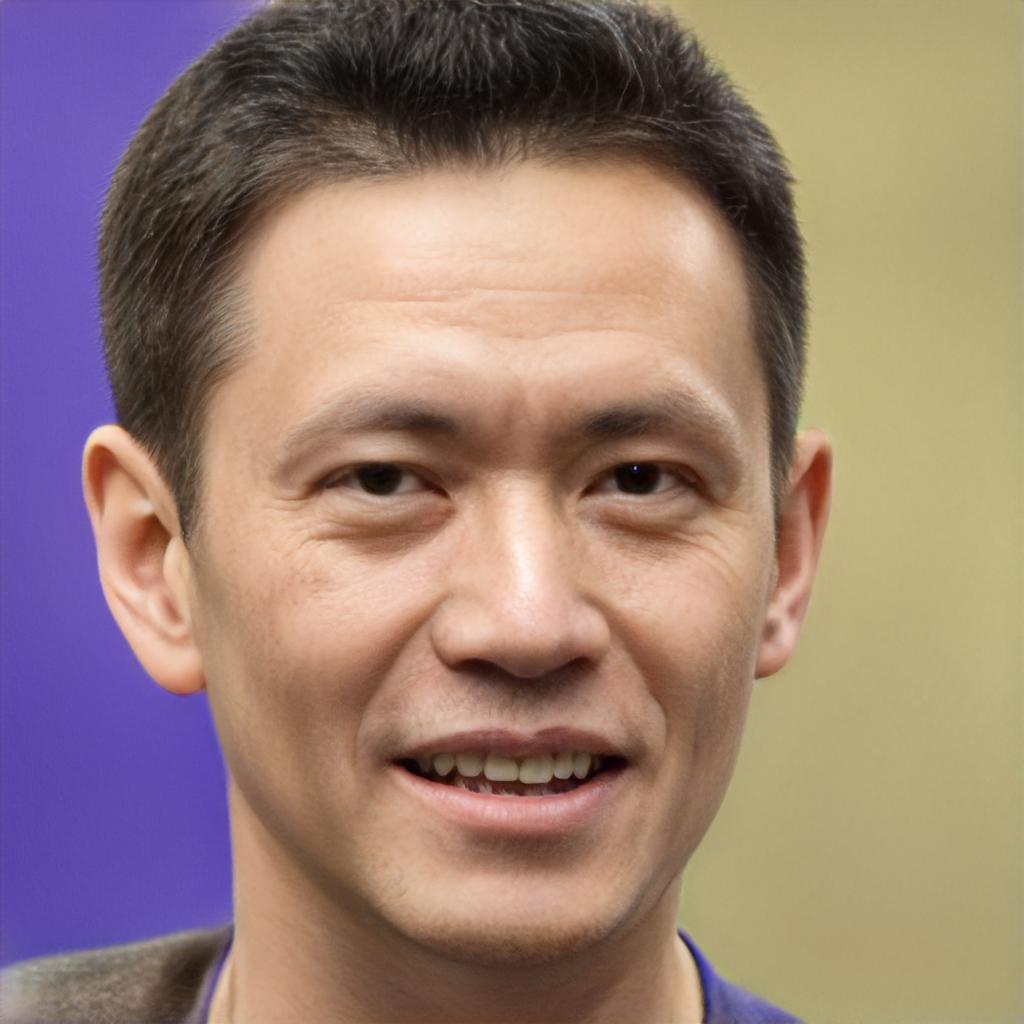} &
        \includegraphics[width=0.075\textwidth]{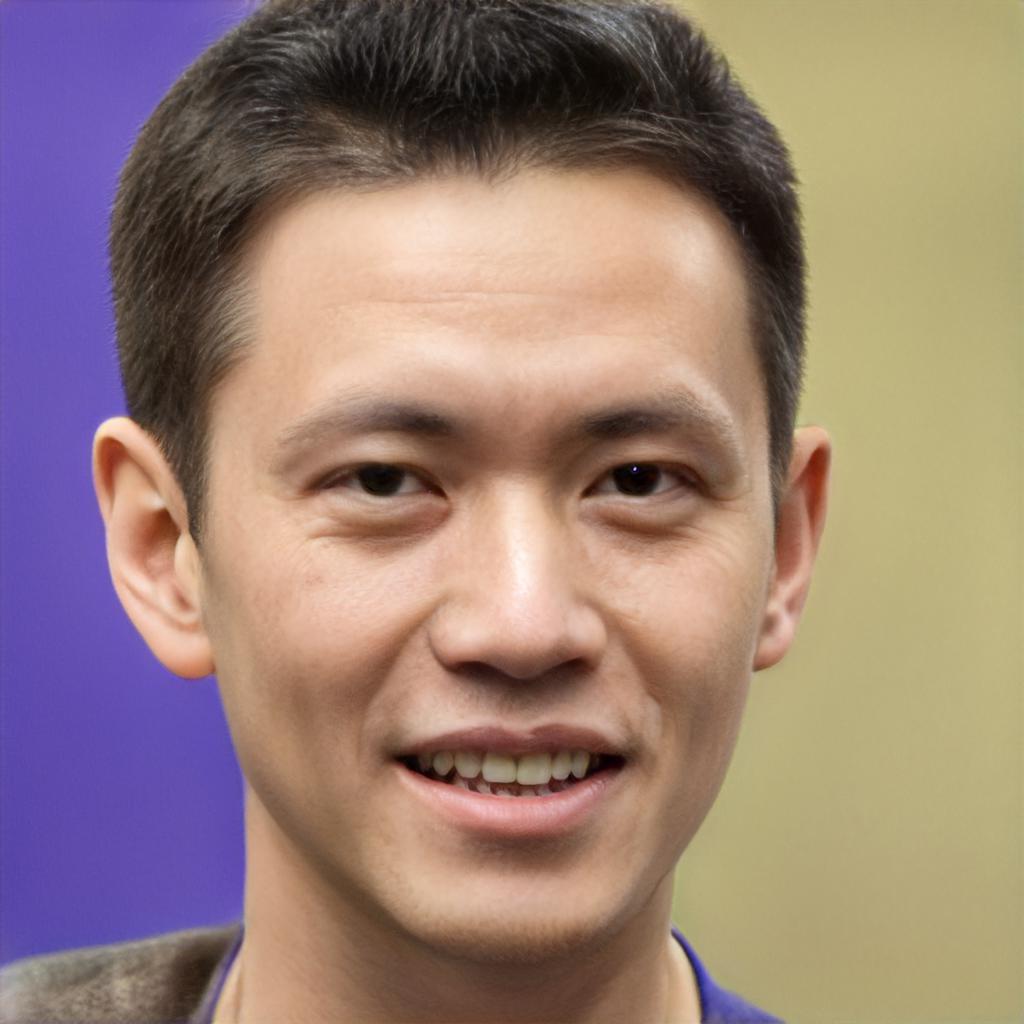} &
        \includegraphics[width=0.075\textwidth]{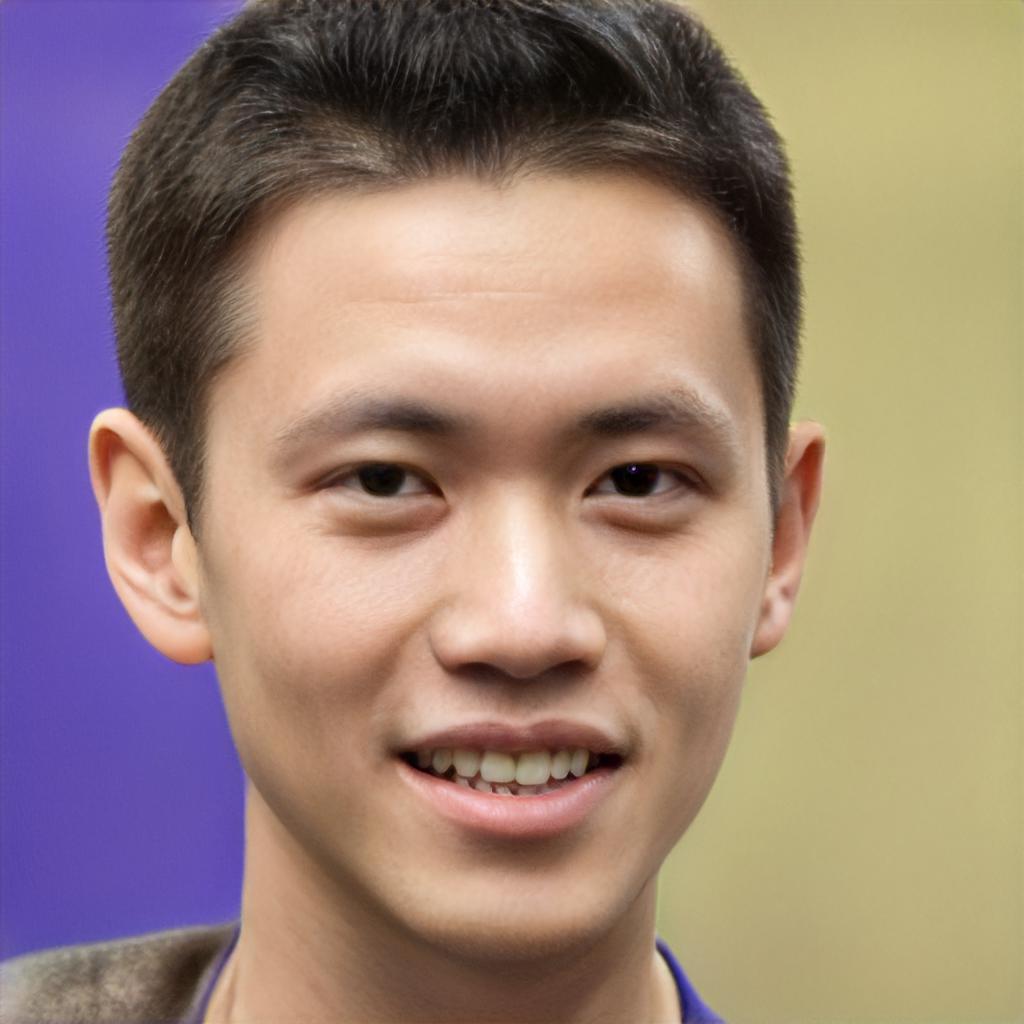} &
        \includegraphics[width=0.075\textwidth]{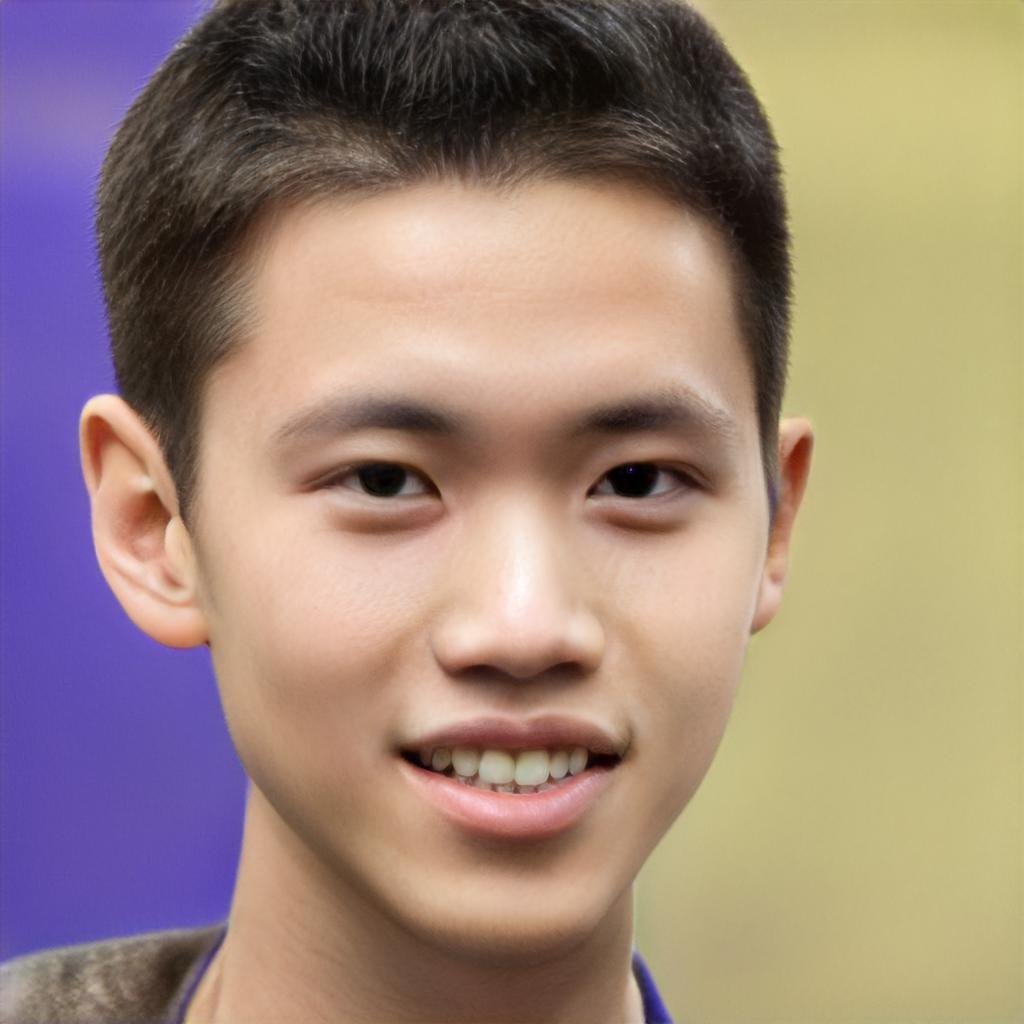} \\
        
        & \multicolumn{5}{c}{ Age }
        
    \end{tabular}
    
    &
    
        \begin{tabular}{c c c c c c}

        \raisebox{0.08in}{\rotatebox{90}{\footnotesize StyleFlow}} &
        \includegraphics[width=0.075\textwidth]{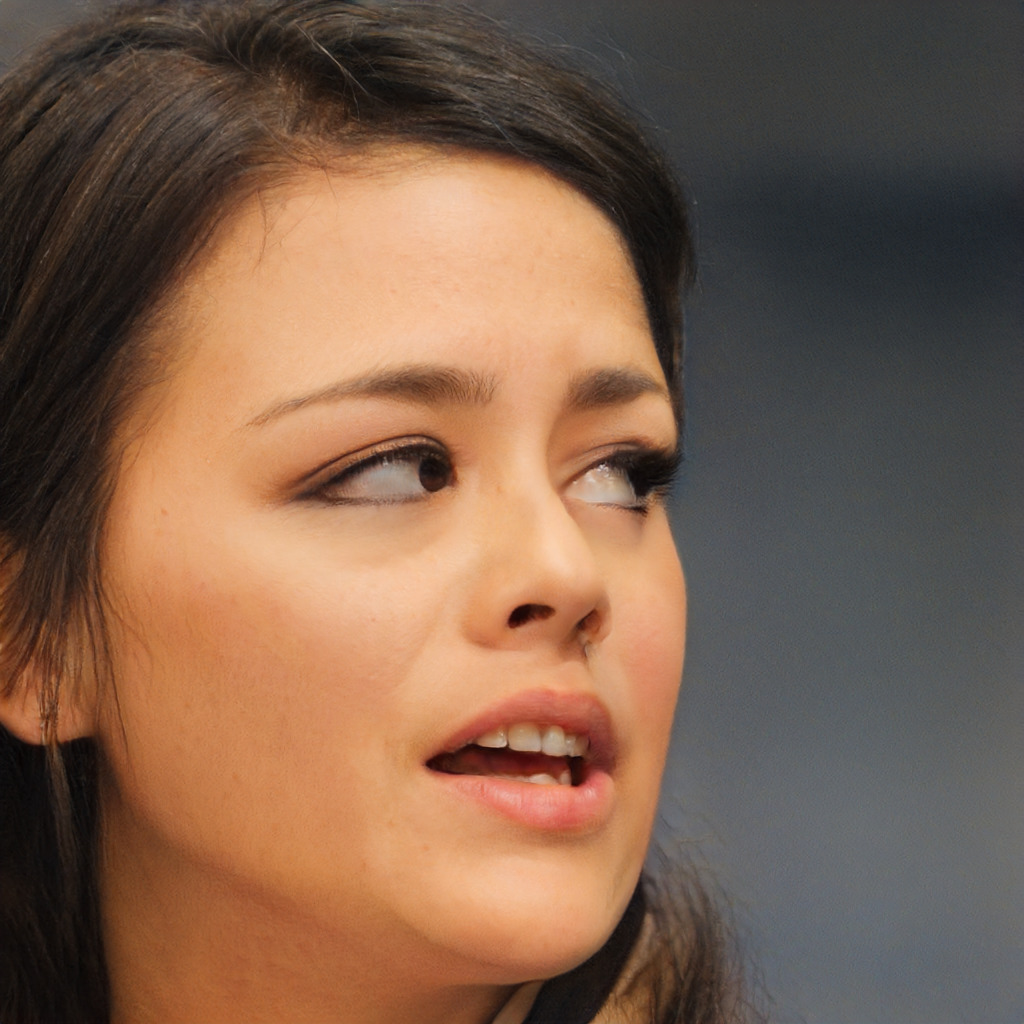} &
        \includegraphics[width=0.075\textwidth]{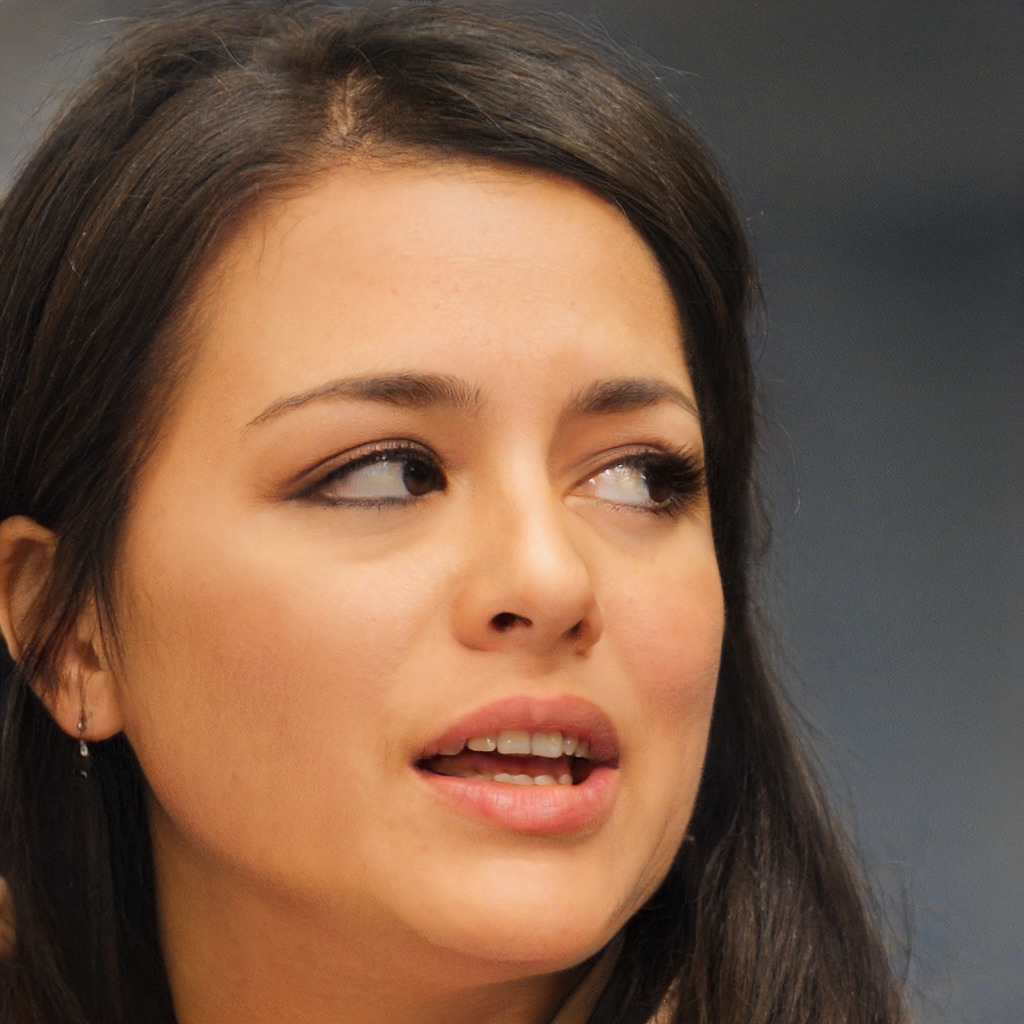} &
        \includegraphics[width=0.075\textwidth]{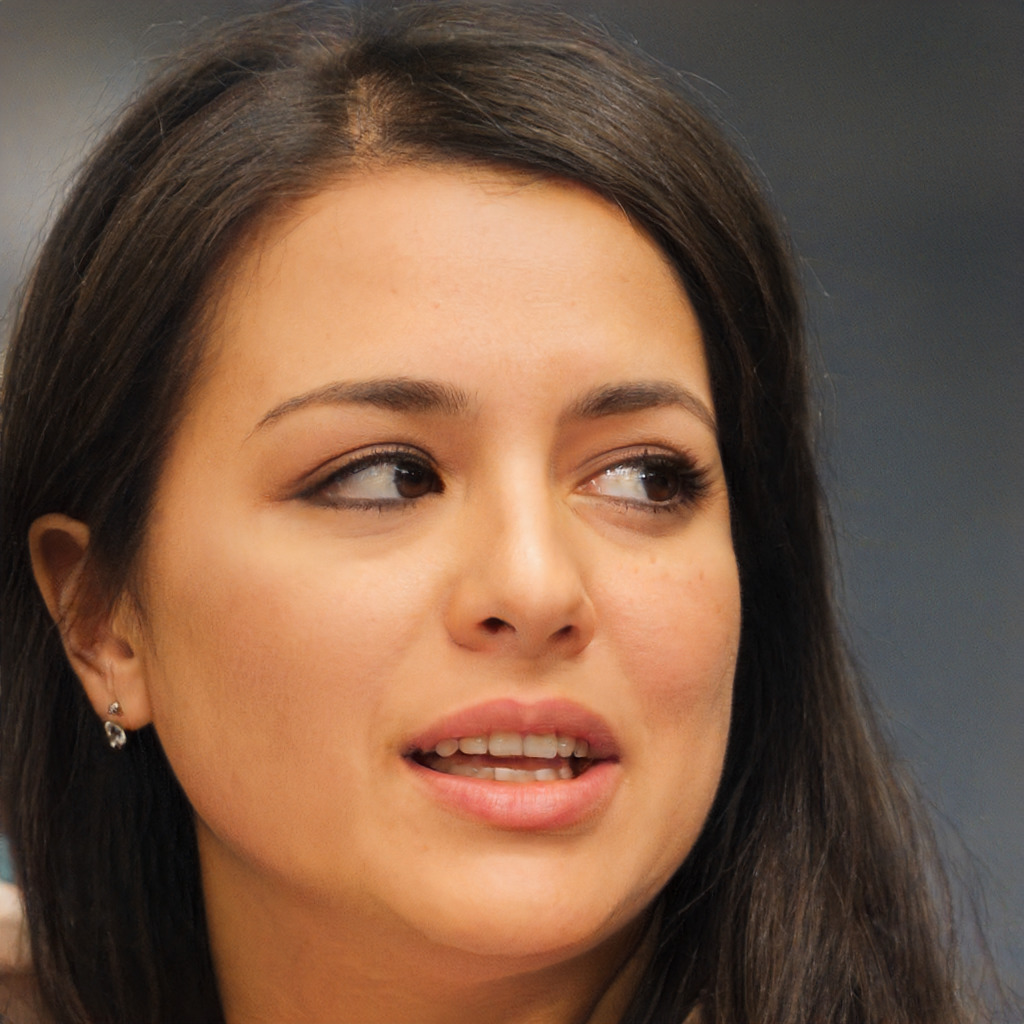} &
        \includegraphics[width=0.075\textwidth]{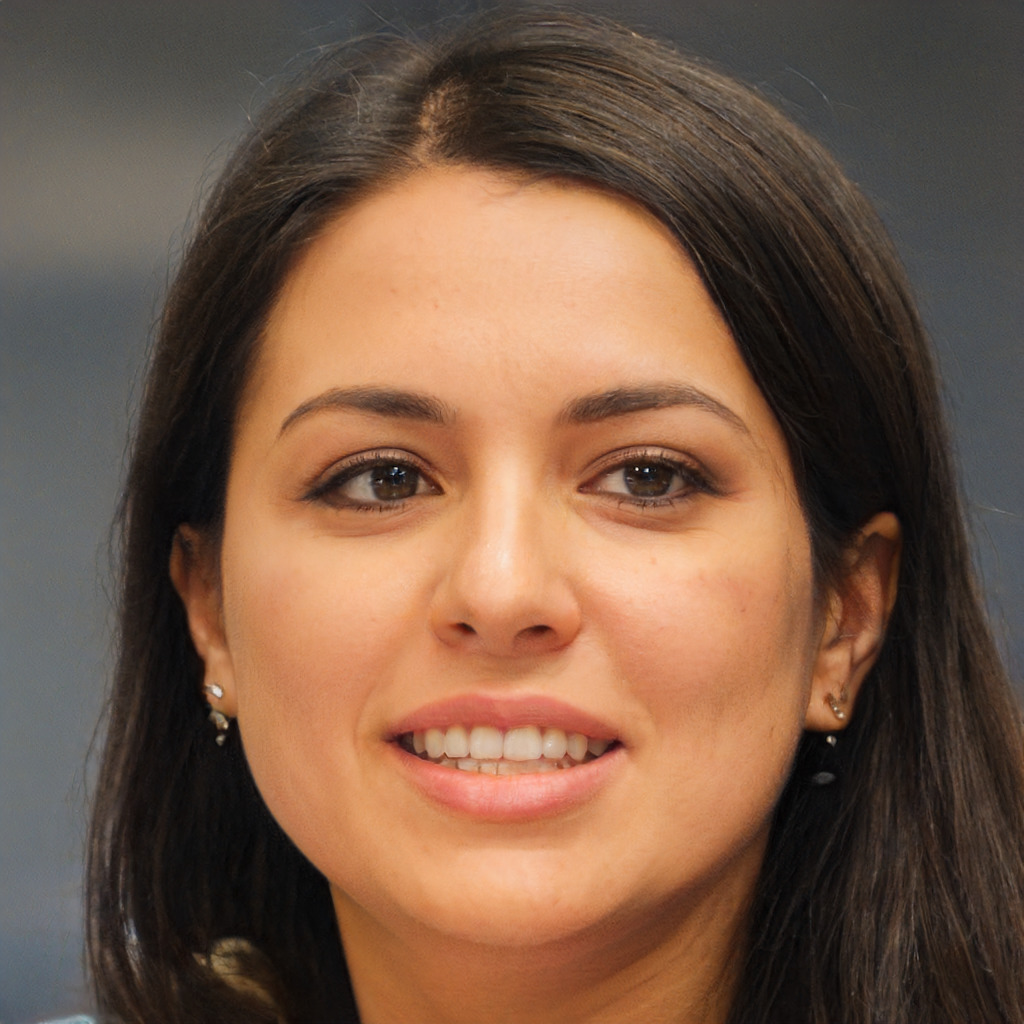} &
        \includegraphics[width=0.075\textwidth]{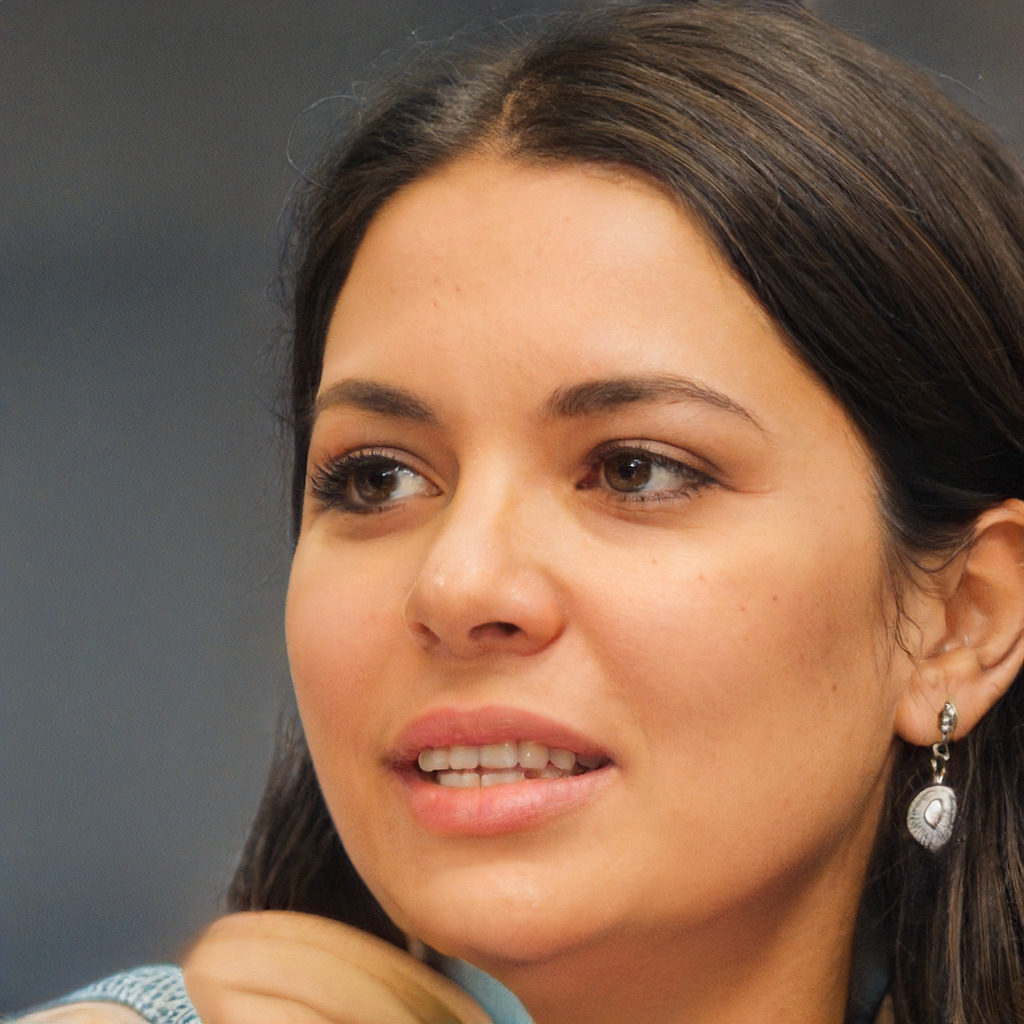} \\
        
        \raisebox{0.155in}{\rotatebox{90}{Ours}} &
        \includegraphics[width=0.075\textwidth]{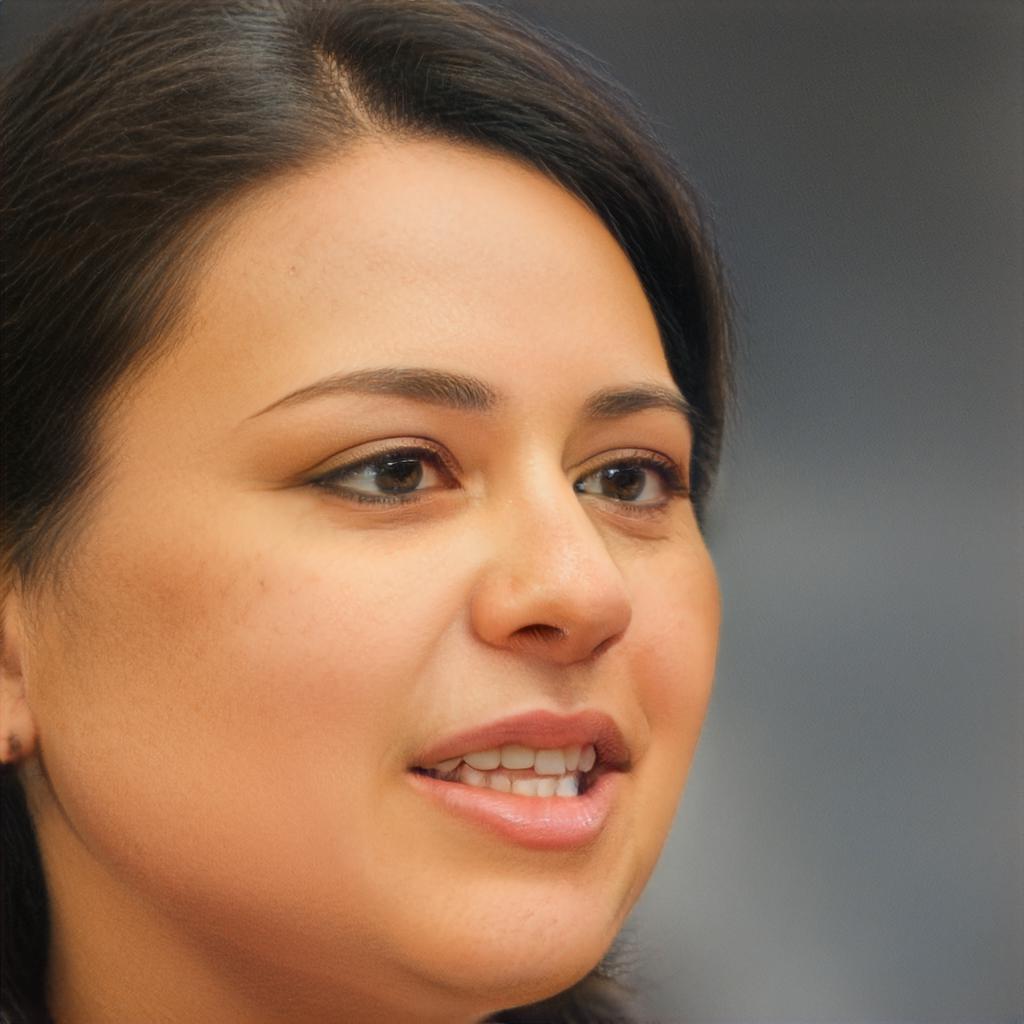} &
        \includegraphics[width=0.075\textwidth]{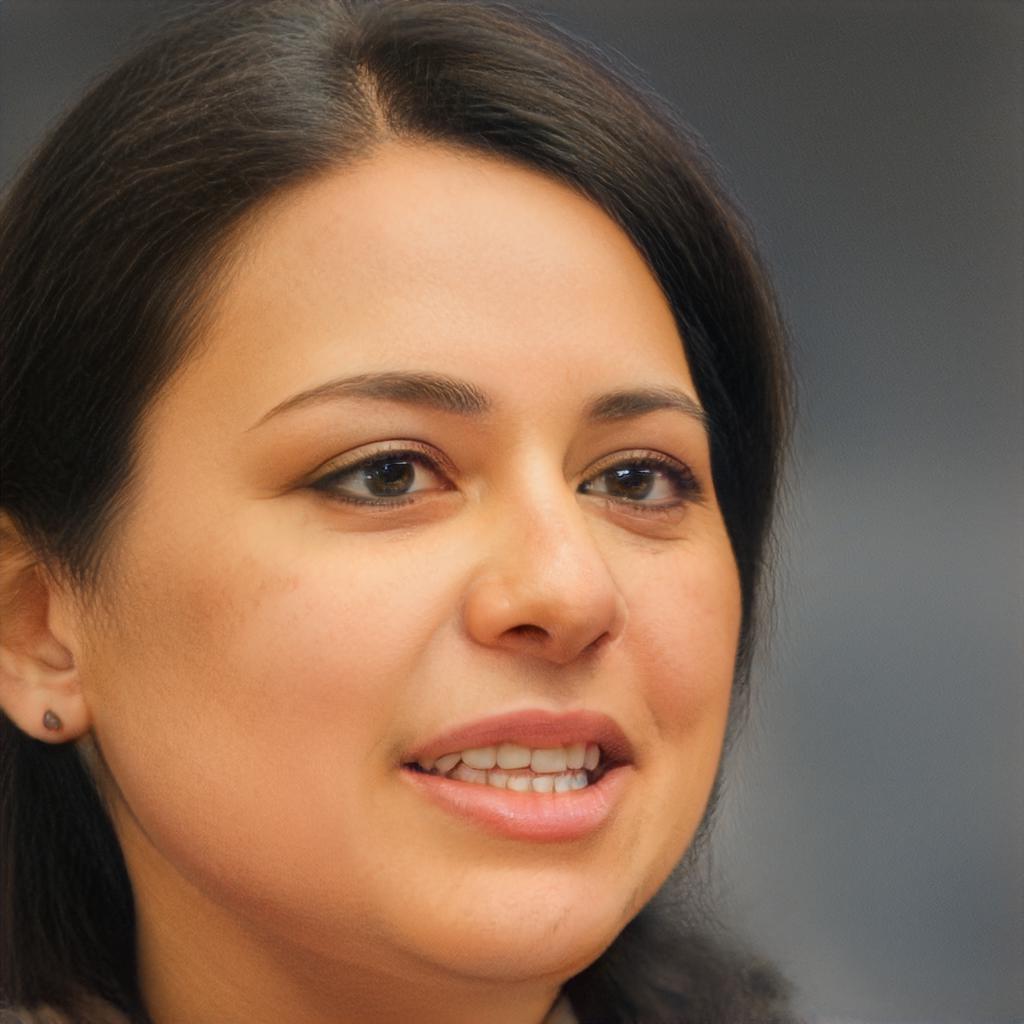} &
        \includegraphics[width=0.075\textwidth]{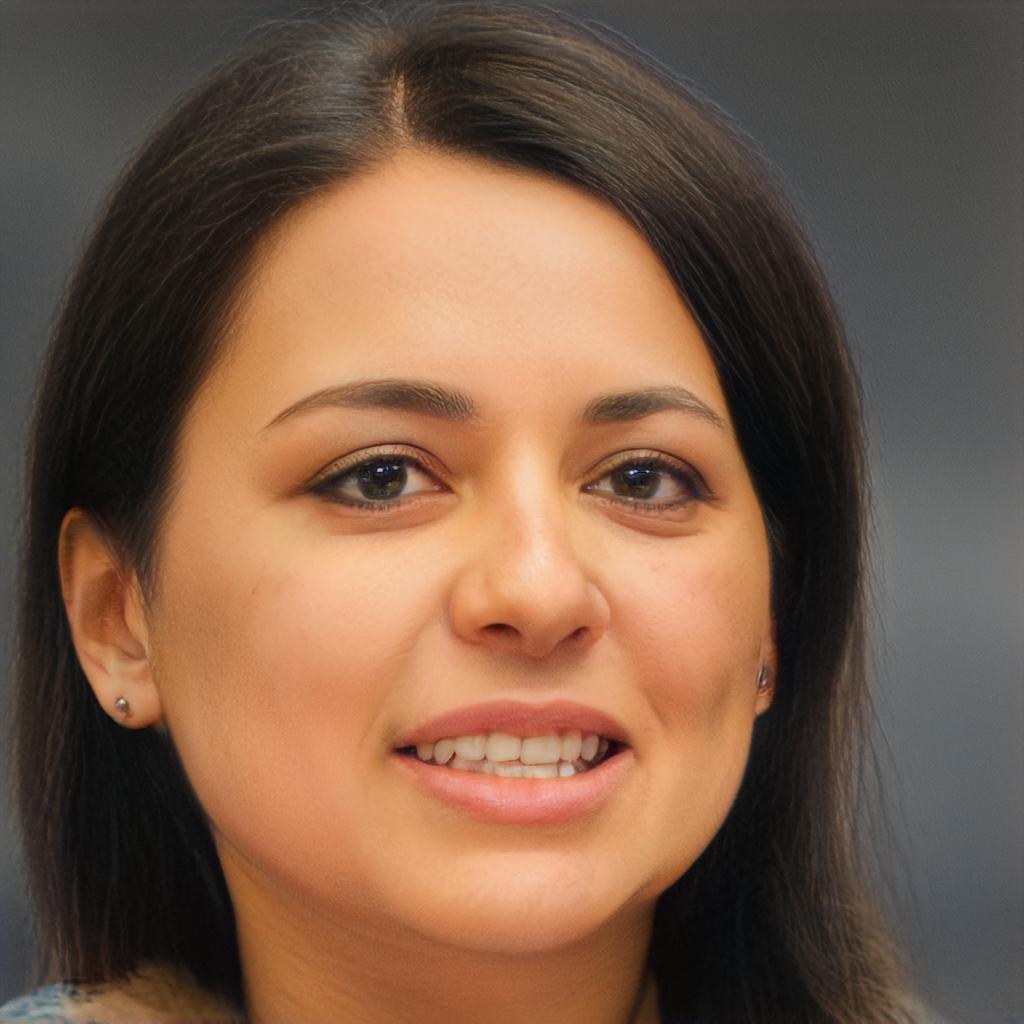} &
        \includegraphics[width=0.075\textwidth]{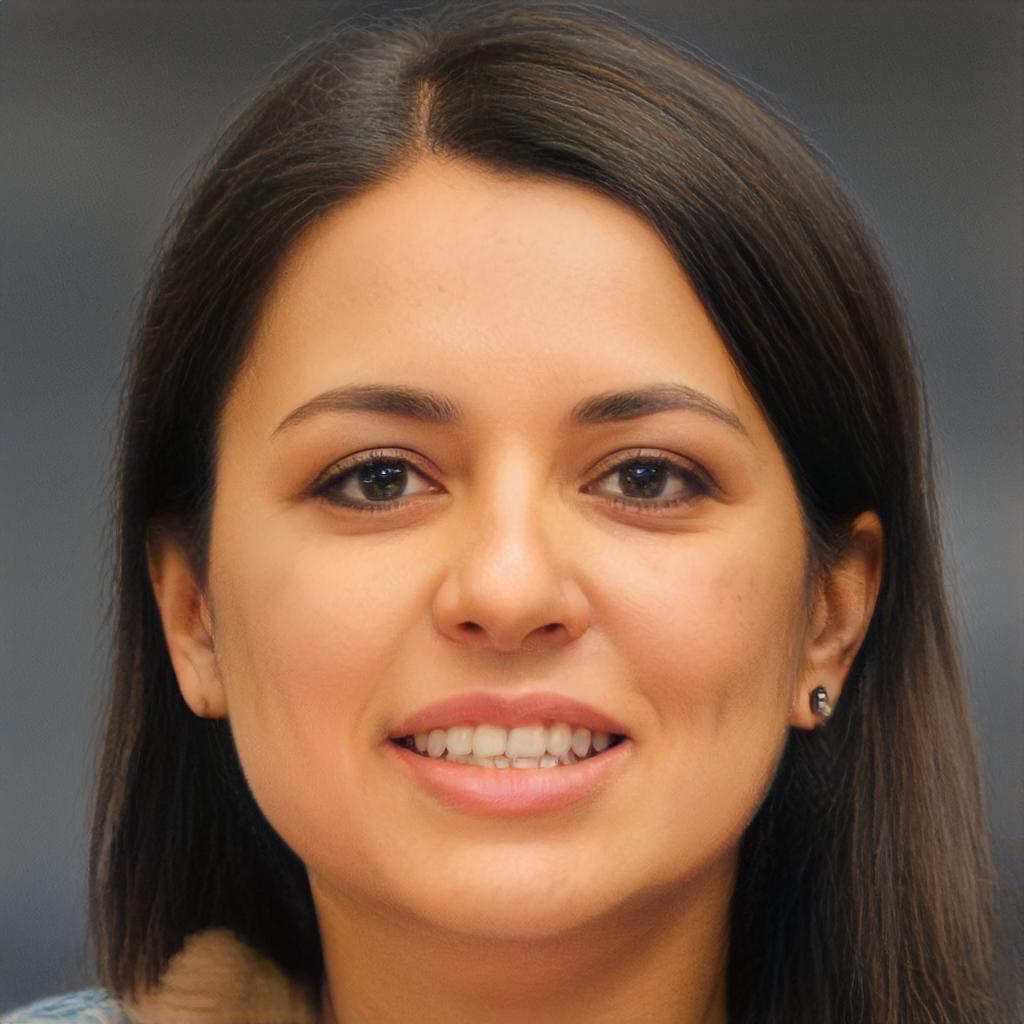} &
        \includegraphics[width=0.075\textwidth]{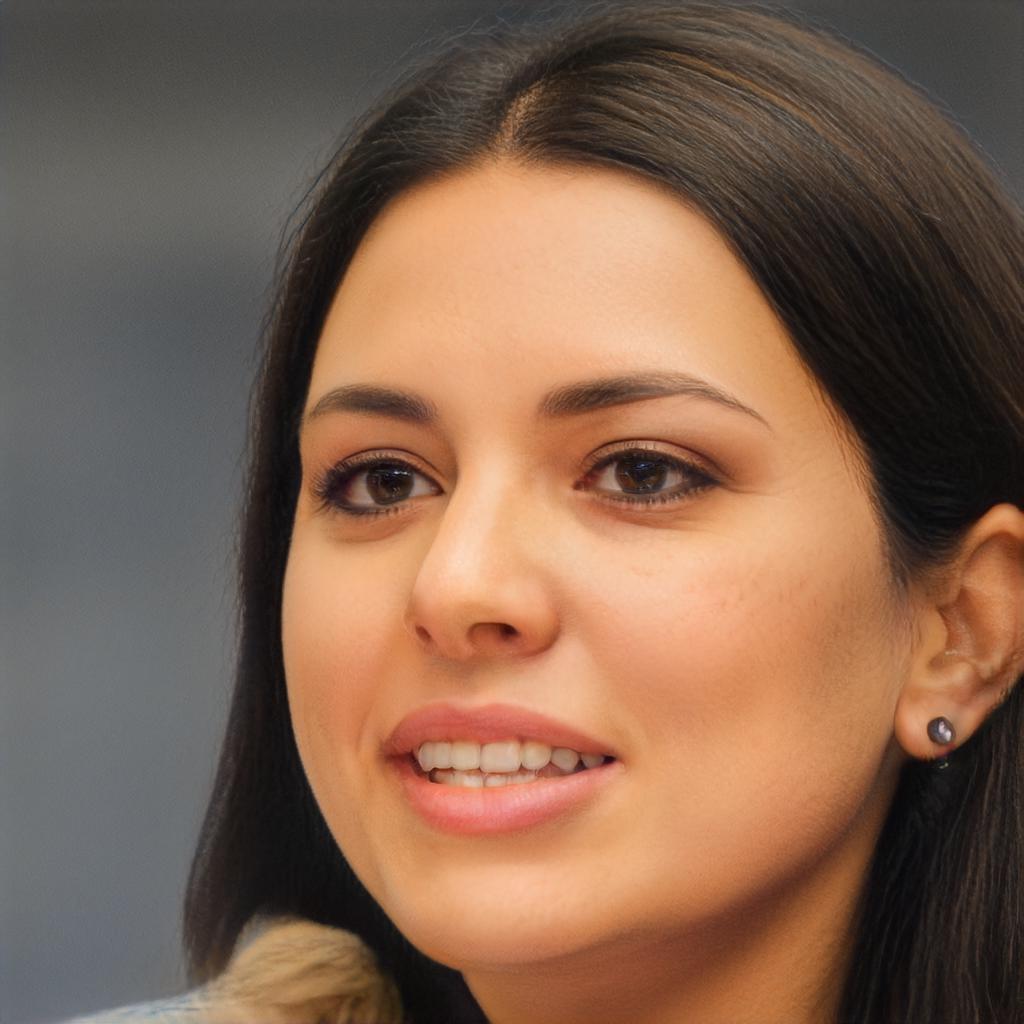} \\
        
        & \multicolumn{5}{c}{ Pose }
        
    \end{tabular}

    \end{tabular}
    
    }
    \vspace{-0.27cm}
    \caption{Comparisons to non-linear editing methods. Our model maintains better identity when adjusting age, particularly in the case of racial minorities. When modifying the pose, our model shows reduced corruption for large changes.}
    \label{fig:non_linear_comparisons}
\end{figure*}

%% file: resources/figures/edit_comparisons_inversions_arxiv.tex
\begin{figure*}
    \centering
    \setlength{\belowcaptionskip}{-8pt}
    \setlength{\tabcolsep}{3.25pt}
    {\small
    \begin{tabular}{c c | c | c c c c c }

        \raisebox{0.08in}{\rotatebox{90}{\footnotesize InterFaceGAN}} &
        \includegraphics[width=0.107\textwidth]{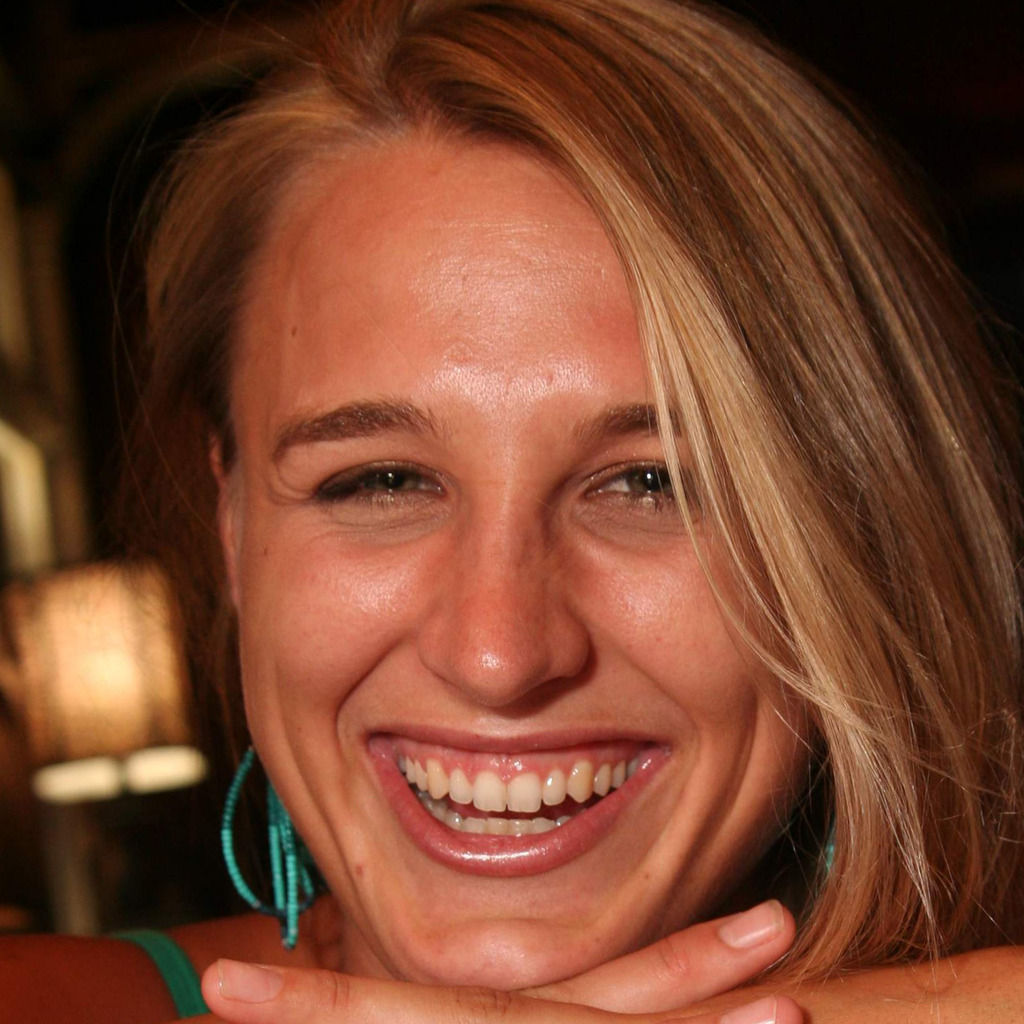} &
        \includegraphics[width=0.107\textwidth]{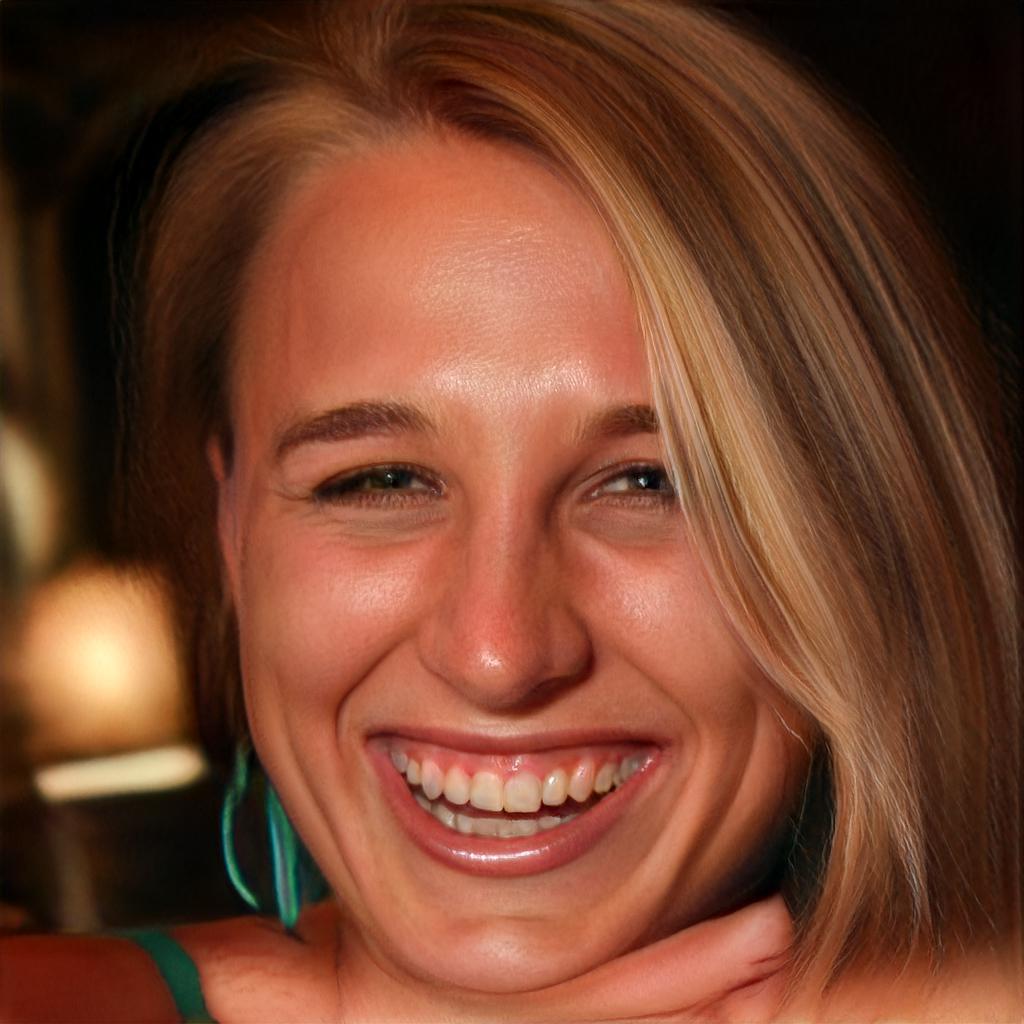} &
        \includegraphics[width=0.107\textwidth]{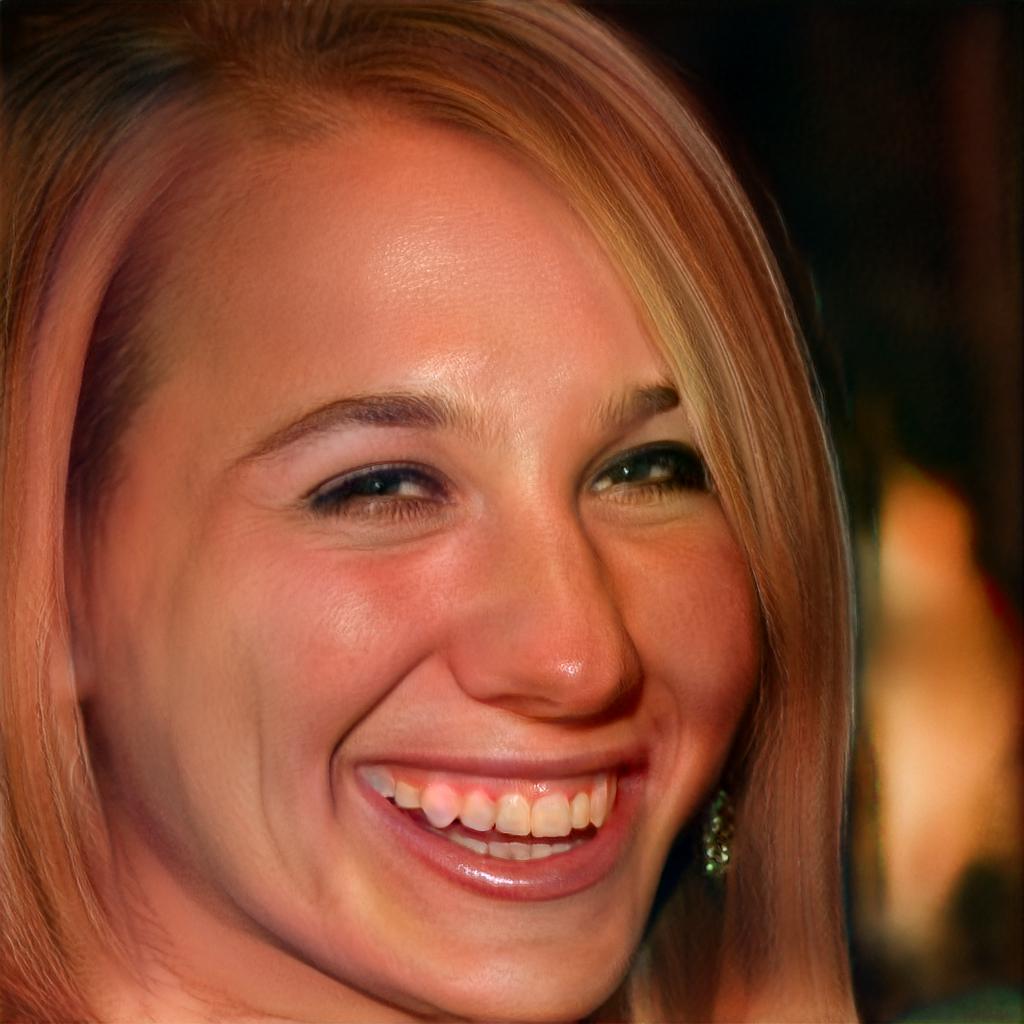} &
        \includegraphics[width=0.107\textwidth]{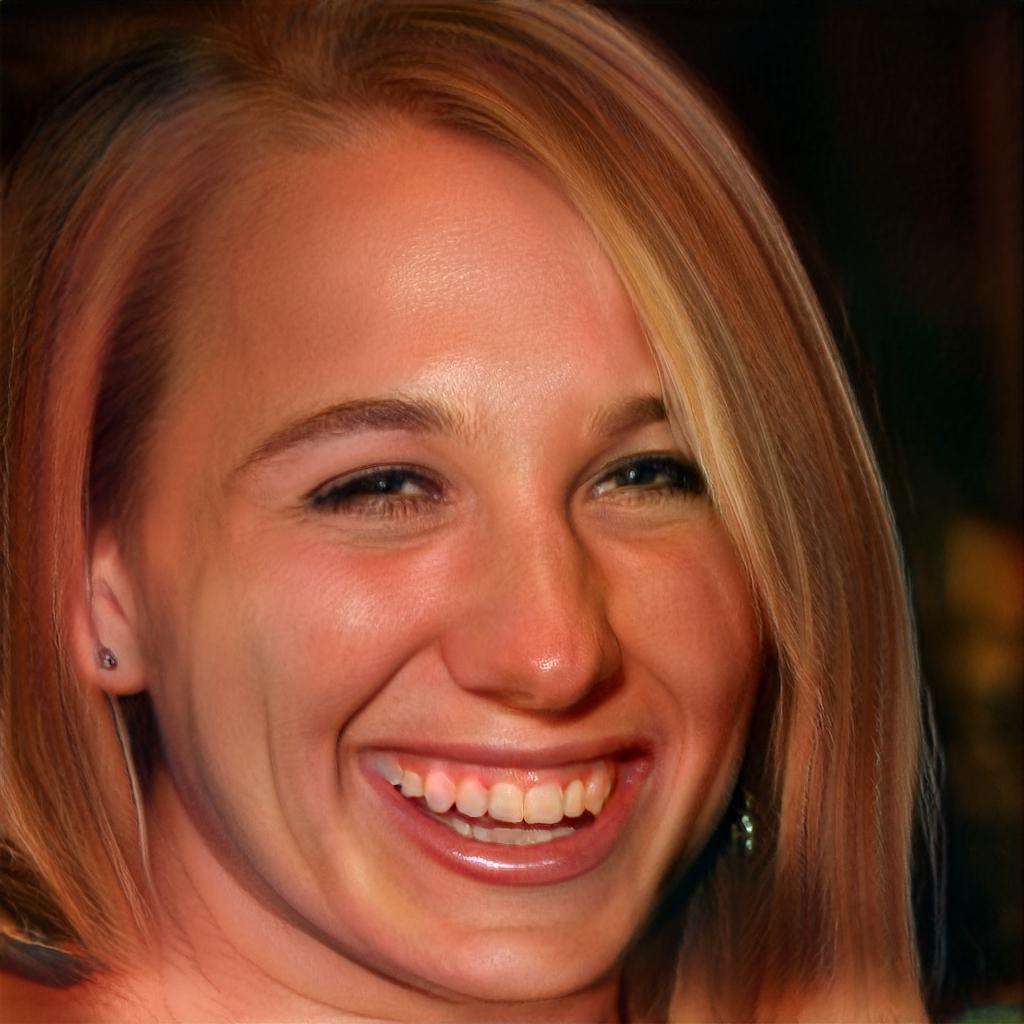} &
        \includegraphics[width=0.107\textwidth]{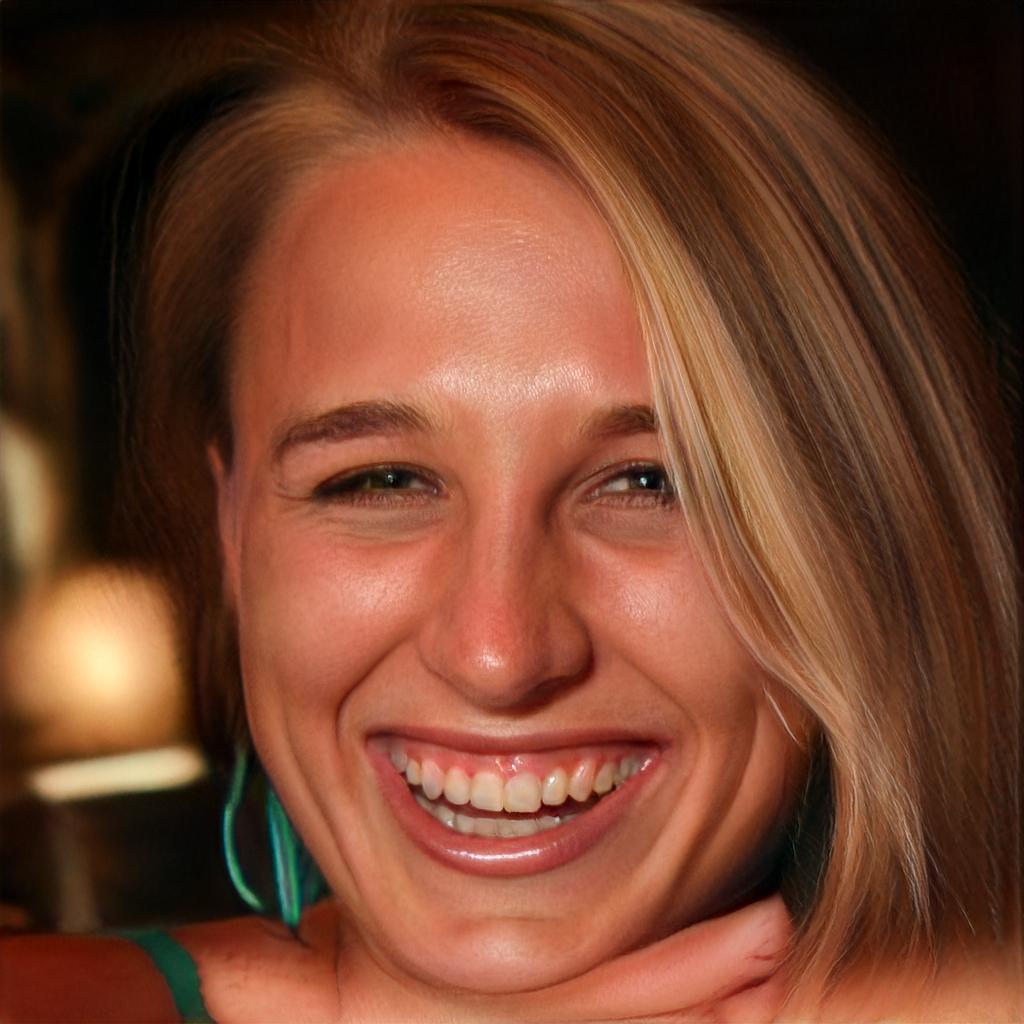} &
        \includegraphics[width=0.107\textwidth]{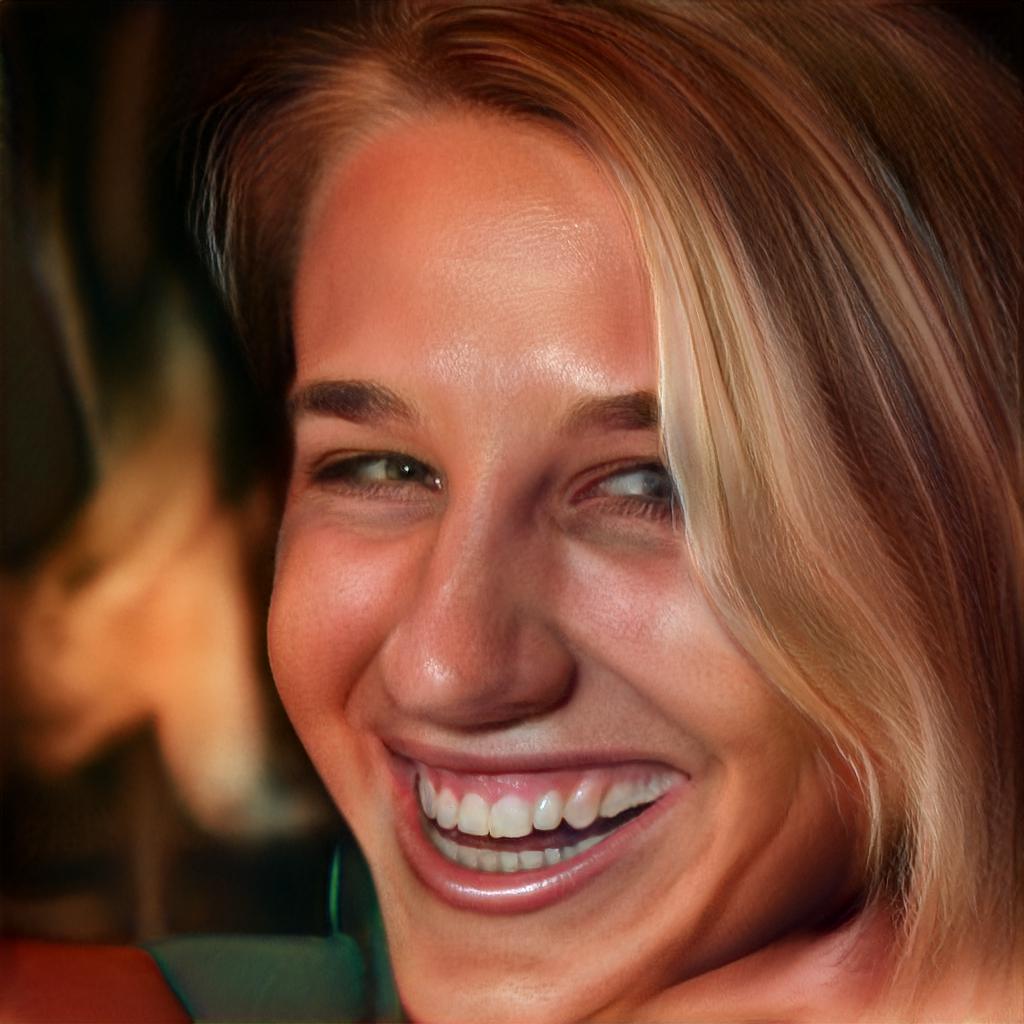} &
        \includegraphics[width=0.107\textwidth]{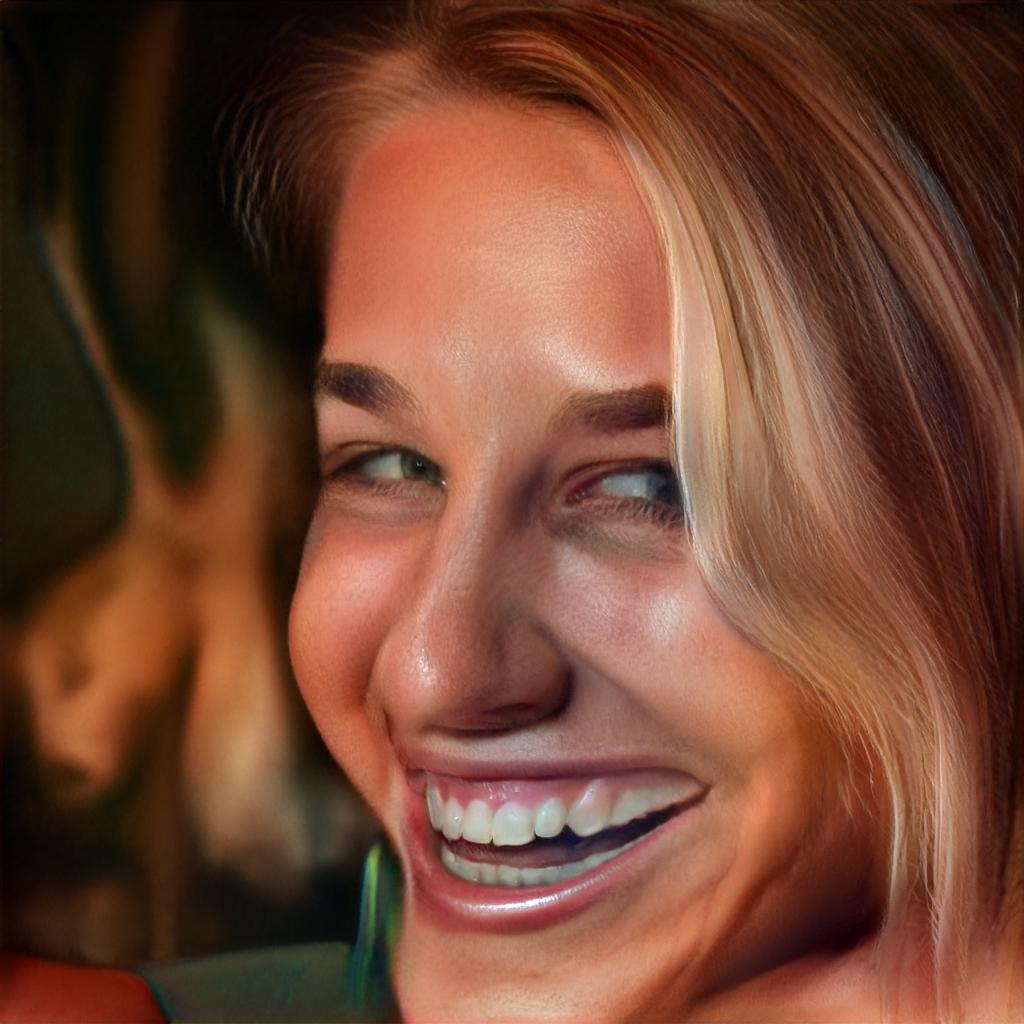} \\
        
        \raisebox{0.265in}{\rotatebox{90}{Ours}} &
        \includegraphics[width=0.107\textwidth]{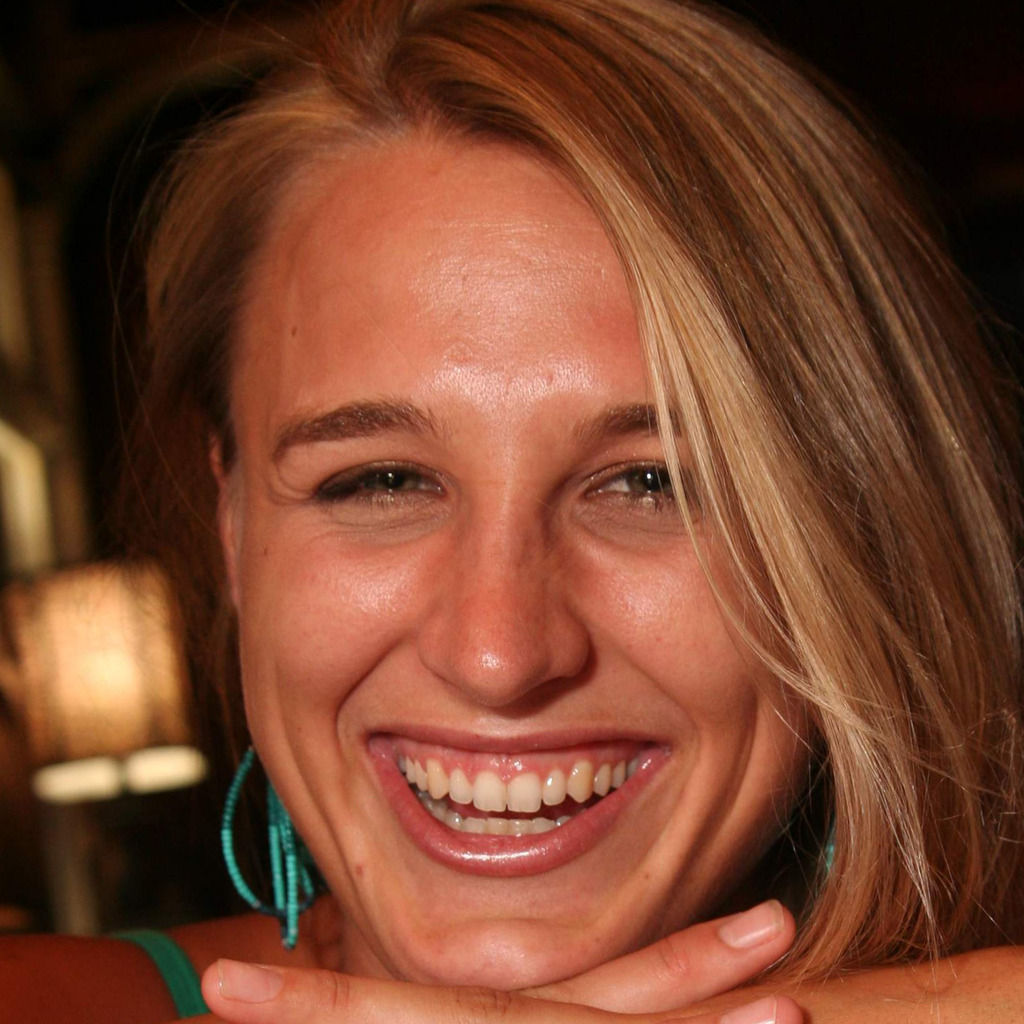} &
        \includegraphics[width=0.107\textwidth]{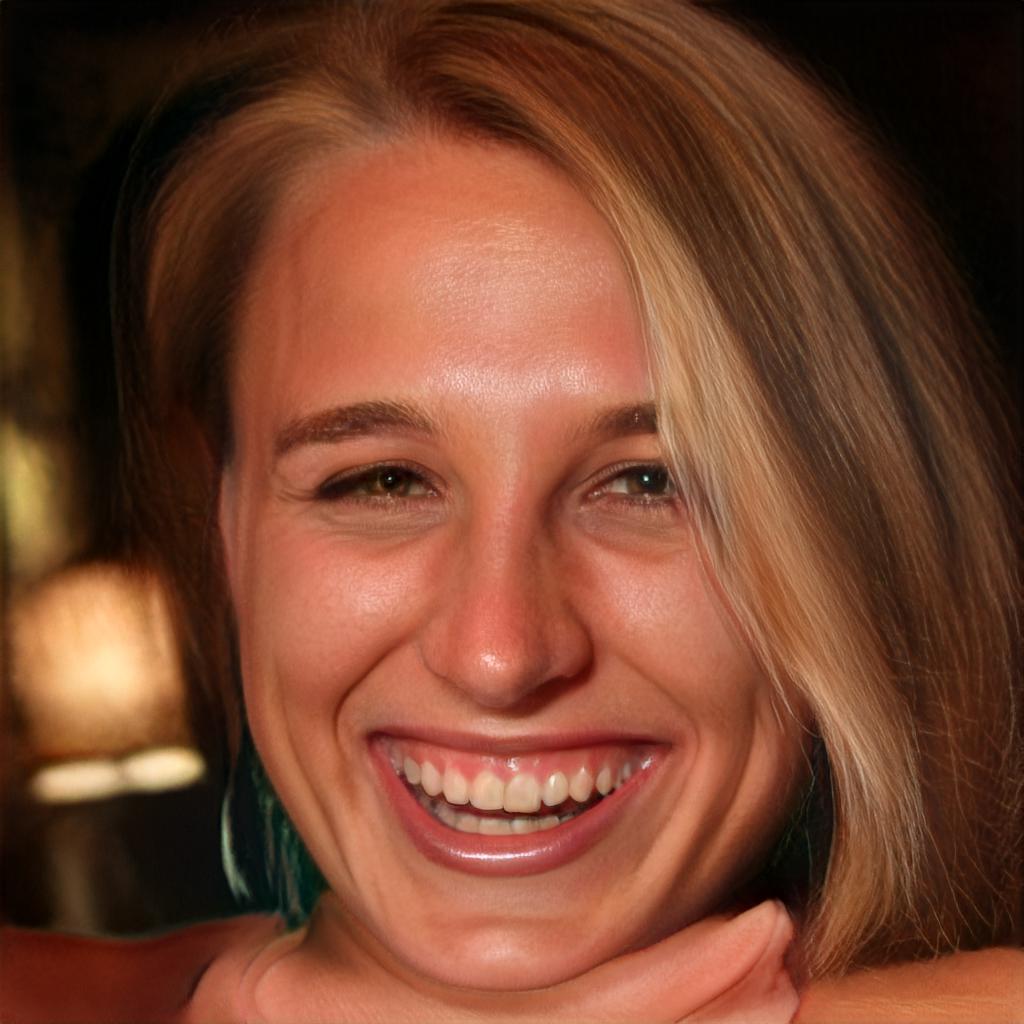} &
        \includegraphics[width=0.107\textwidth]{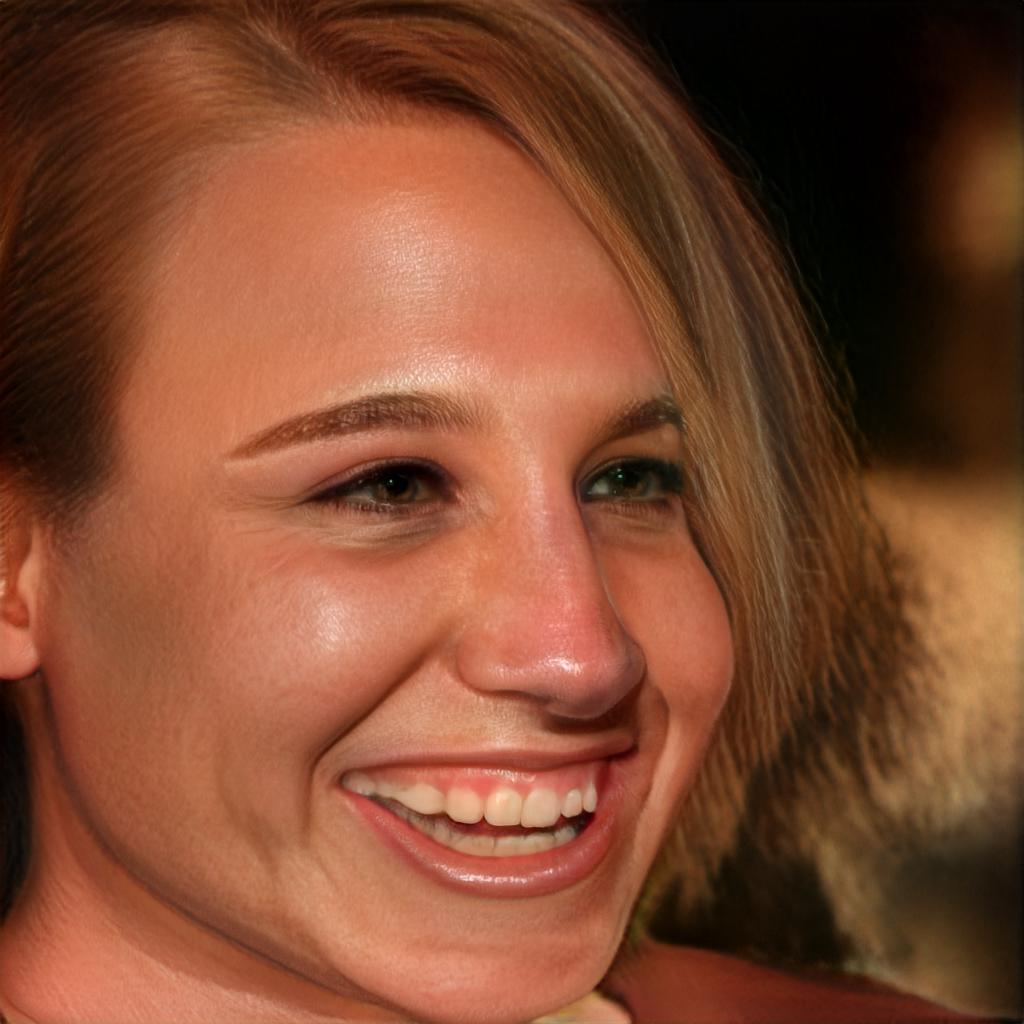} &
        \includegraphics[width=0.107\textwidth]{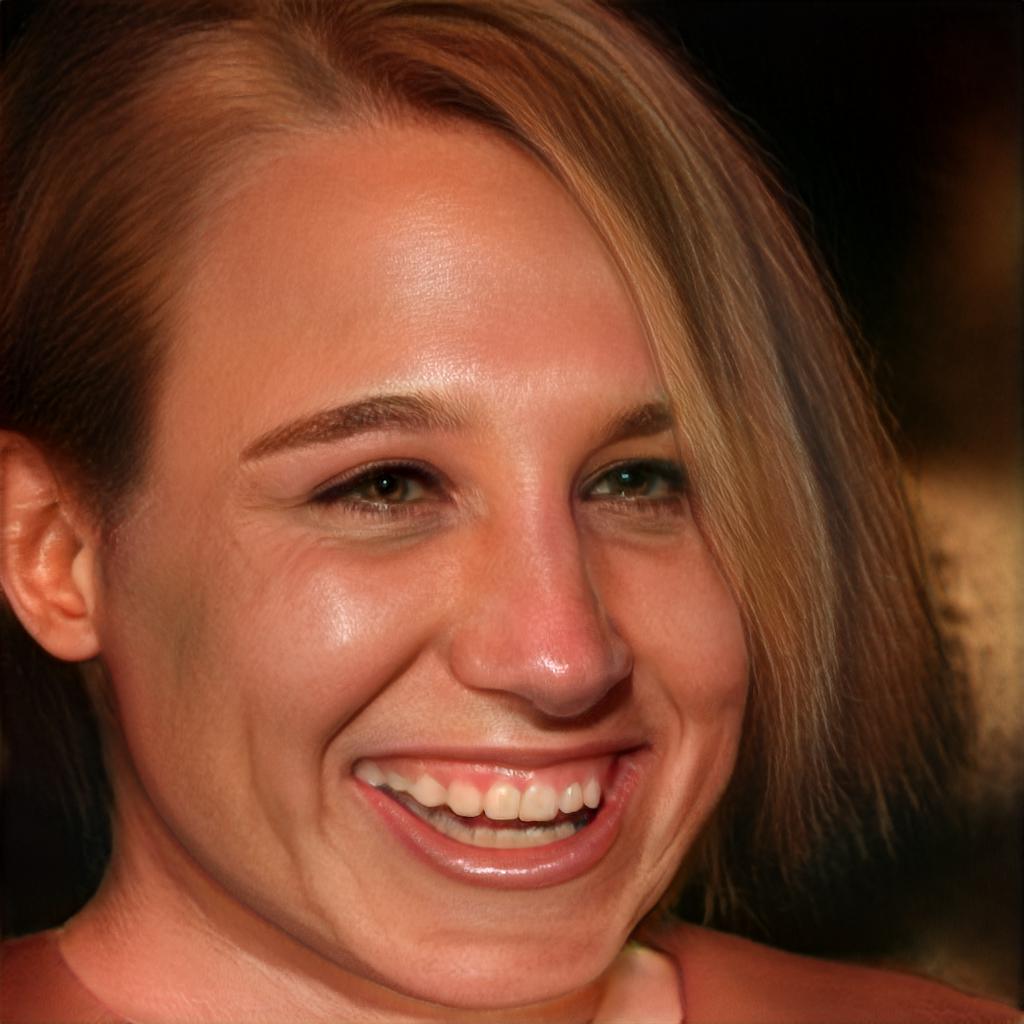} &
        \includegraphics[width=0.107\textwidth]{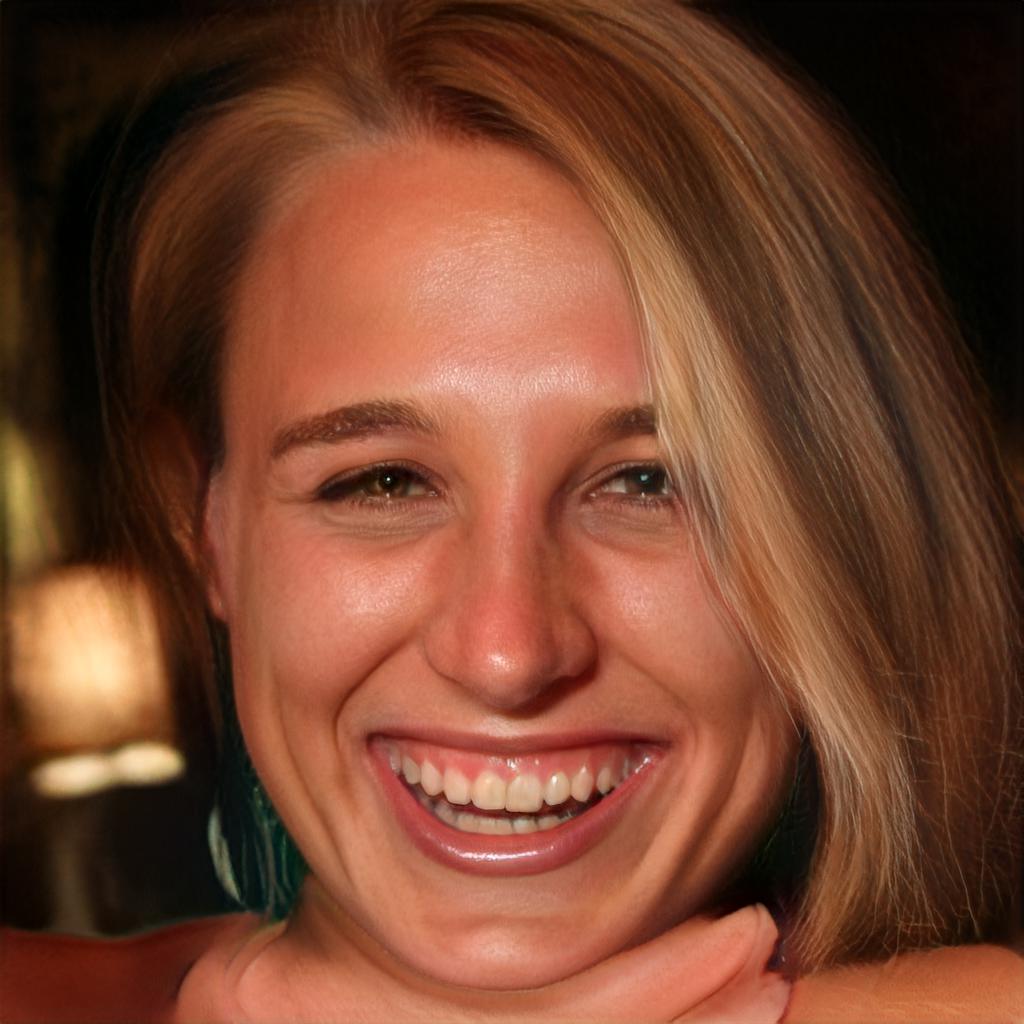} &
        \includegraphics[width=0.107\textwidth]{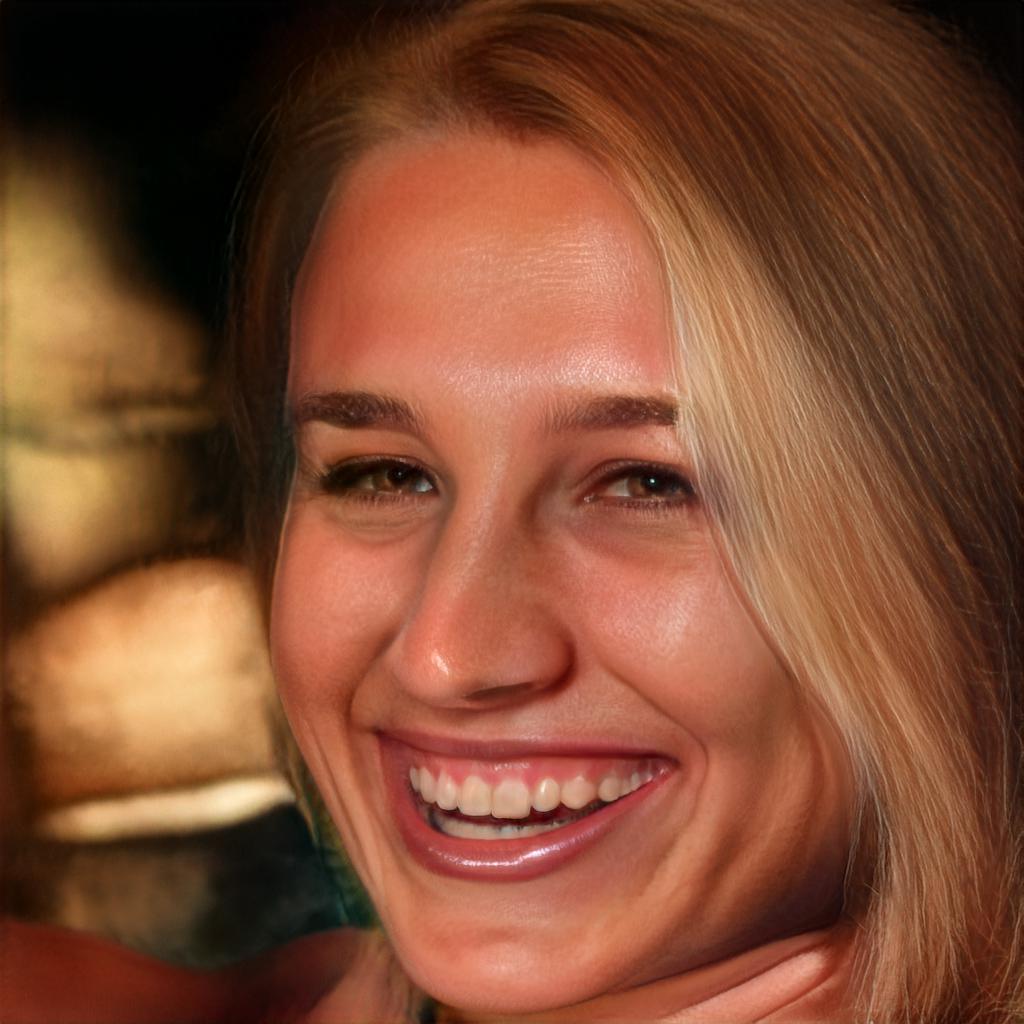} &
        \includegraphics[width=0.107\textwidth]{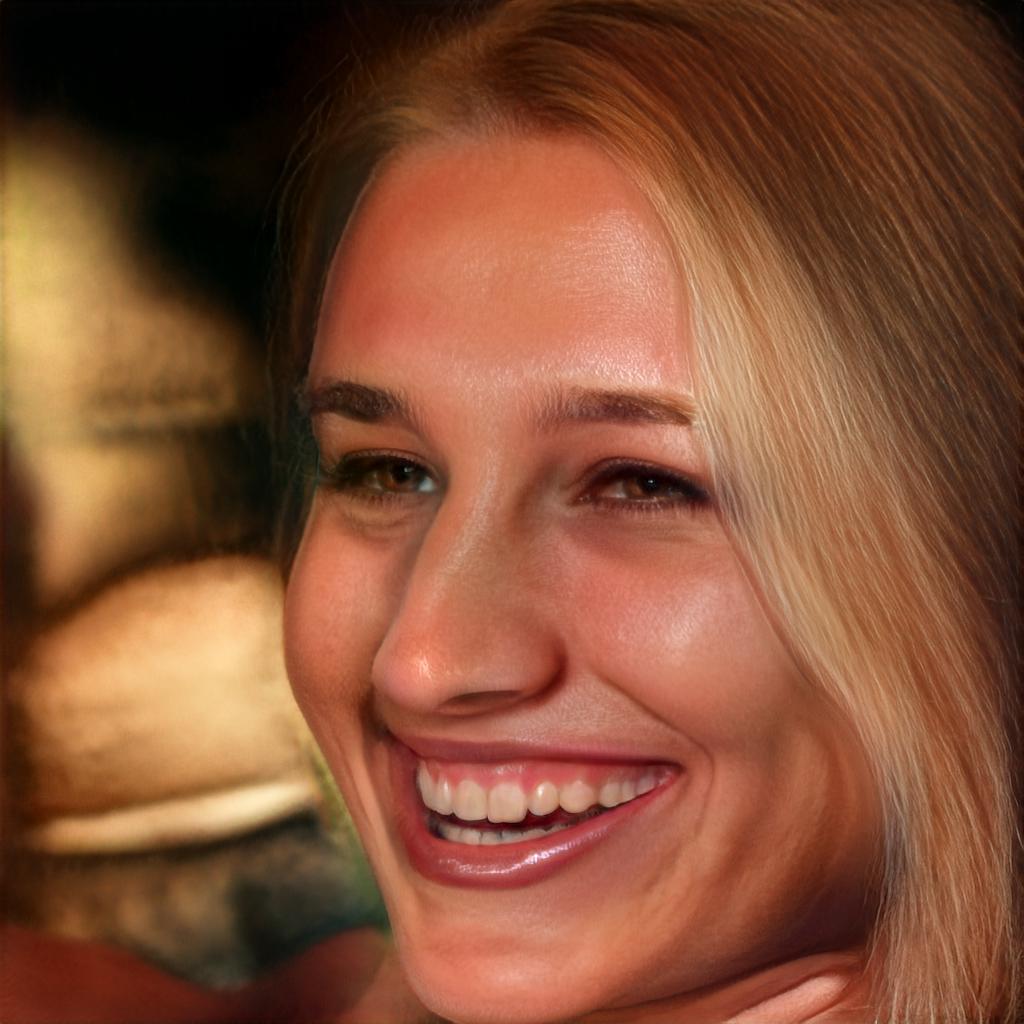} 

        \\

        \raisebox{0.265in}{\rotatebox{90}{\footnotesize SeFa}} &
        \includegraphics[width=0.107\textwidth]{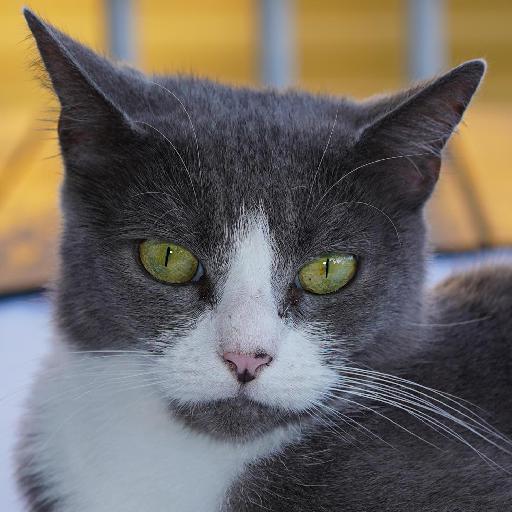} &
        \includegraphics[width=0.107\textwidth]{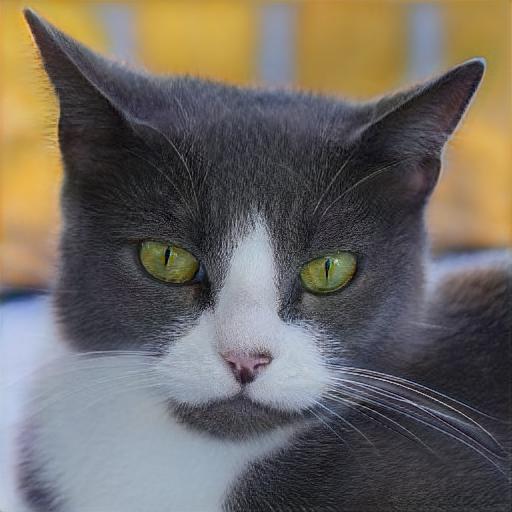} &
        \includegraphics[width=0.107\textwidth]{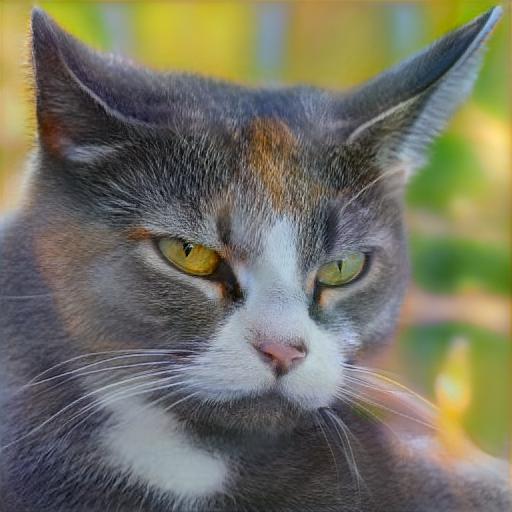} &
        \includegraphics[width=0.107\textwidth]{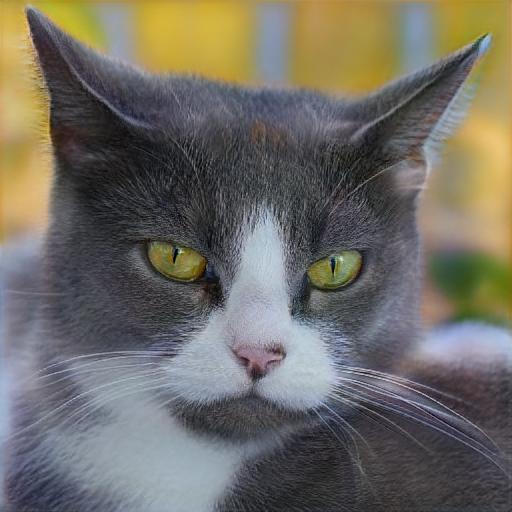} &
        \includegraphics[width=0.107\textwidth]{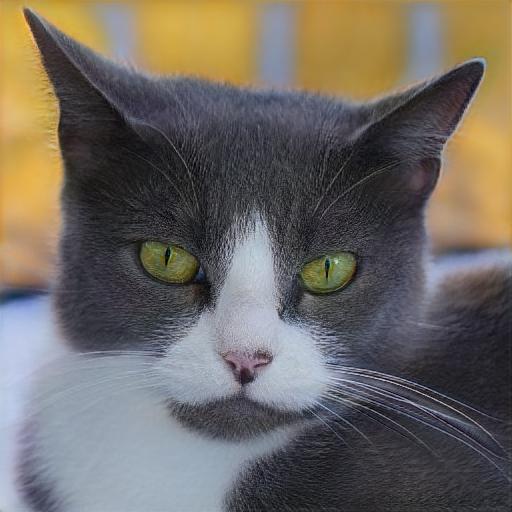} &
        \includegraphics[width=0.107\textwidth]{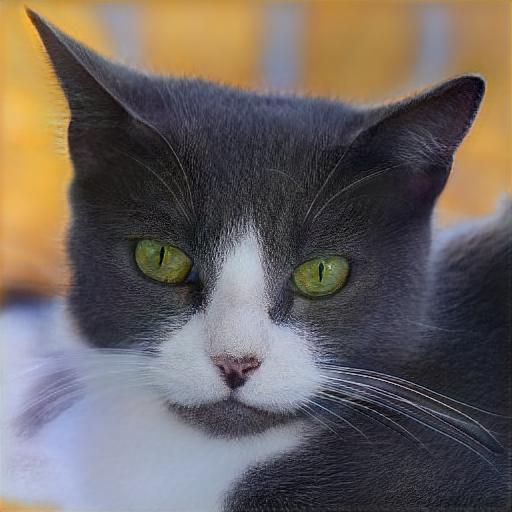} &
        \includegraphics[width=0.107\textwidth]{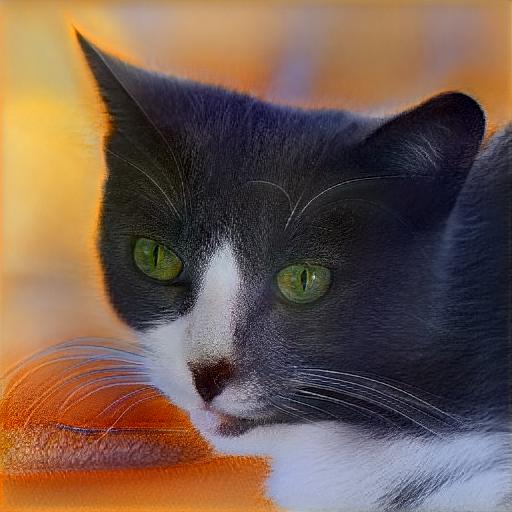} \\
        
        \raisebox{0.265in}{\rotatebox{90}{Ours}} &
        \includegraphics[width=0.107\textwidth]{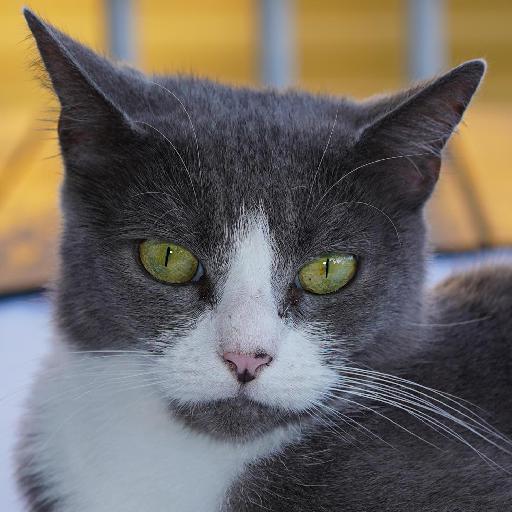} &
        \includegraphics[width=0.107\textwidth]{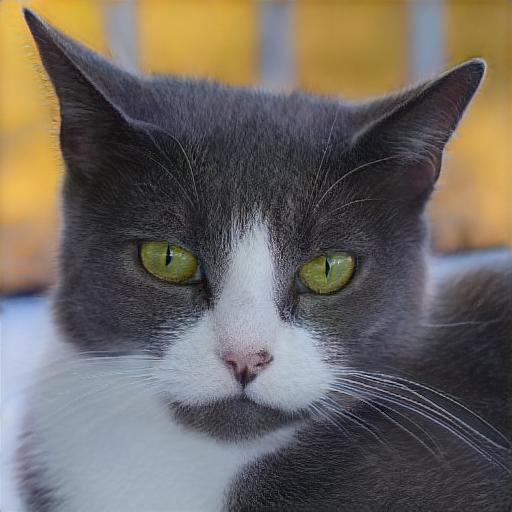} &
        \includegraphics[width=0.107\textwidth]{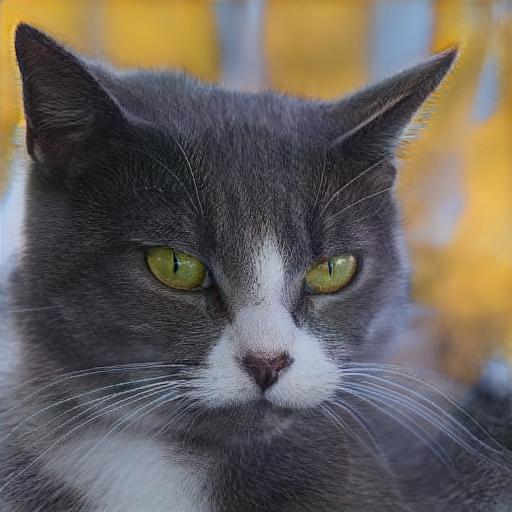} &
        \includegraphics[width=0.107\textwidth]{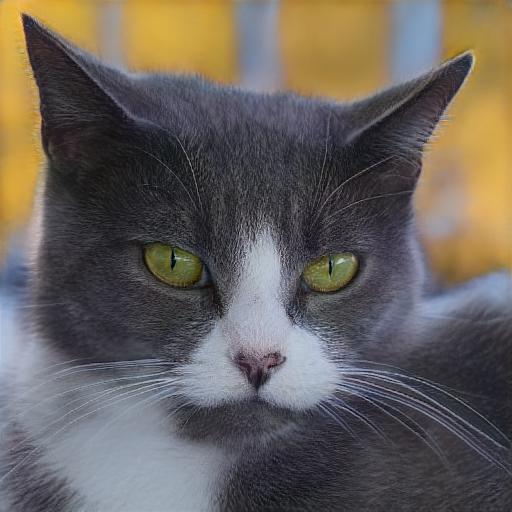} &
        \includegraphics[width=0.107\textwidth]{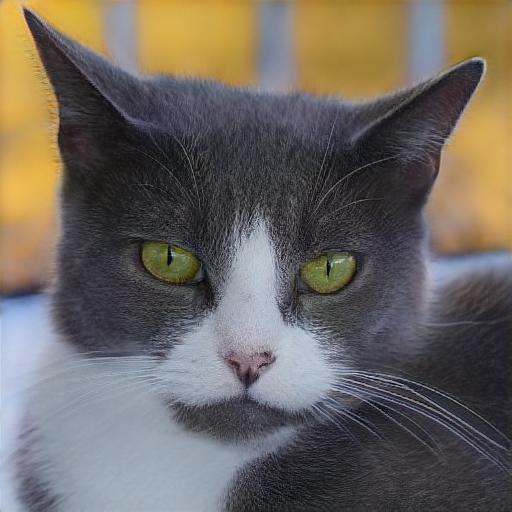} &
        \includegraphics[width=0.107\textwidth]{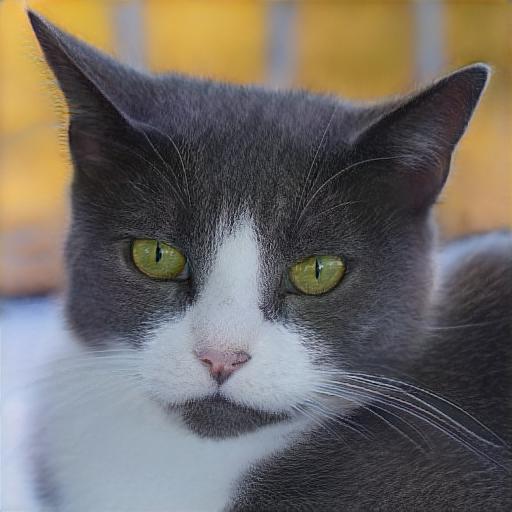} &
        \includegraphics[width=0.107\textwidth]{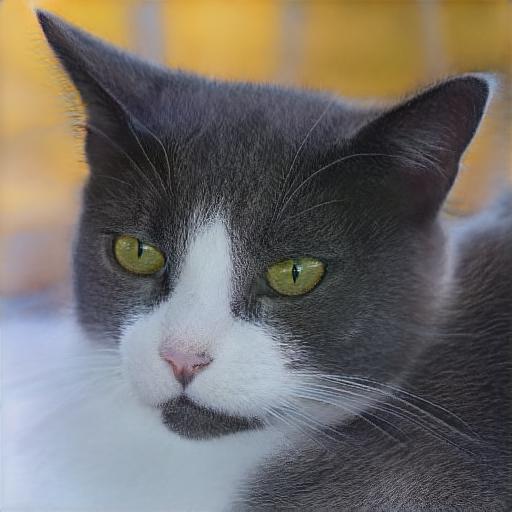}

        \\ & Input & Inversion & \multicolumn{5}{c}{$\myleftarrow$~Pose~$\myarrow$ } \\

        \raisebox{0.15in}{\rotatebox{90}{\footnotesize StyleCLIP}} &
        \includegraphics[width=0.107\textwidth]{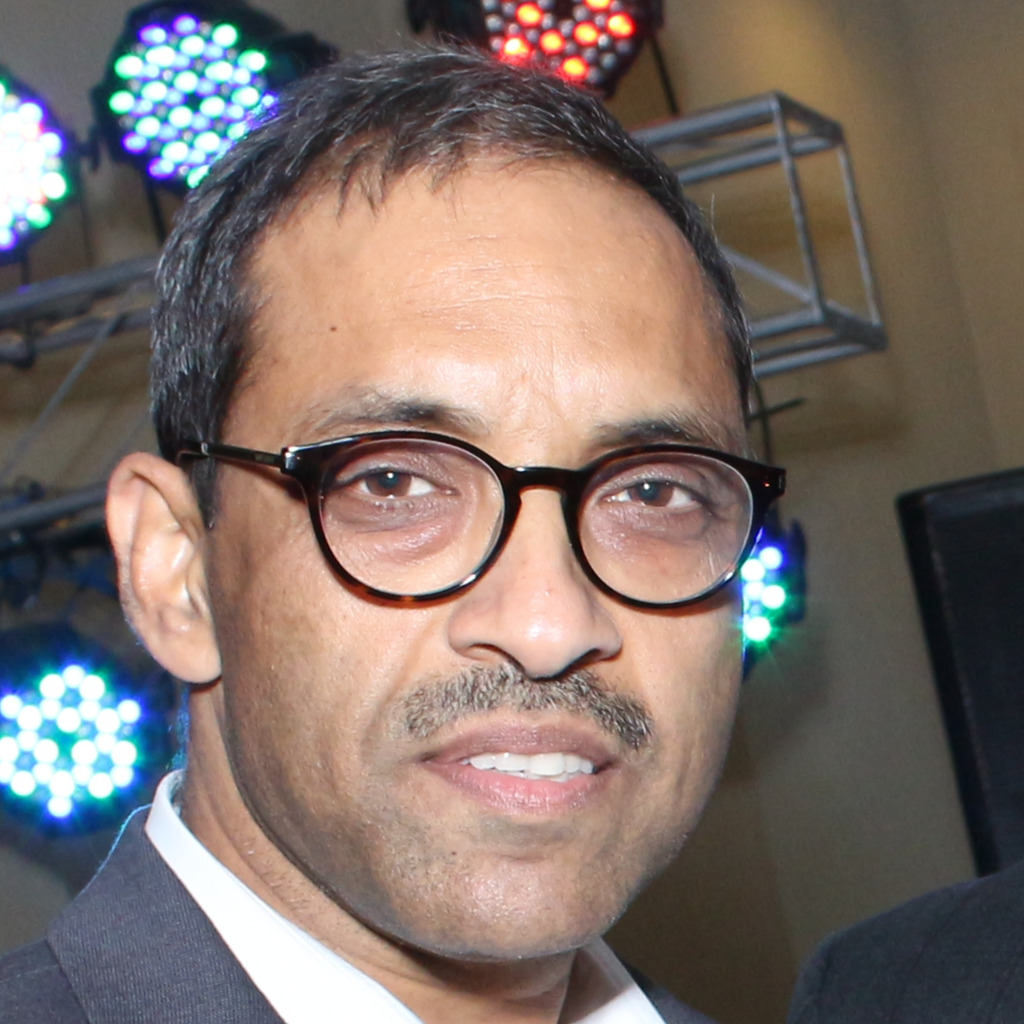} &
        \includegraphics[width=0.107\textwidth]{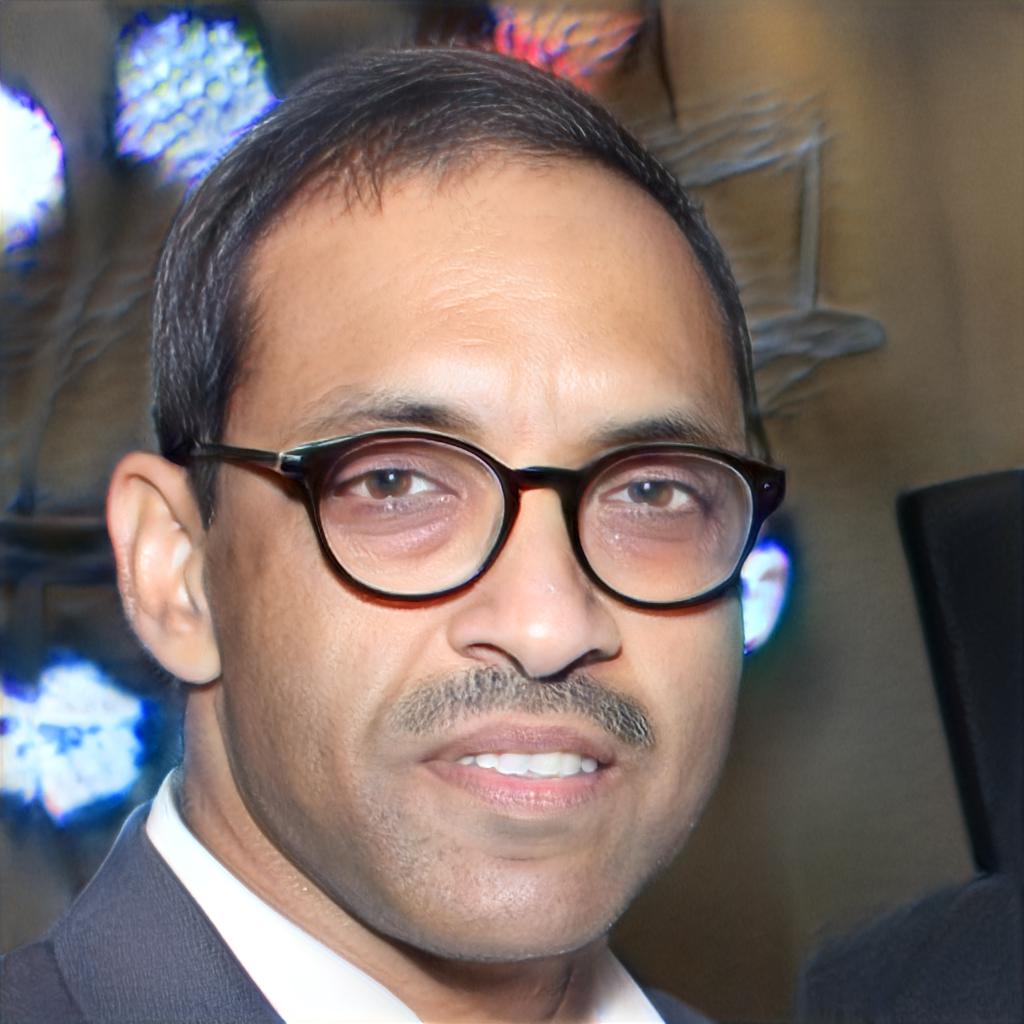} &
        \includegraphics[width=0.107\textwidth]{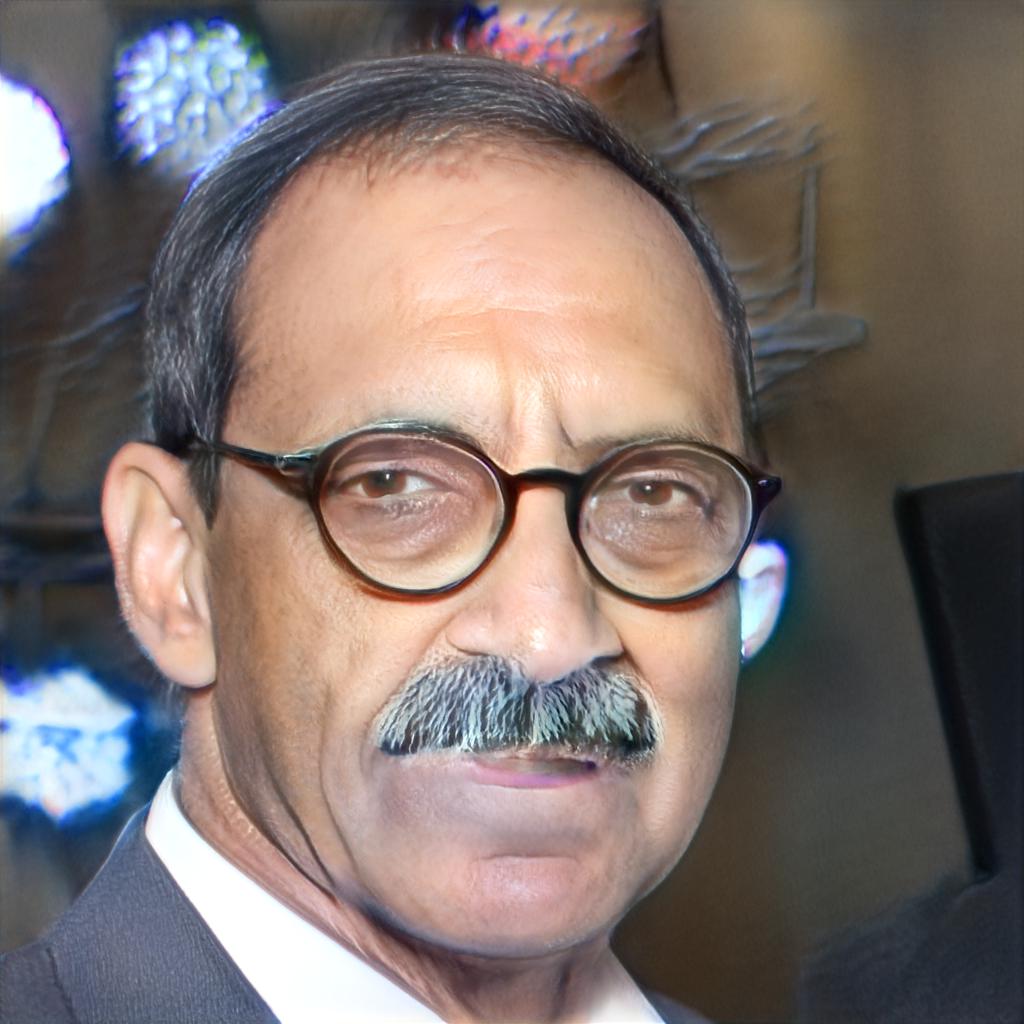} &
        \includegraphics[width=0.107\textwidth]{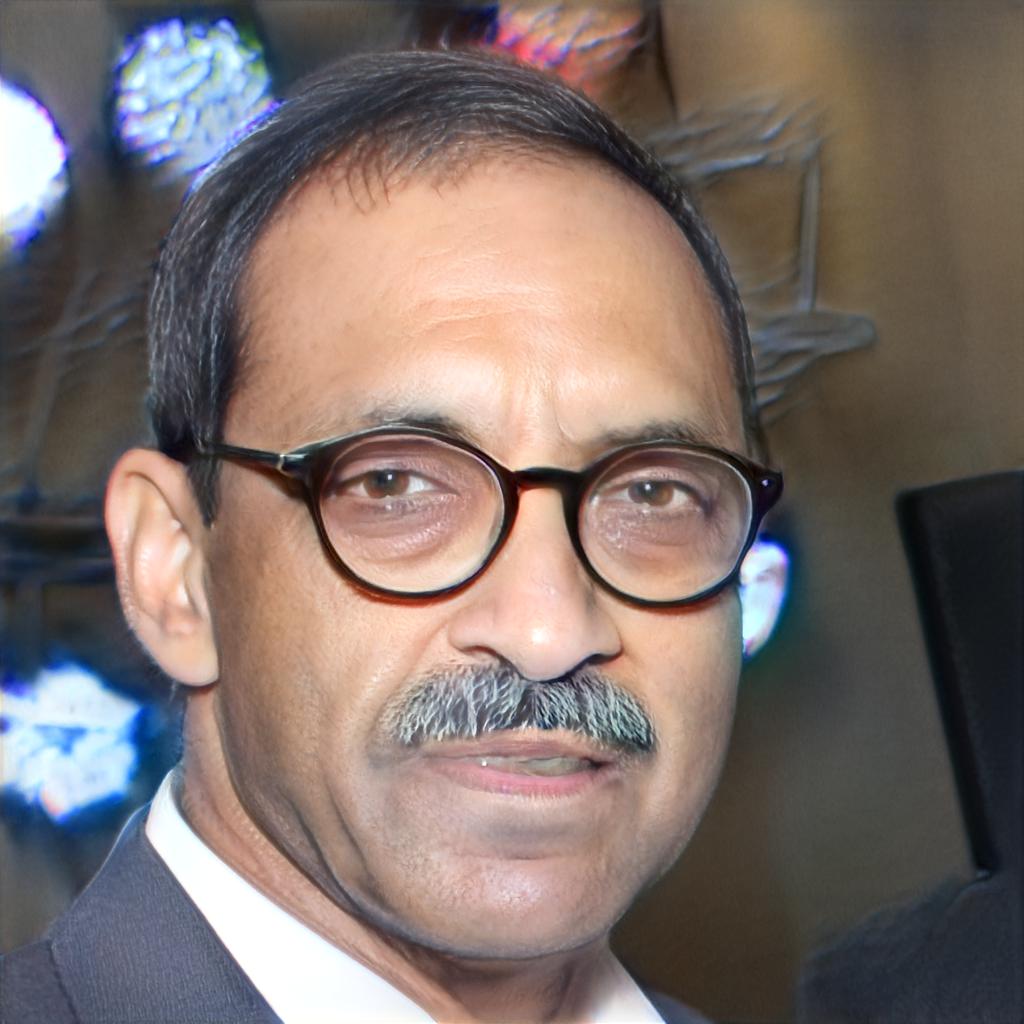} &
        \includegraphics[width=0.107\textwidth]{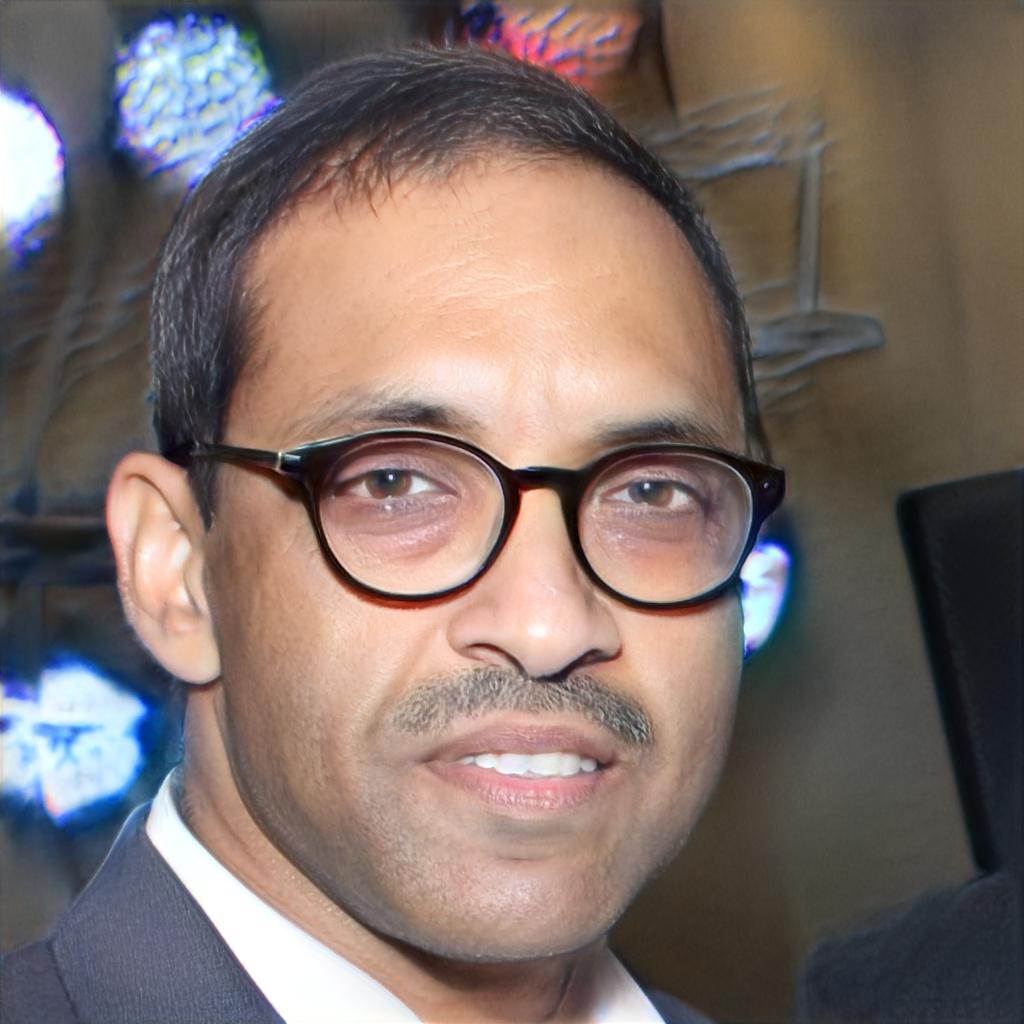} &
        \includegraphics[width=0.107\textwidth]{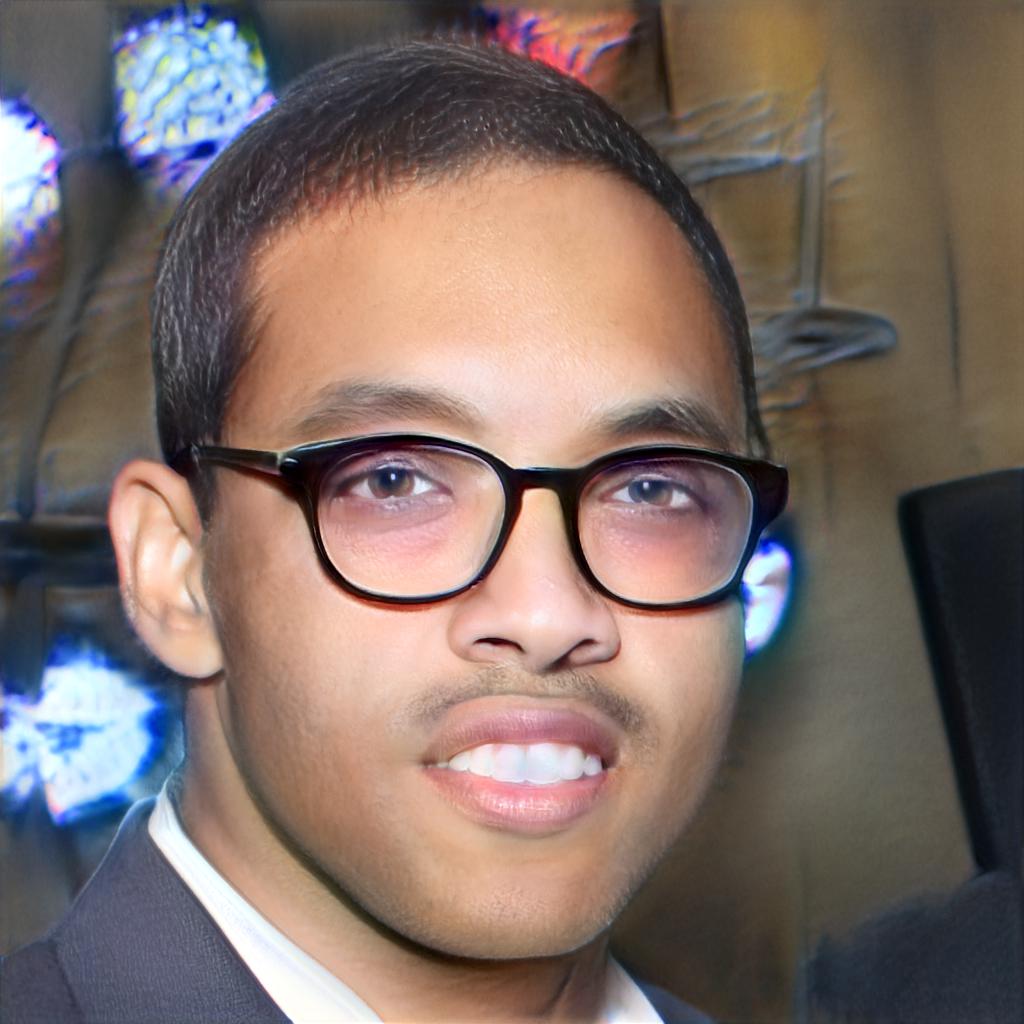} &
        \includegraphics[width=0.107\textwidth]{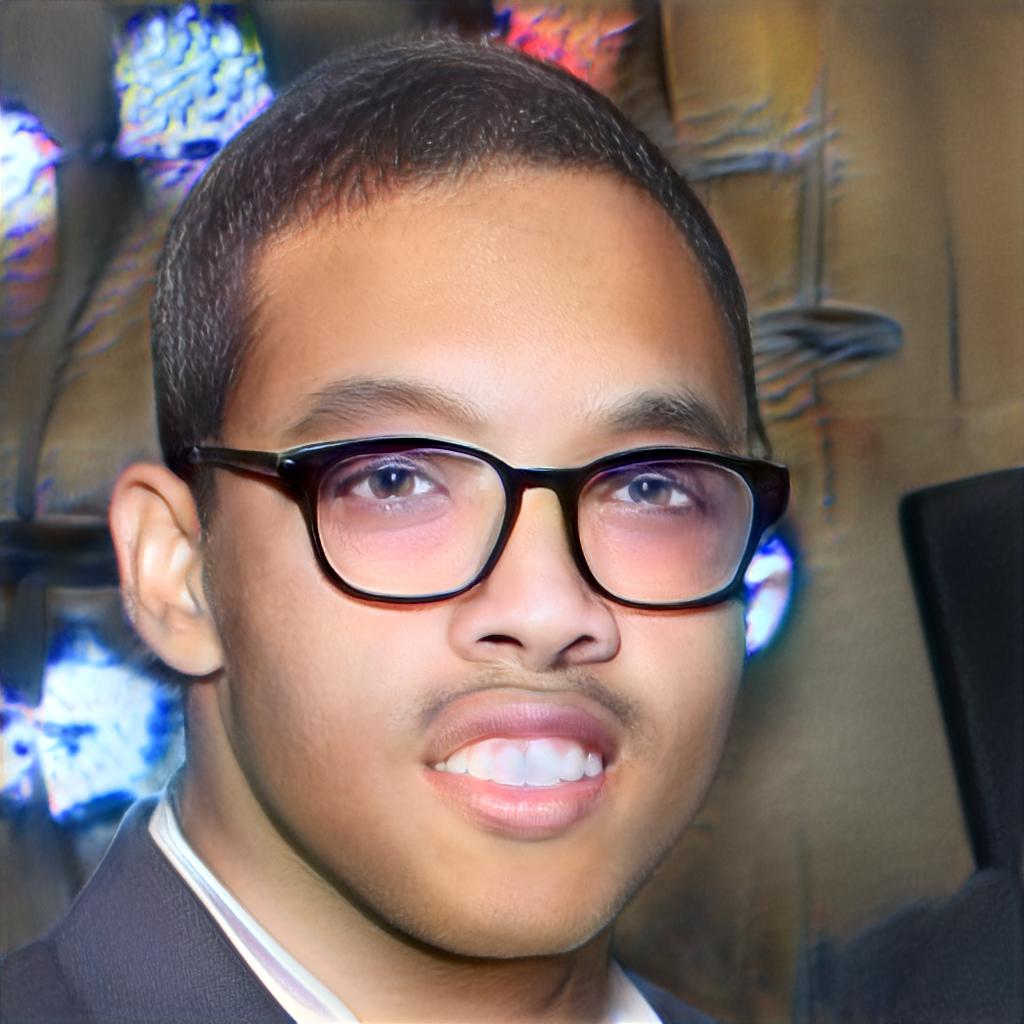} \\
        
        \raisebox{0.265in}{\rotatebox{90}{Ours}} &
        \includegraphics[width=0.107\textwidth]{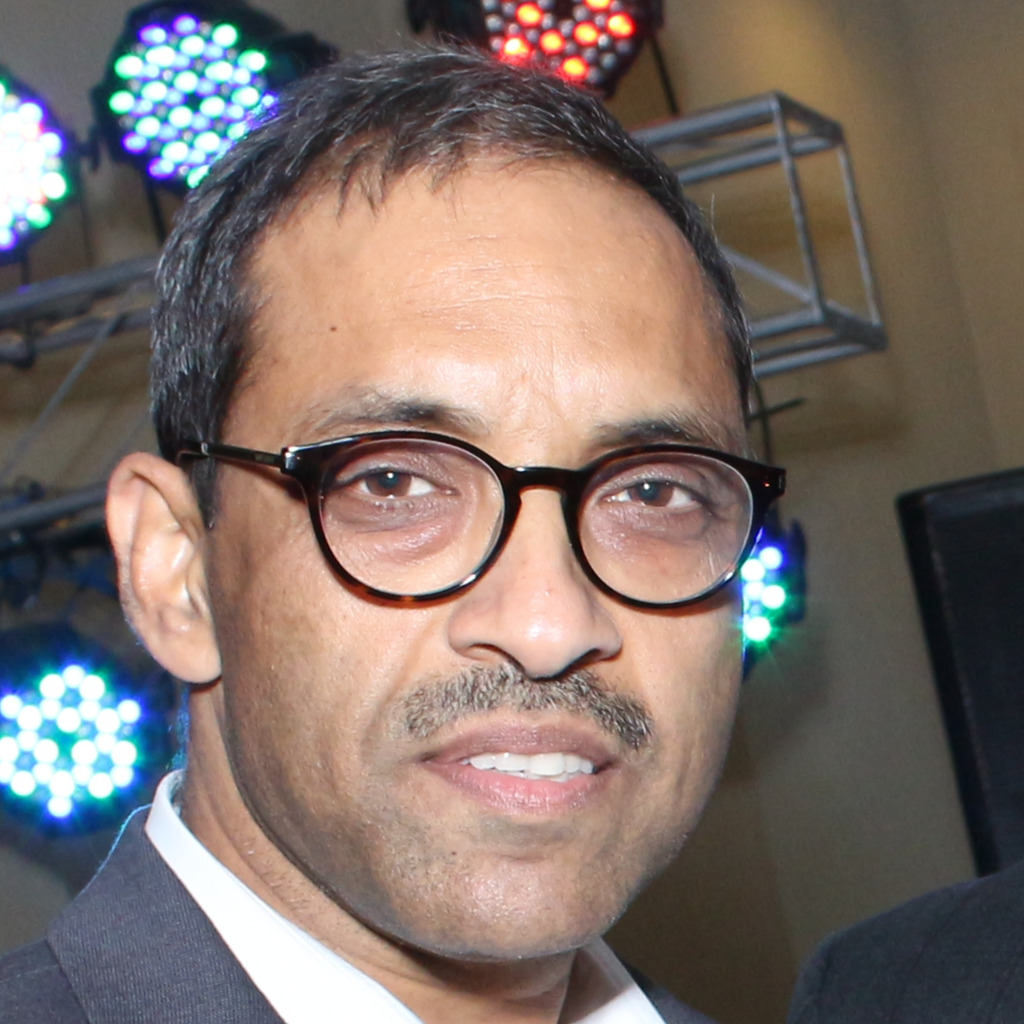} &
        \includegraphics[width=0.107\textwidth]{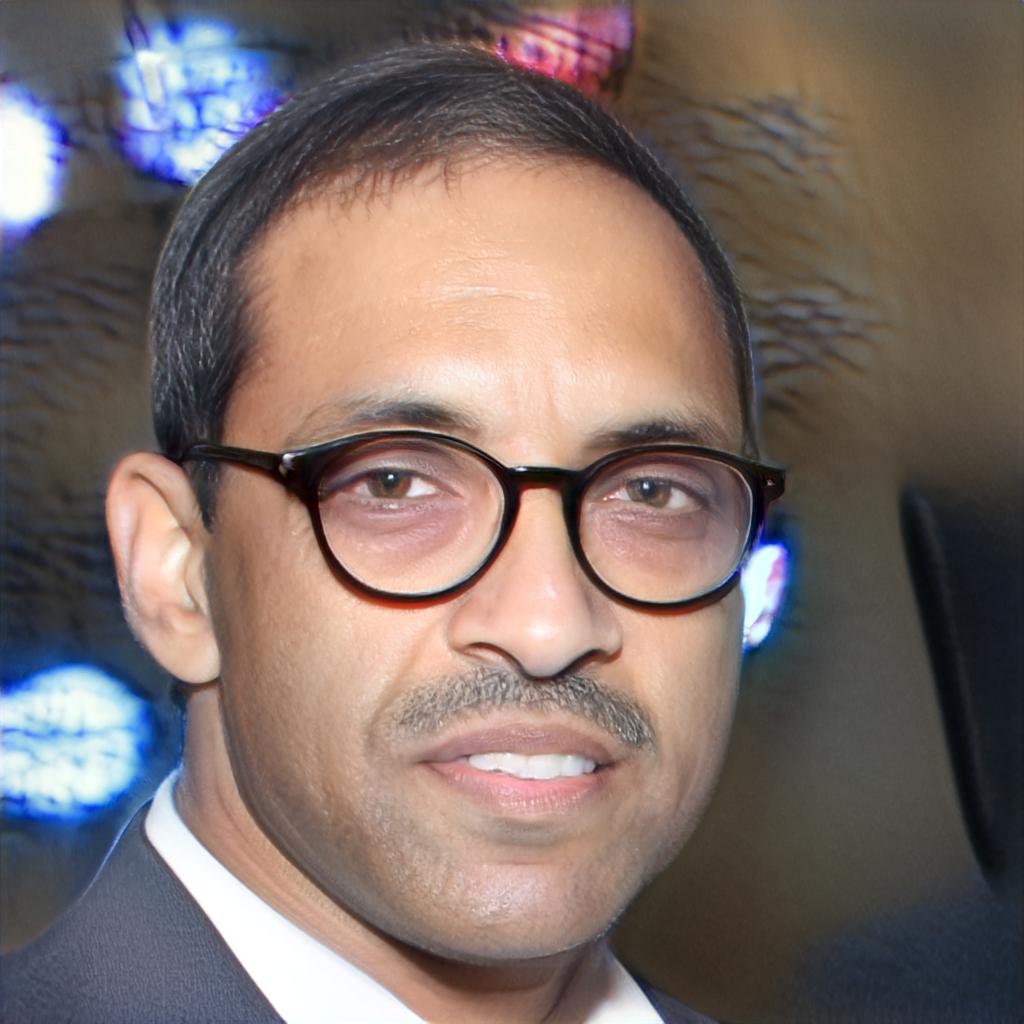} &
        \includegraphics[width=0.107\textwidth]{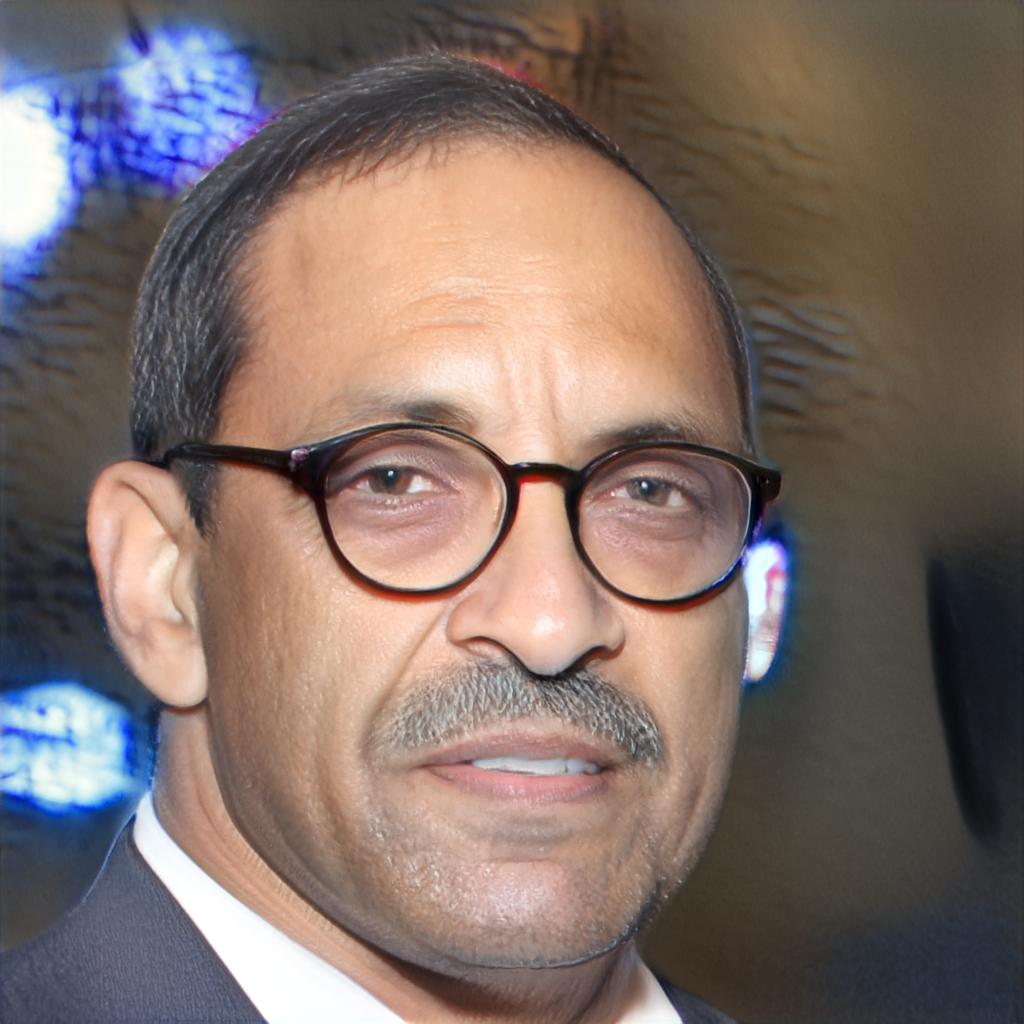} &
        \includegraphics[width=0.107\textwidth]{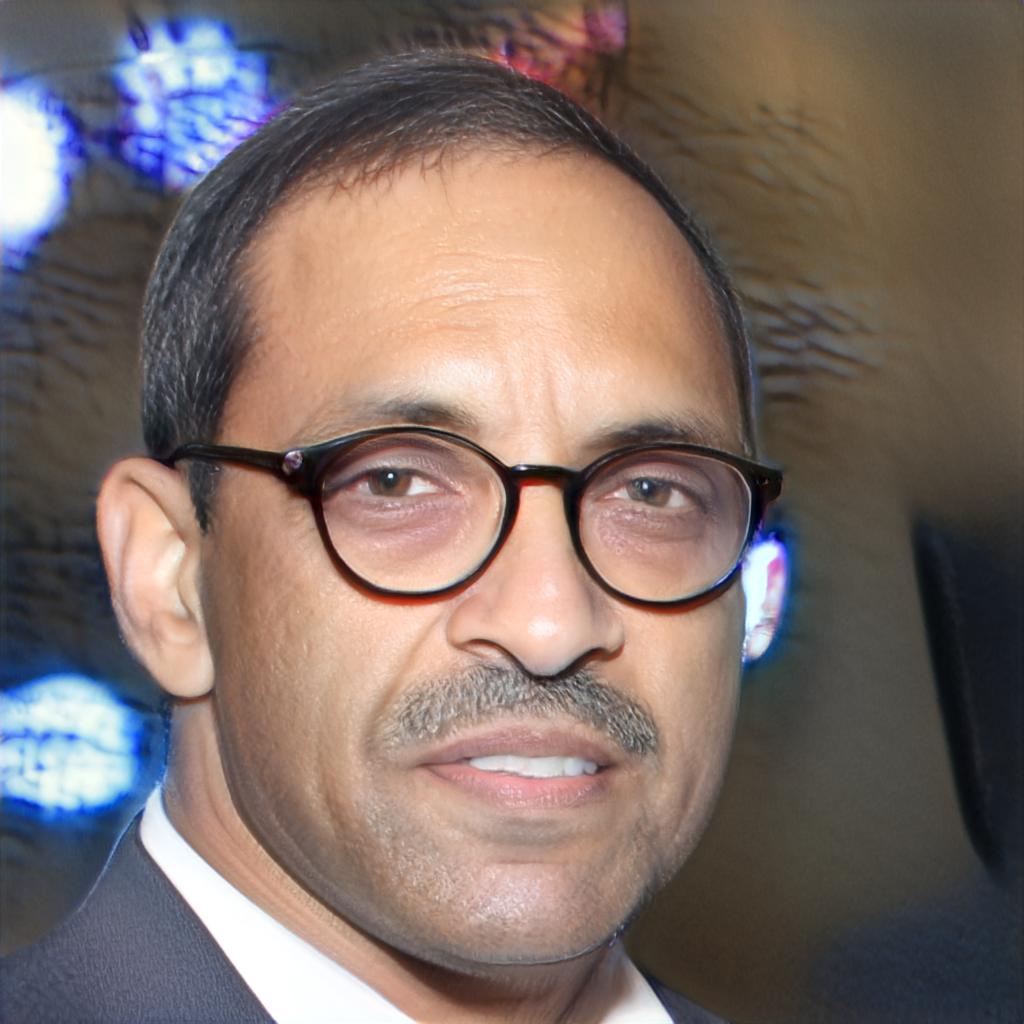} &
        \includegraphics[width=0.107\textwidth]{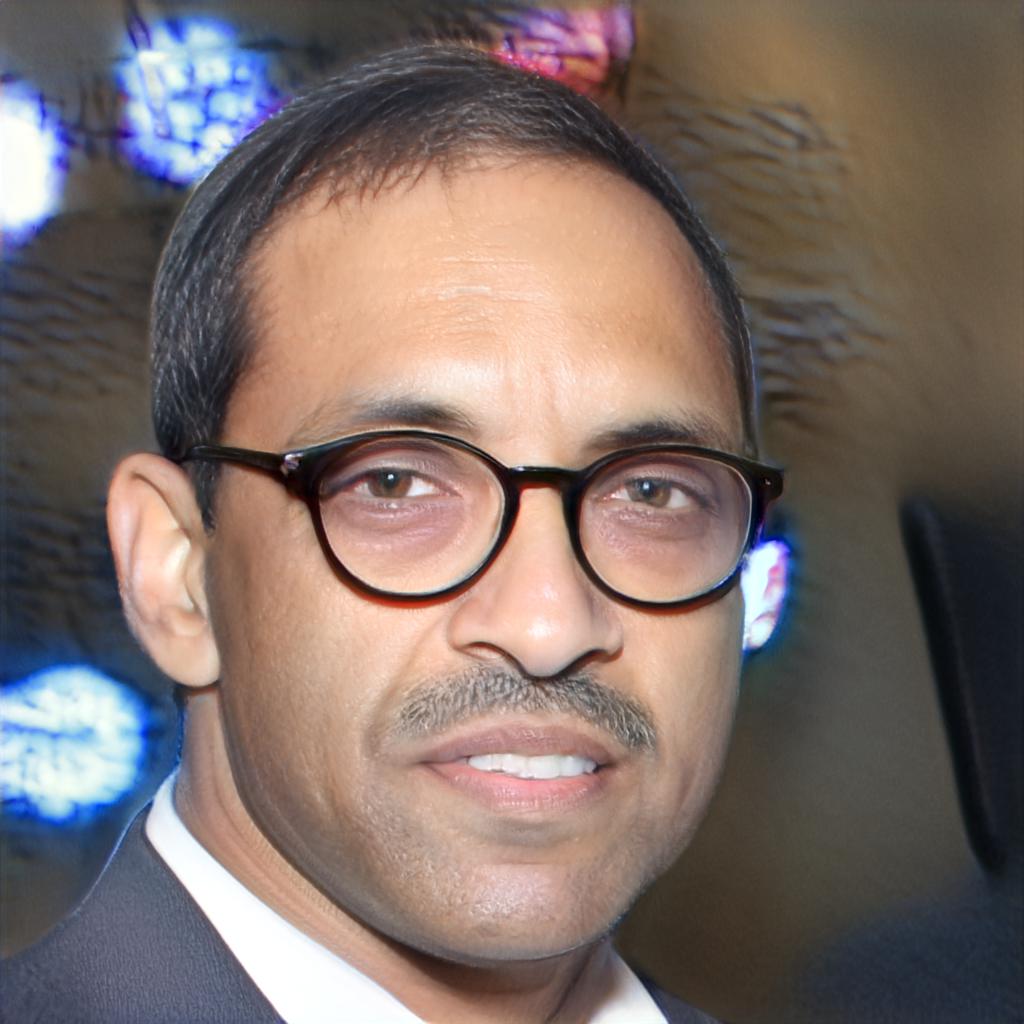} &
        \includegraphics[width=0.107\textwidth]{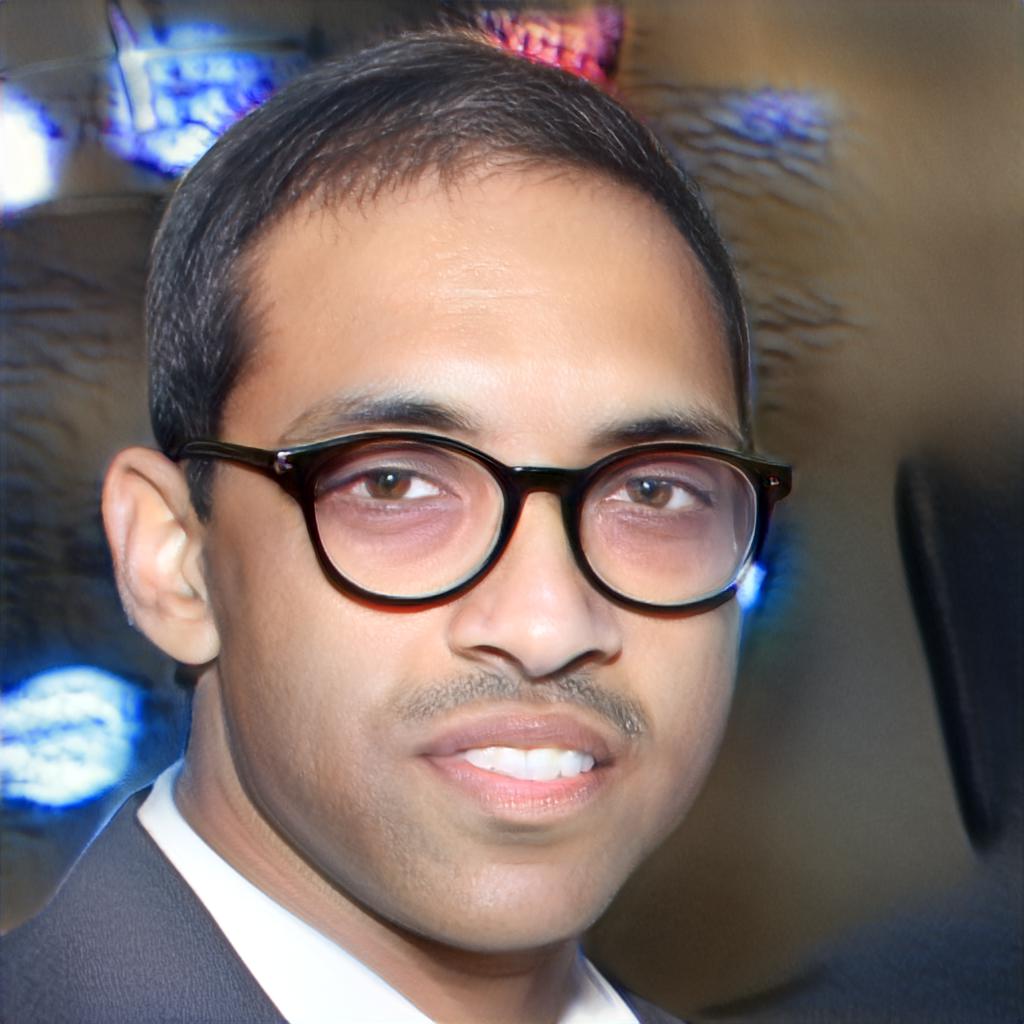} &
        \includegraphics[width=0.107\textwidth]{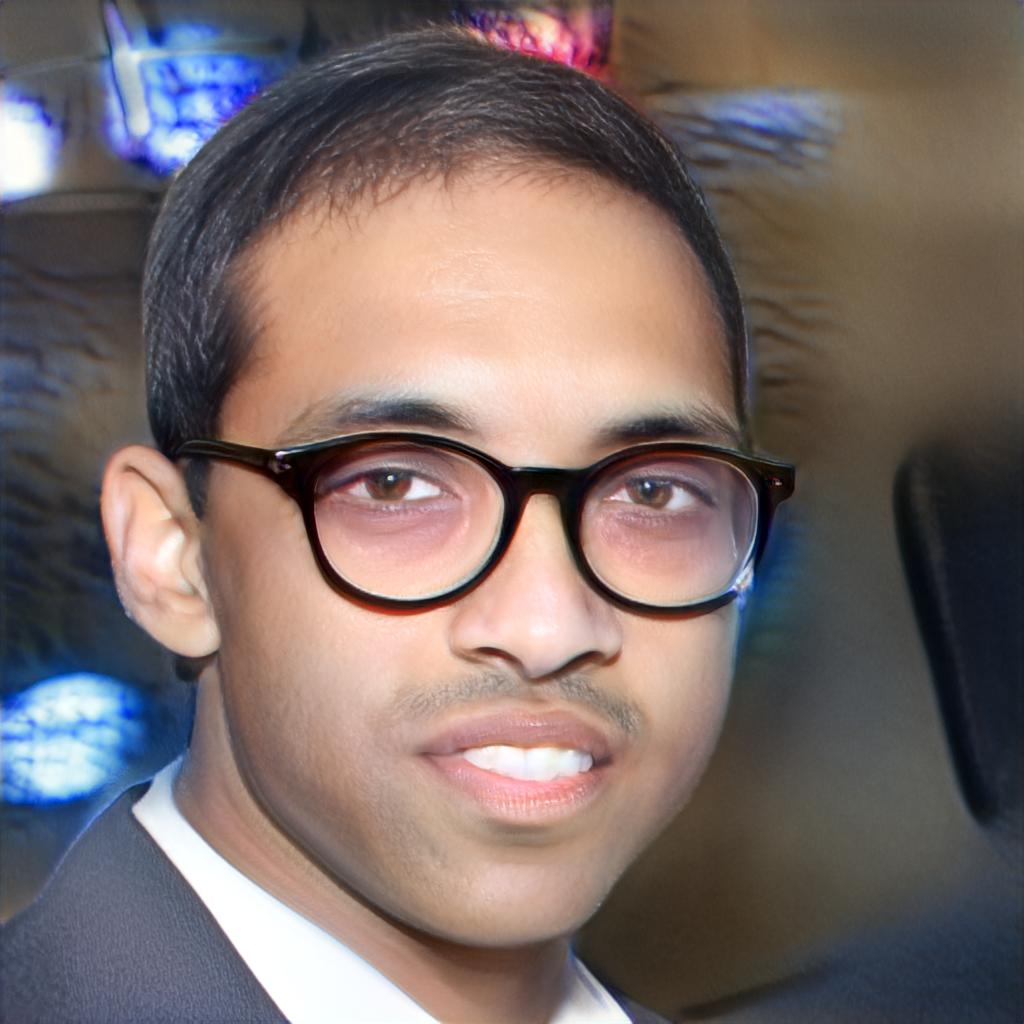} \\
        
        \raisebox{0.265in}{\rotatebox{90}{\footnotesize SeFa}} &
        \includegraphics[width=0.107\textwidth]{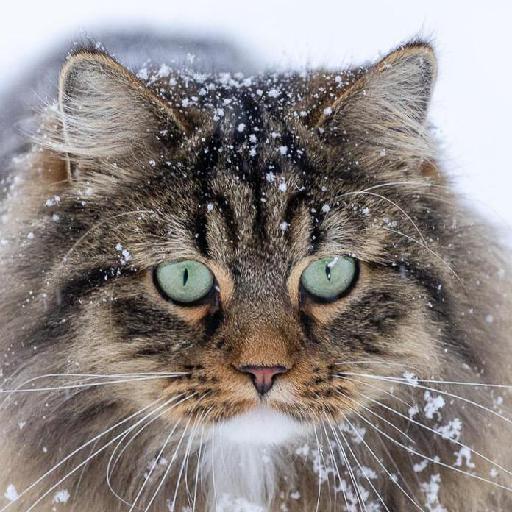} &
        \includegraphics[width=0.107\textwidth]{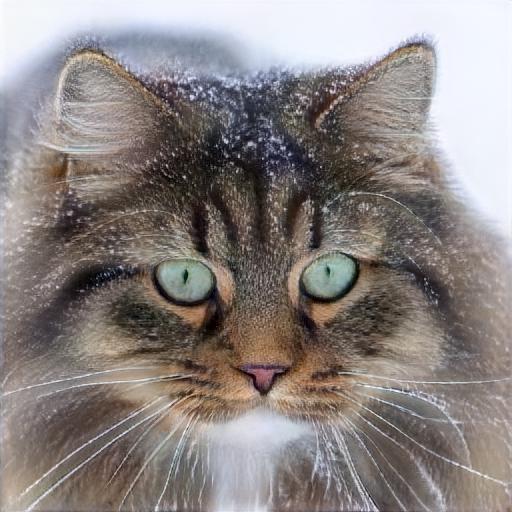} &
        \includegraphics[width=0.107\textwidth]{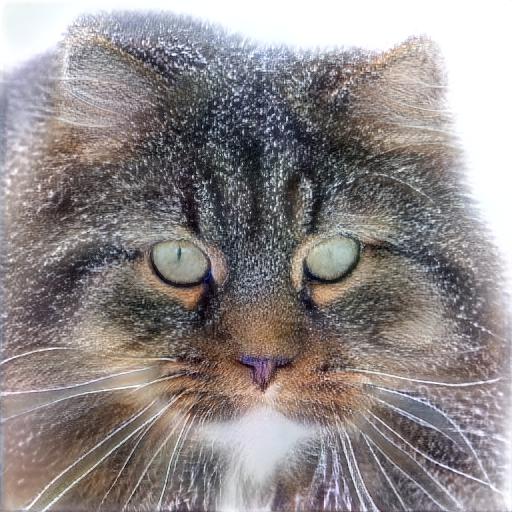} &
        \includegraphics[width=0.107\textwidth]{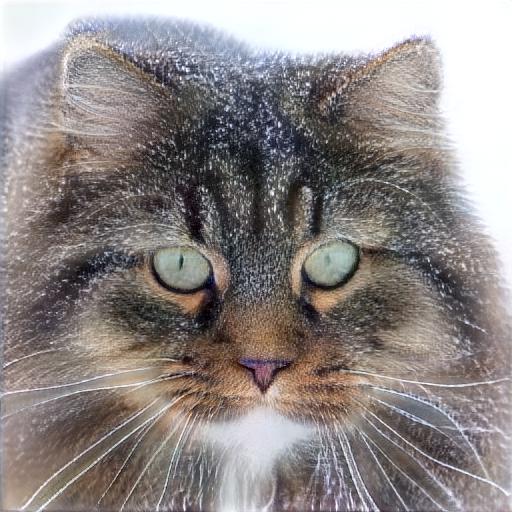} &
        \includegraphics[width=0.107\textwidth]{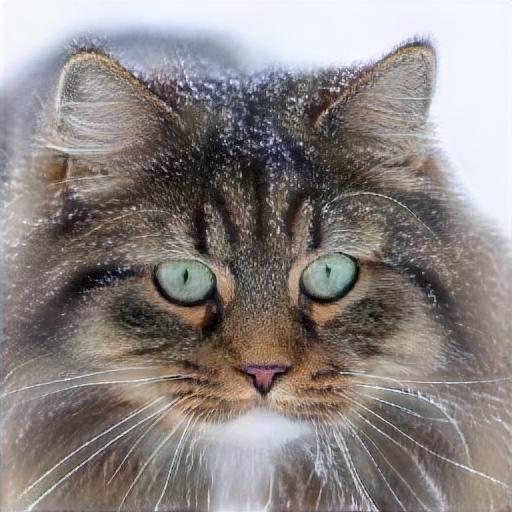} &
        \includegraphics[width=0.107\textwidth]{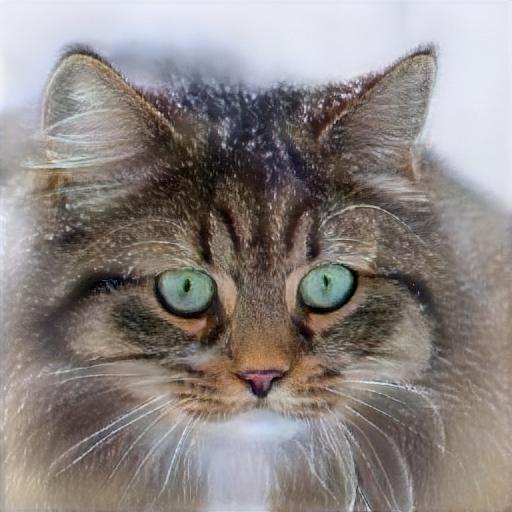} &
        \includegraphics[width=0.107\textwidth]{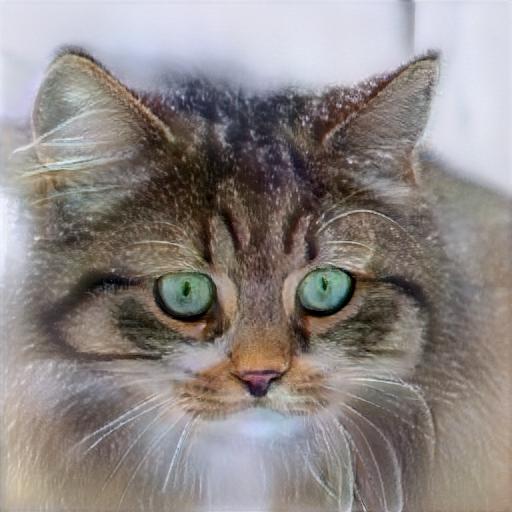} \\
        
        \raisebox{0.265in}{\rotatebox{90}{Ours}} &
        \includegraphics[width=0.107\textwidth]{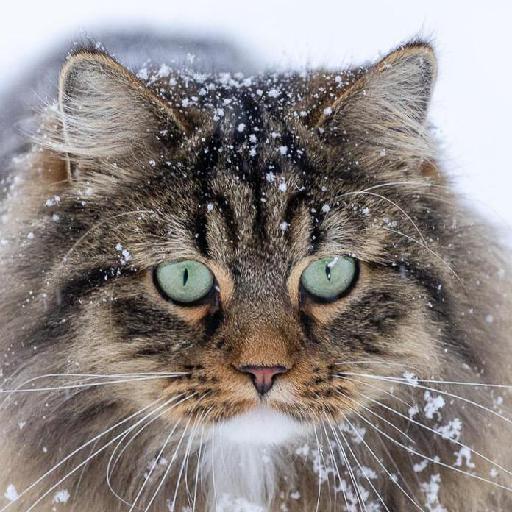} &
        \includegraphics[width=0.107\textwidth]{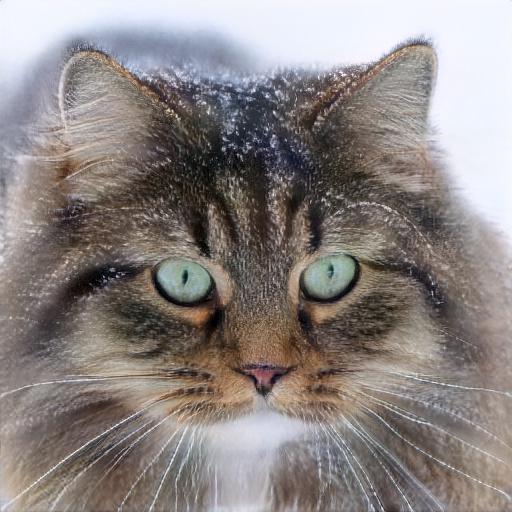} &
        \includegraphics[width=0.107\textwidth]{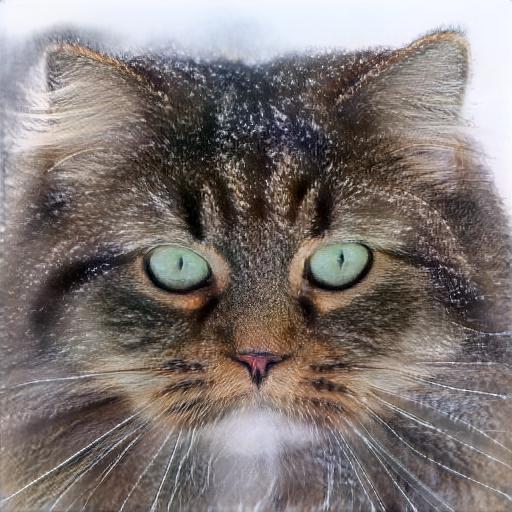} &
        \includegraphics[width=0.107\textwidth]{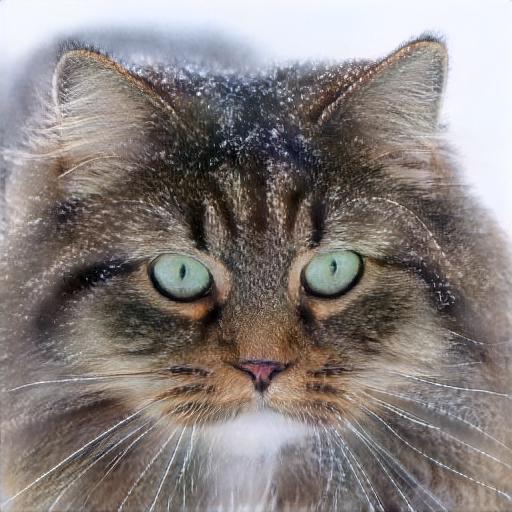} &
        \includegraphics[width=0.107\textwidth]{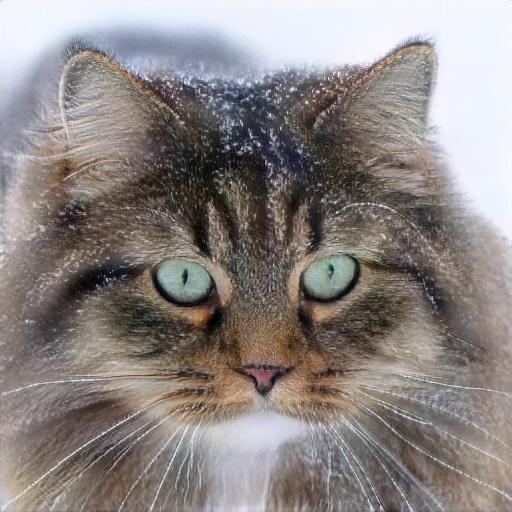} &
        \includegraphics[width=0.107\textwidth]{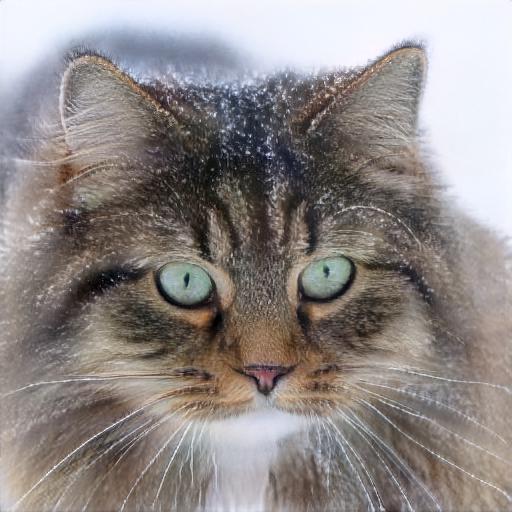} &
        \includegraphics[width=0.107\textwidth]{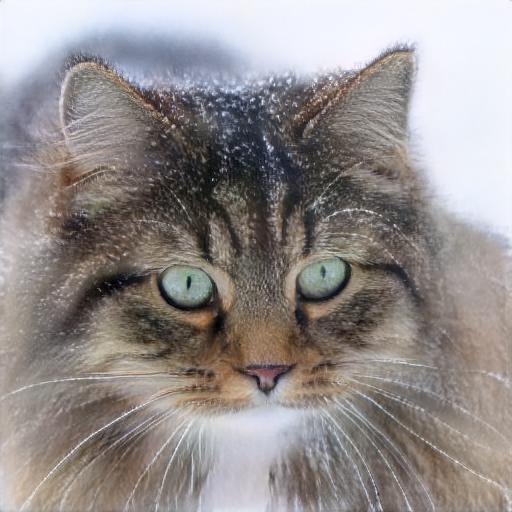} 

        \\ & Input & Inversion & \multicolumn{5}{c}{$\myleftarrow$~Age~$\myarrow$ } \\
        
        \raisebox{0.15in}{\rotatebox{90}{\footnotesize StyleCLIP}} &
        \includegraphics[width=0.107\textwidth]{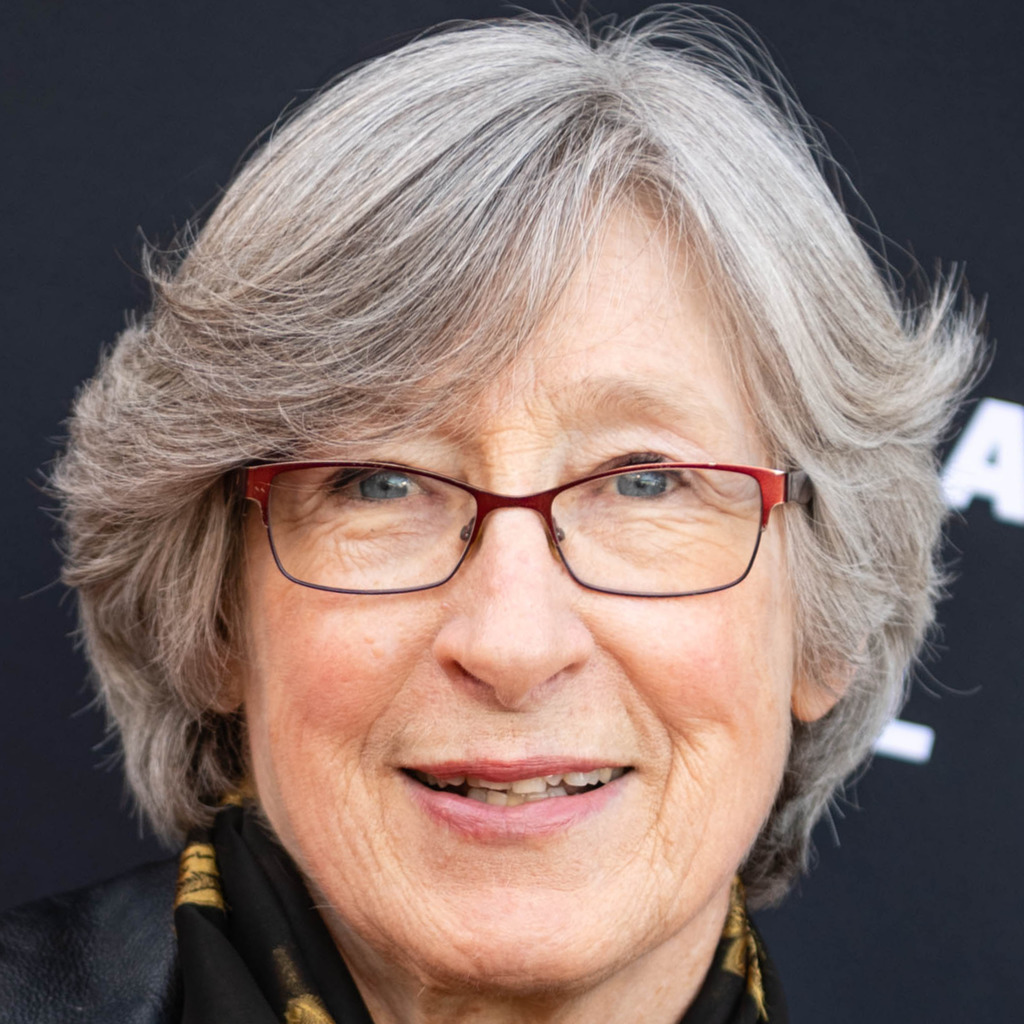} &
        \includegraphics[width=0.107\textwidth]{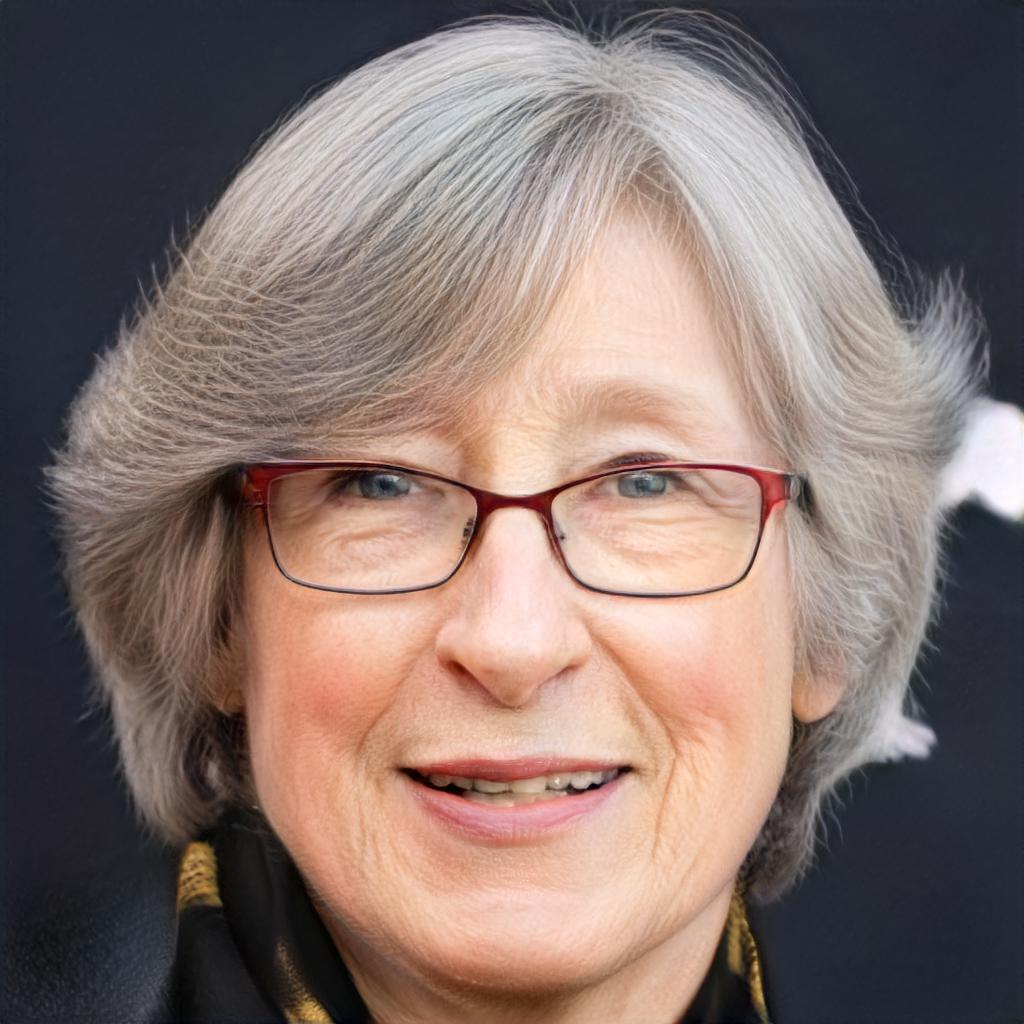} &
        \includegraphics[width=0.107\textwidth]{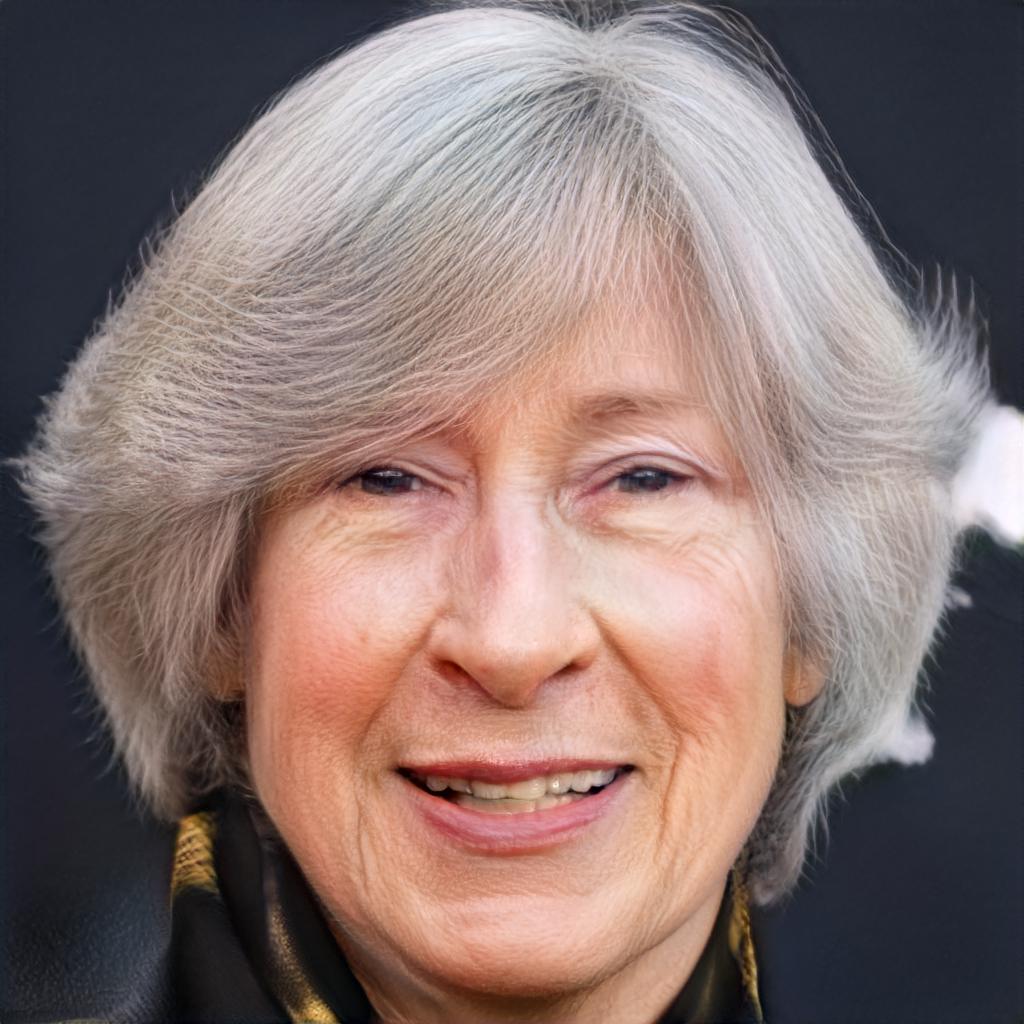} &
        \includegraphics[width=0.107\textwidth]{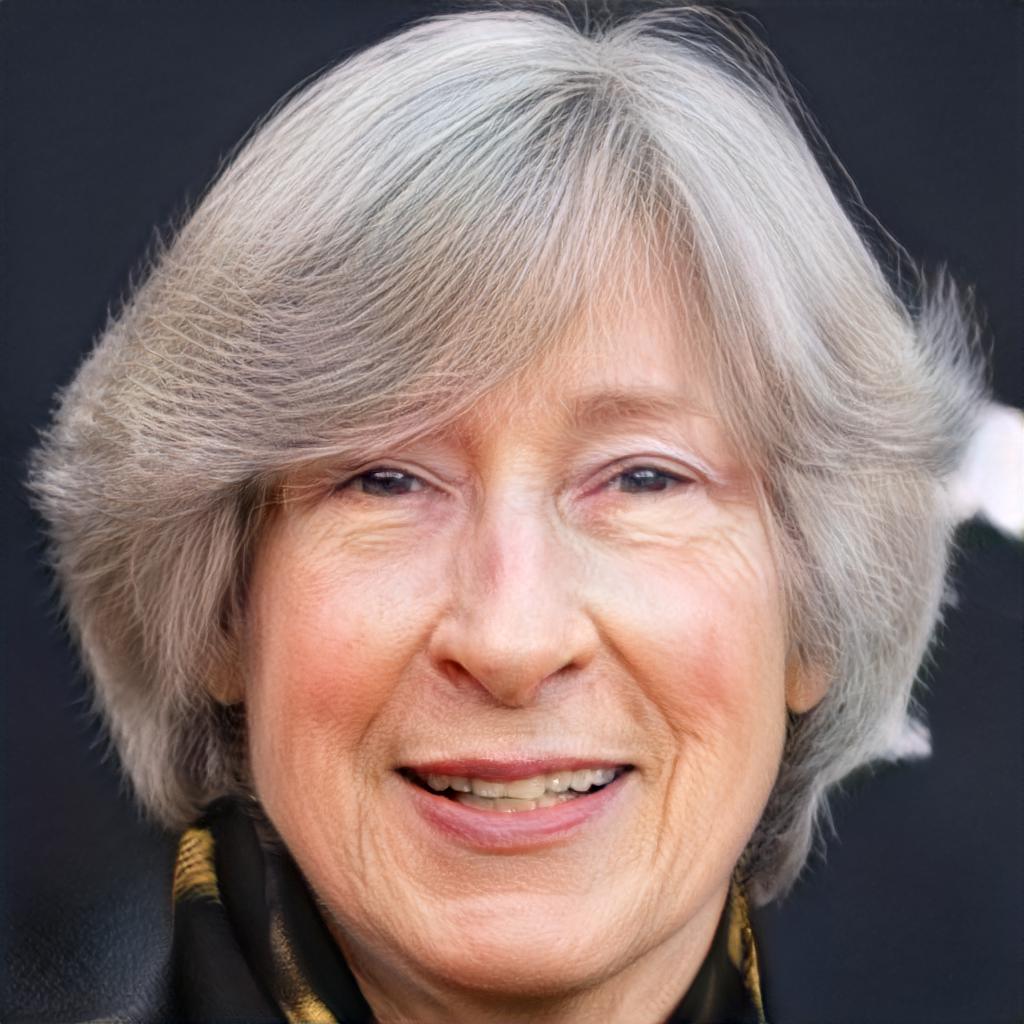} &
        \includegraphics[width=0.107\textwidth]{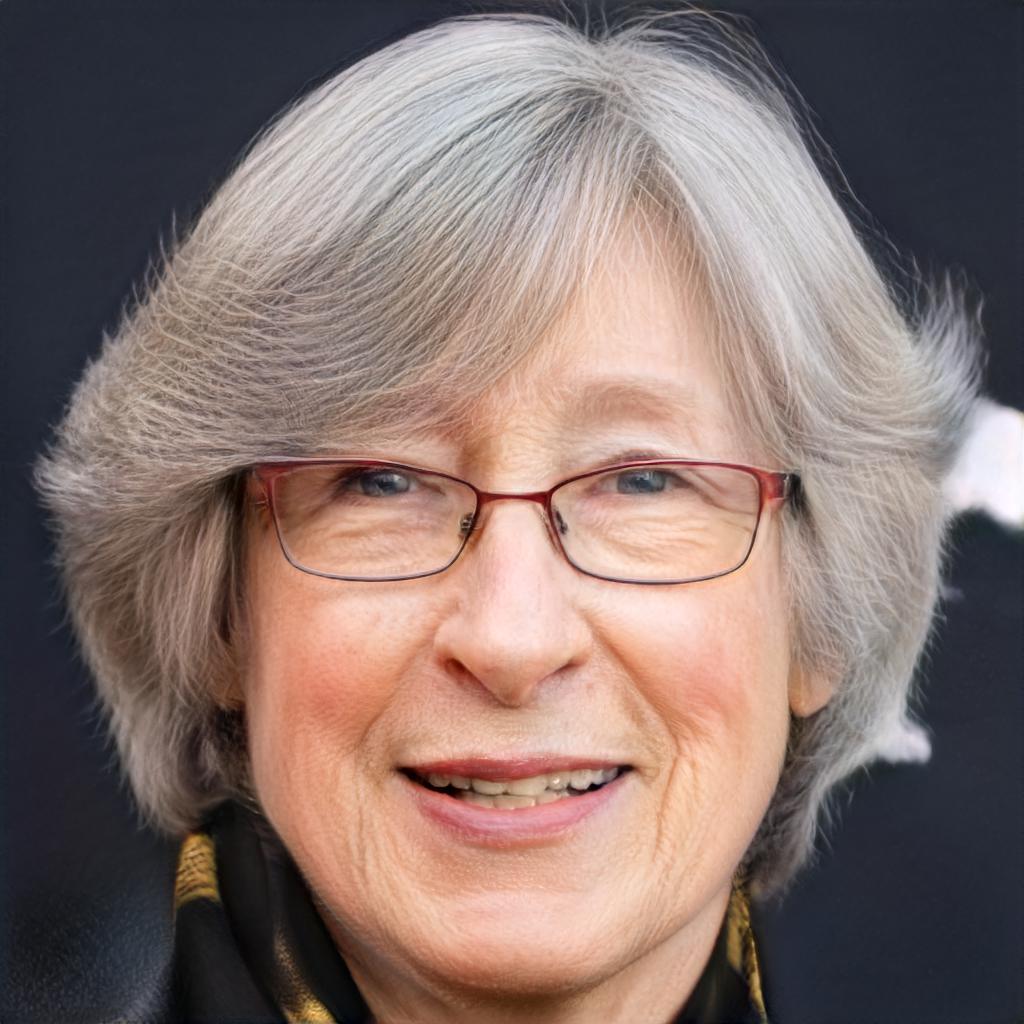} &
        \includegraphics[width=0.107\textwidth]{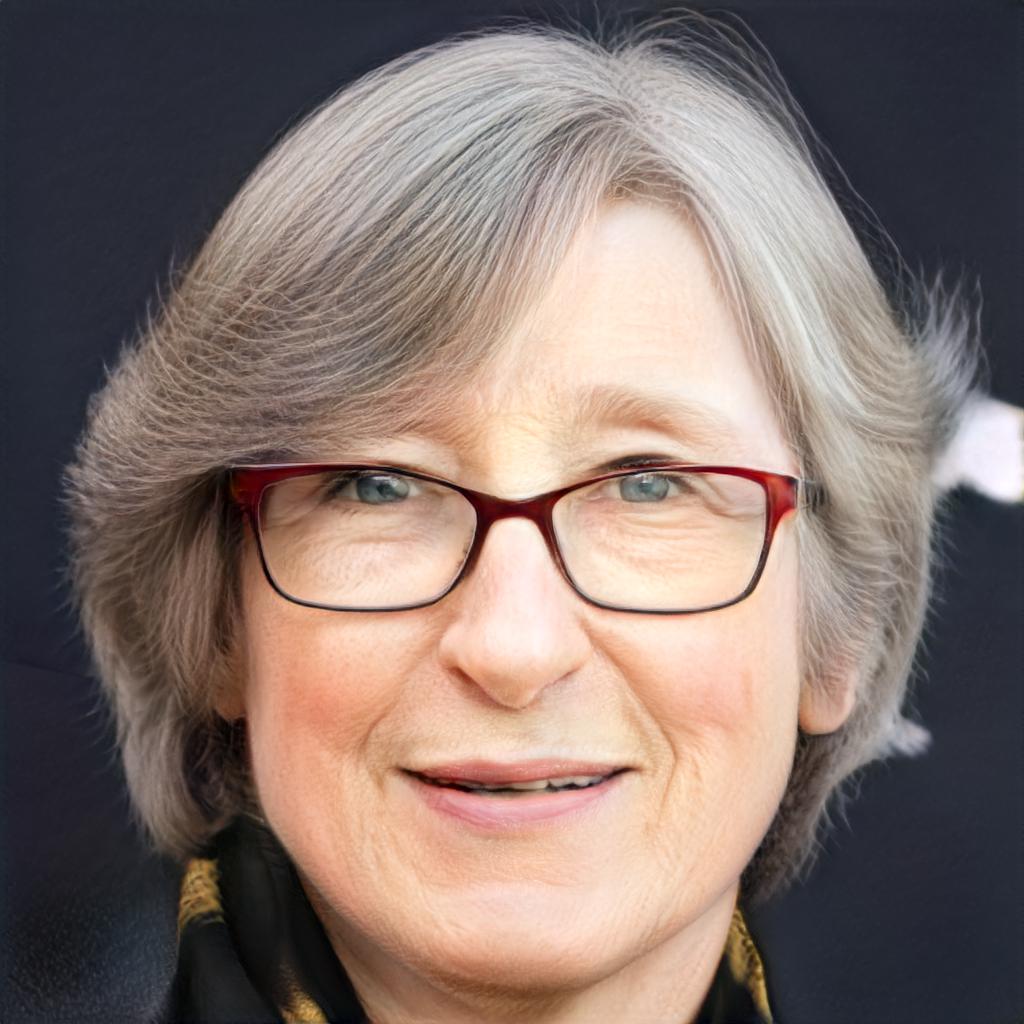} &
        \includegraphics[width=0.107\textwidth]{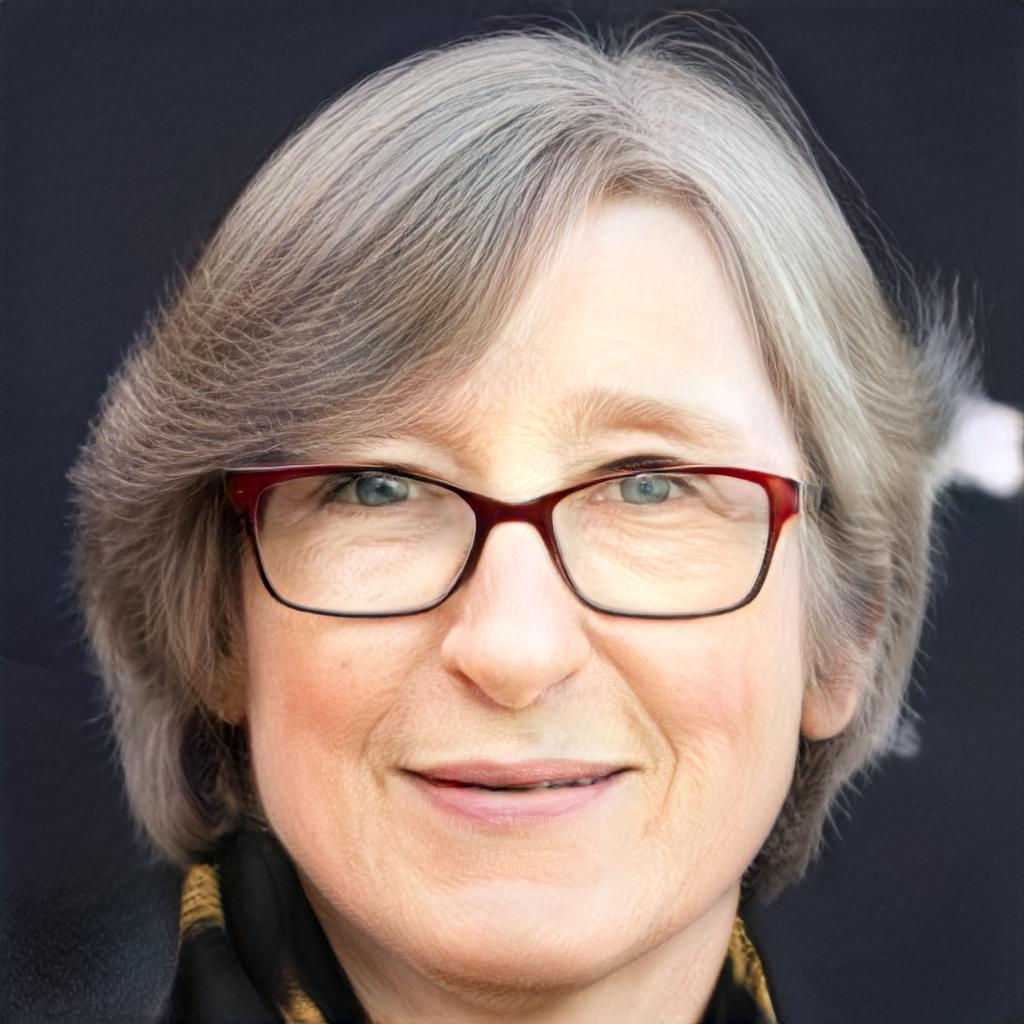} \\
        
        \raisebox{0.265in}{\rotatebox{90}{Ours}} &
        \includegraphics[width=0.107\textwidth]{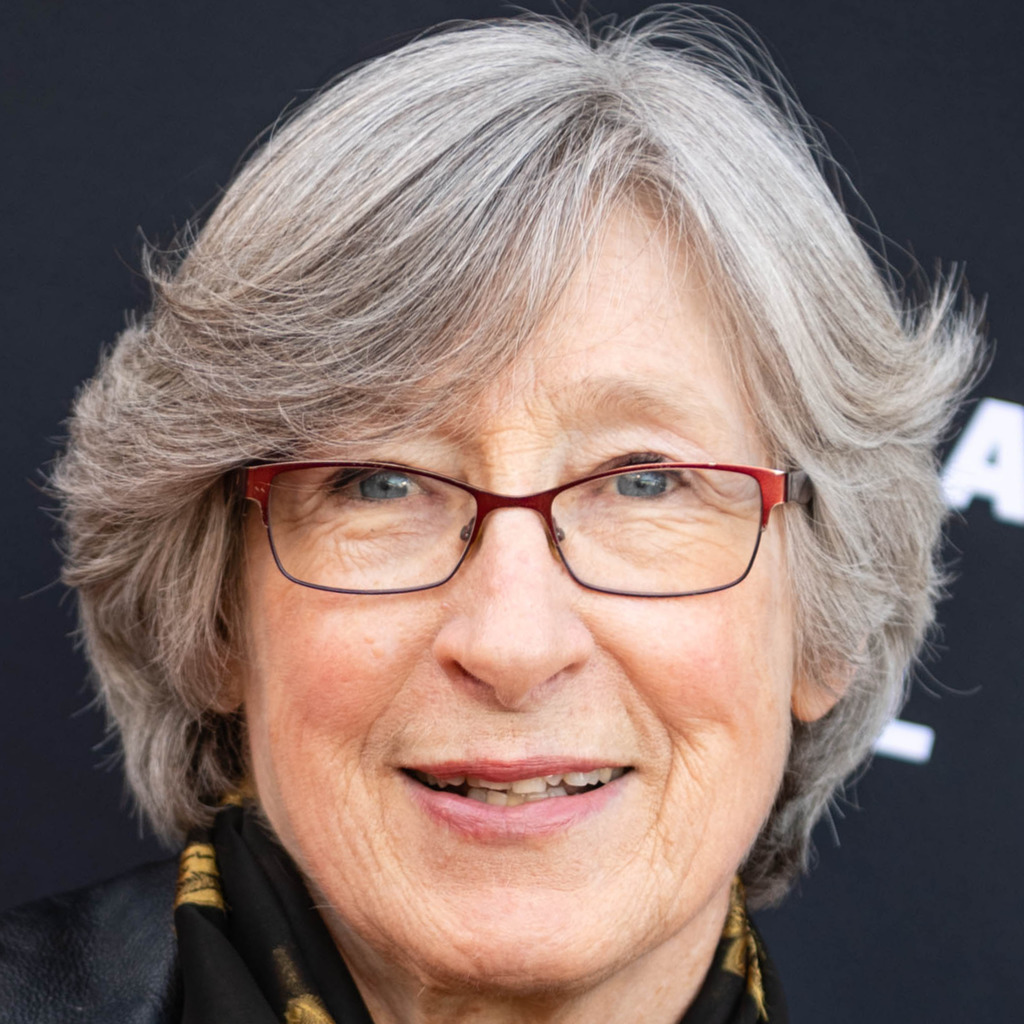} &
        \includegraphics[width=0.107\textwidth]{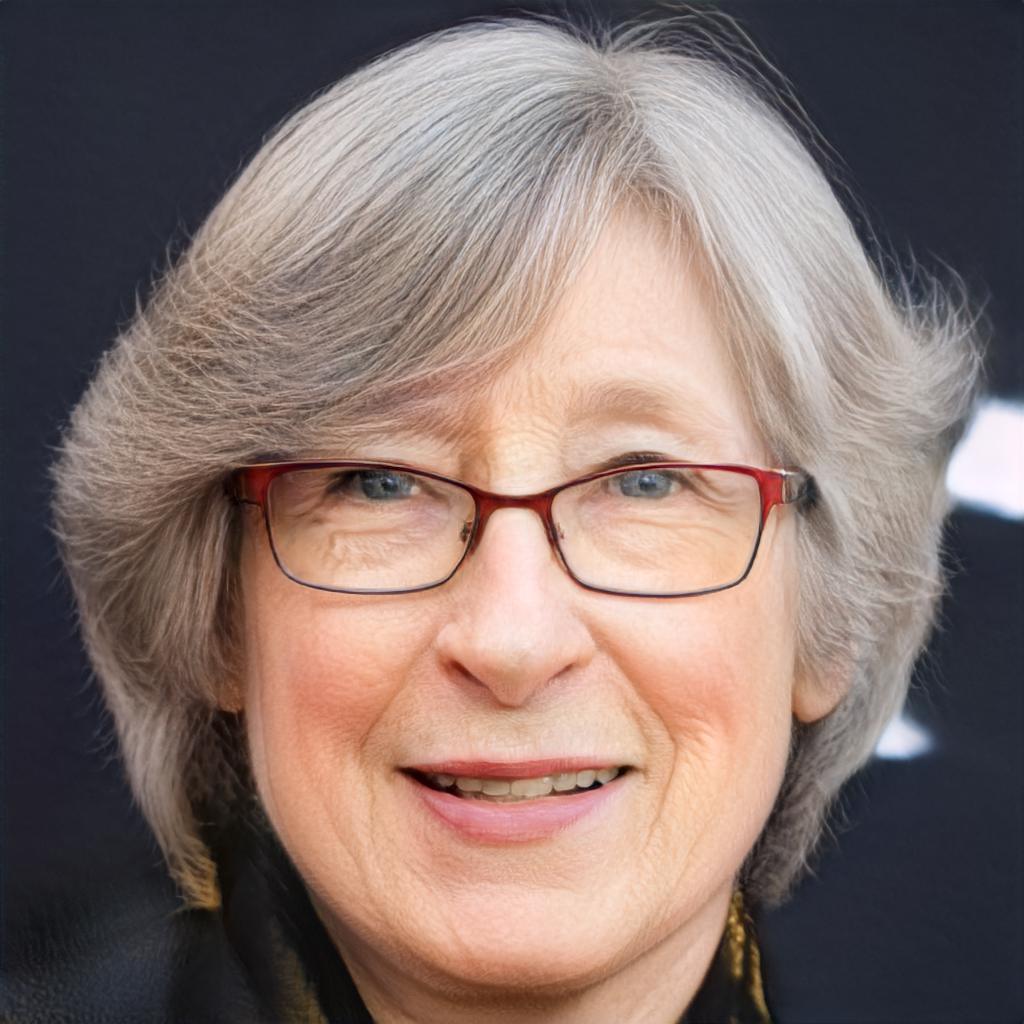} &
        \includegraphics[width=0.107\textwidth]{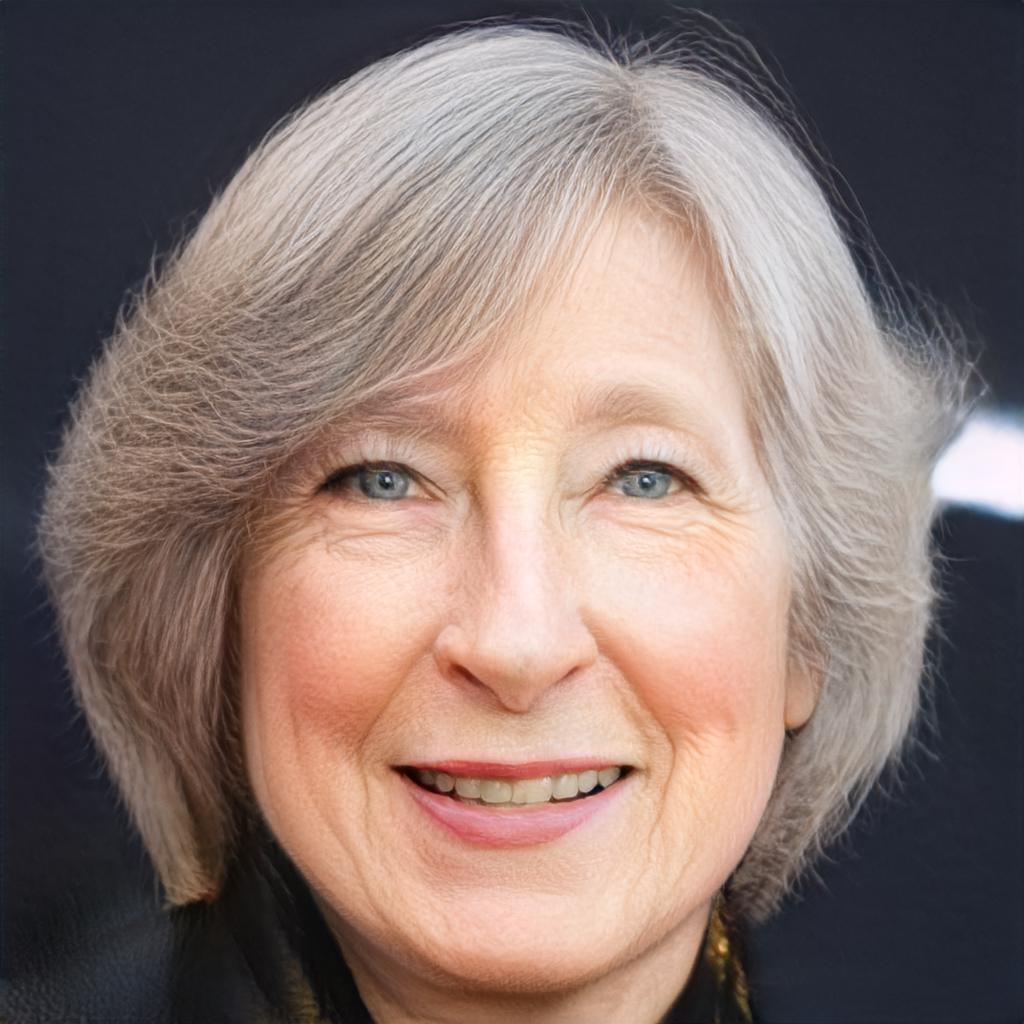} &
        \includegraphics[width=0.107\textwidth]{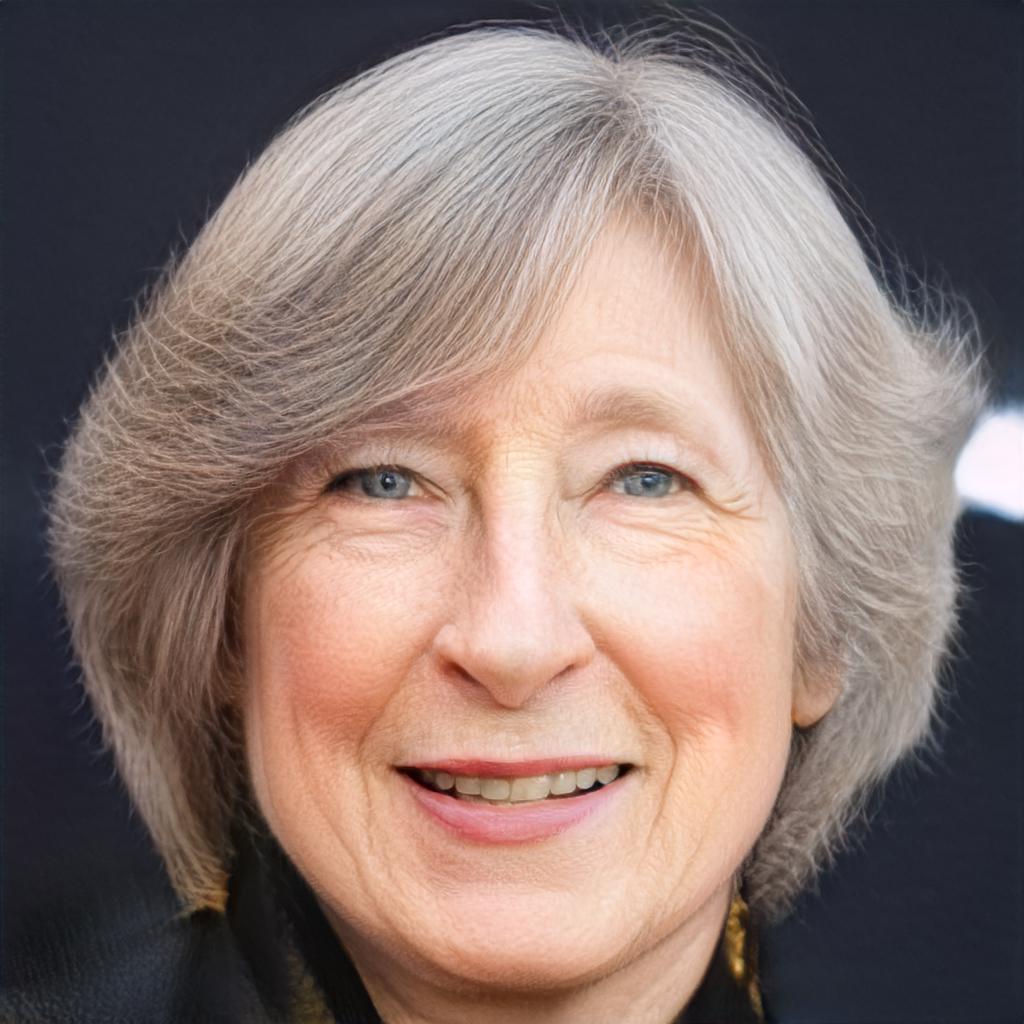} &
        \includegraphics[width=0.107\textwidth]{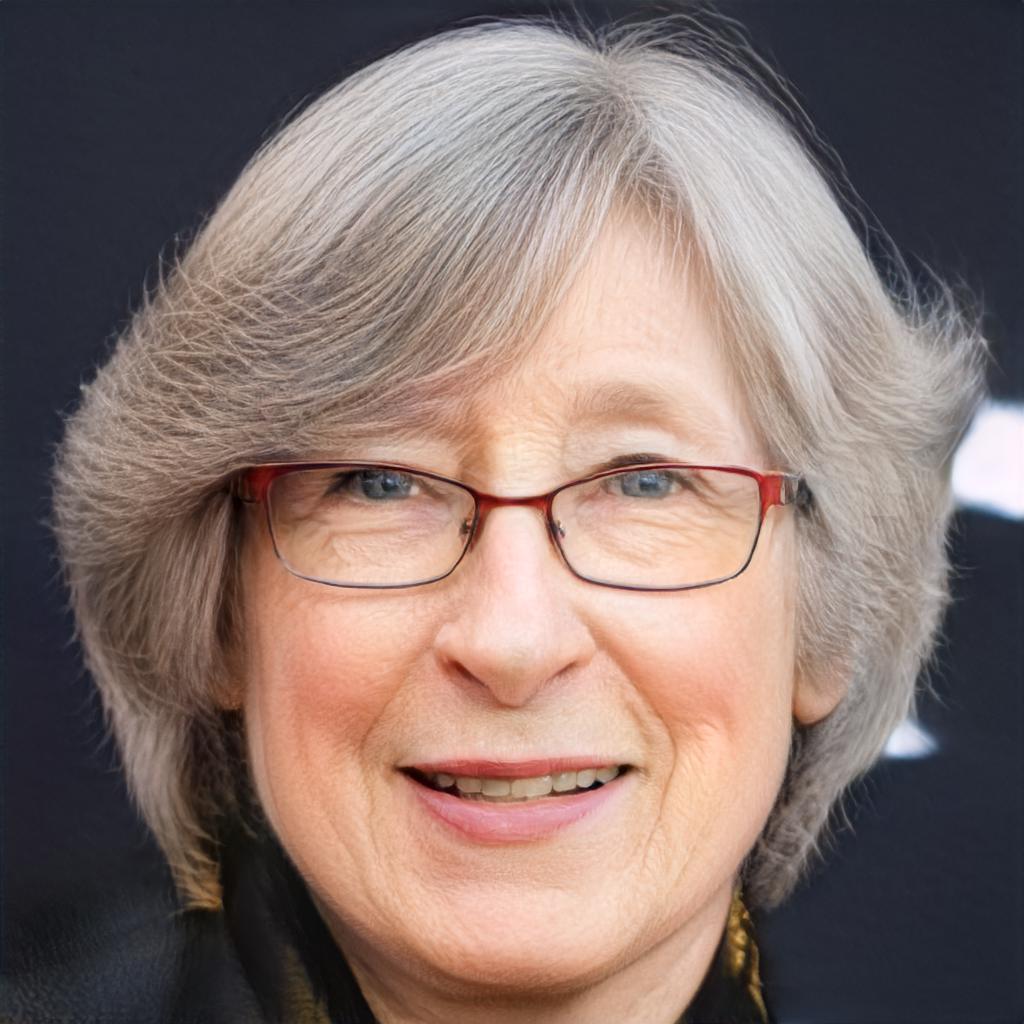} &
        \includegraphics[width=0.107\textwidth]{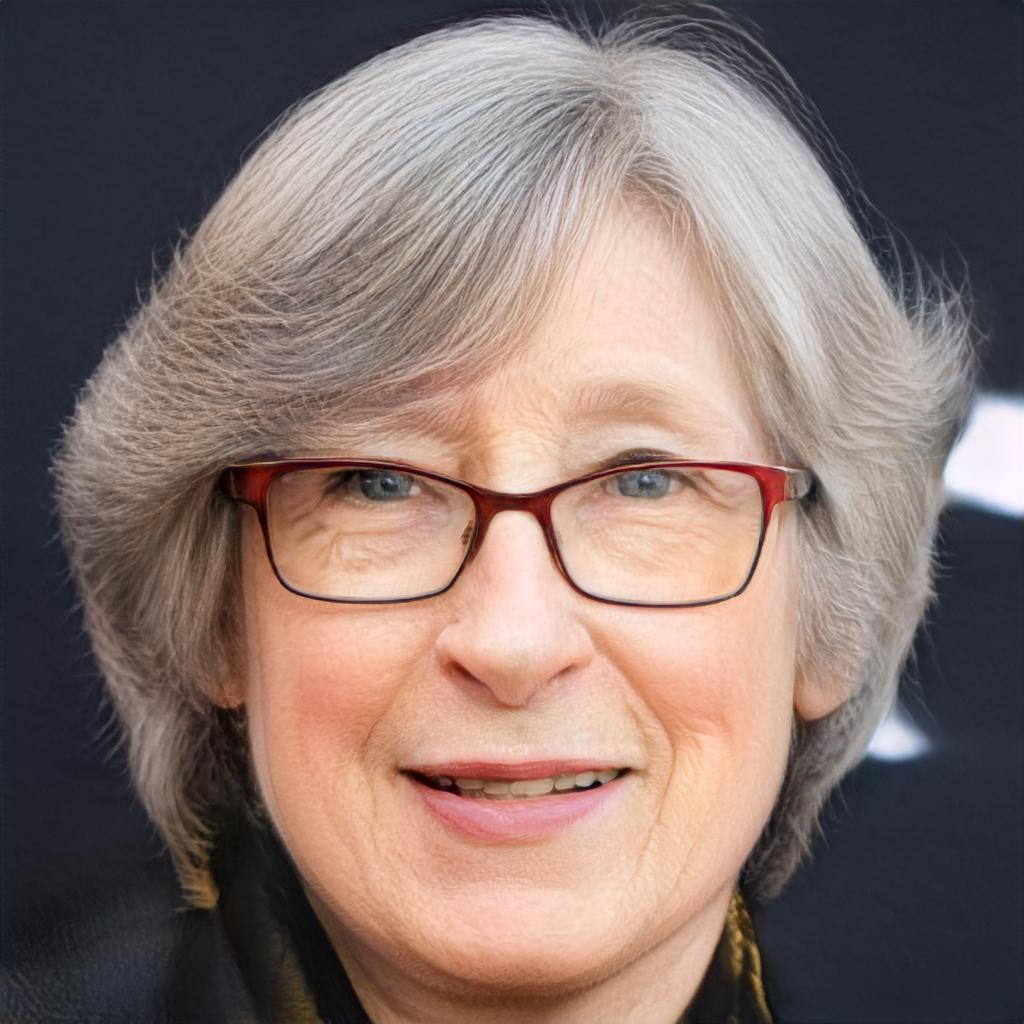} &
        \includegraphics[width=0.107\textwidth]{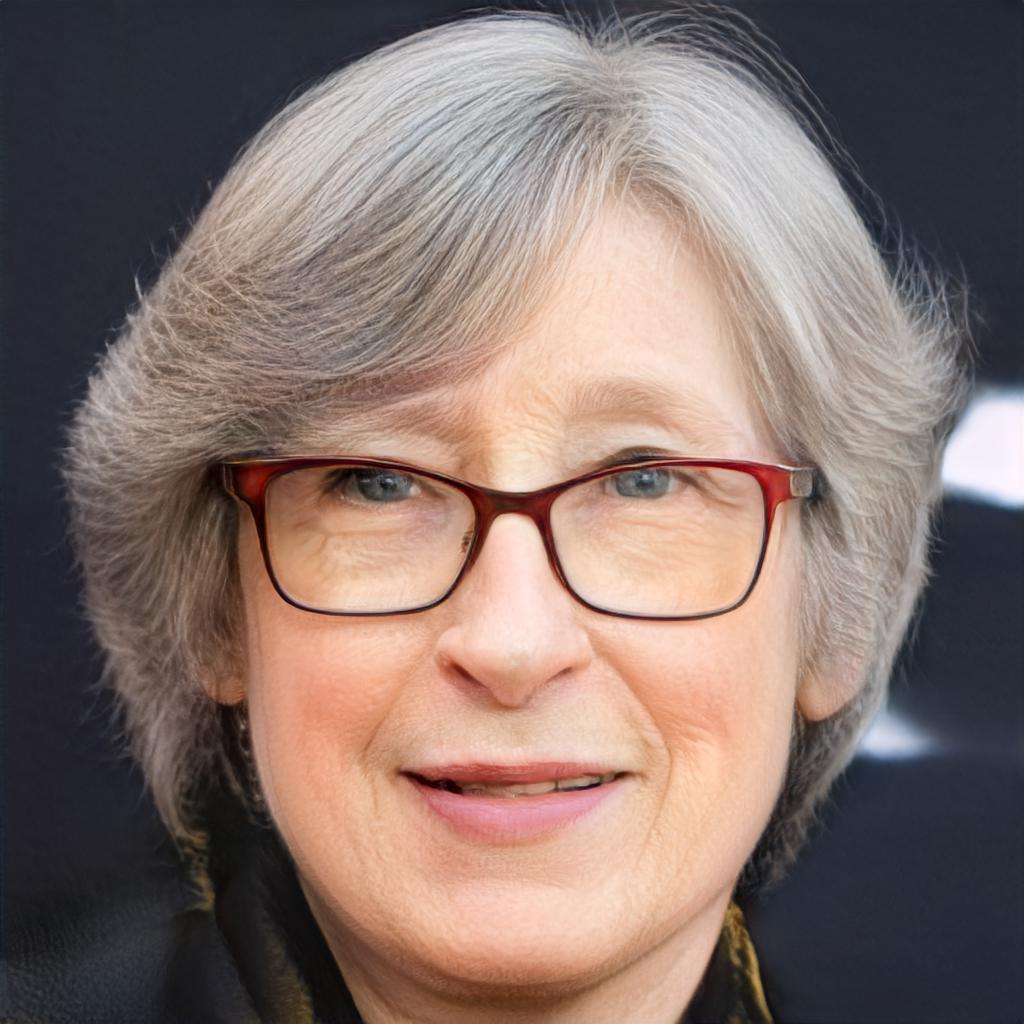} 
        
        \\ & Input & Inversion & \multicolumn{5}{c}{ $\myleftarrow$~Glasses~$\myarrow$ } \\

    \end{tabular}
    
    }
    \vspace{-0.21cm}
    \caption{Linear editing comparisons on real images. In each pair of rows, we compare our method against the initial latent direction which was used to extract our self-conditioning labels. In all cases, our method better preserves the identity and allows for more significant manipulations before image quality suffers drastically. In the glasses example (bottom), continued movement in the negative direction on an image without glasses leads to an increase in age. Our model avoids this problem and produces the same identity.}
    \vspace{-0.17cm}
    \label{fig:editing_comparisons}
\end{figure*}

%% file: resources/figures/non_linear_comparisons_arxiv.tex
\begin{figure*}
    \centering
    \setlength{\belowcaptionskip}{-8pt}
    \setlength{\tabcolsep}{2.0pt}
    {\small
    \begin{tabular}{c c}
    
    \begin{tabular}{c c c c c c}

        \raisebox{0.00in}{\rotatebox{90}{\scriptsize Local Basis\cite{choi2021escape}}} &
        \includegraphics[width=0.0875\textwidth]{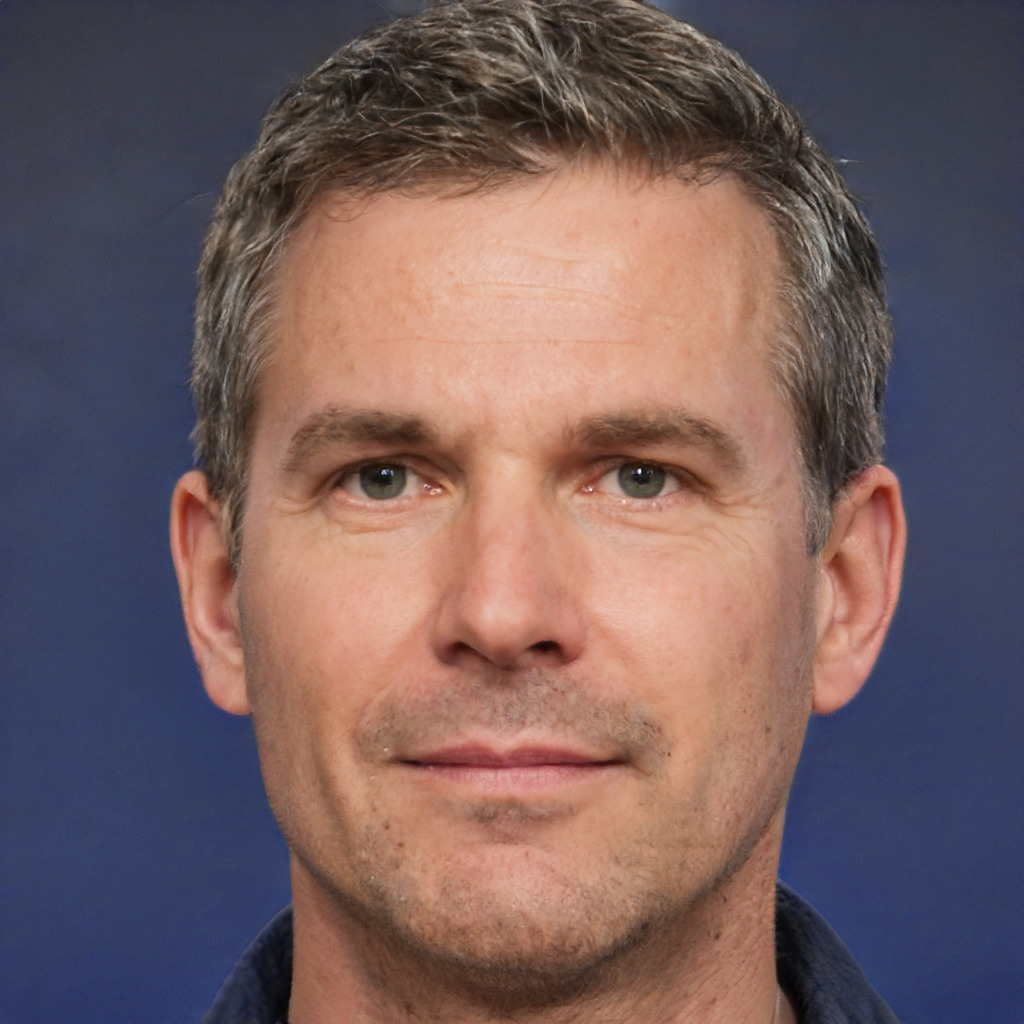} &
        \includegraphics[width=0.0875\textwidth]{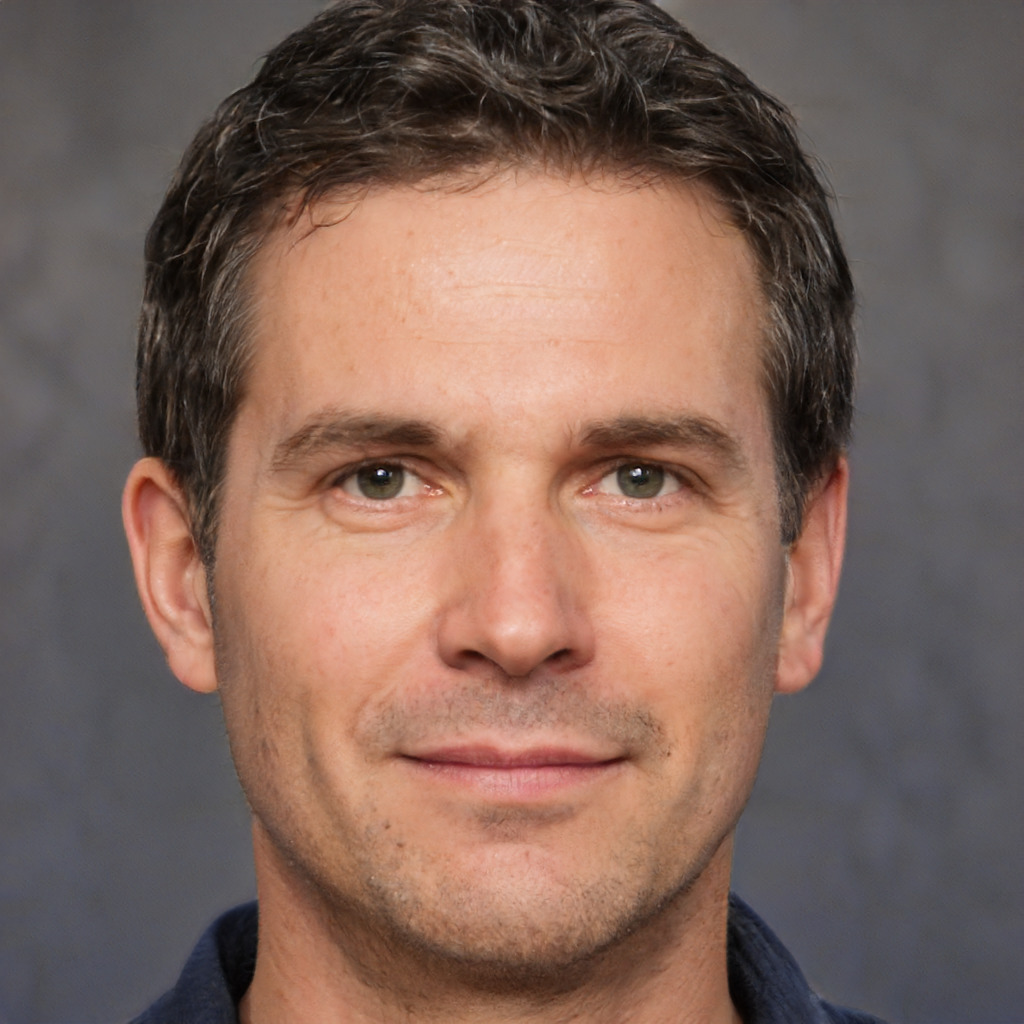} &
        \includegraphics[width=0.0875\textwidth]{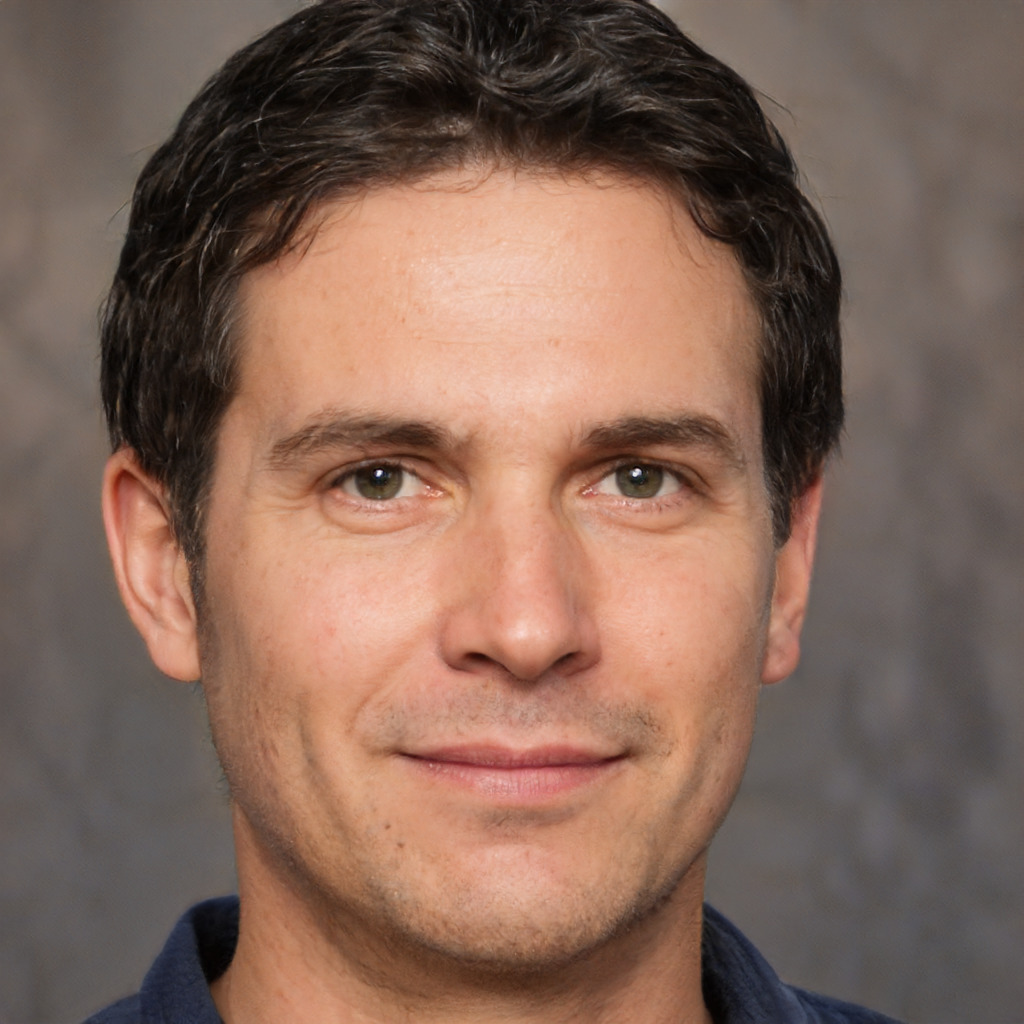} &
        \includegraphics[width=0.0875\textwidth]{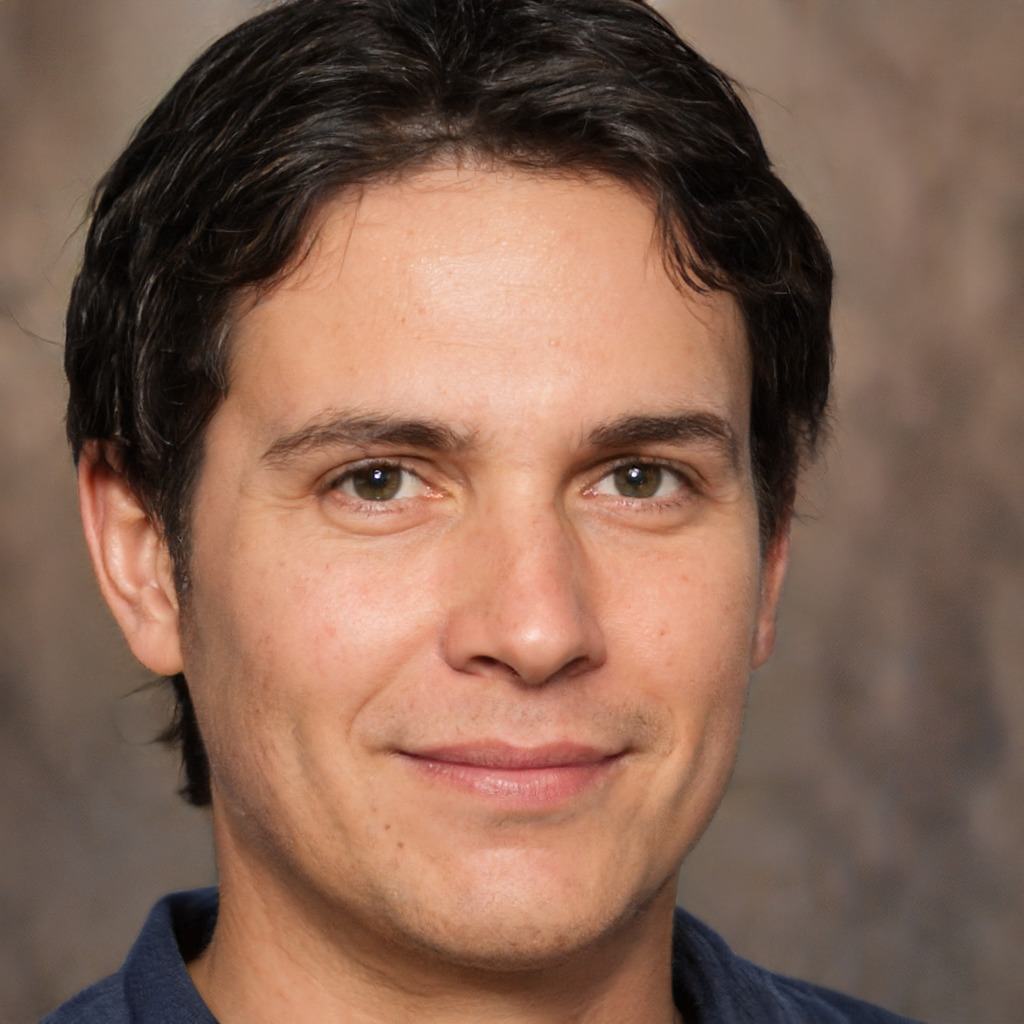} &
        \includegraphics[width=0.0875\textwidth]{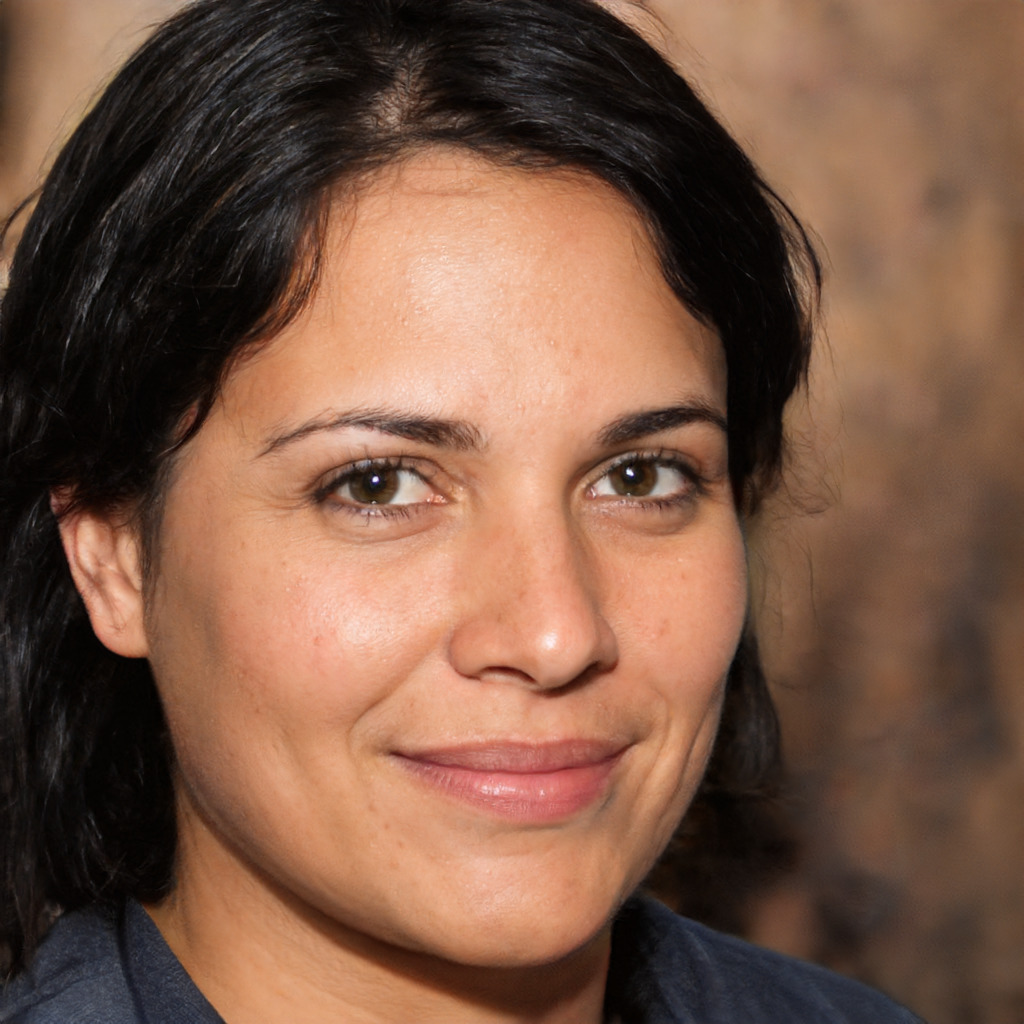} \\
        
        \raisebox{0.155in}{\rotatebox{90}{Ours}} &
        \includegraphics[width=0.0875\textwidth]{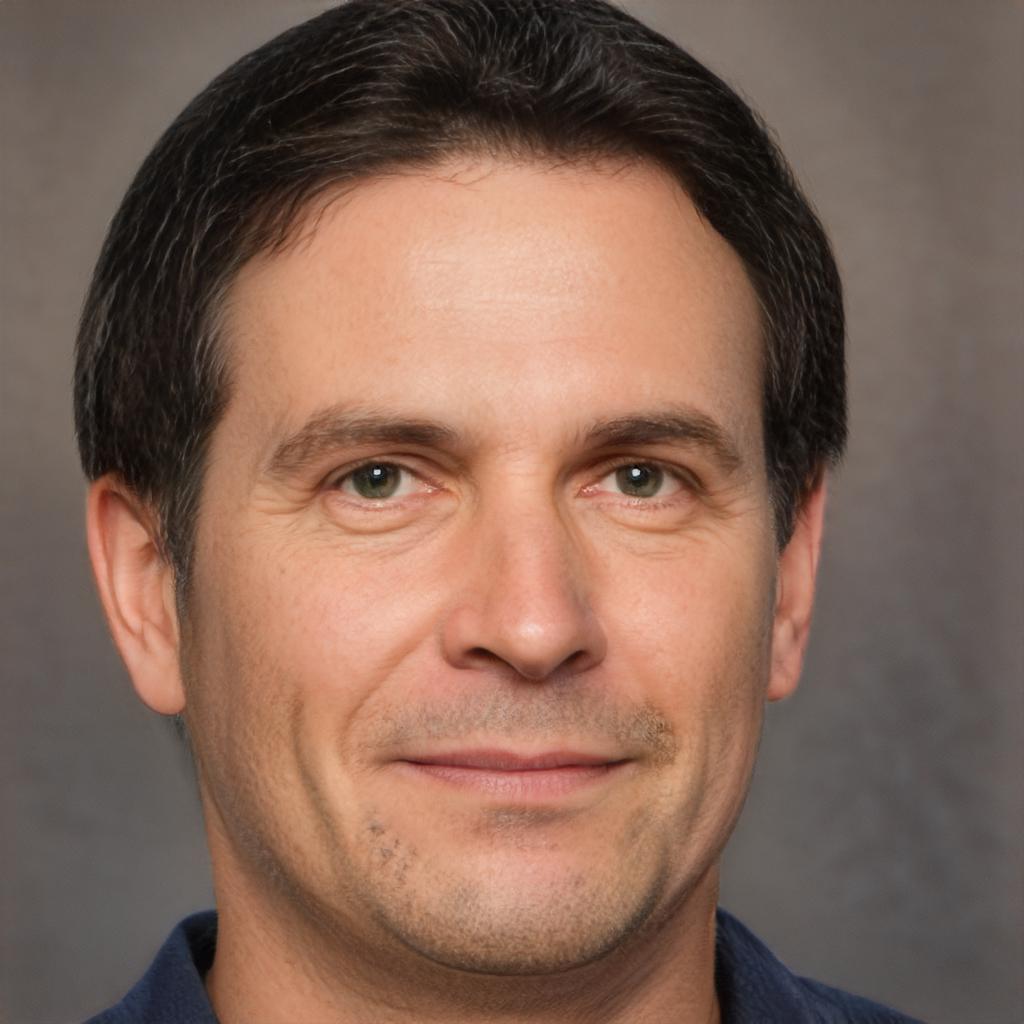} &
        \includegraphics[width=0.0875\textwidth]{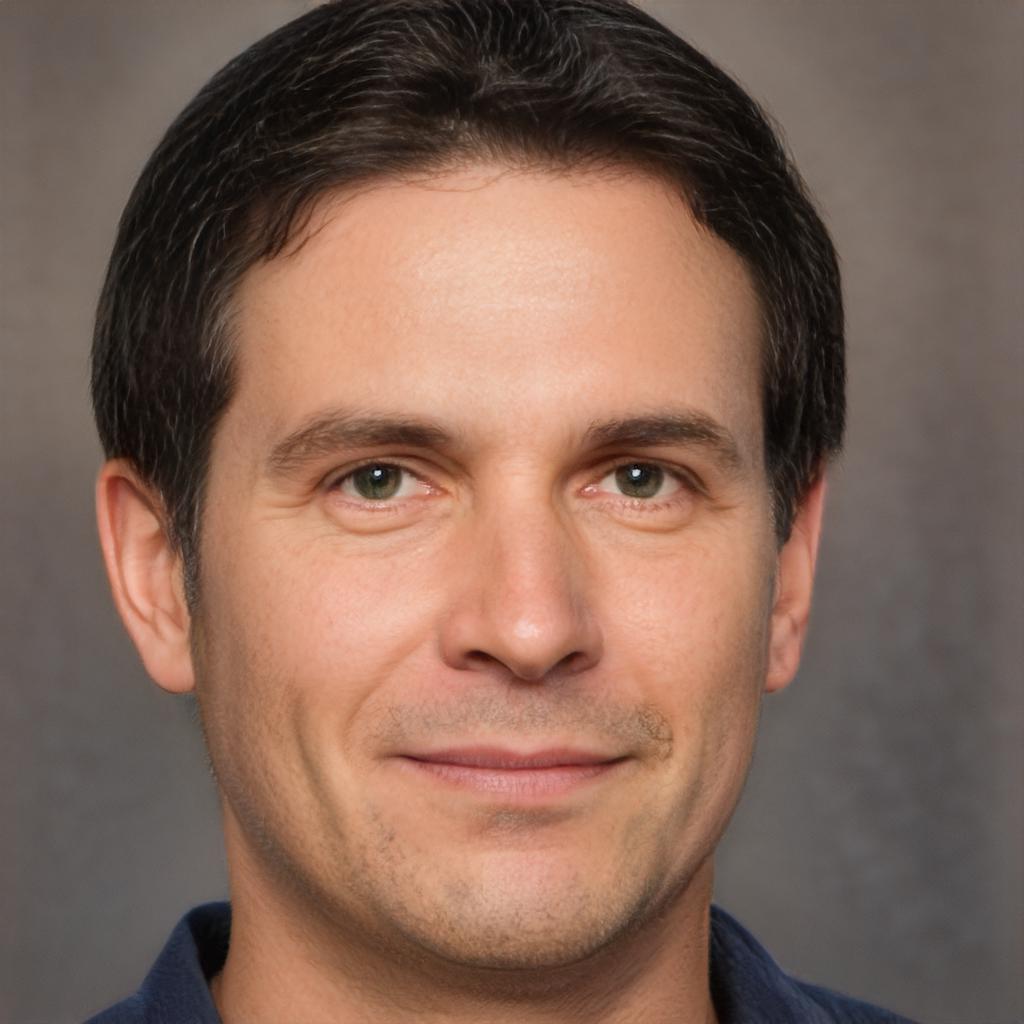} &
        \includegraphics[width=0.0875\textwidth]{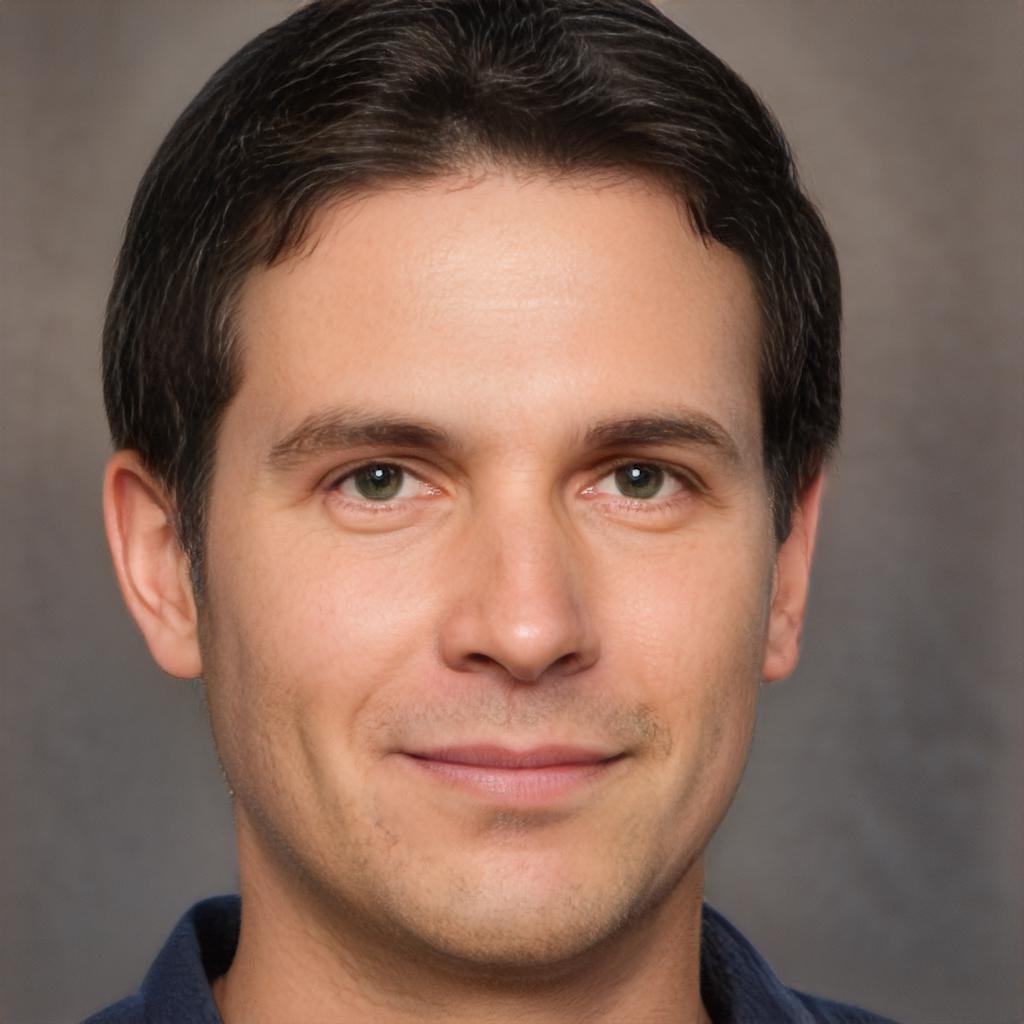} &
        \includegraphics[width=0.0875\textwidth]{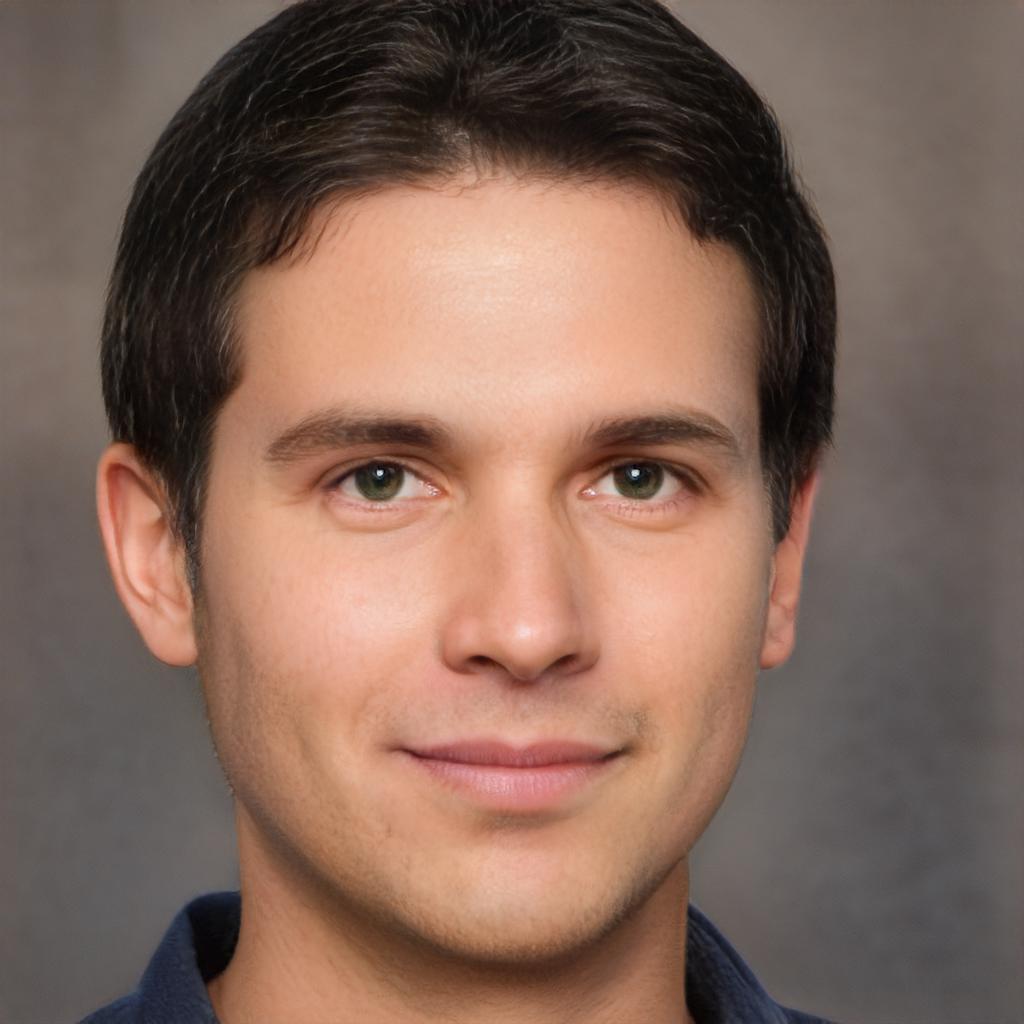} &
        \includegraphics[width=0.0875\textwidth]{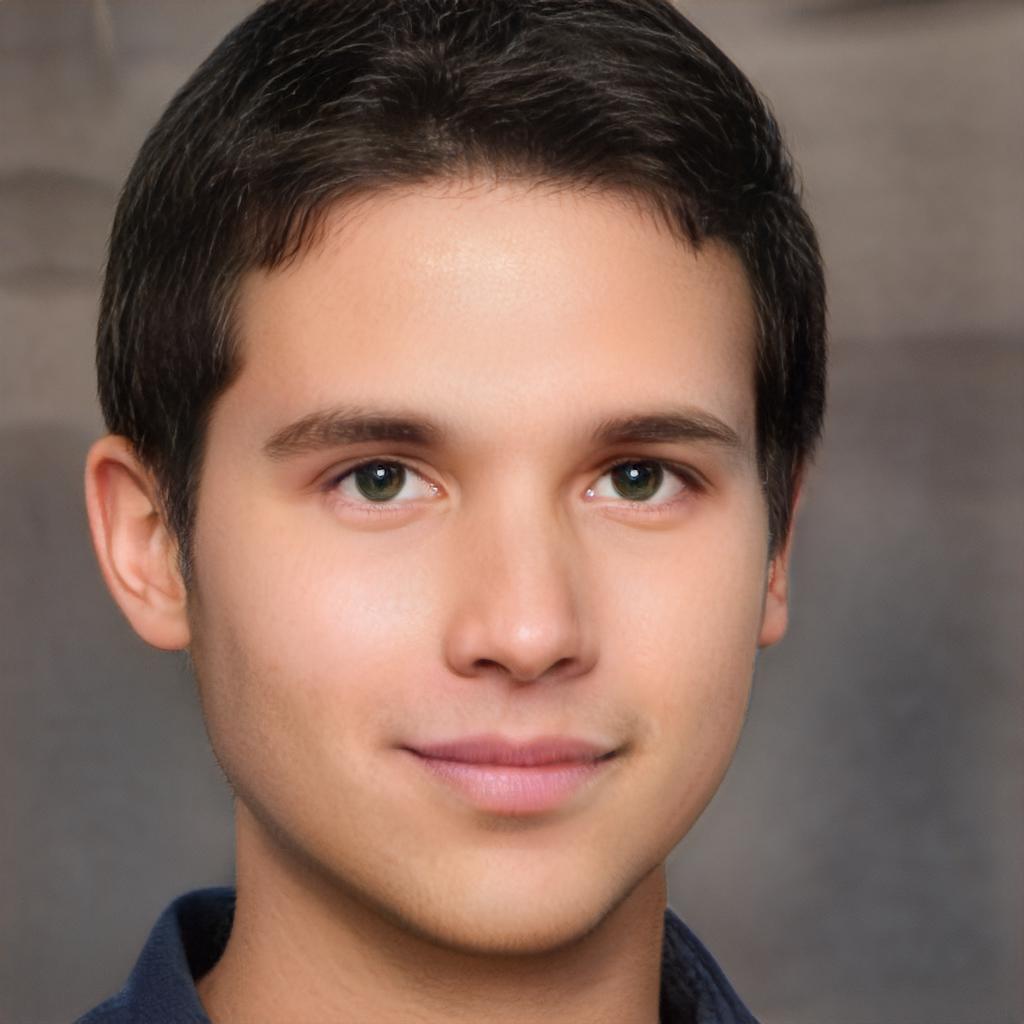} \\
        
        & \multicolumn{5}{c}{ Age }
        
    \end{tabular} 
    
    &
    
        \begin{tabular}{c c c c c c}

        \raisebox{0.00in}{\rotatebox{90}{\scriptsize Local Basis\cite{choi2021escape}}} &
        \includegraphics[width=0.0875\textwidth]{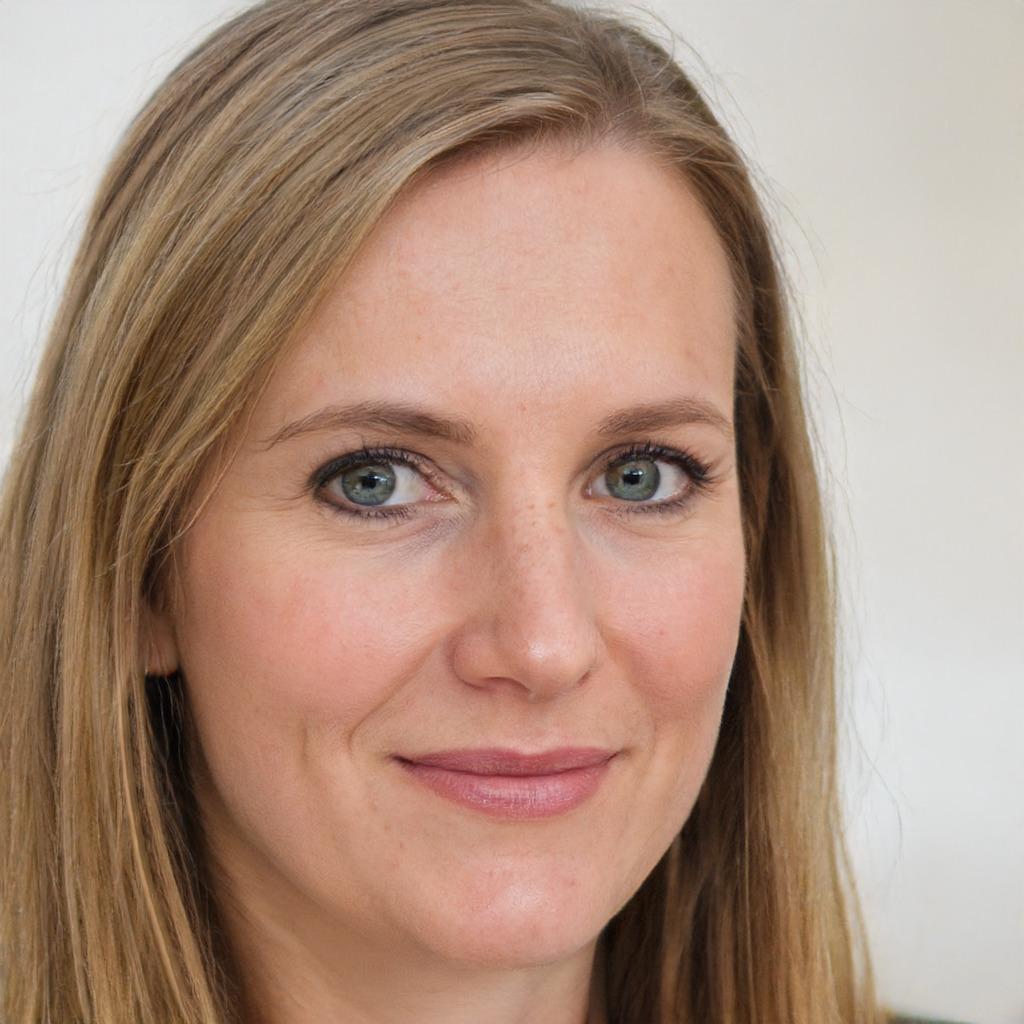} &
        \includegraphics[width=0.0875\textwidth]{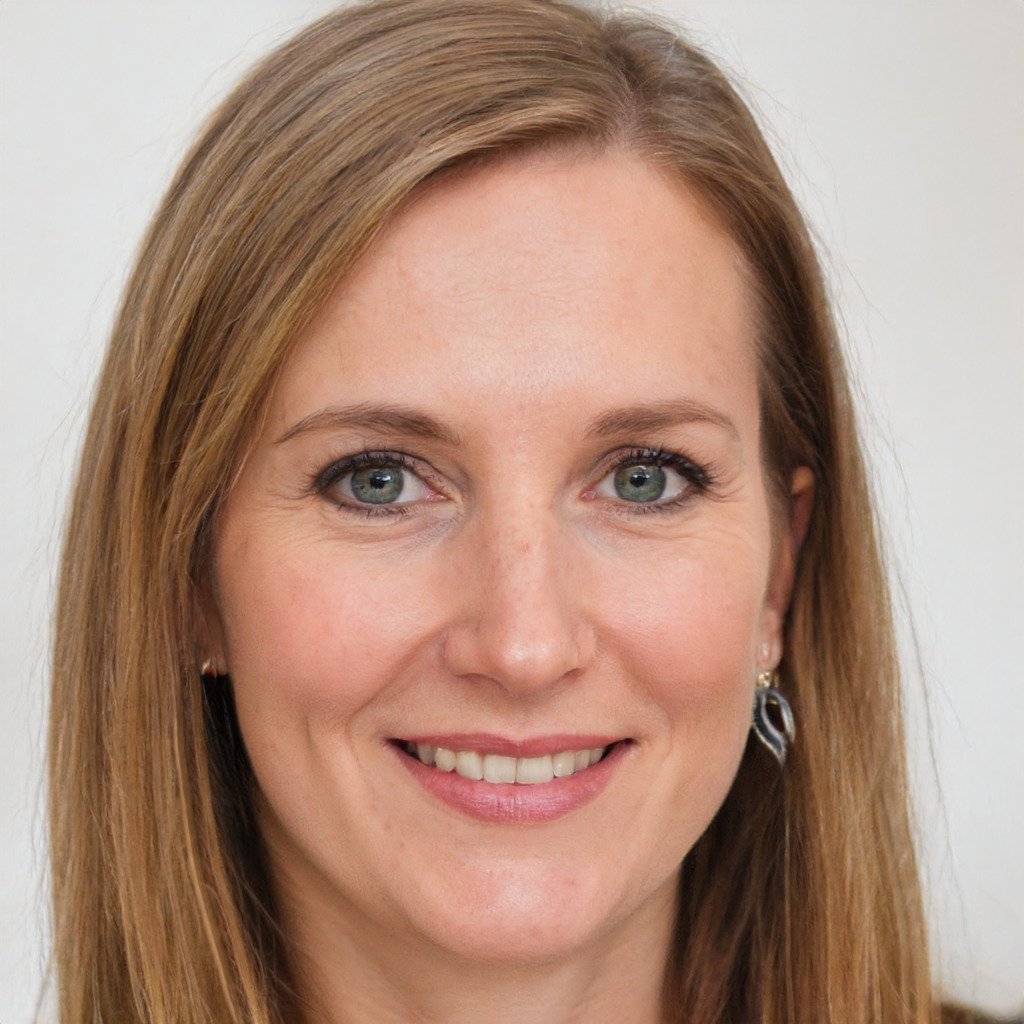} &
        \includegraphics[width=0.0875\textwidth]{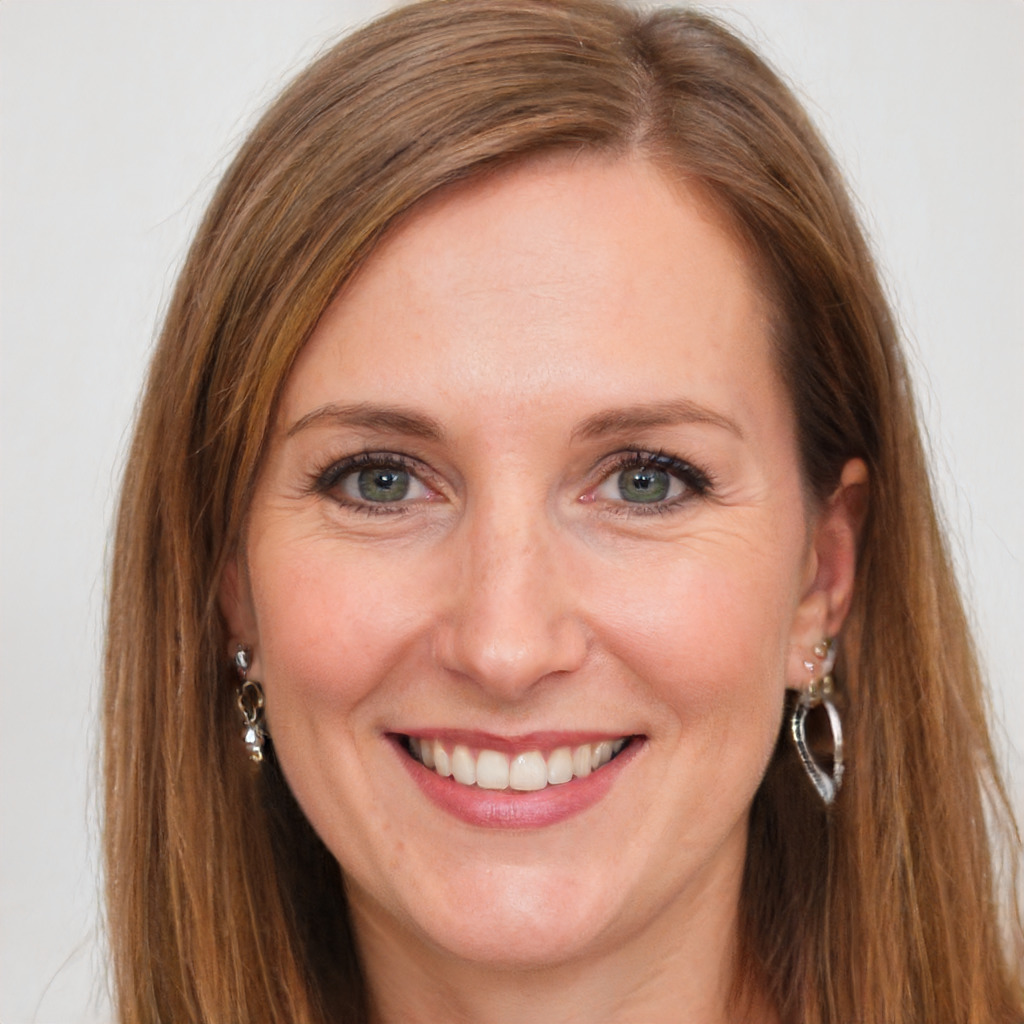} &
        \includegraphics[width=0.0875\textwidth]{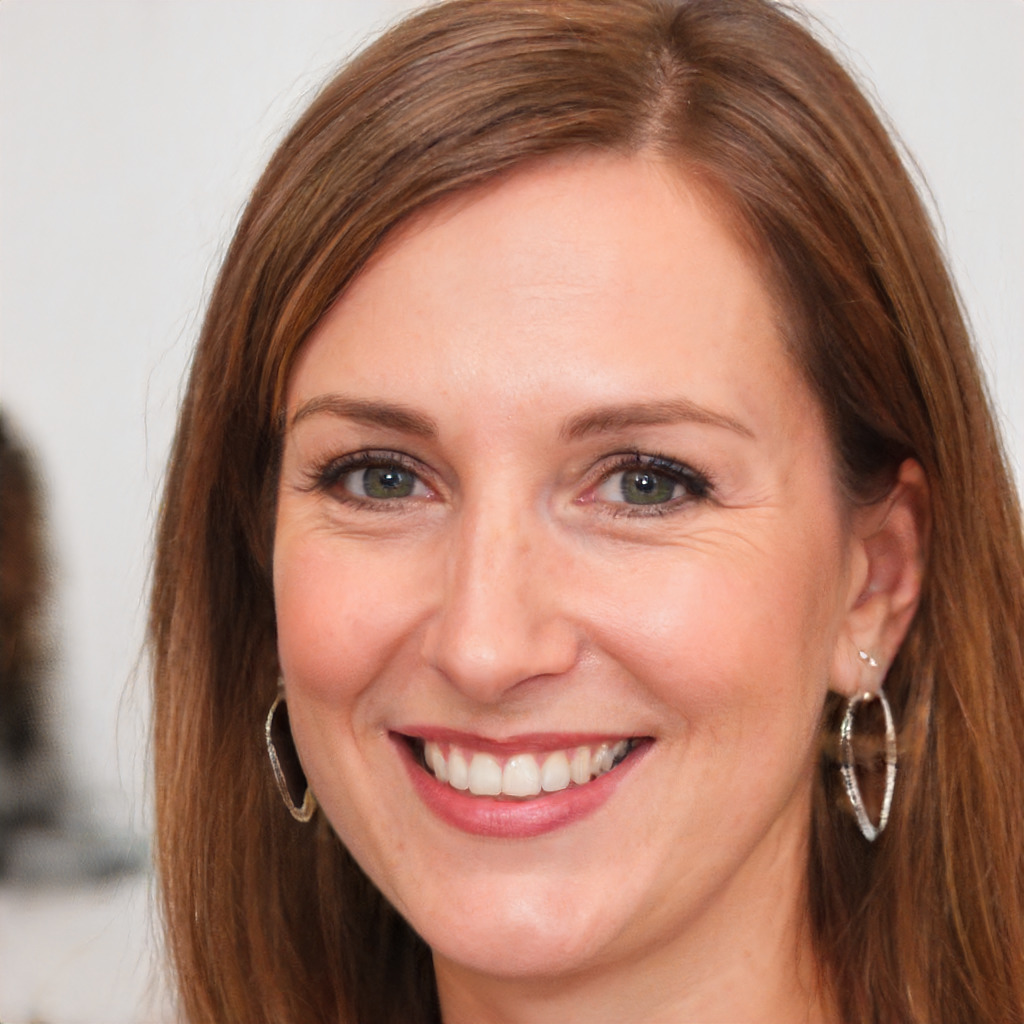} &
        \includegraphics[width=0.0875\textwidth]{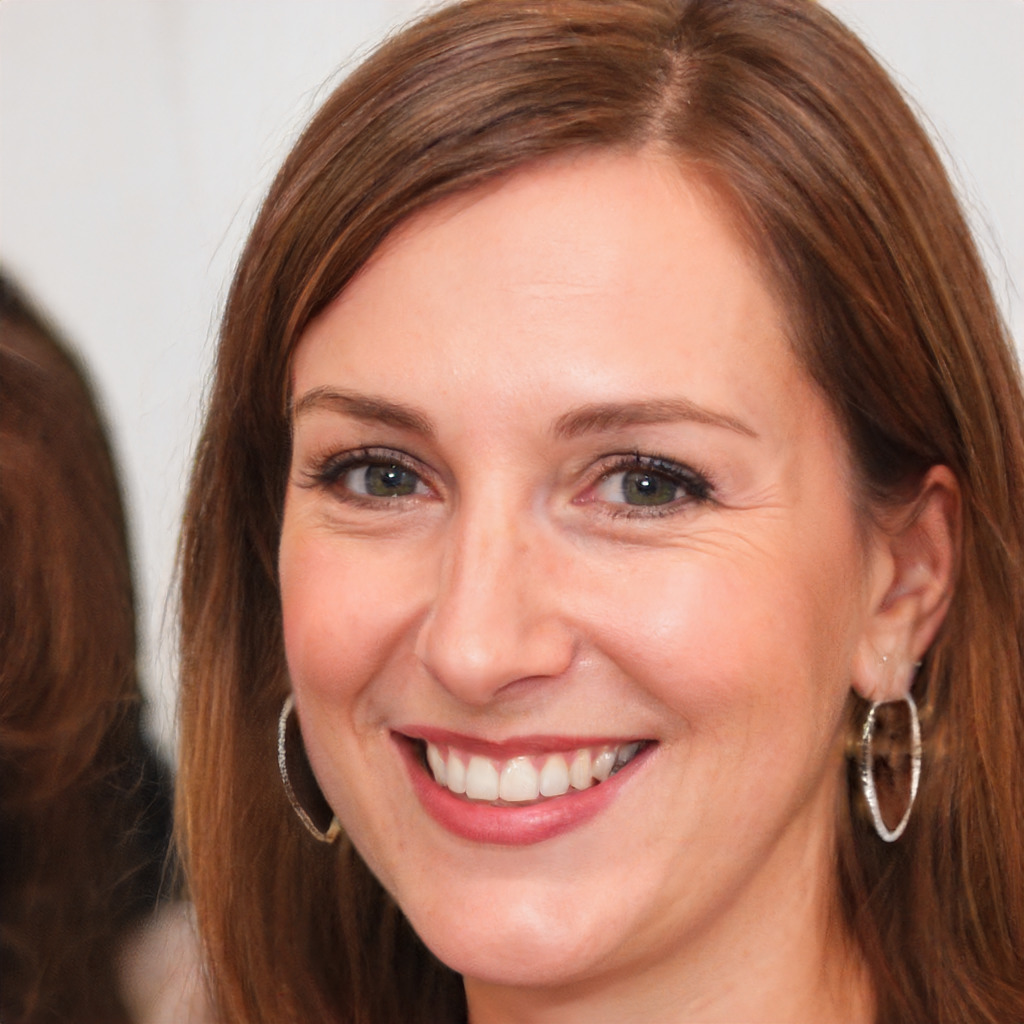} \\
        
        \raisebox{0.155in}{\rotatebox{90}{Ours}} &
        \includegraphics[width=0.0875\textwidth]{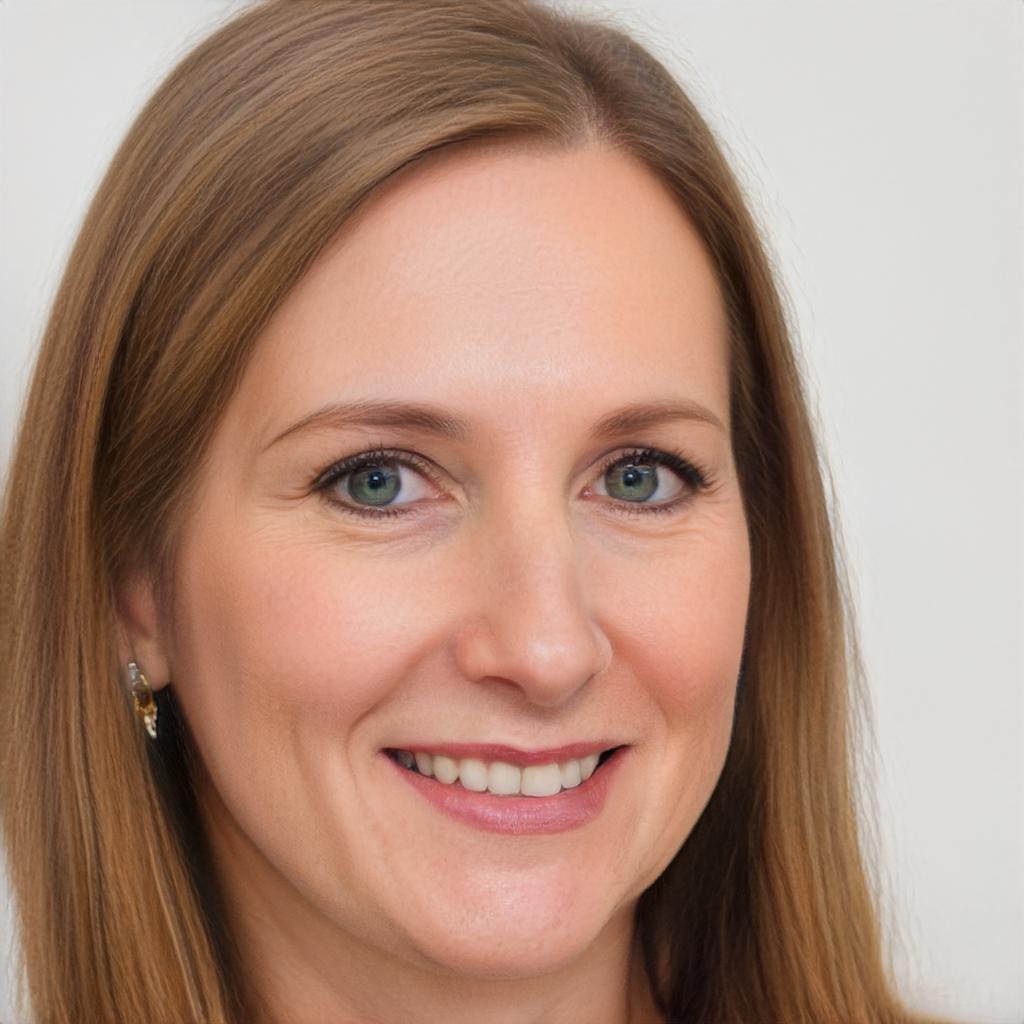} &
        \includegraphics[width=0.0875\textwidth]{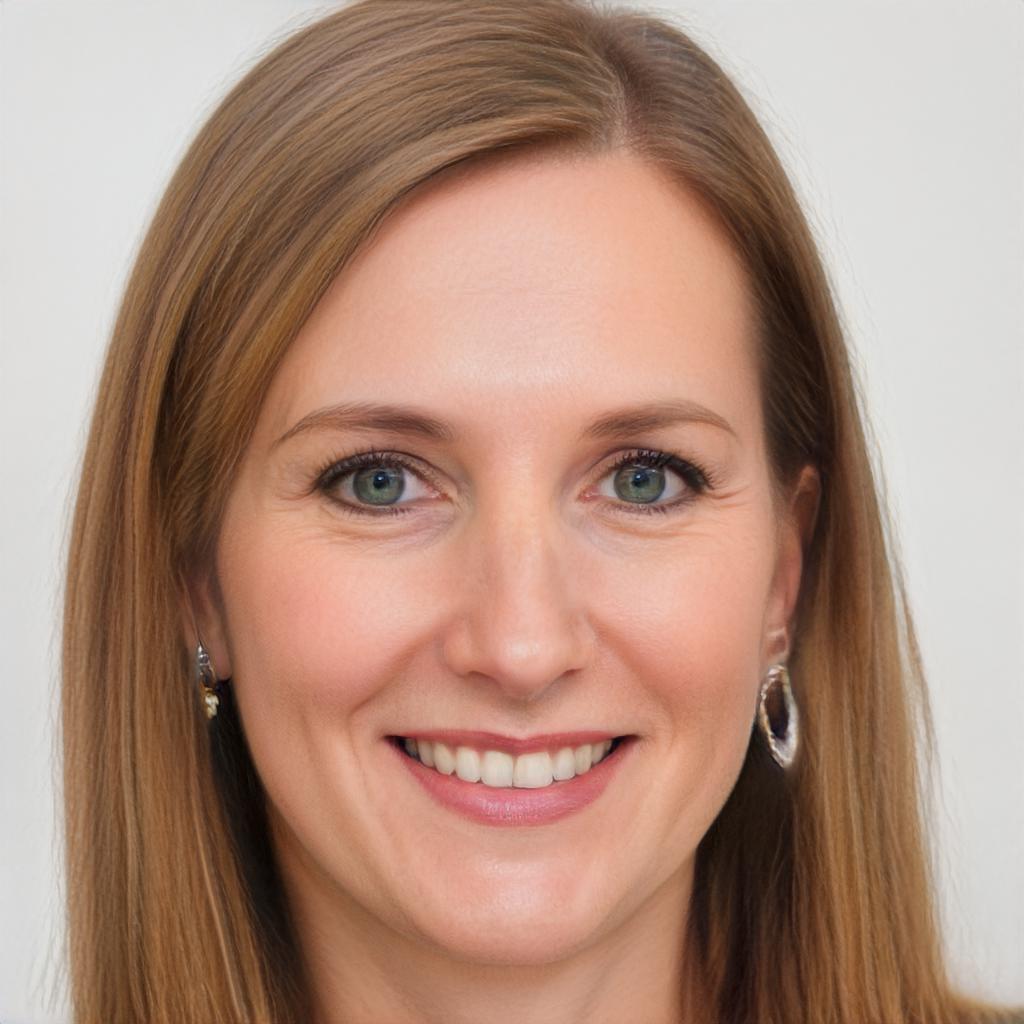} &
        \includegraphics[width=0.0875\textwidth]{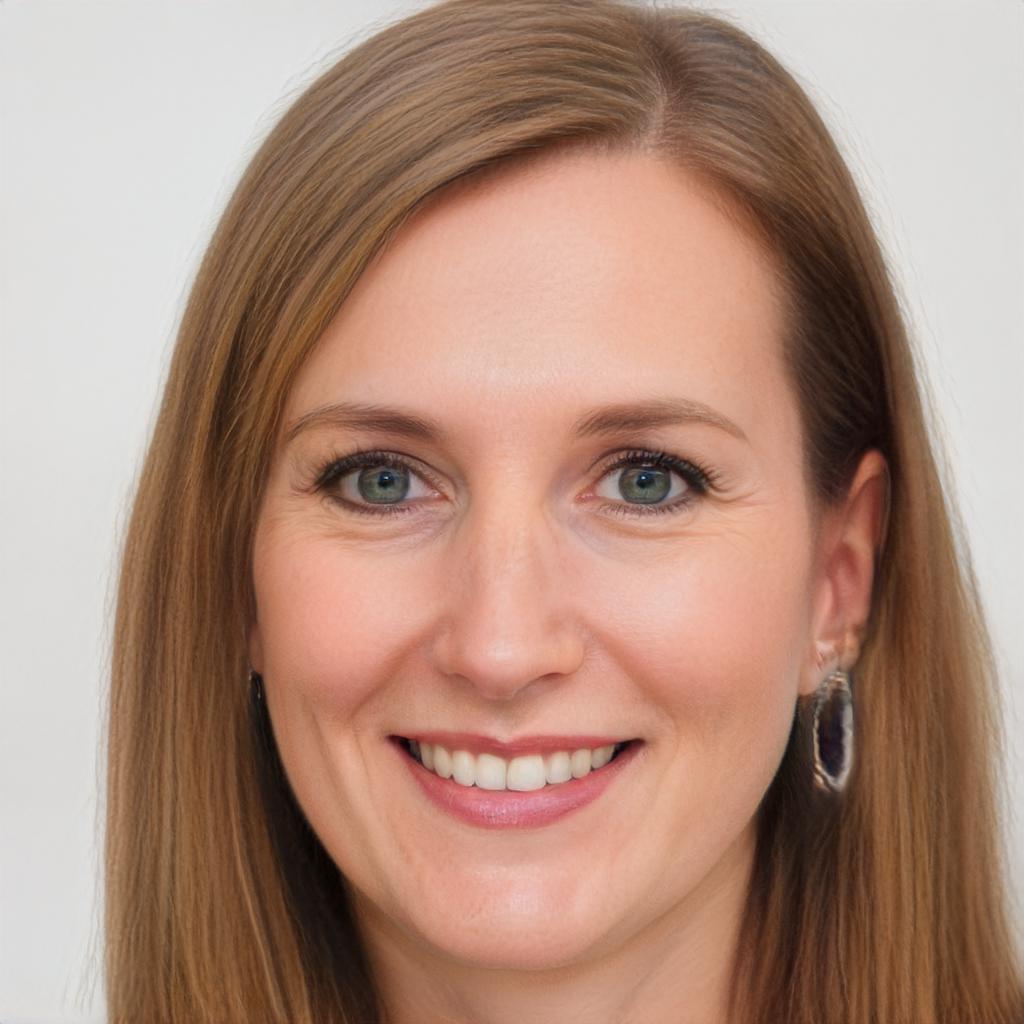} &
        \includegraphics[width=0.0875\textwidth]{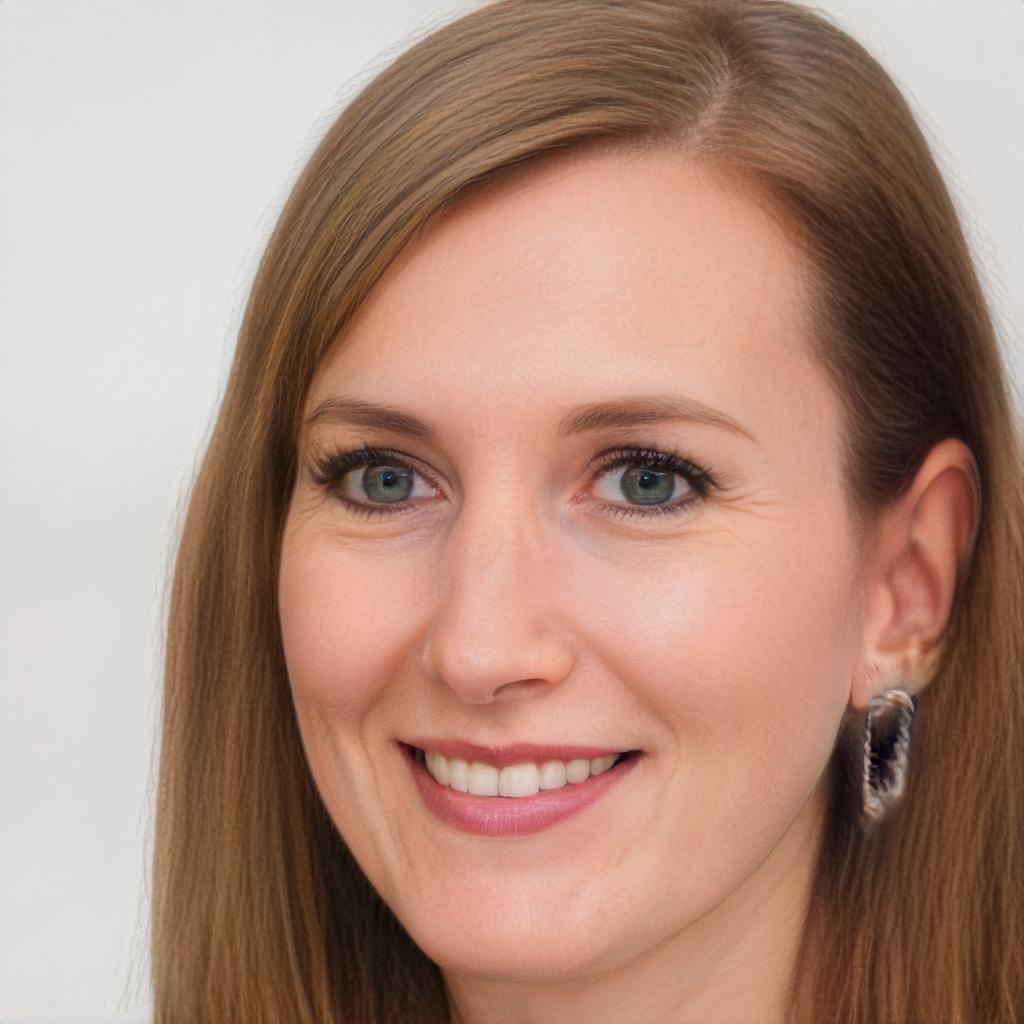} &
        \includegraphics[width=0.0875\textwidth]{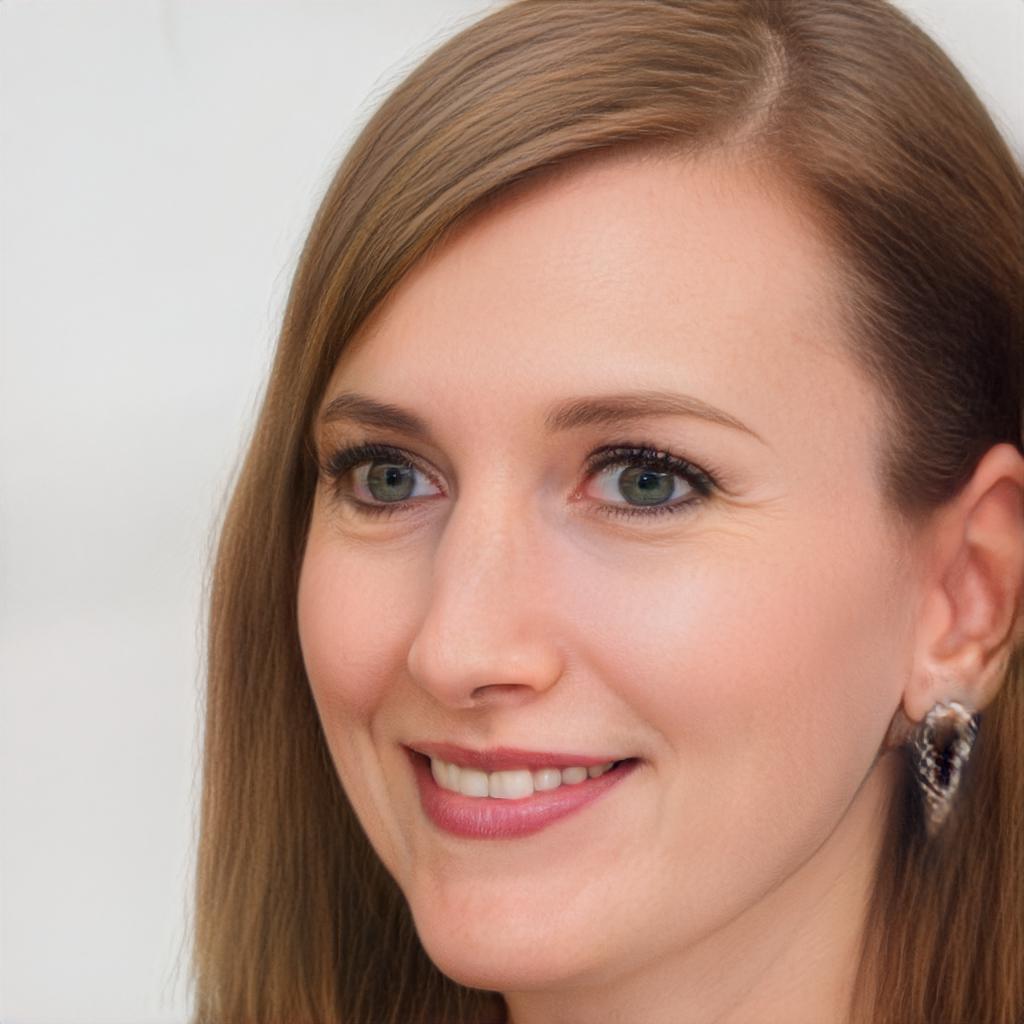} \\
        
        & \multicolumn{5}{c}{ Pose }
        
    \end{tabular}
    
    \\
    
        \begin{tabular}{c c c c c c}

        \raisebox{0.08in}{\rotatebox{90}{\footnotesize StyleFlow}} &
        \includegraphics[width=0.0875\textwidth]{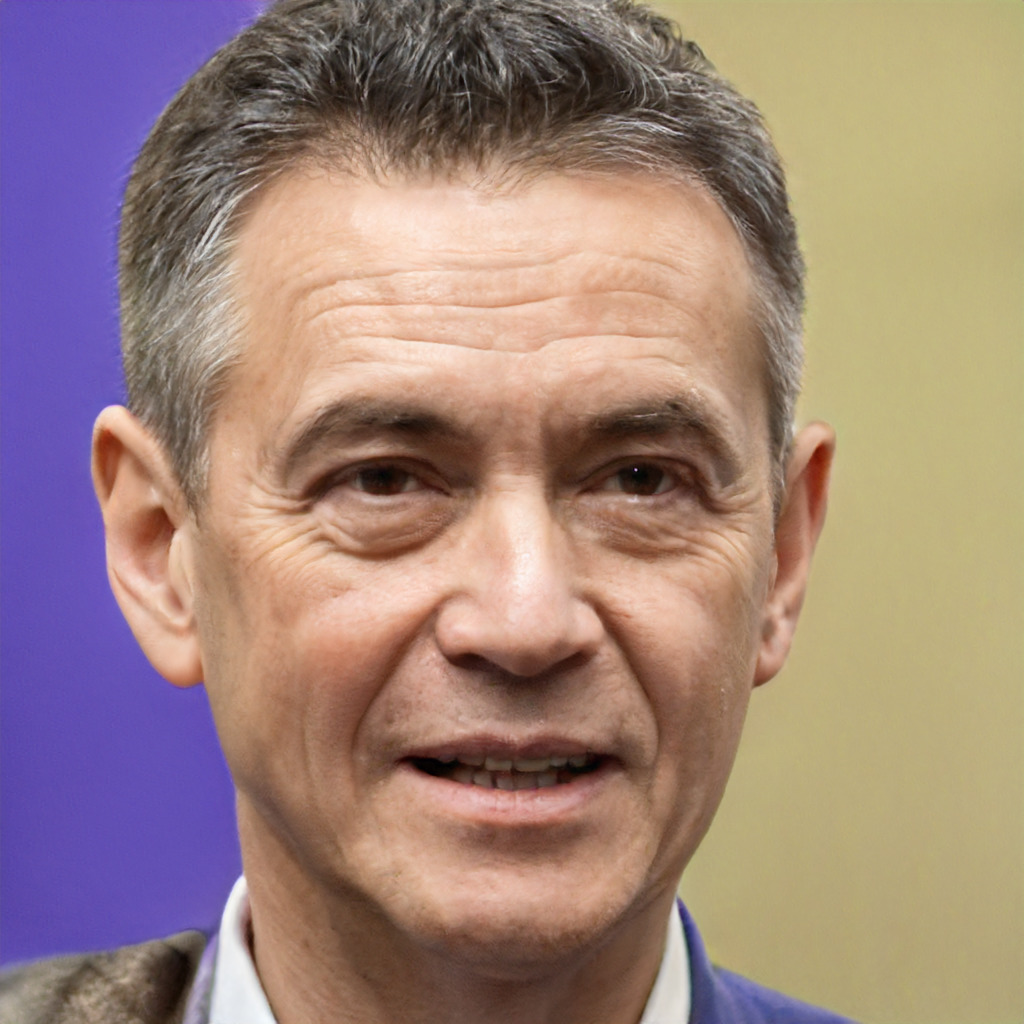} &
        \includegraphics[width=0.0875\textwidth]{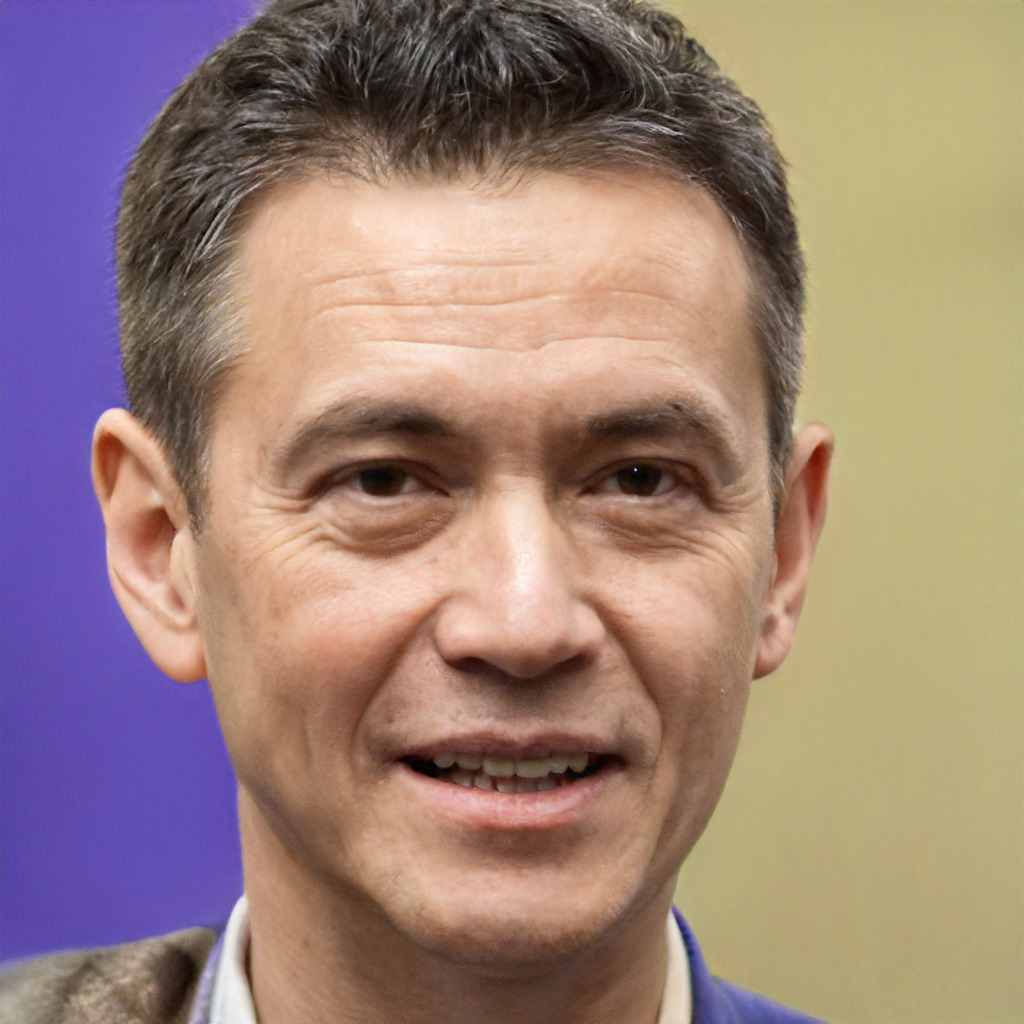} &
        \includegraphics[width=0.0875\textwidth]{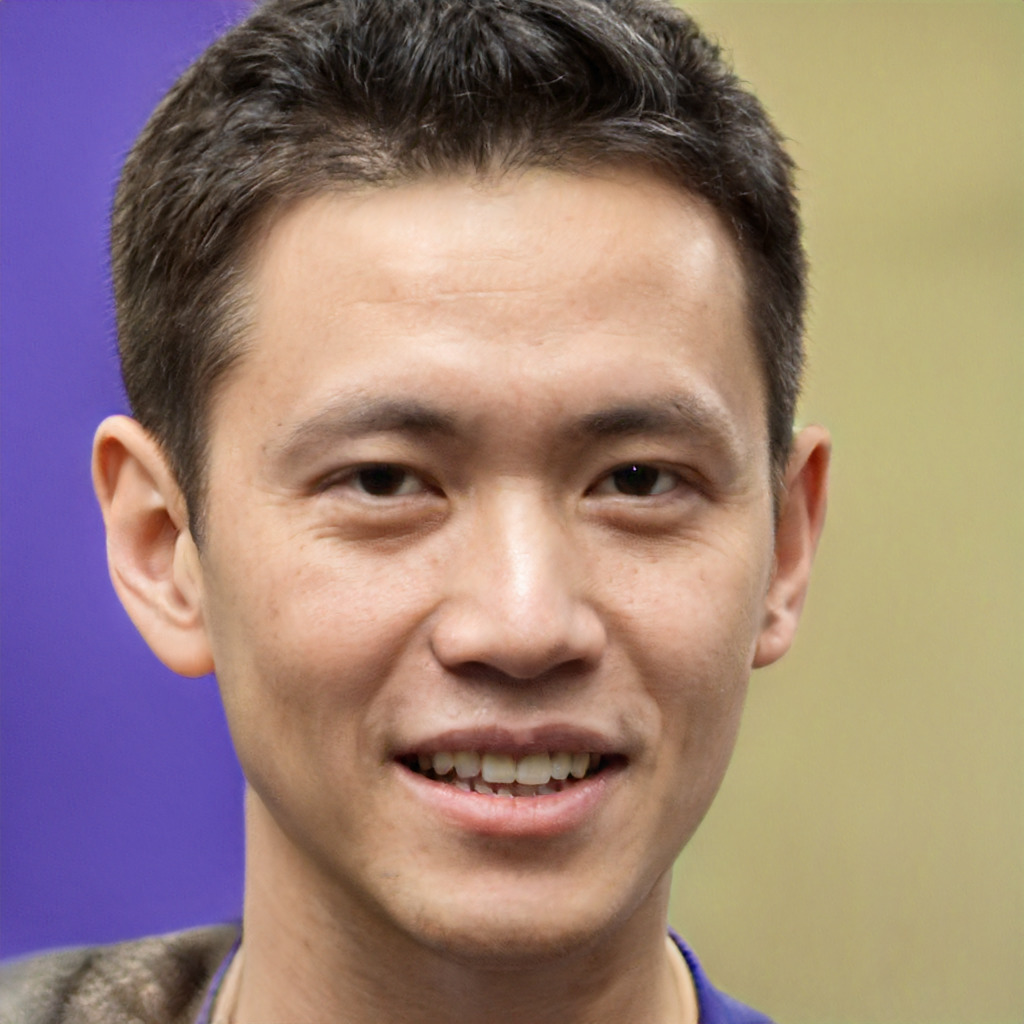} &
        \includegraphics[width=0.0875\textwidth]{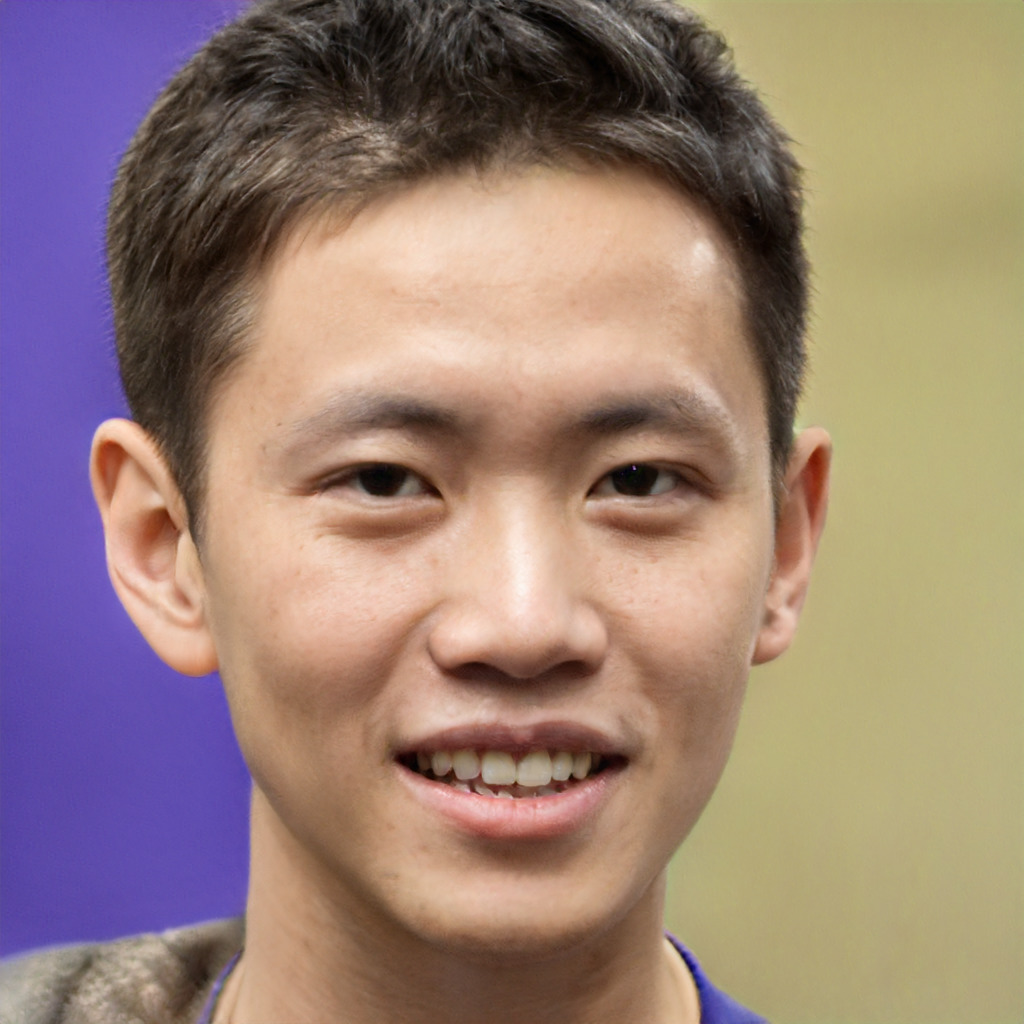} &
        \includegraphics[width=0.0875\textwidth]{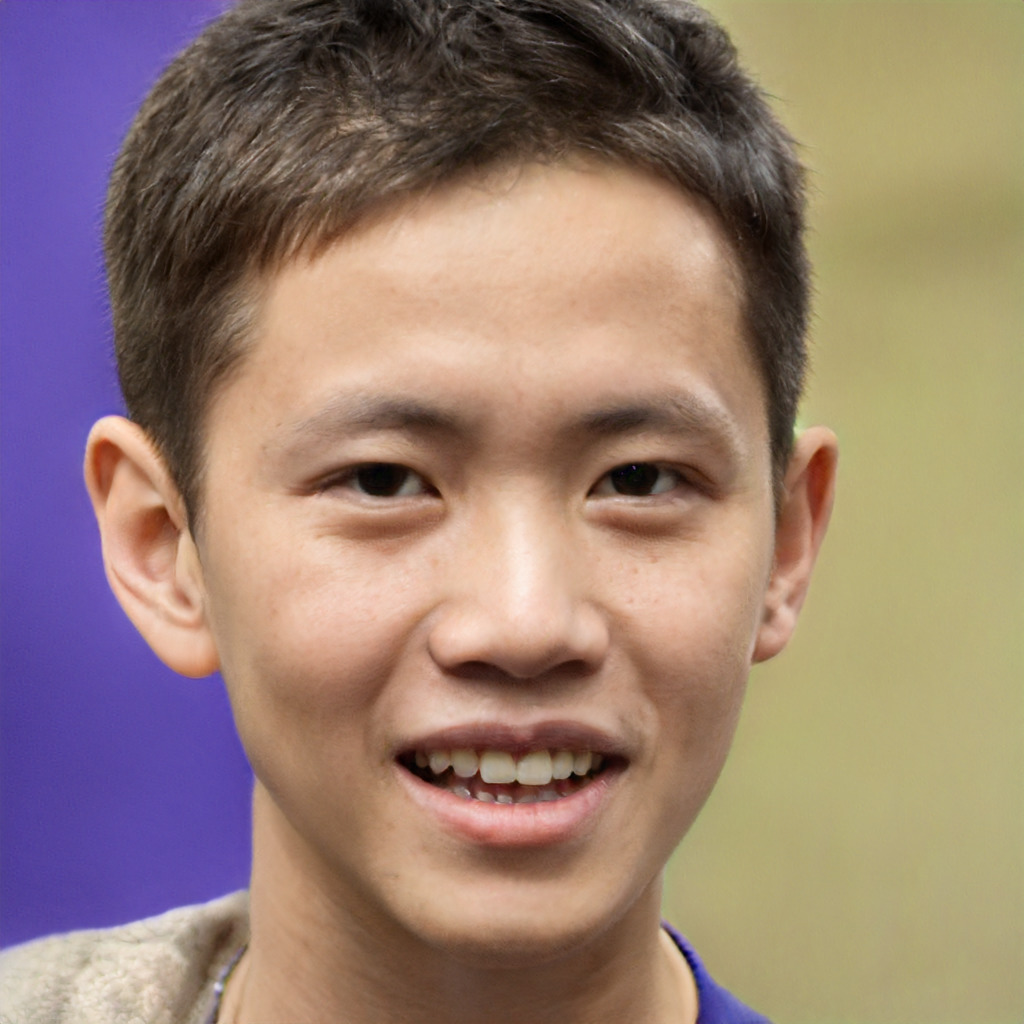} \\
        
        \raisebox{0.155in}{\rotatebox{90}{Ours}} &
        \includegraphics[width=0.0875\textwidth]{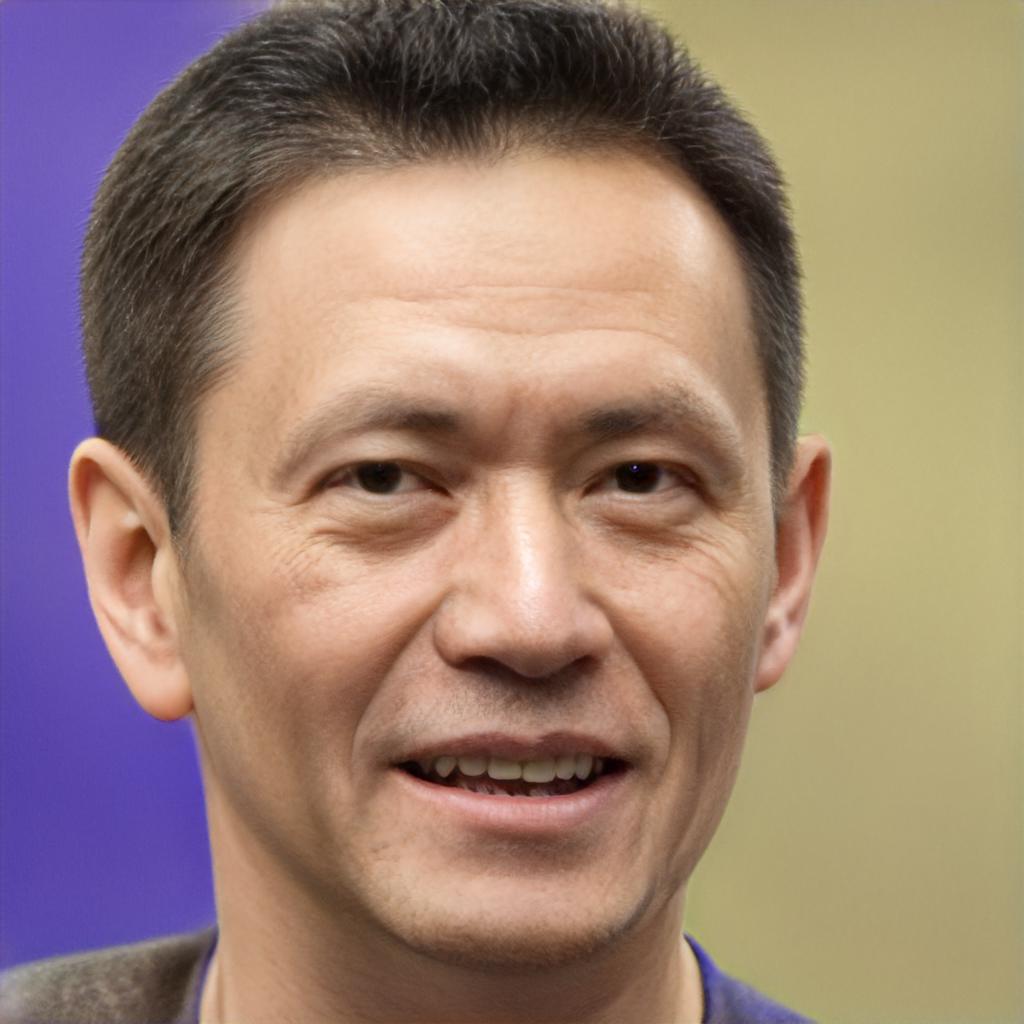} &
        \includegraphics[width=0.0875\textwidth]{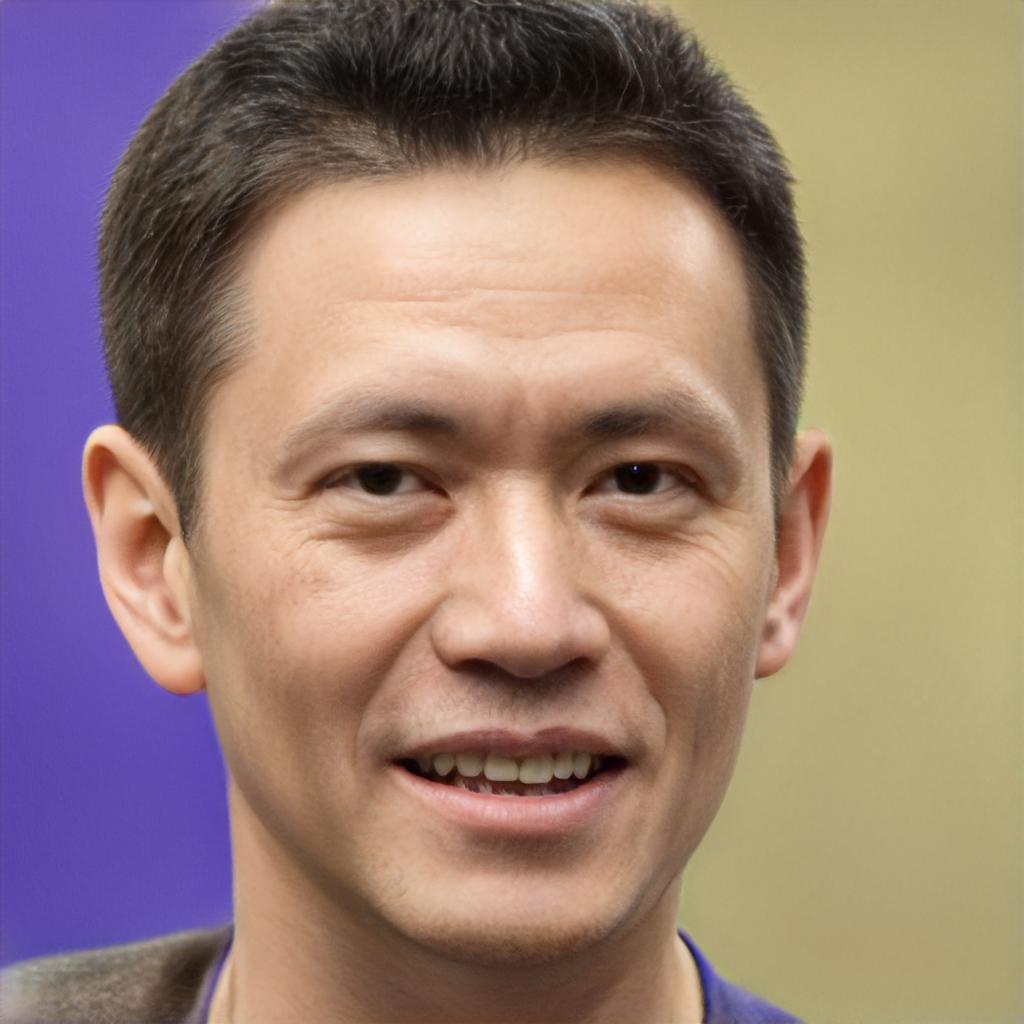} &
        \includegraphics[width=0.0875\textwidth]{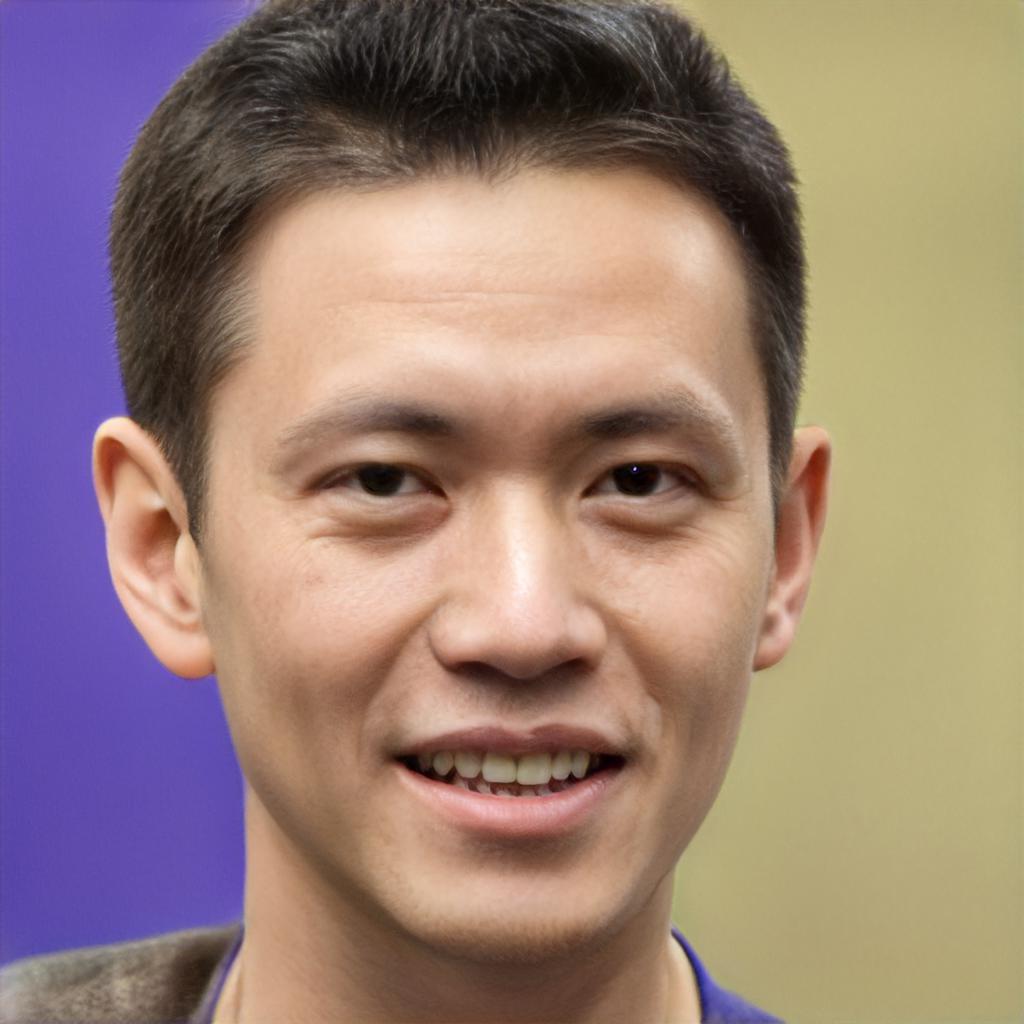} &
        \includegraphics[width=0.0875\textwidth]{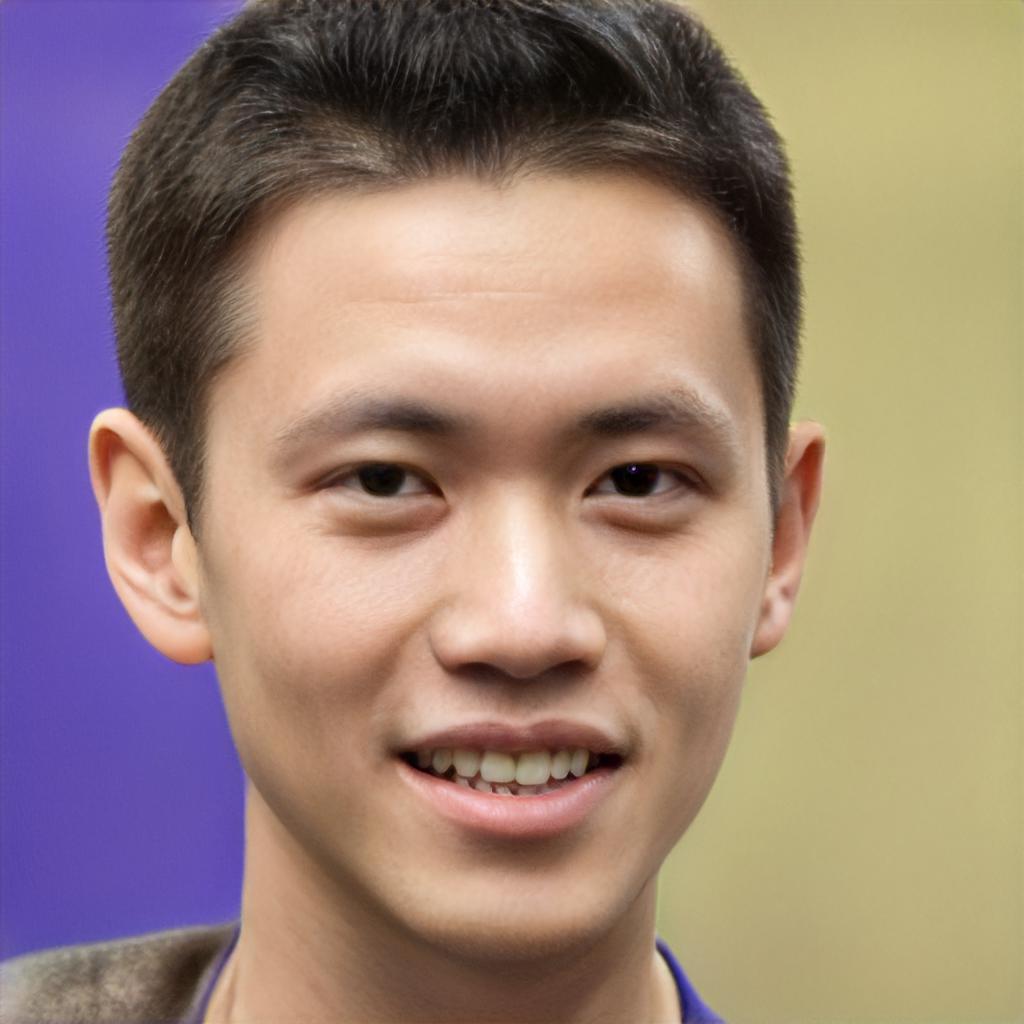} &
        \includegraphics[width=0.0875\textwidth]{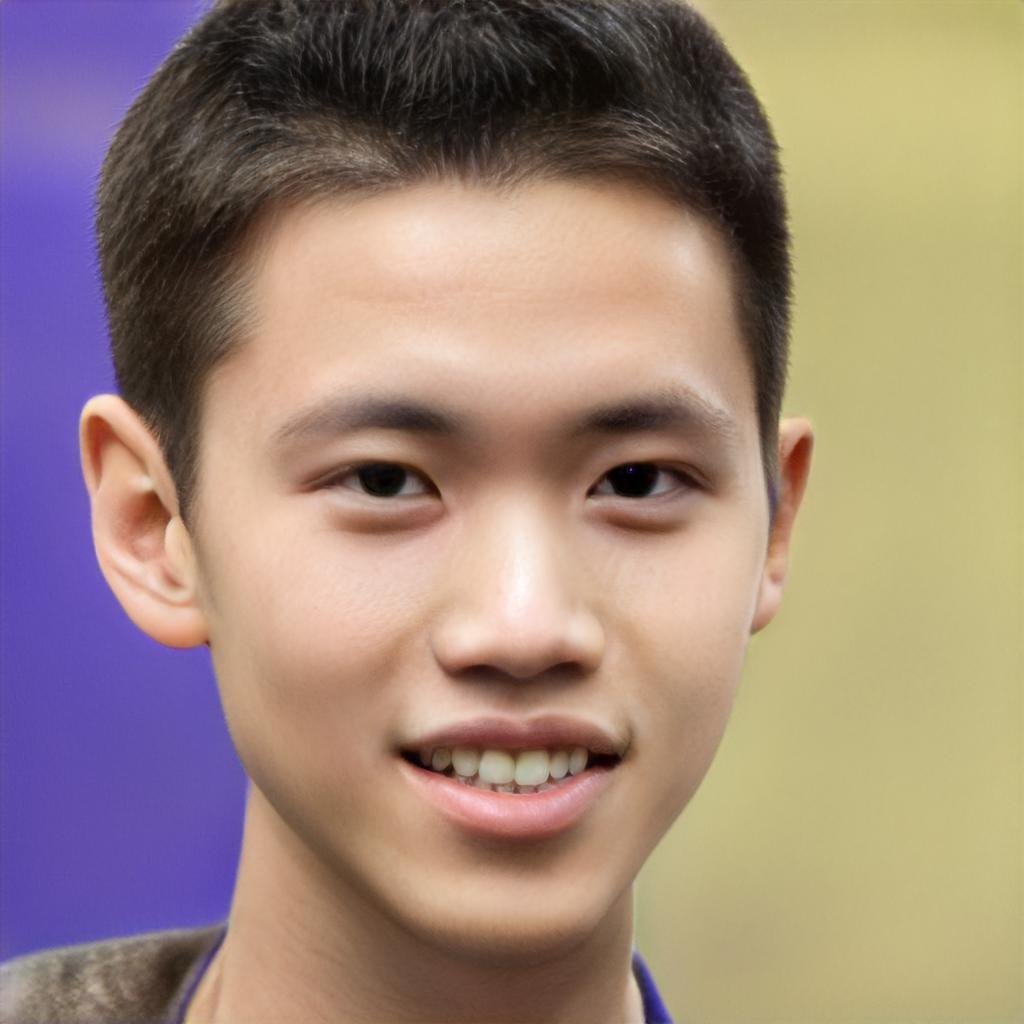} \\
        
        & \multicolumn{5}{c}{ Age }
        
    \end{tabular}
    
    &
    
        \begin{tabular}{c c c c c c}

        \raisebox{0.08in}{\rotatebox{90}{\footnotesize StyleFlow}} &
        \includegraphics[width=0.0875\textwidth]{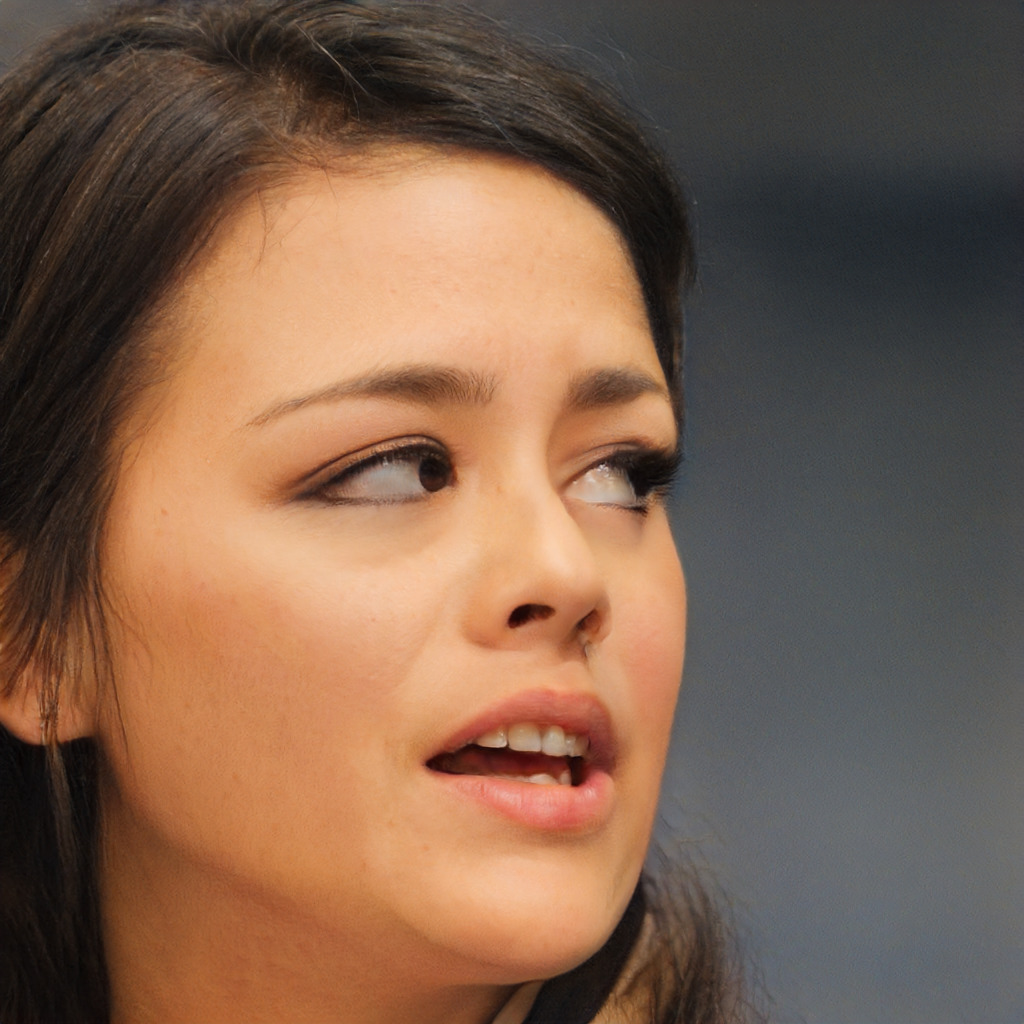} &
        \includegraphics[width=0.0875\textwidth]{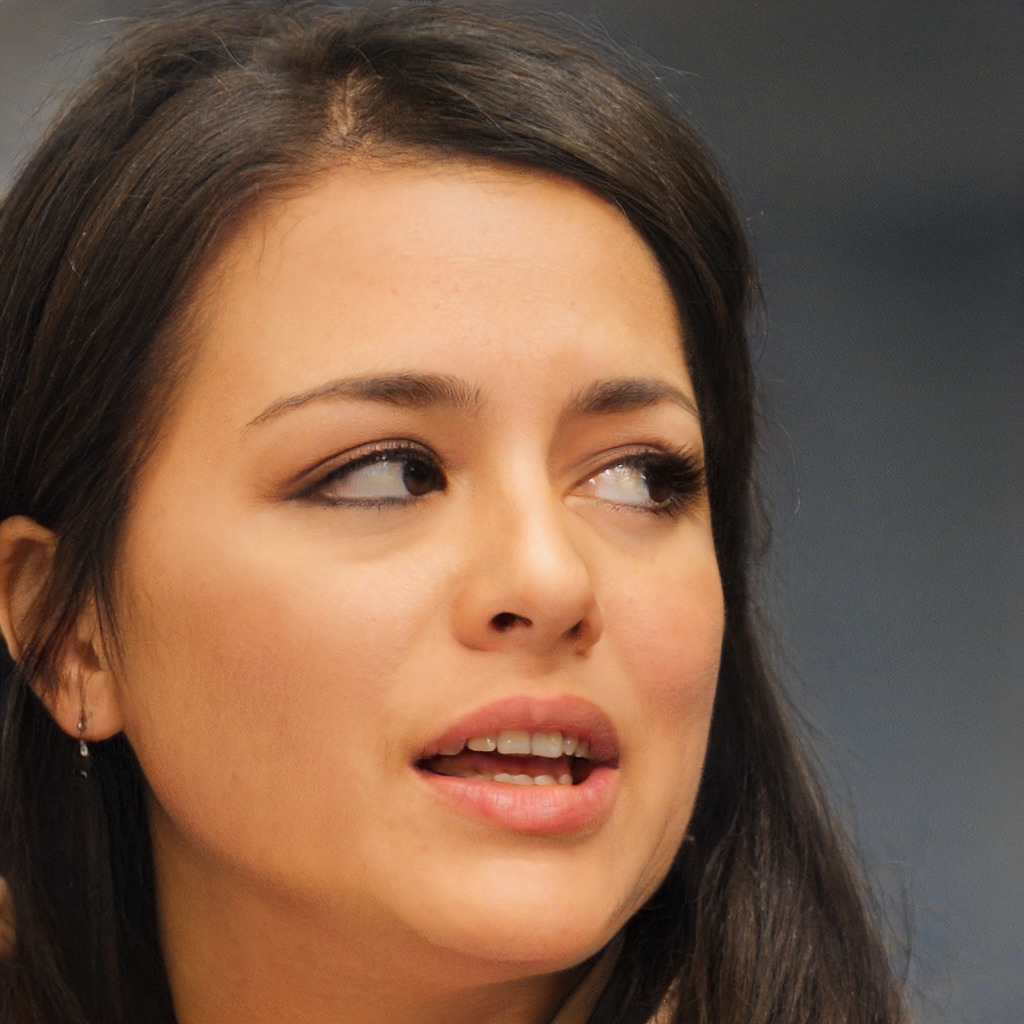} &
        \includegraphics[width=0.0875\textwidth]{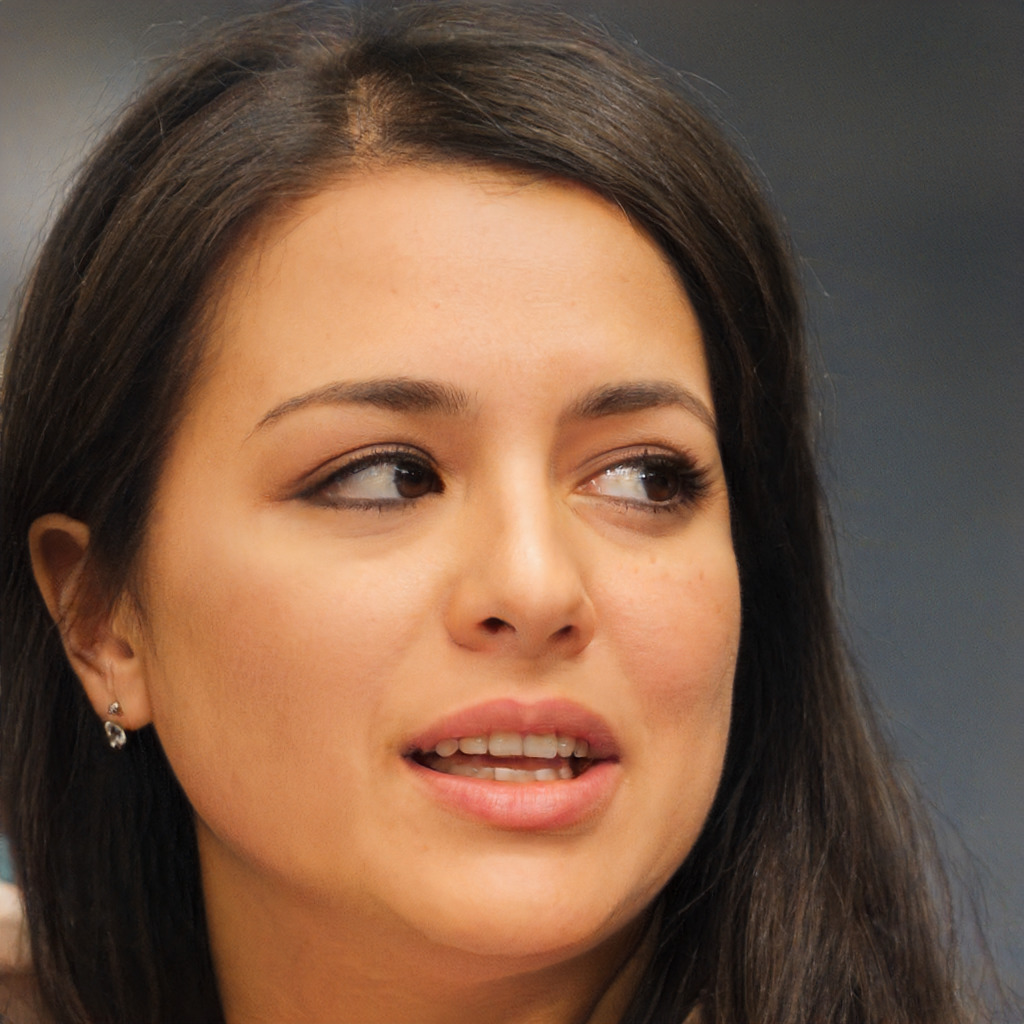} &
        \includegraphics[width=0.0875\textwidth]{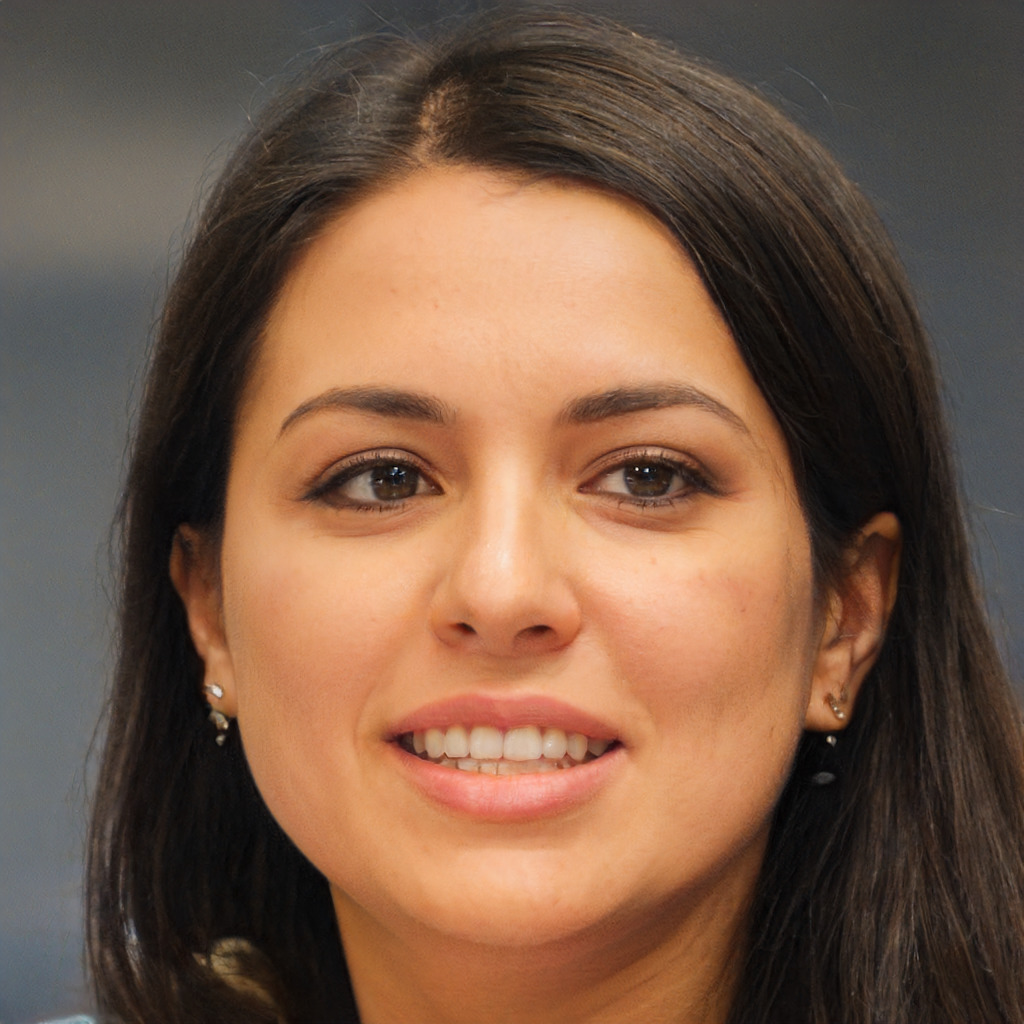} &
        \includegraphics[width=0.0875\textwidth]{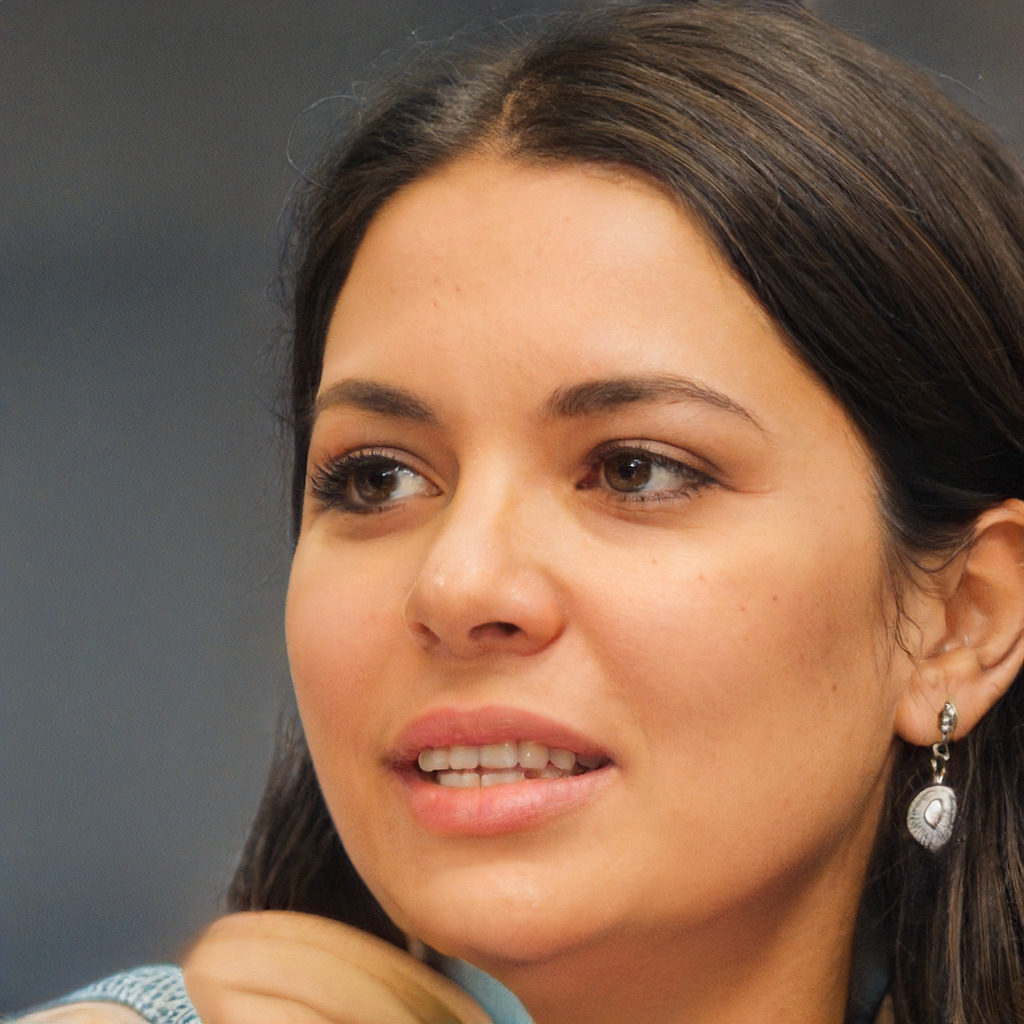} \\
        
        \raisebox{0.155in}{\rotatebox{90}{Ours}} &
        \includegraphics[width=0.0875\textwidth]{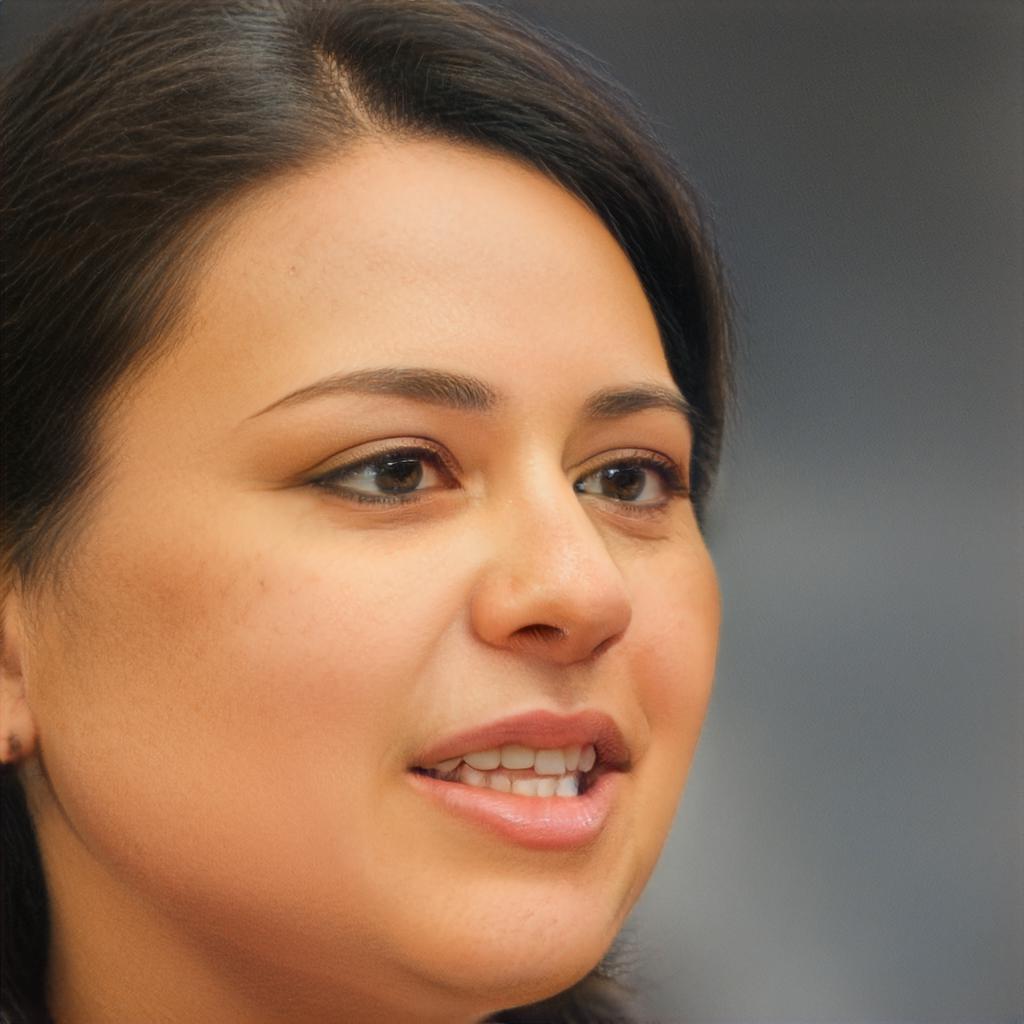} &
        \includegraphics[width=0.0875\textwidth]{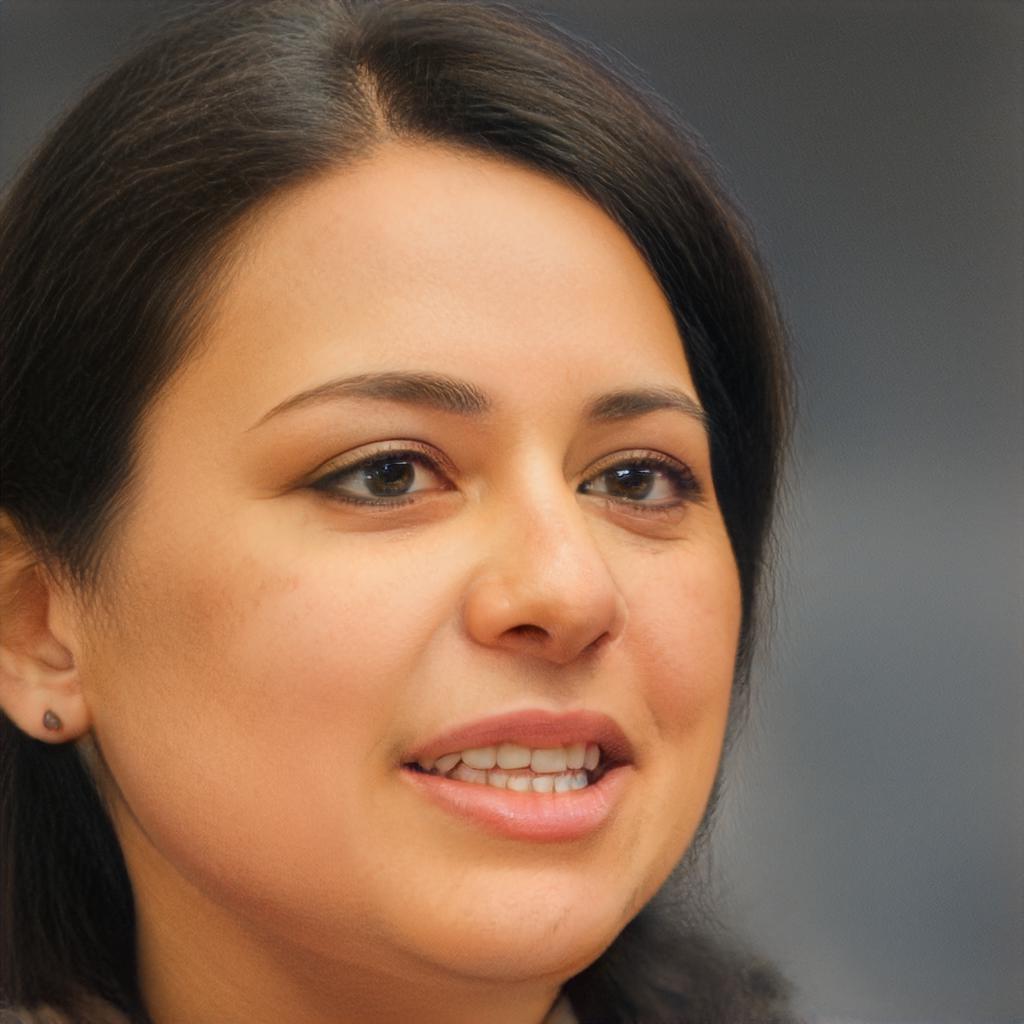} &
        \includegraphics[width=0.0875\textwidth]{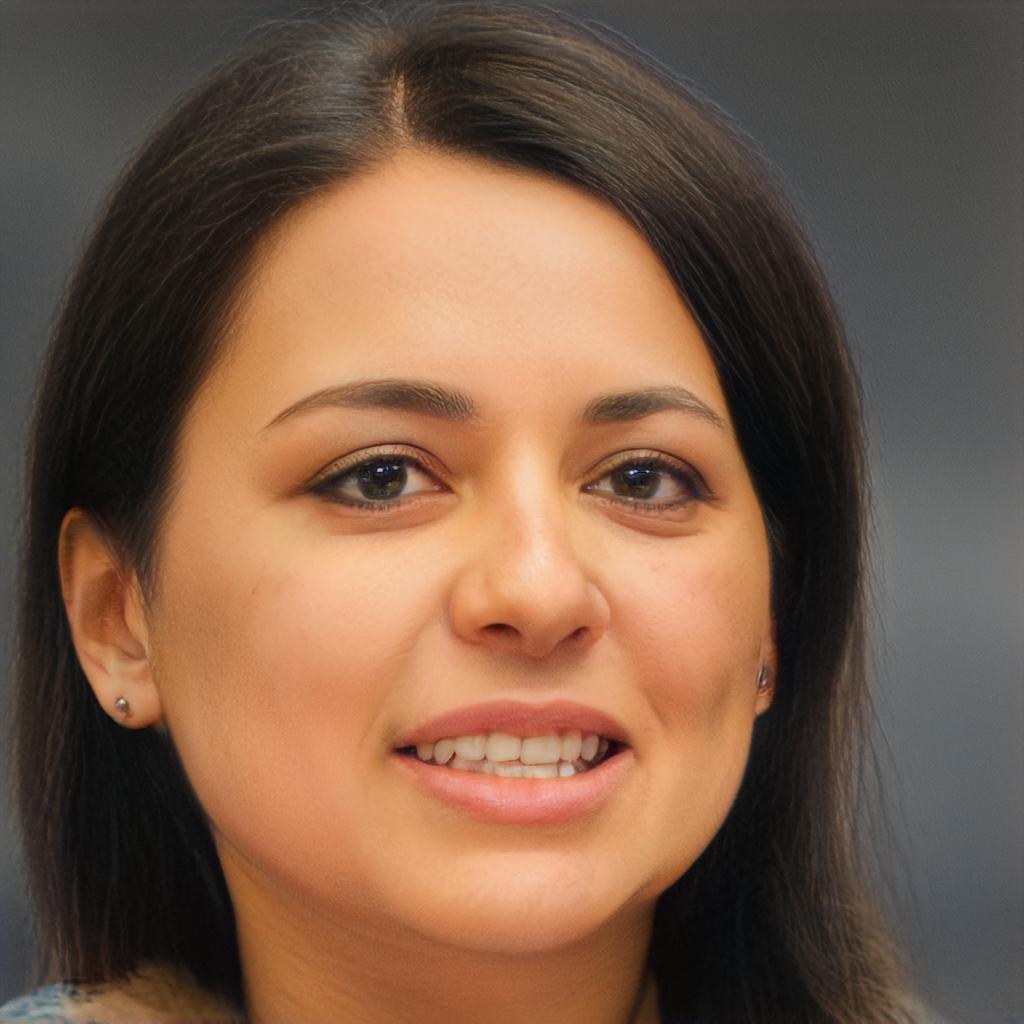} &
        \includegraphics[width=0.0875\textwidth]{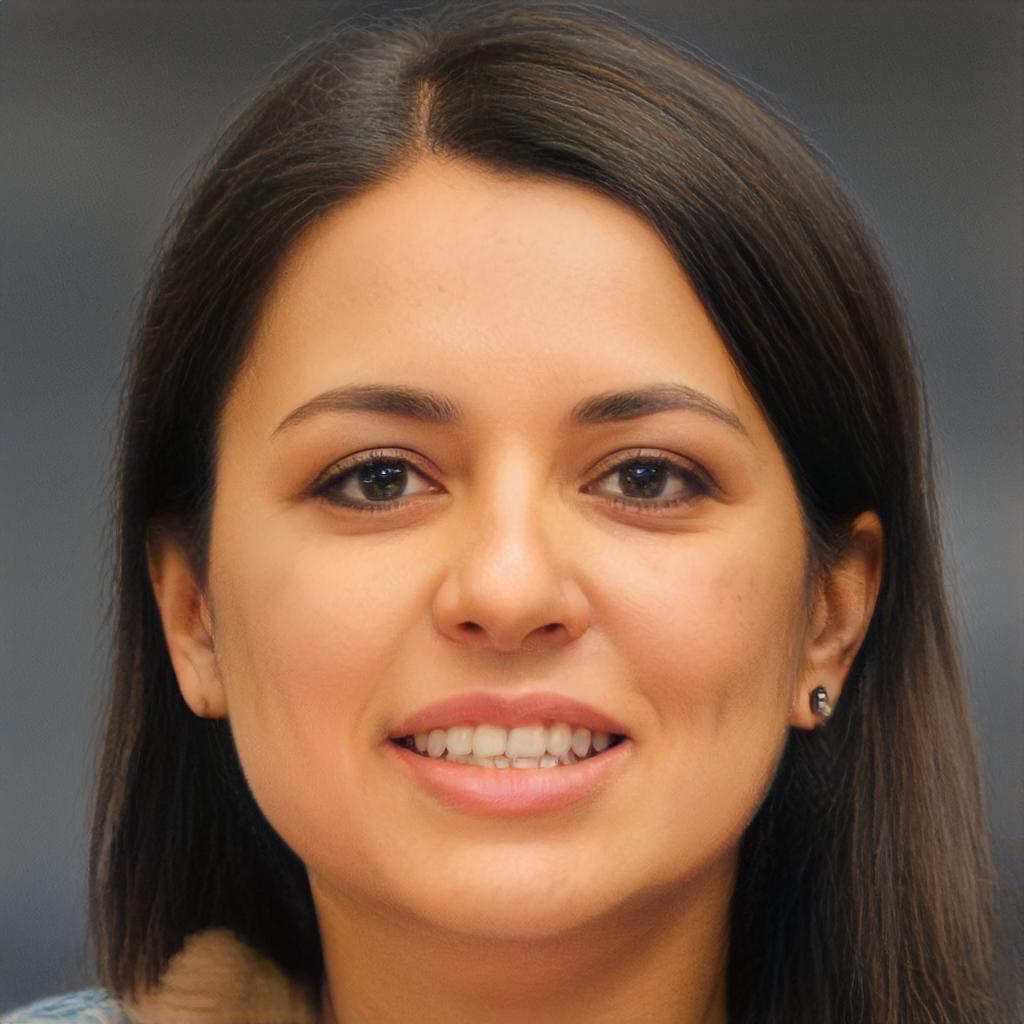} &
        \includegraphics[width=0.0875\textwidth]{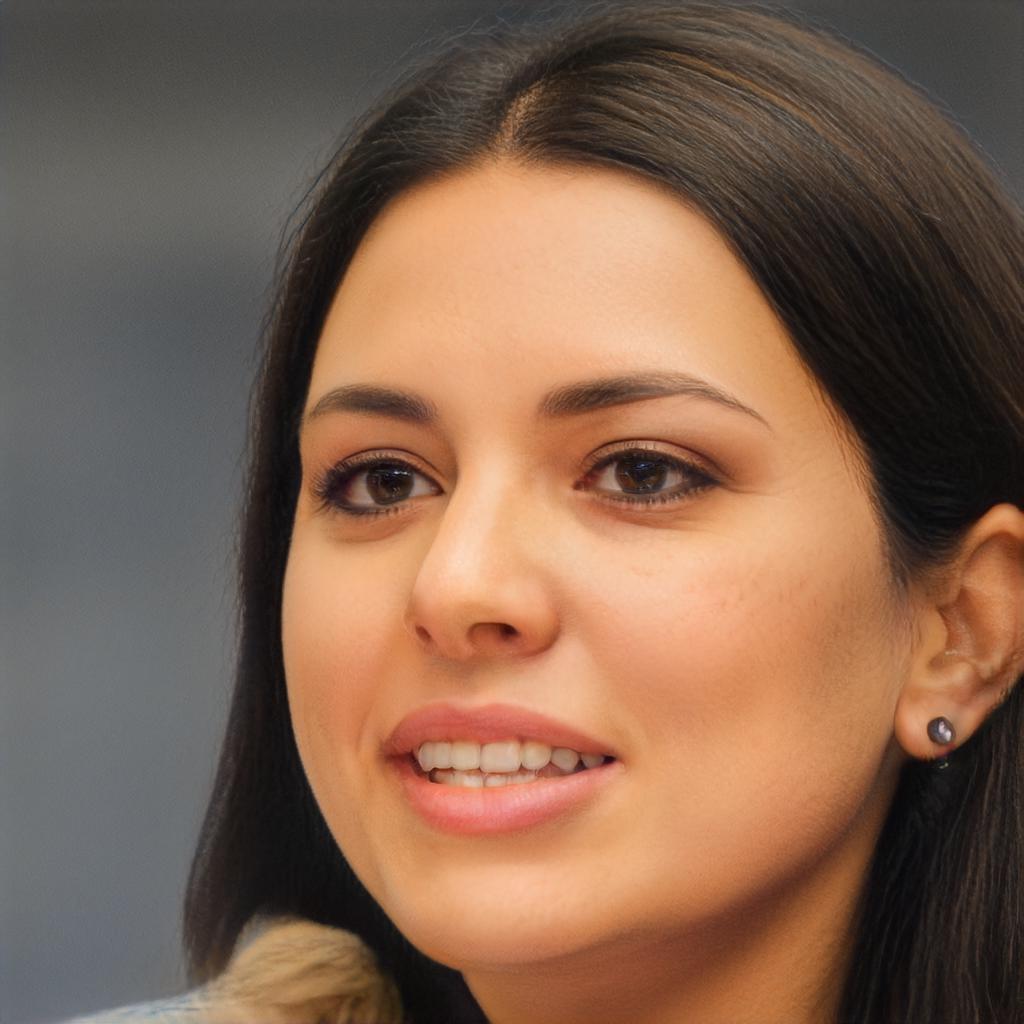} \\
        
        & \multicolumn{5}{c}{ Pose }
        
    \end{tabular}

    \end{tabular}
    
    }
    \vspace{-0.27cm}
    \caption{Comparisons to non-linear editing methods. Our model maintains better identity when adjusting age, particularly in the case of racial minorities. When modifying the pose, our model shows reduced corruption for large changes.}
    \label{fig:non_linear_comparisons}
\end{figure*}

%% file: resources/figures/id_comp.tex
\begin{figure}[!hbt]
    \centering
    \setlength{\belowcaptionskip}{-2.5pt}
    \setlength{\tabcolsep}{1pt}
    {
    \begin{tabular}{c c}
    
        \includegraphics[width=0.48\linewidth]{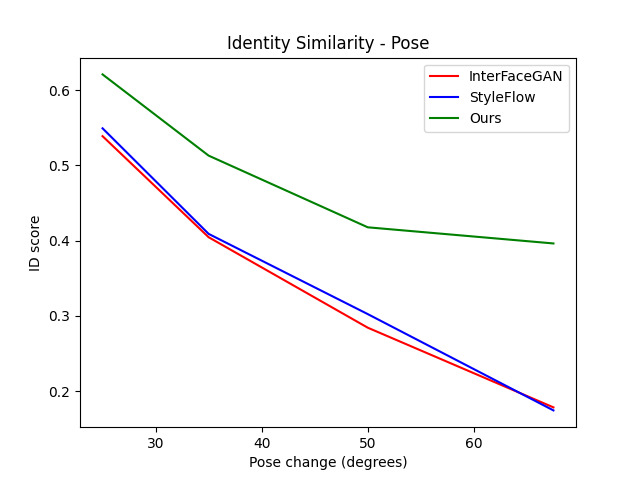} &
        \includegraphics[width=0.48\linewidth]{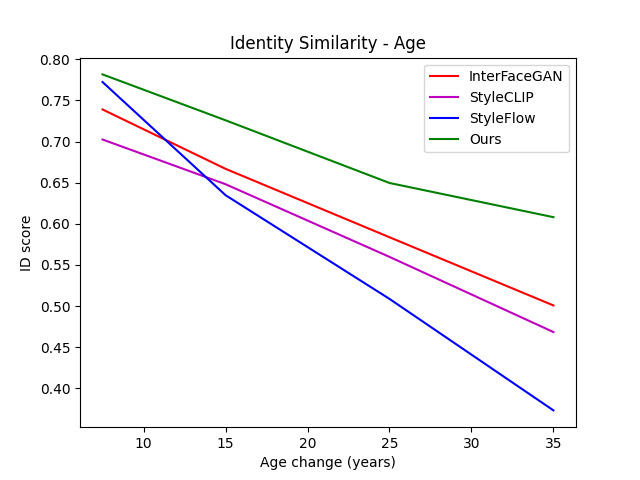} \\
        
         \includegraphics[width=0.48\linewidth]{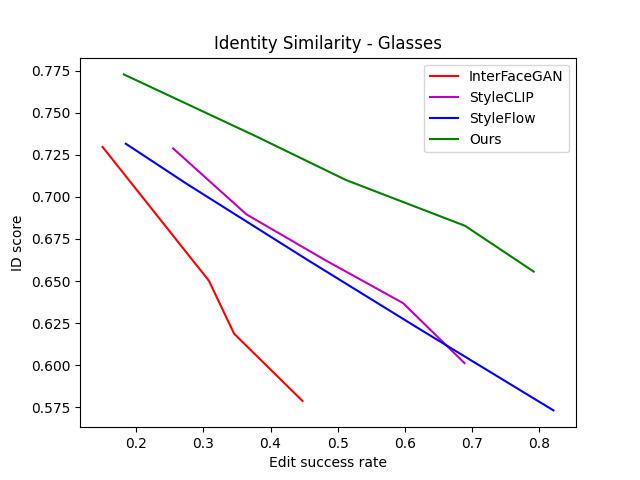} &
        \includegraphics[width=0.48\linewidth]{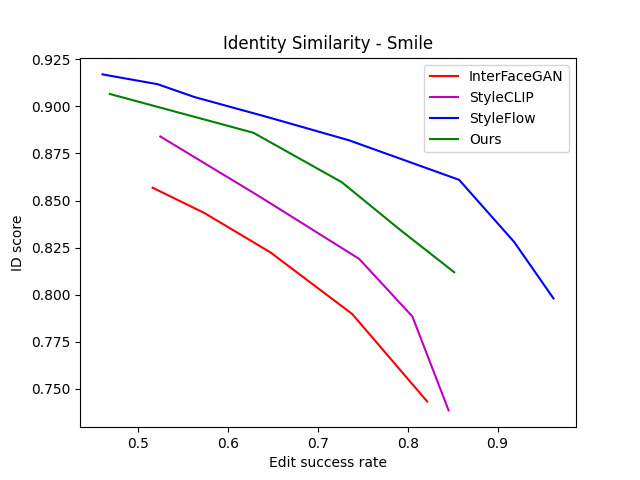}
        
    \end{tabular}}
    \vspace{-4pt}
    \caption{Identity preservation graphs. Our model maintains better identity for meaningful changes, and shows more moderate decline as we edit towards regions where training data was scarce. For non-minority attributes like smile, our model achieves comparable performance to non-linear editing methods, and outperforms the linear editing directions from which our self-labels were drawn.}
    \label{fig:id_comps} \vspace{-5pt}
\end{figure}

%% file: resources/figures/ablation.tex
\begin{figure}[!hbt]
    \centering
    \setlength{\belowcaptionskip}{-2.5pt}
    \setlength{\tabcolsep}{1pt}
    {
    
    \begin{tabular}{c c}
    
    \begin{tabular}{c c c c}
    
        \raisebox{0.105in}{\rotatebox{90}{Ours}} &
        \includegraphics[width=0.135\linewidth]{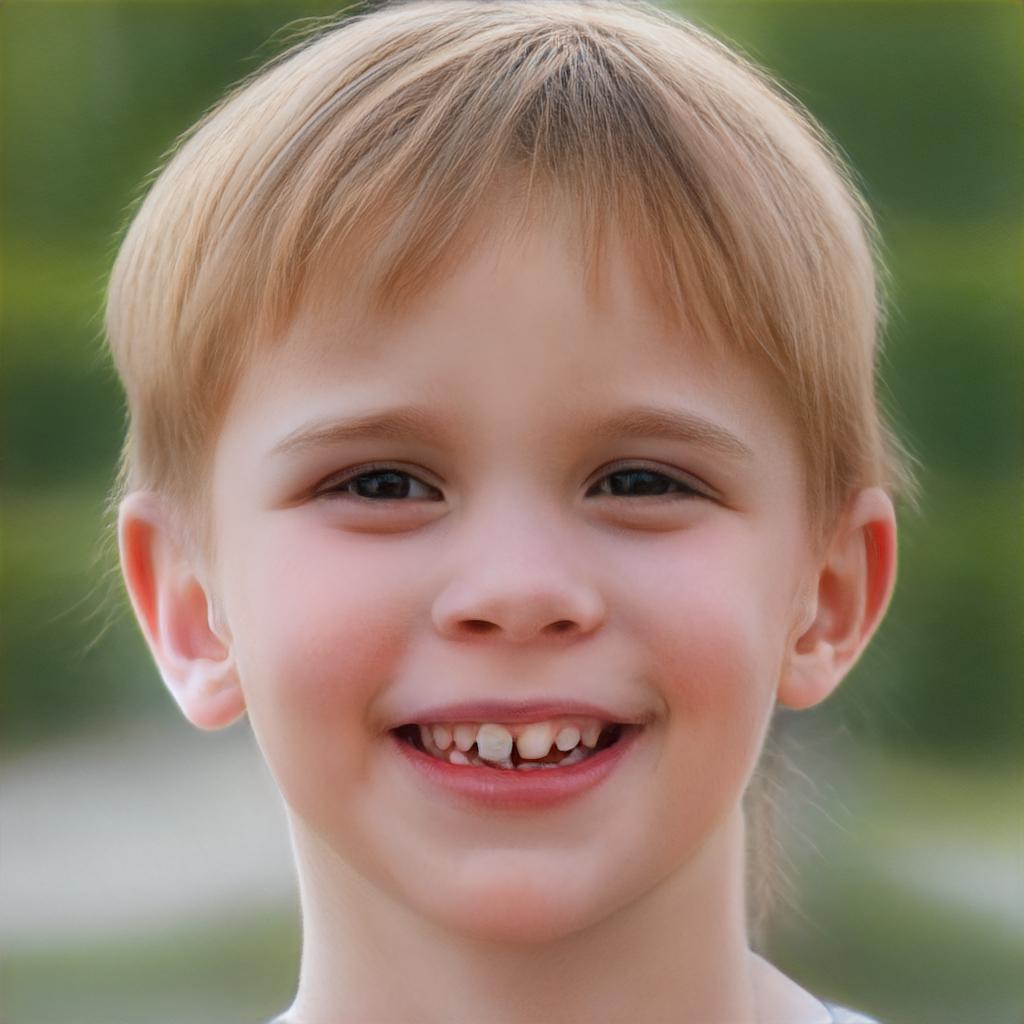} &
        \includegraphics[width=0.135\linewidth]{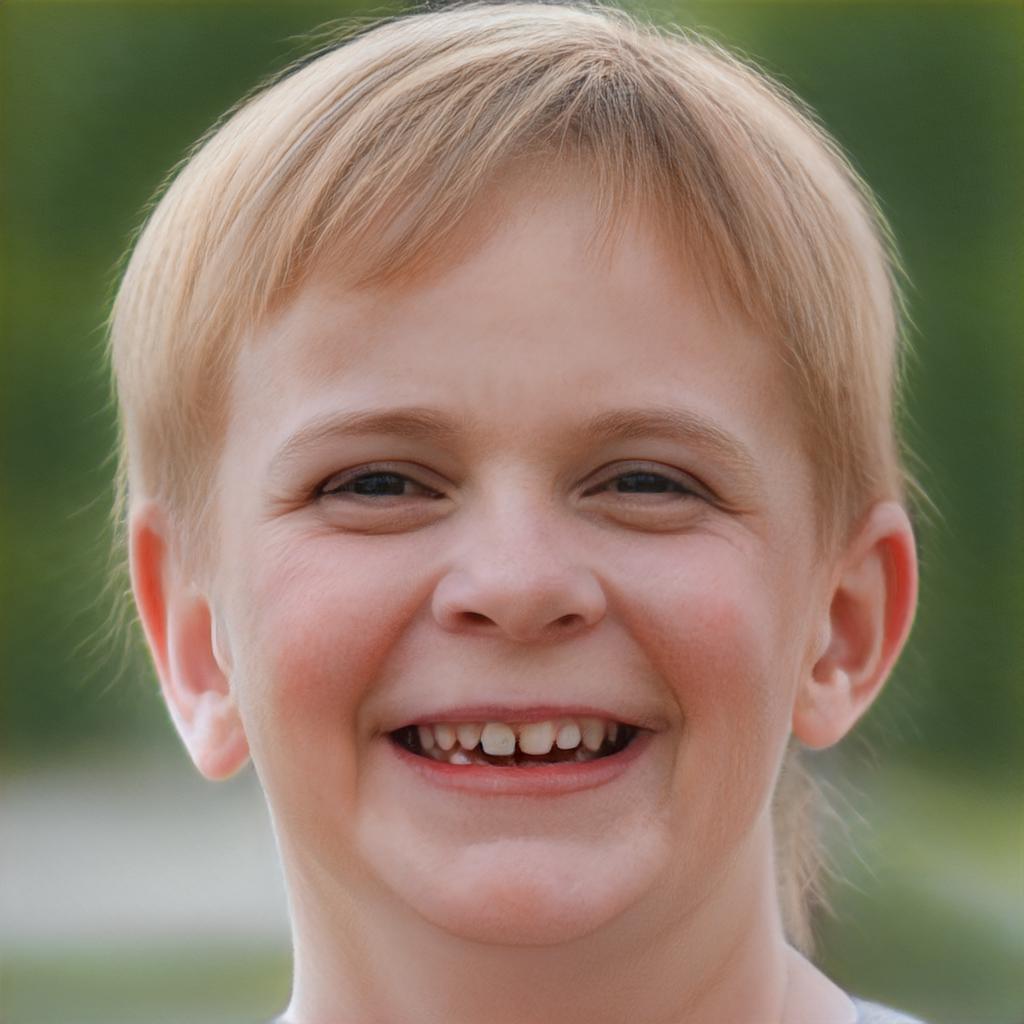} &
        \includegraphics[width=0.135\linewidth]{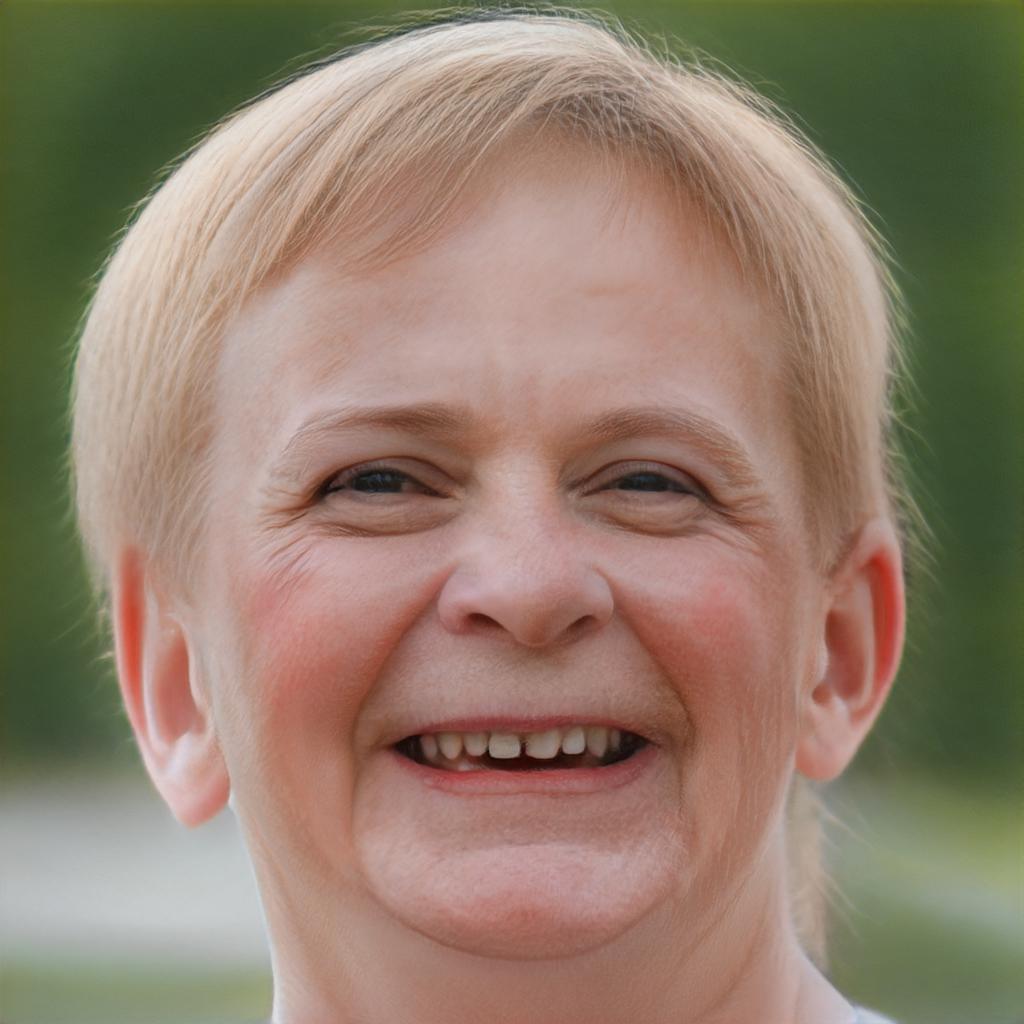} \\
        
        \raisebox{0.055in}{\rotatebox{90}{\tiny\begin{tabular}{c@{}c@{}}Conditional \\ Mapping \end{tabular}}} & 
        \includegraphics[width=0.135\linewidth]{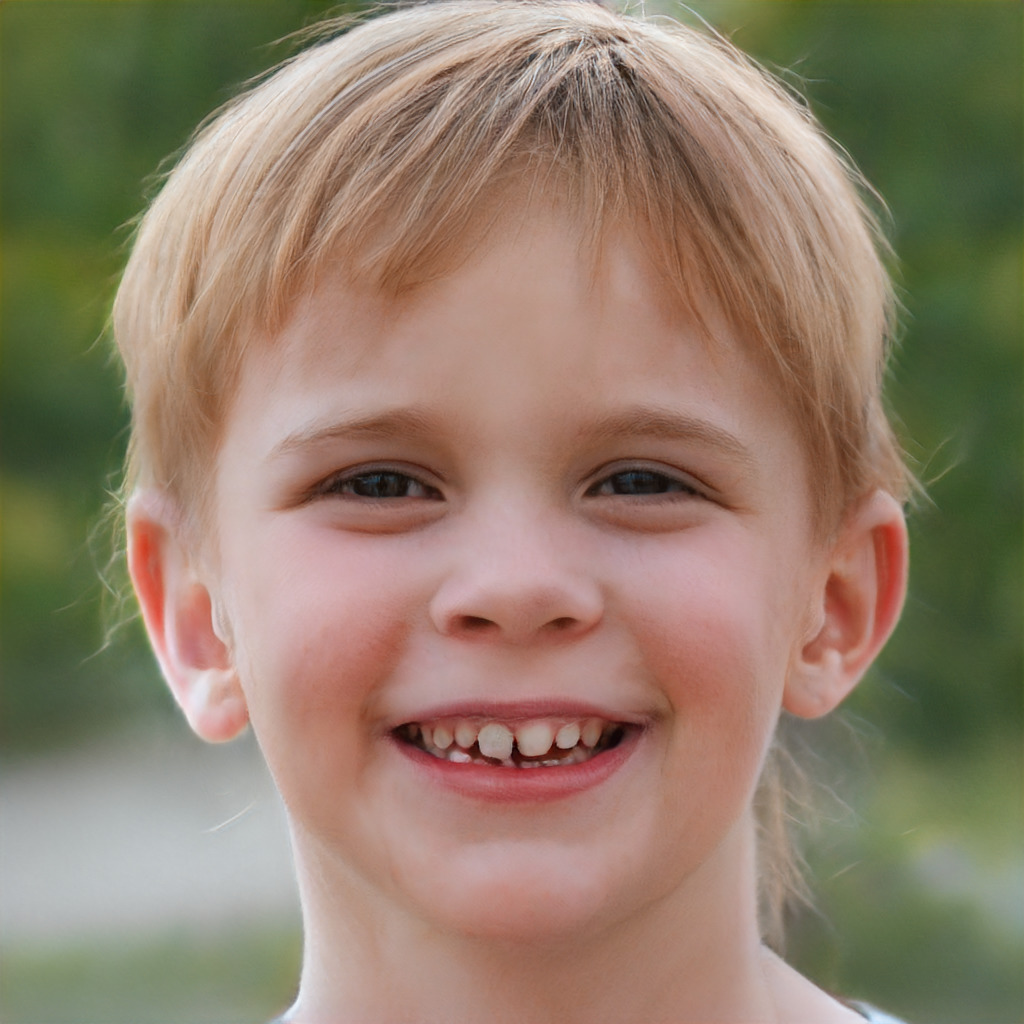} &
        \includegraphics[width=0.135\linewidth]{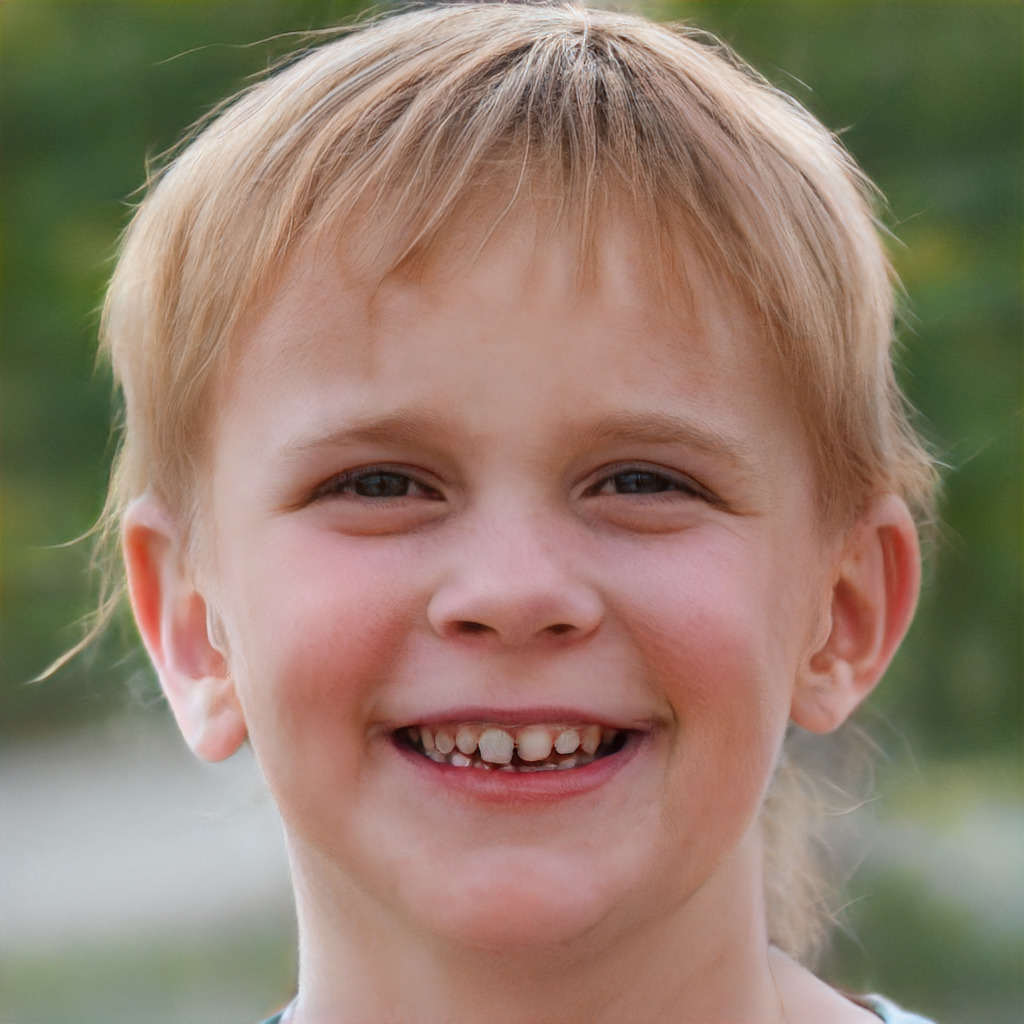} &
        \includegraphics[width=0.135\linewidth]{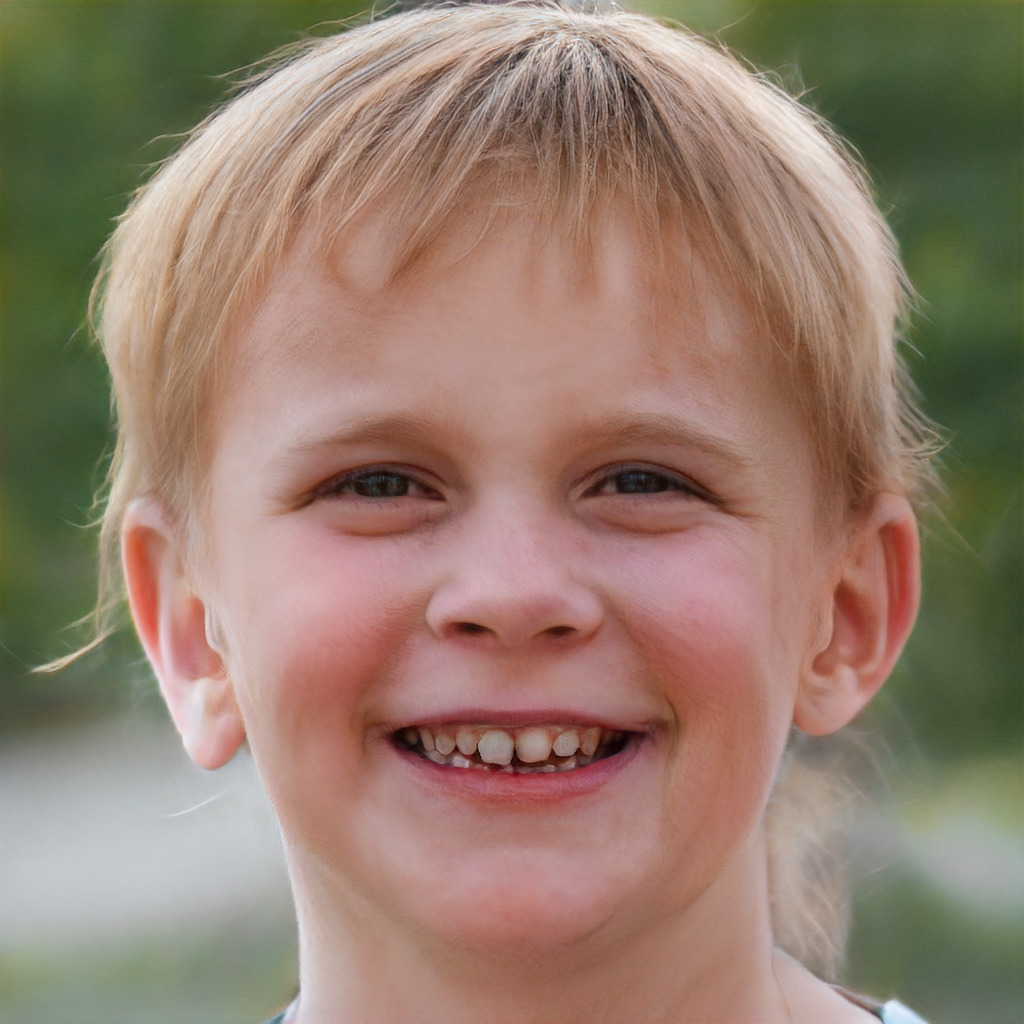} \\

        \raisebox{0.105in}{\rotatebox{90}{Ours}} &
        \includegraphics[width=0.135\linewidth]{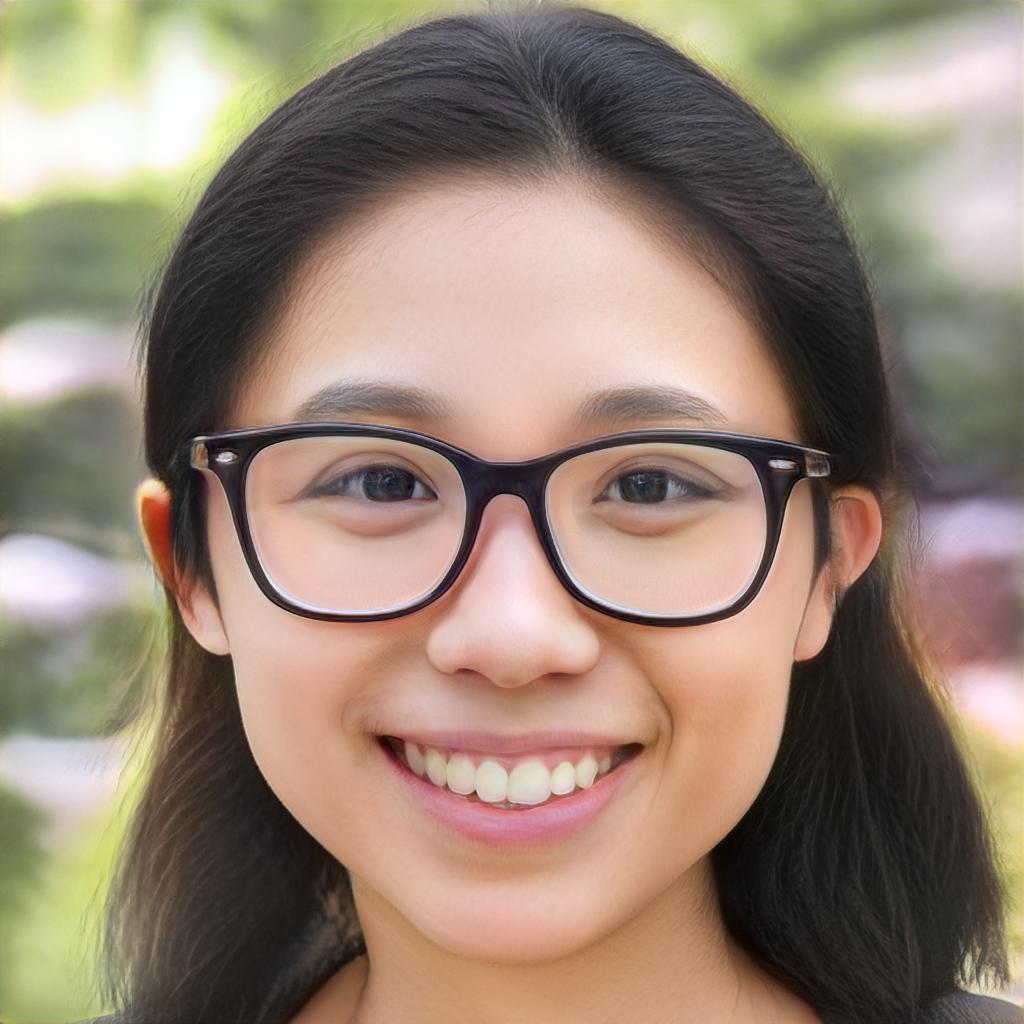} &
        \includegraphics[width=0.135\linewidth]{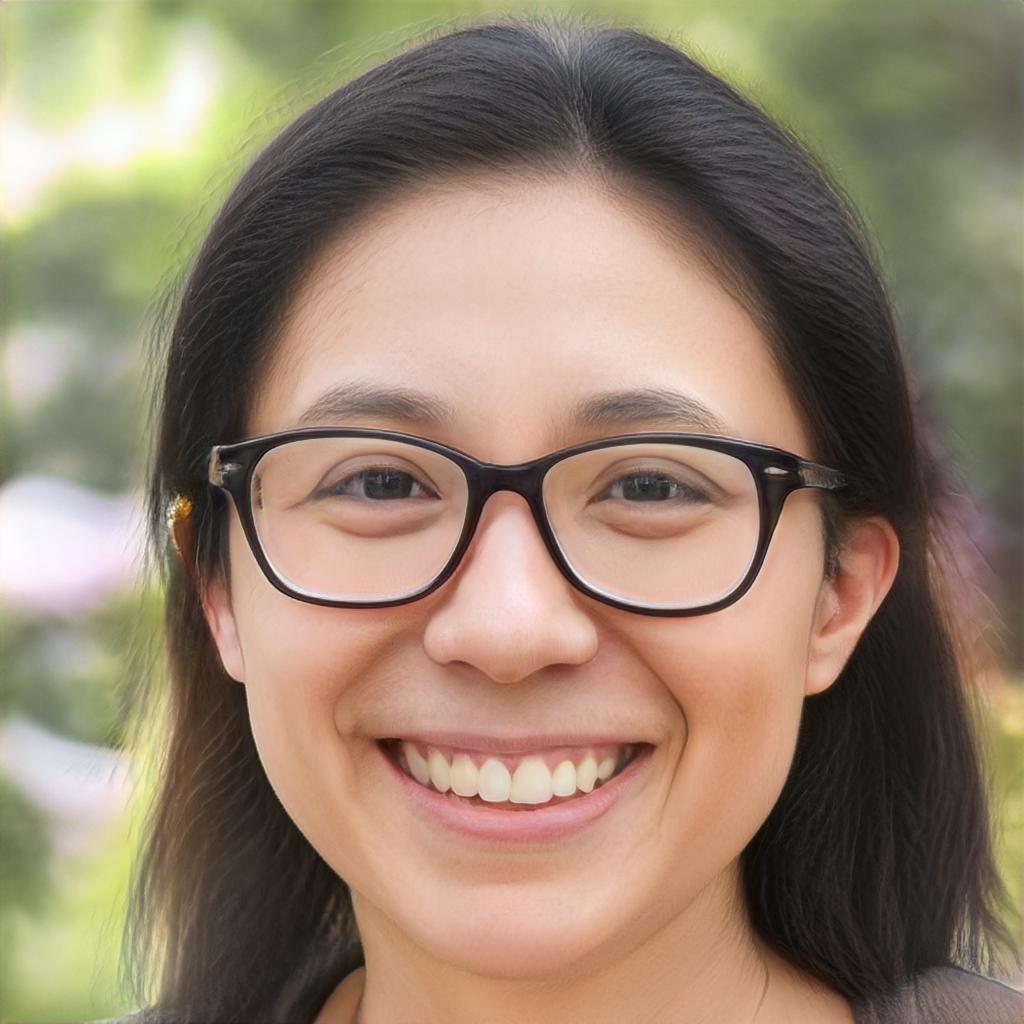} &
        \includegraphics[width=0.135\linewidth]{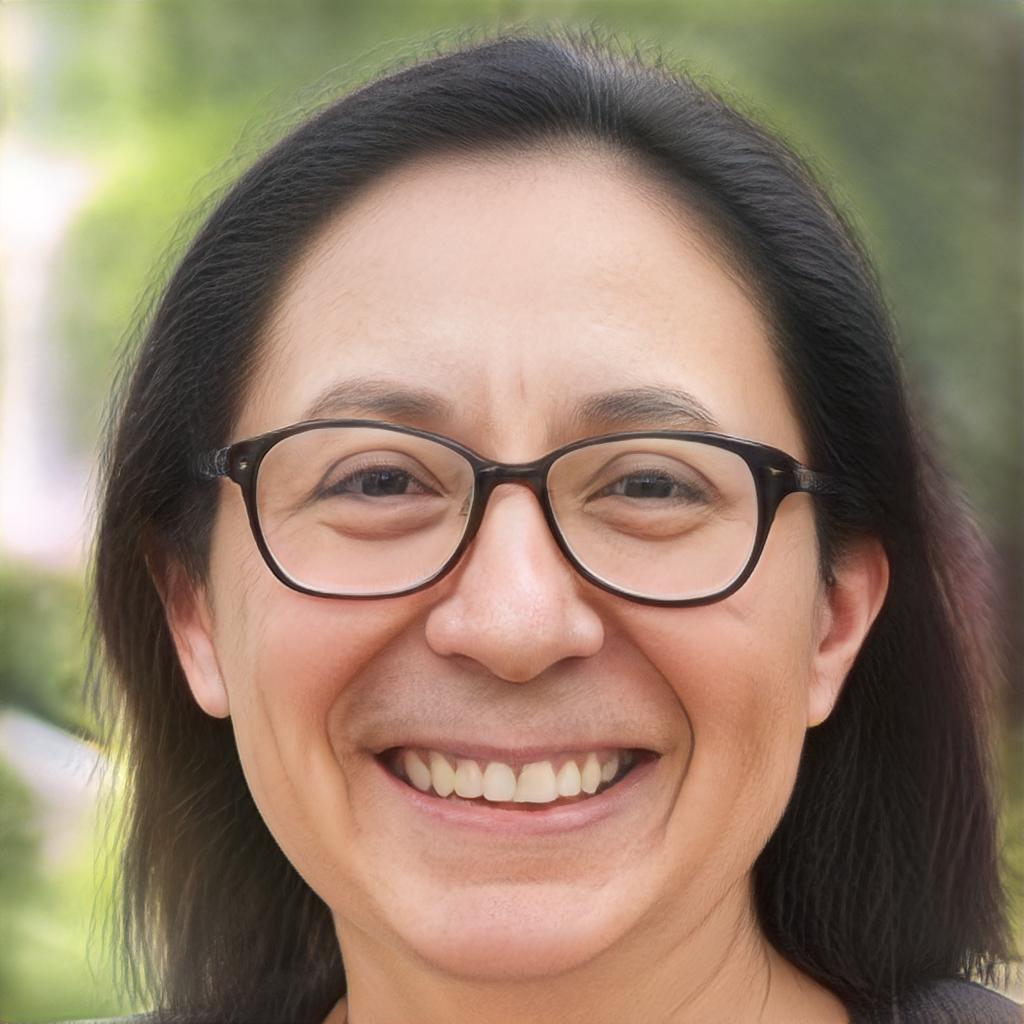} \\
        
        \raisebox{0.075in}{\rotatebox{90}{\tiny\begin{tabular}{c@{}c@{}}Classifier \\ Labels \end{tabular}}} & 
        \includegraphics[width=0.135\linewidth]{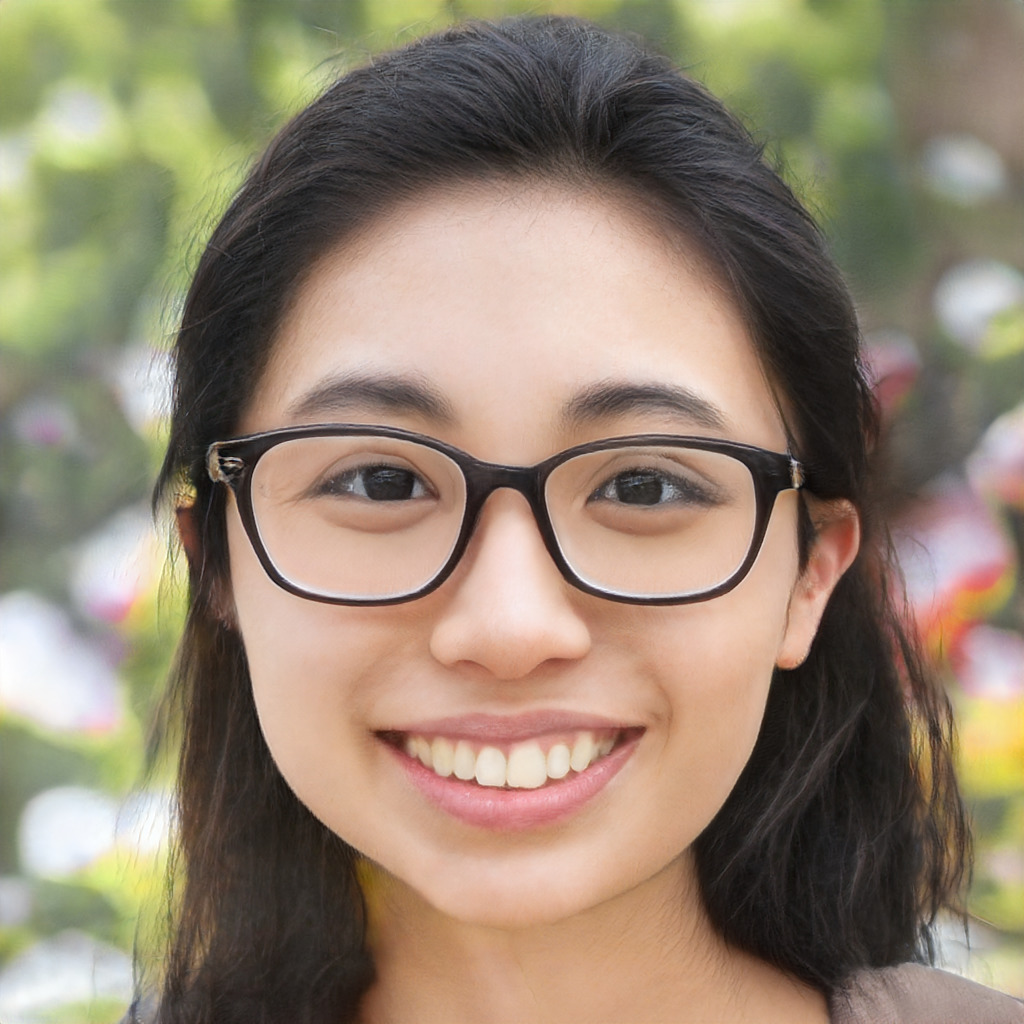} &
        \includegraphics[width=0.135\linewidth]{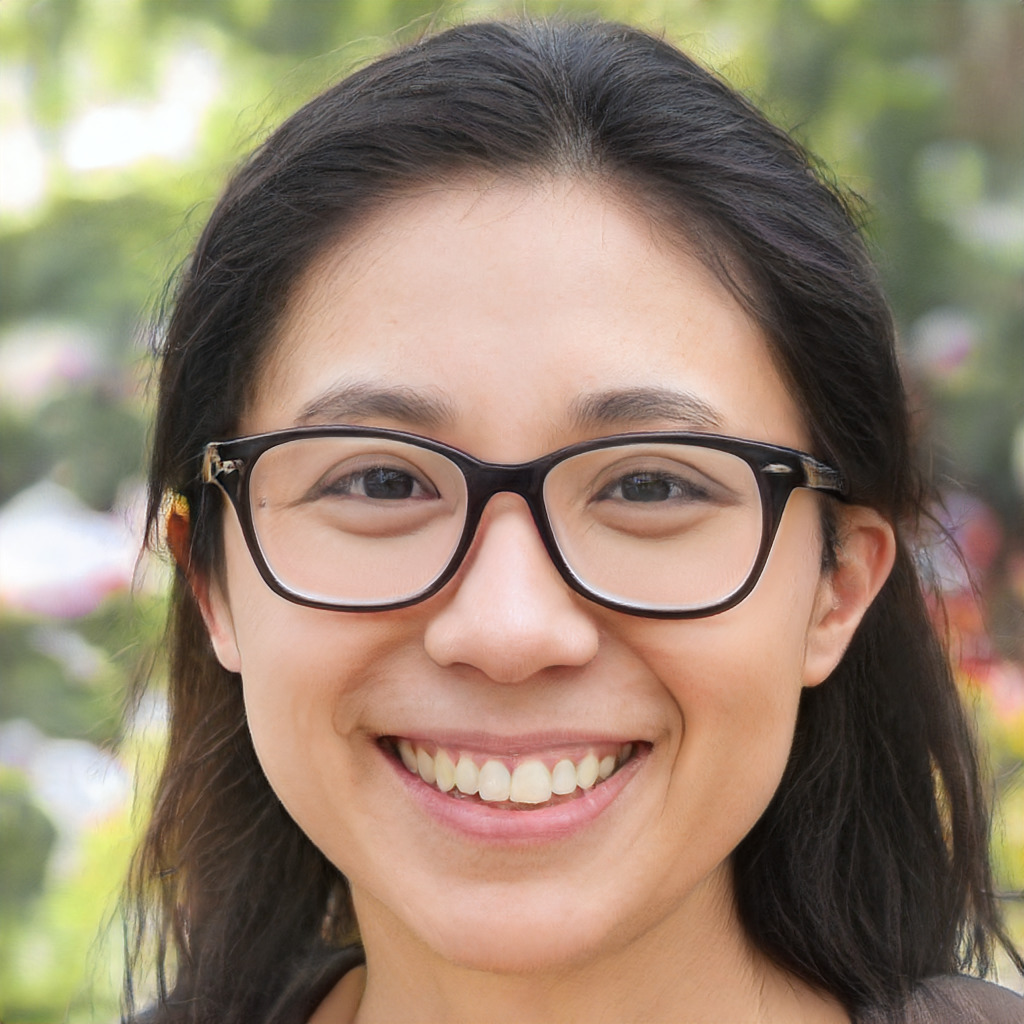} &
        \includegraphics[width=0.135\linewidth]{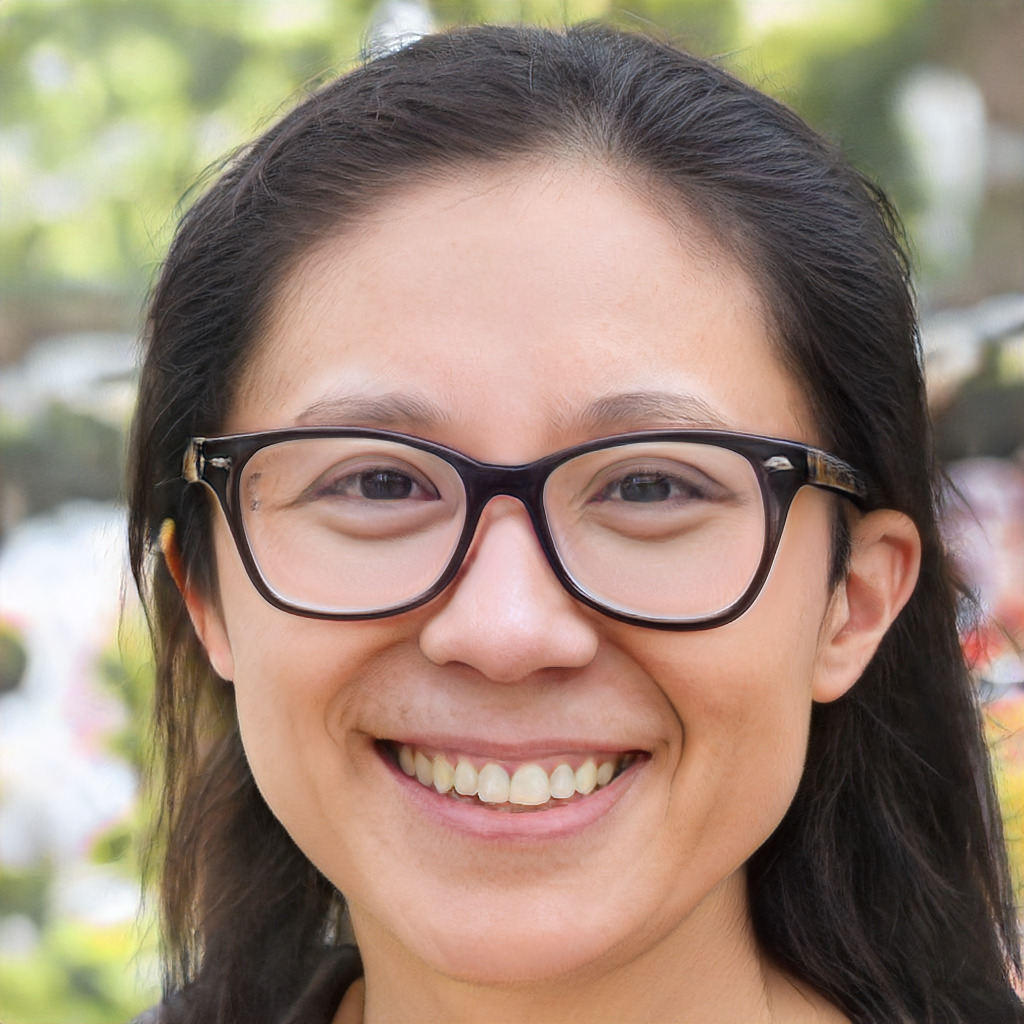} \\
        & \multicolumn{3}{c}{Age $\myarrow$} 
        
    \end{tabular} & 
    \begin{tabular}{c c c c}
        \raisebox{0.105in}{\rotatebox{90}{Ours}} &
        \includegraphics[width=0.135\linewidth]{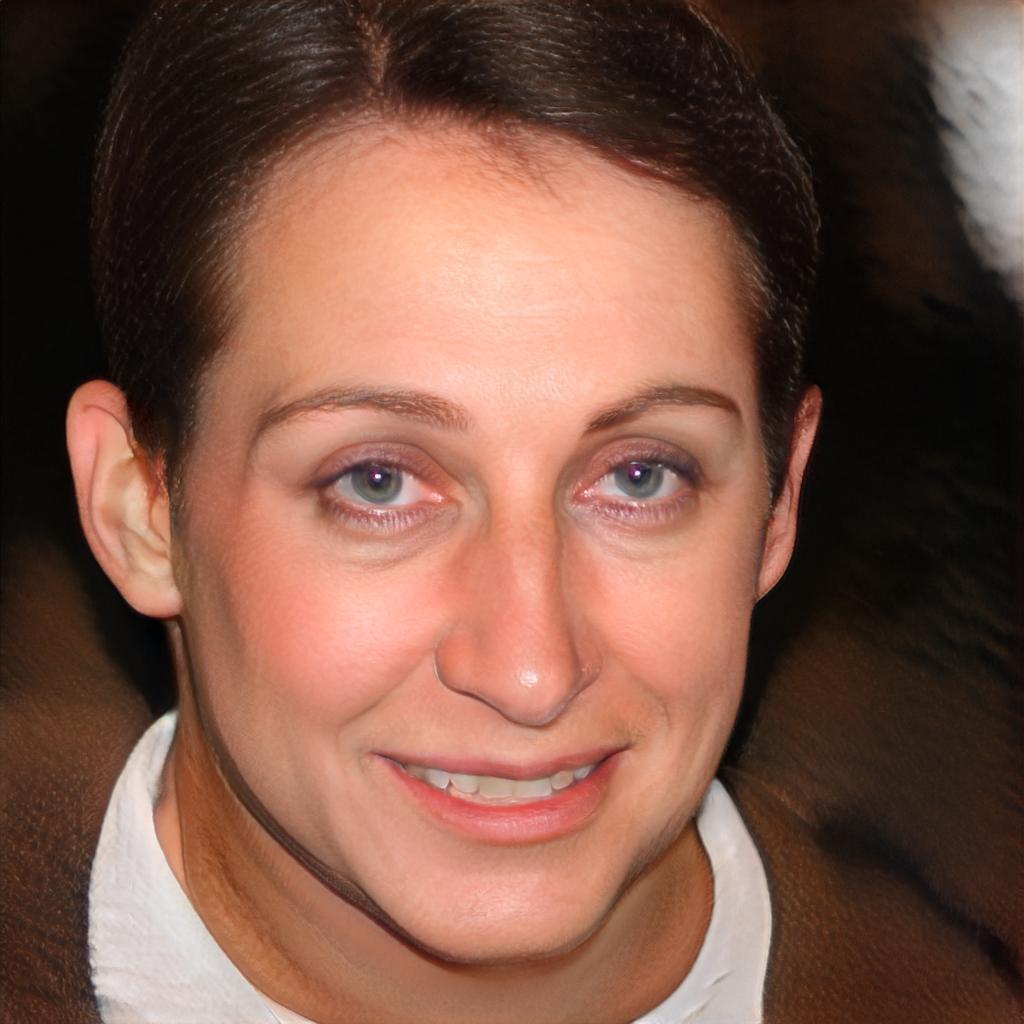} &
        \includegraphics[width=0.135\linewidth]{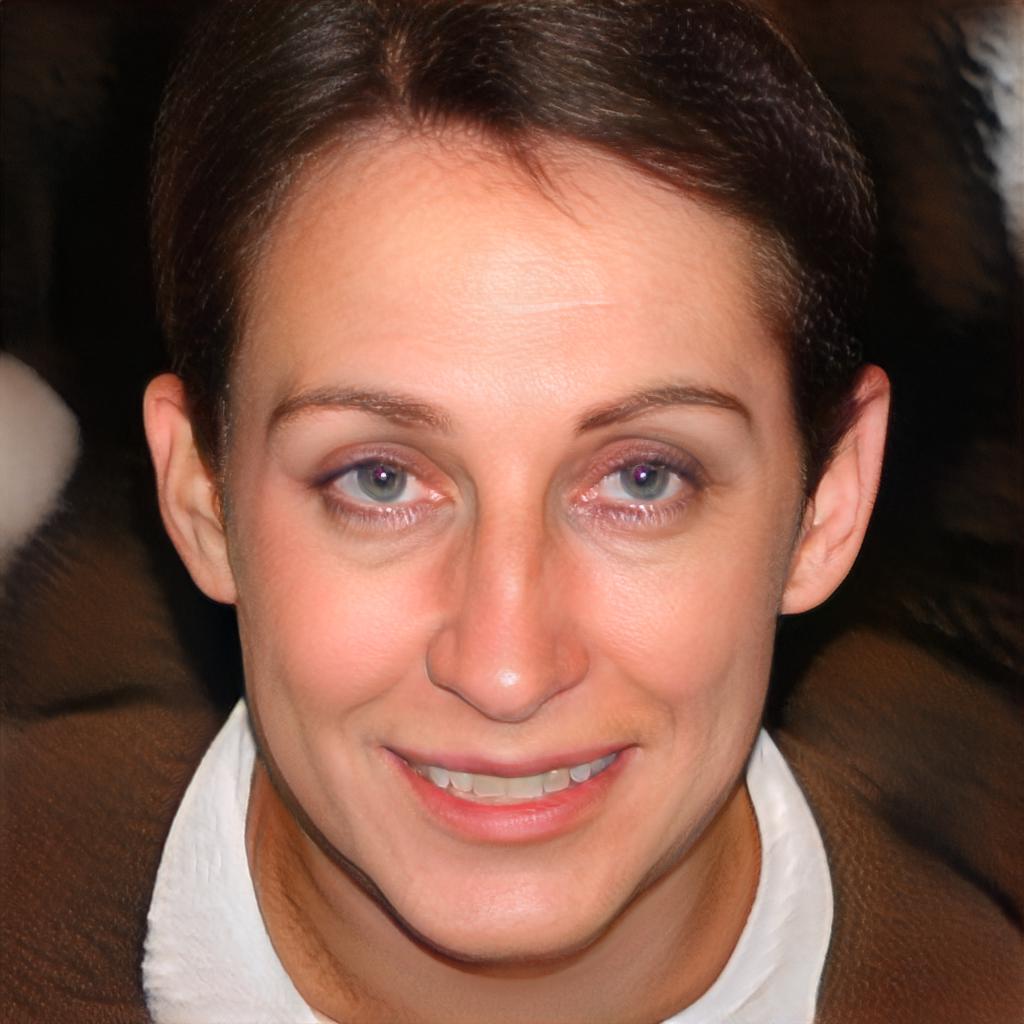} &
        \includegraphics[width=0.135\linewidth]{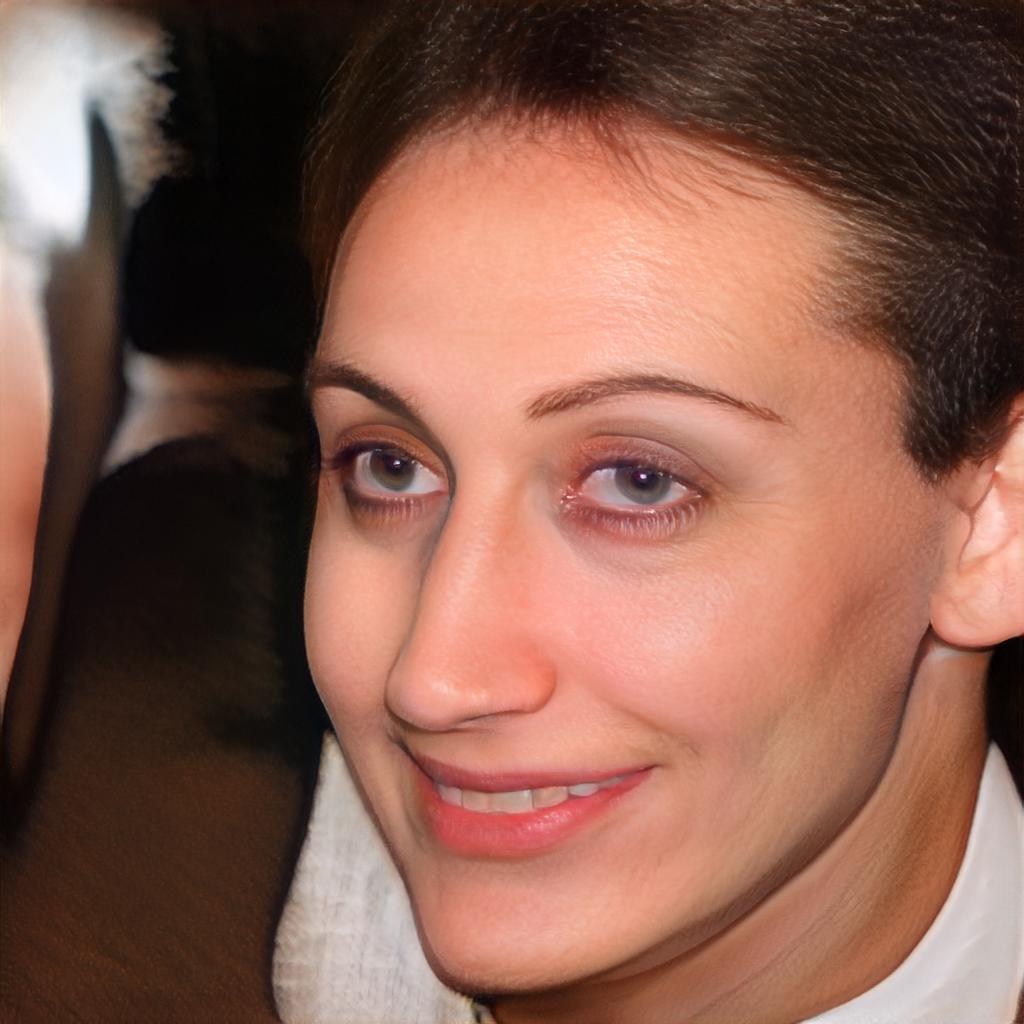} \\
        
        \raisebox{0.085in}{\rotatebox{90}{\tiny\begin{tabular}{c@{}c@{}}Frozen \\ Generator \end{tabular}}} & 
        \includegraphics[width=0.135\linewidth]{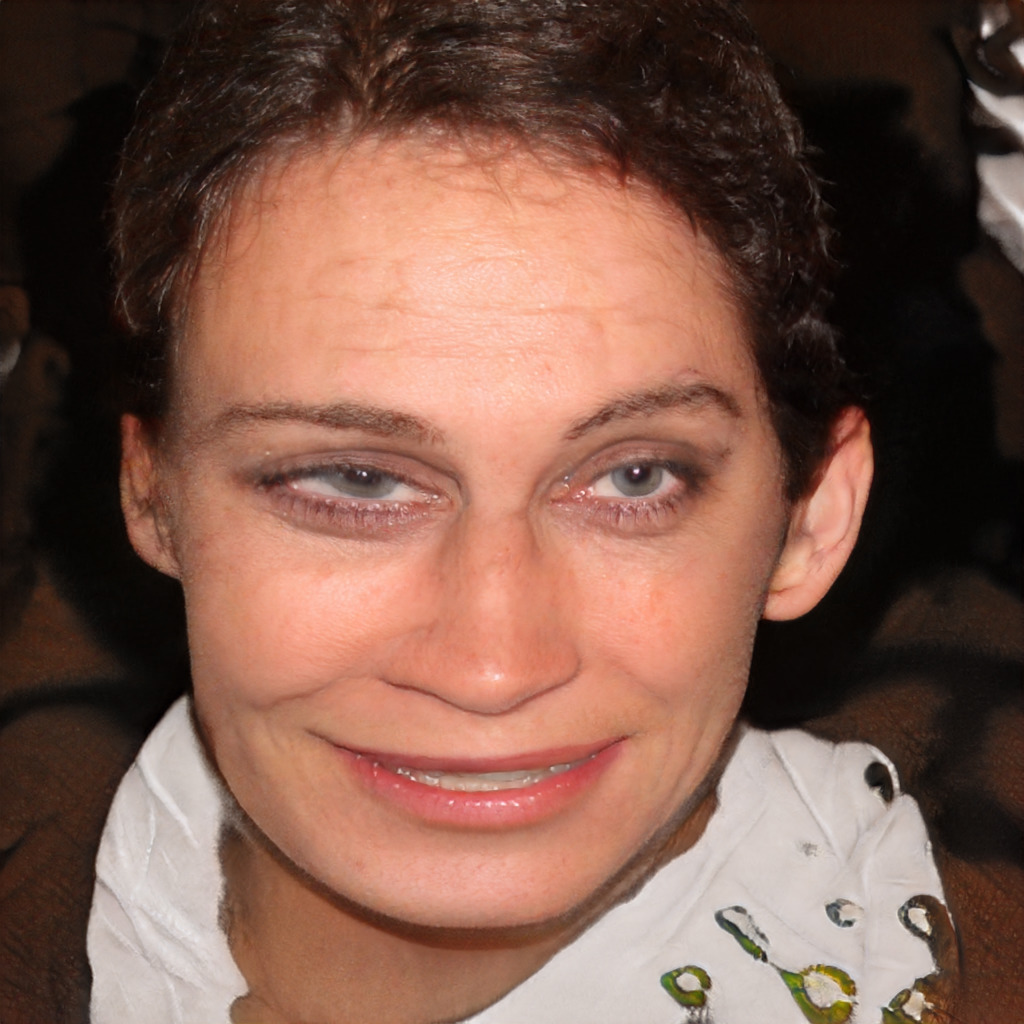} &
        \includegraphics[width=0.135\linewidth]{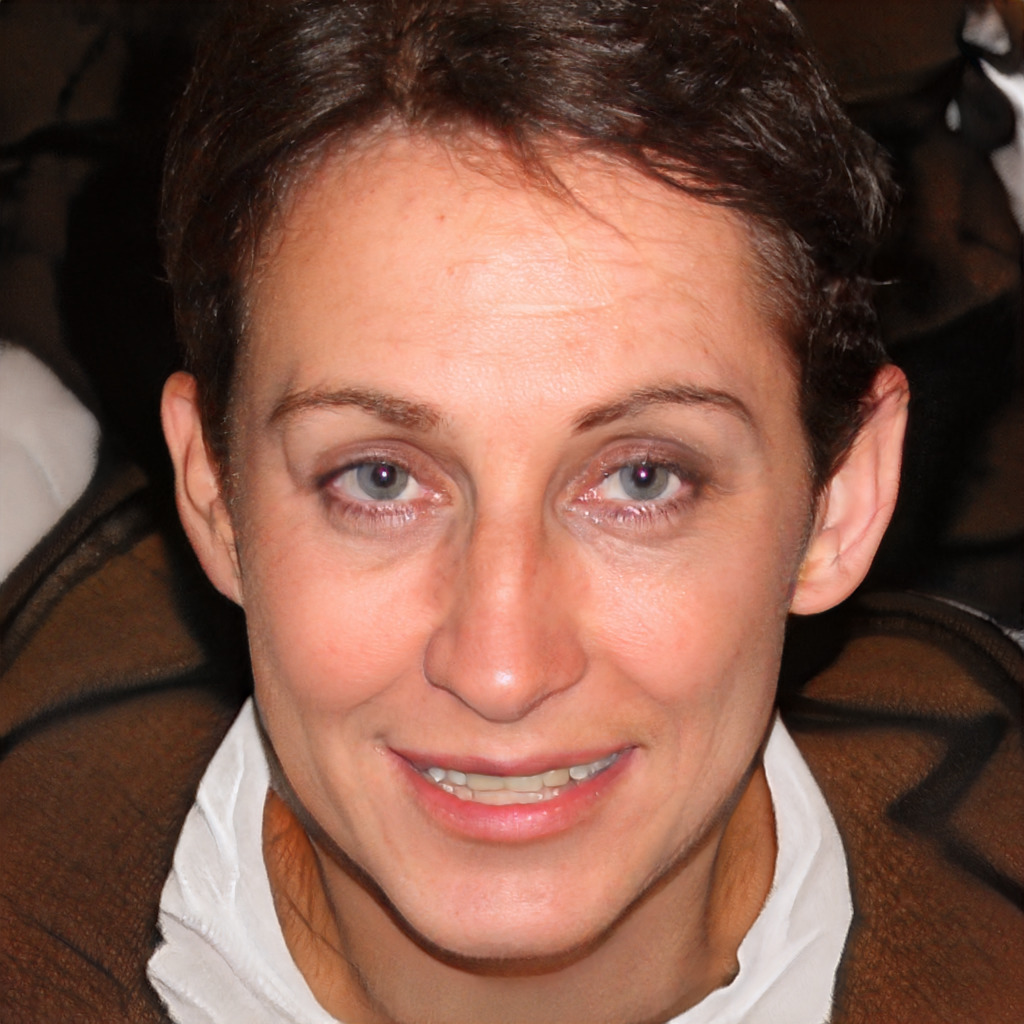} &
        \includegraphics[width=0.135\linewidth]{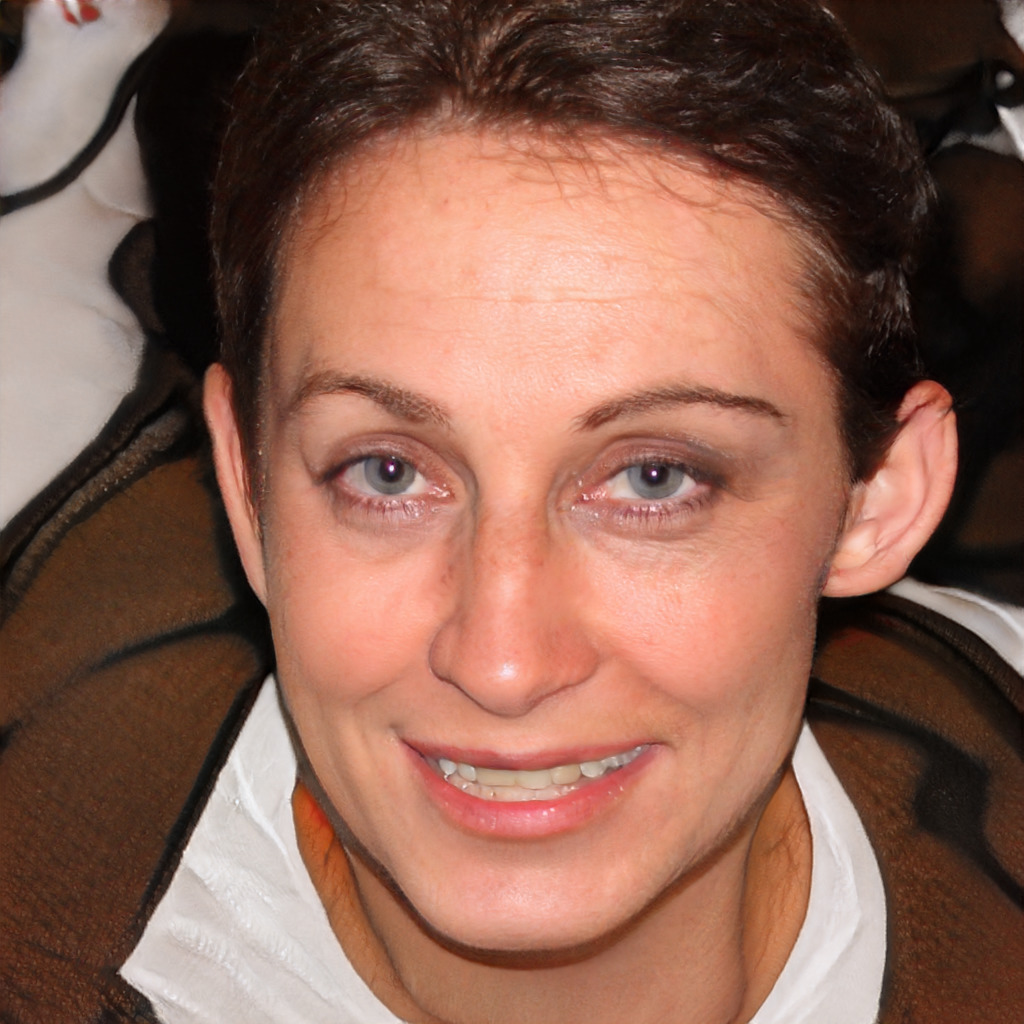} \\

        \raisebox{0.105in}{\rotatebox{90}{Ours}} &
        \includegraphics[width=0.135\linewidth]{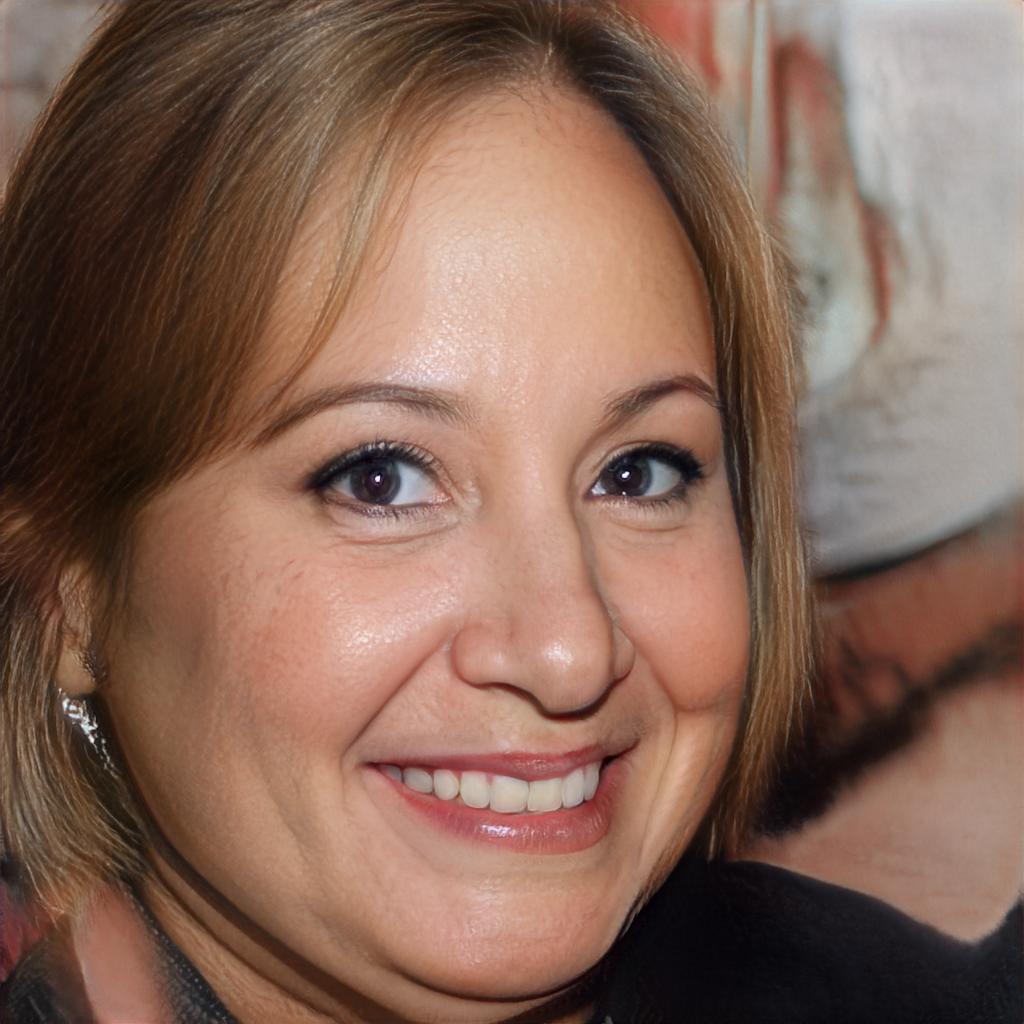} &
        \includegraphics[width=0.135\linewidth]{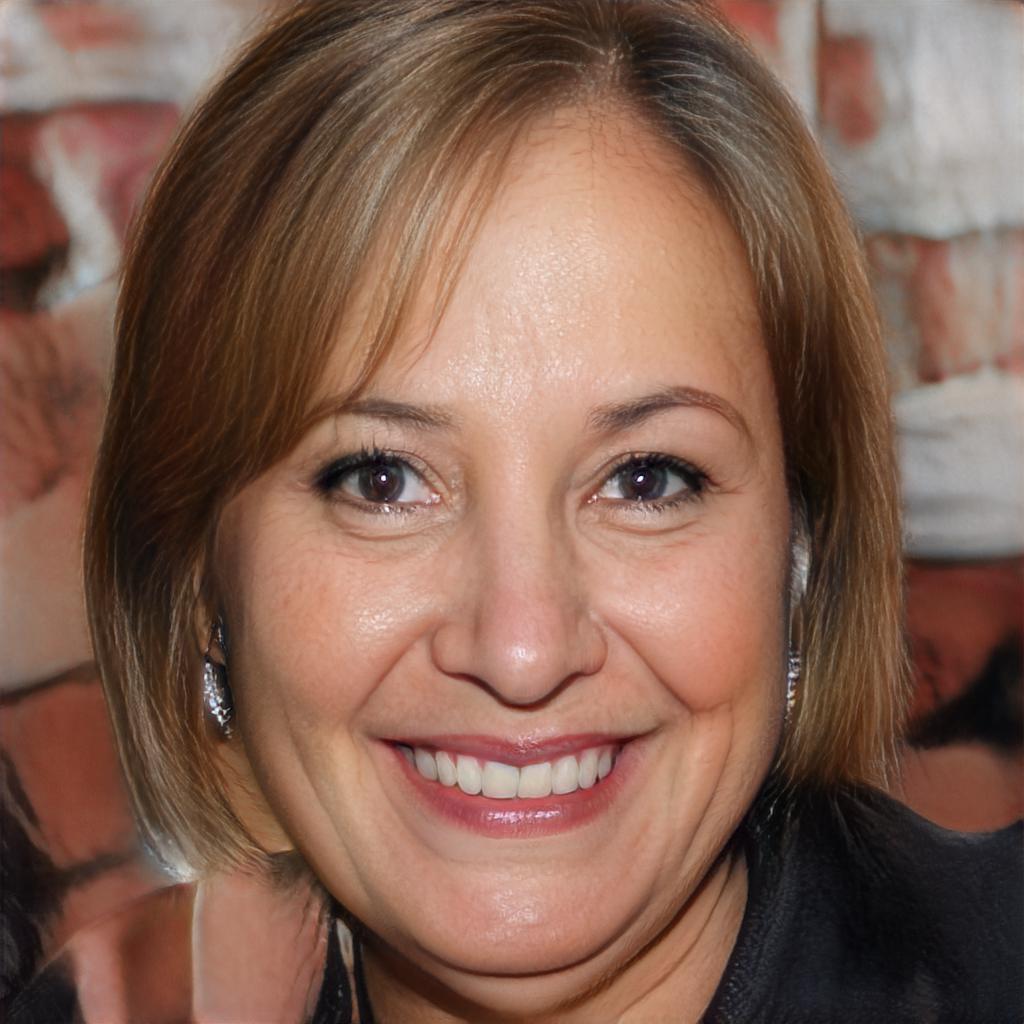} &
        \includegraphics[width=0.135\linewidth]{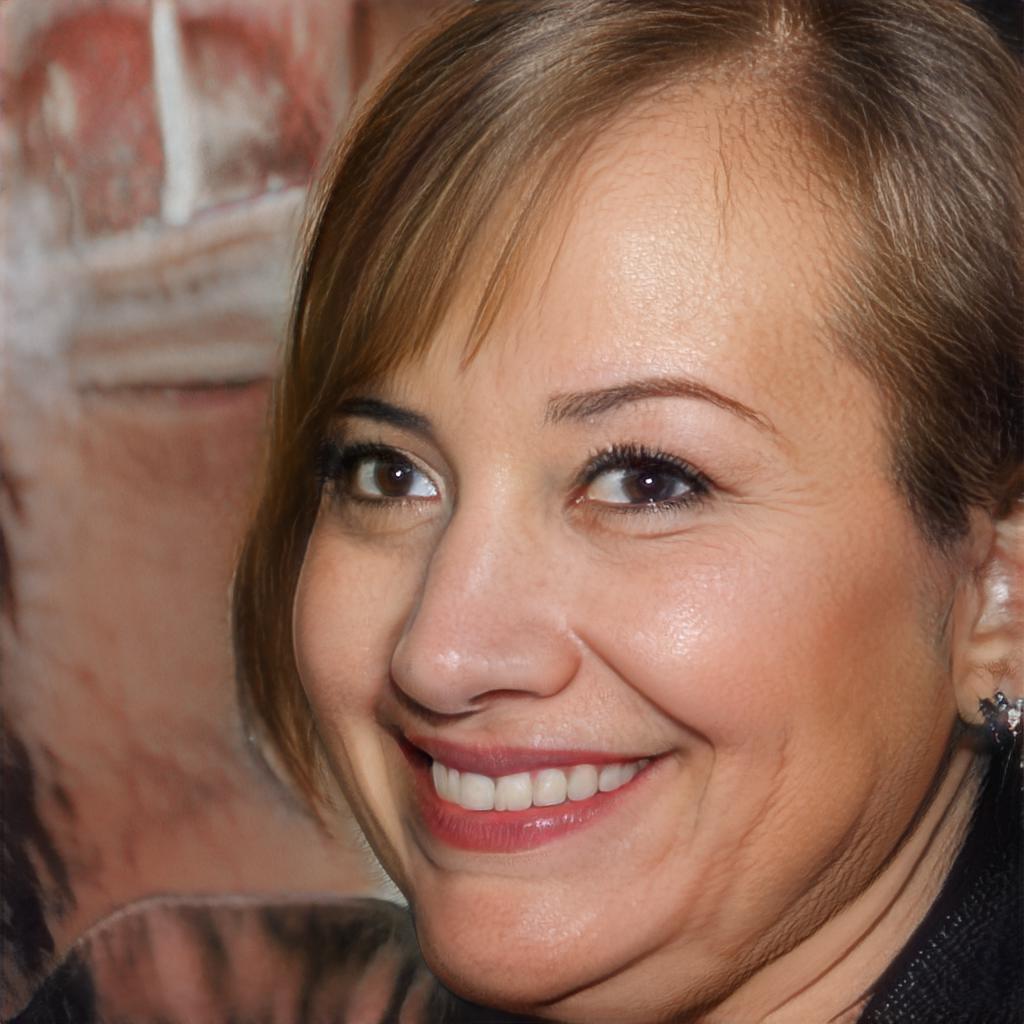} \\
        
        \raisebox{0.035in}{\rotatebox{90}{\tiny\begin{tabular}{c@{}c@{}}Non-Uniform \\ Sampling \end{tabular}}} & 
        \includegraphics[width=0.135\linewidth]{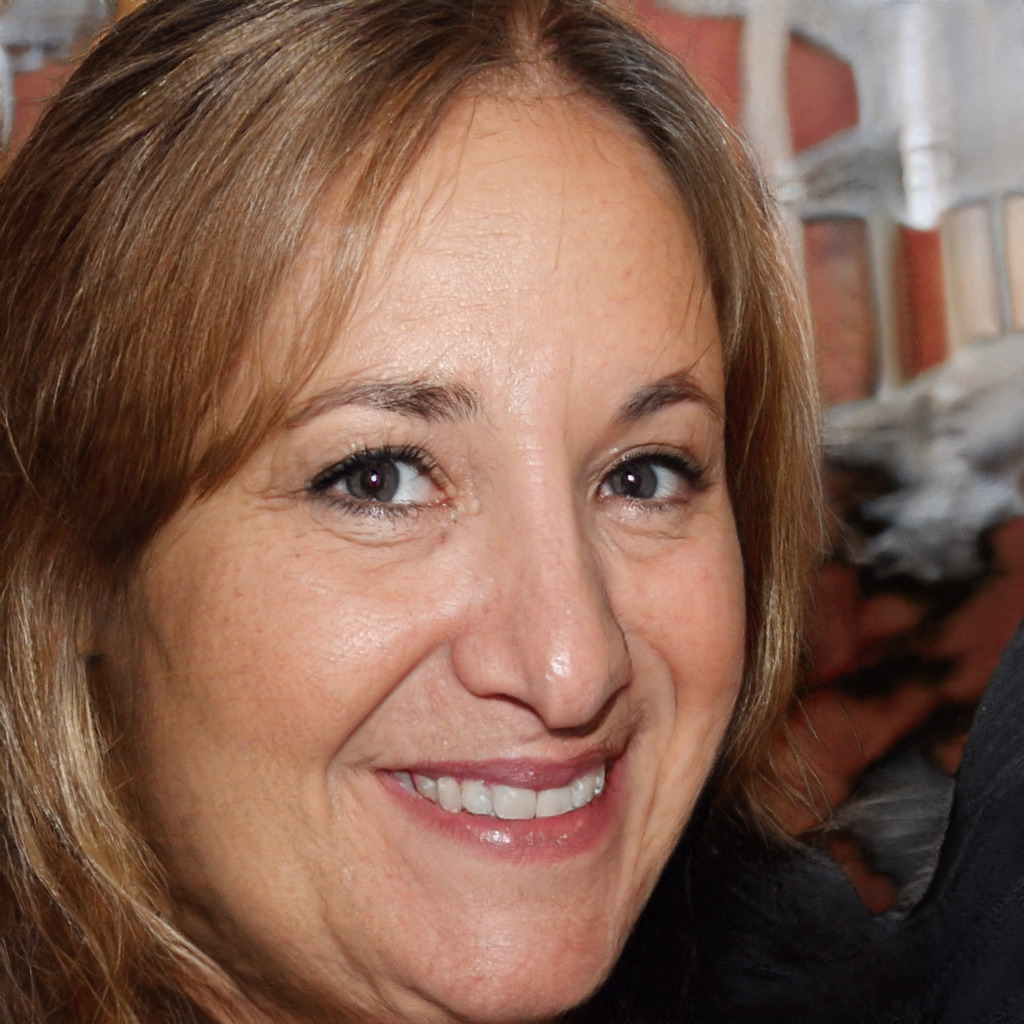} &
        \includegraphics[width=0.135\linewidth]{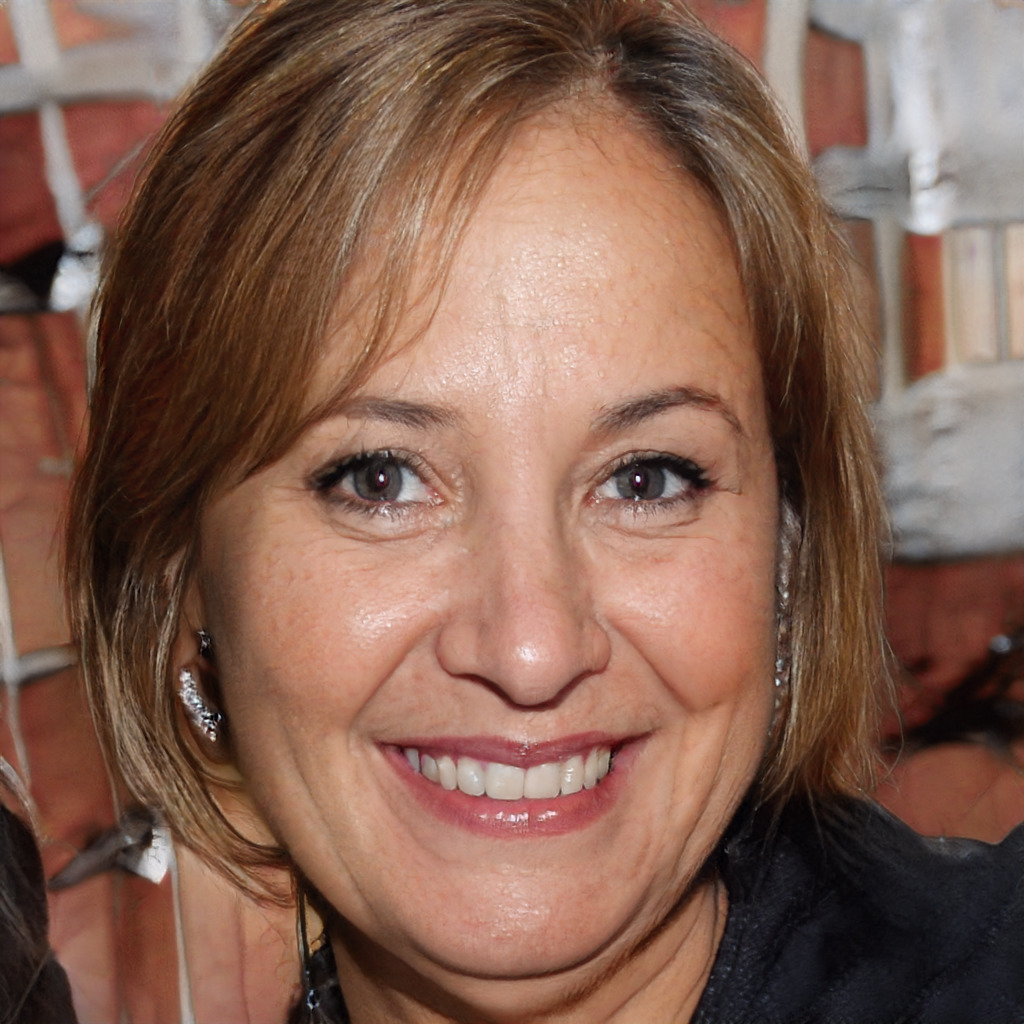} &
        \includegraphics[width=0.135\linewidth]{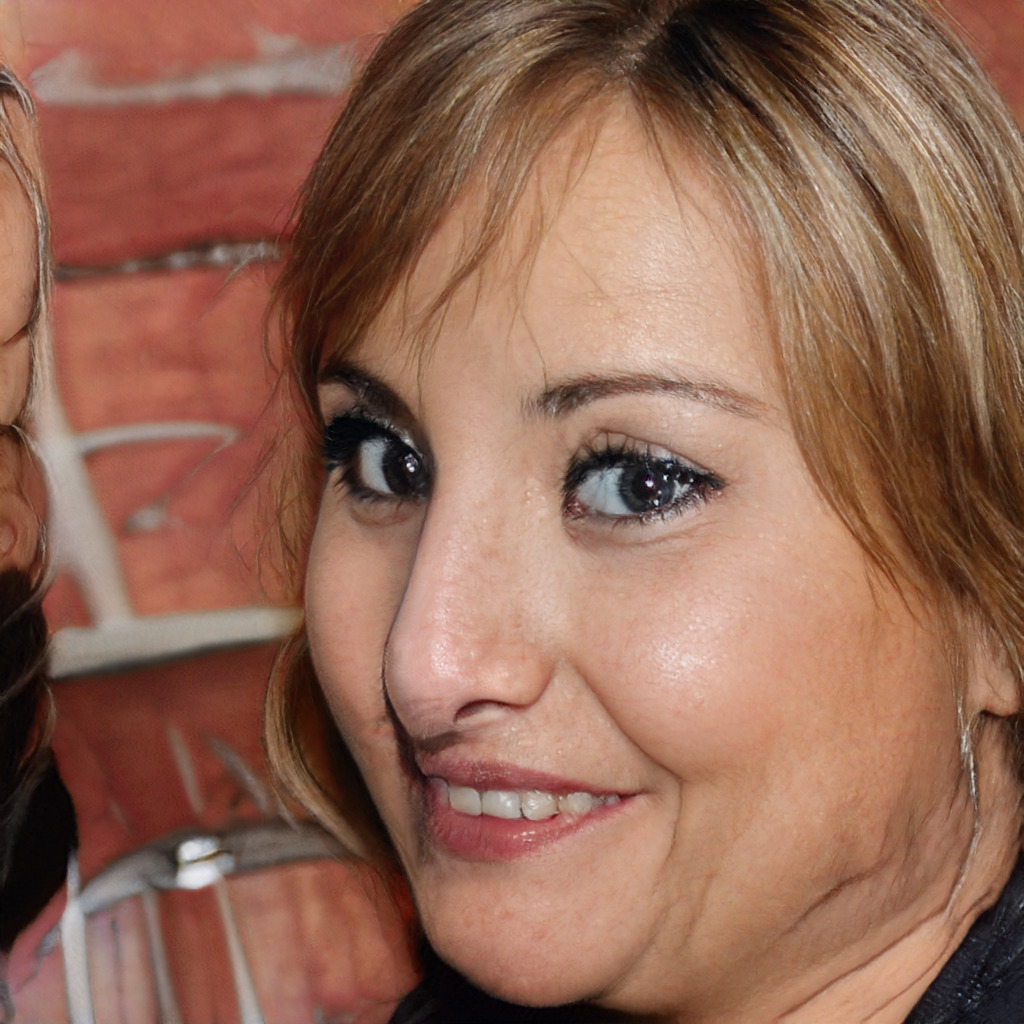} \\
        & \multicolumn{3}{c}{Pose $\myarrow$} \\

    \end{tabular}
    
    \end{tabular}
    }
    \vspace{-4pt}
    \caption{Qualitative ablation results. Using a conditional mapping network or binary classifier labels leads to reduced range of control. Non-uniform sampling leads to worsened performance for rare modalities. Not fine-tuning the generator leads to lack of control or severe image corruptions when attempting to modify the constant.}
    \label{fig:ablation} \vspace{-5pt}
\end{figure} 

%% file: future.tex
\section{Discussion}

We presented a self-conditioned generative model, designed to tackle the inherent Generative Bias of GANs and improve image editing. Our method leverages existing linear latent traversal methods, and empowers them to properly deal with minority modalities. In doing so, it achieves better identity preservation and direct control over image attributes.

Our results demonstrate the benefits of keeping fairness considerations in mind when dealing with generative tasks. More importantly, they provide further proof that there exist venues through which these, often painful, biases can be mitigated --- \textit{without} having to collect additional, unbiased data.

While our network demonstrated improved performance in regions that suffer from the Generative Bias -- \ie, the GAN's tendency towards mode collapse around minority modalities -- we observe that in some cases it can increase the network's susceptibility to the other kind of bias which affects the network - entanglement of attributes in the dataset. However, in many cases this effect can be alleviated by training a multi-attribute model to further condition on the entangled attribute. 

In the future, we would like to further develop mechanisms that better exploit the existing data in order to promote better control over specific attributes of interest.
We hope that our work serves as additional motivation for further explorations into fairer generative models, as these can impact both ethical \textit{and} practical concerns.